%% file: main.tex
\documentclass{report}
\usepackage[online]{suthesis-2e}

\usepackage[pdftex]{graphicx}
\usepackage[small]{caption}
\usepackage{subcaption}
\graphicspath{./figures}
\DeclareGraphicsExtensions{.pdf,.jpeg,.png}
\usepackage{wrapfig}
\usepackage{mathtools,fancyhdr,amsmath,amssymb,amsthm,enumitem}
\usepackage{bm,upgreek}
\usepackage{algorithm,setspace}
\usepackage{algpseudocode}
\usepackage{url}
\usepackage{tabularx,makecell,multirow,booktabs,colortbl,array,threeparttable}
\usepackage{thm-restate}
\usepackage[numbers]{natbib}

\input{macros}

\input{definitions}

\dept{Electrical Engineering}

\begin{document}
\title{Learning Preferences\\
       for Interactive Autonomy}
\author{Erdem B\i y\i k}
\principaladvisor{Dorsa Sadigh}
\firstreader{Emma Brunskill}
\secondreader{Chelsea Finn}
 
\beforepreface
\prefacesection{Abstract}
\input{00_thesis/01_abstract}
\prefacesection{Acknowledgments}
\input{00_thesis/02_acknowledgments}
\afterpreface

\chapter{Introduction}
\input{01_introduction/00_intro}
\section{Thesis Approach}
\input{01_introduction/01_thesis_approach}
\section{Contributions}
\input{01_introduction/02_contributions}
\section{Thesis Organization}
\input{01_introduction/03_organization}

\chapter{Background}\label{chap:background}
\section{Reinforcement Learning (RL)}
\input{02_background/01_rl}
\section{Inverse Reinforcement Learning (IRL)}
\input{02_background/02_irl}

\chapter{Learning Reward Functions via Comparative Feedback}\label{chap:learning}
\input{03_learning/00_intro}
\section{Incorporating Comparisons into IRL}
\input{03_learning/01_pairwise_comparisons}
\section{Learning Non-parametric Rewards via Pairwise Comparisons}
\input{03_learning/02_gp}
\section{Incorporating Ordinal Feedback along with Comparisons}
\input{03_learning/03_roial}
\section{More Expressive Feedback: Scale Questions}
\input{03_learning/04_scale}
\section{Learning Multimodal Rewards via Ranking Queries}
\input{03_learning/05_rankings}
\section{Hierarchical Comparison Queries for Non-stationary Rewards}
\input{03_learning/06_hierarchical}

\section{Chapter Summary}
\input{03_learning/07_summary}

\chapter{Active Querying for Comparative Feedback}\label{chap:active}
\input{04_active/00_intro}
\section{Choosing Queries with Volume Removal}
\input{04_active/01_volume_removal}
\section{Choosing Queries with Mutual Information}
\input{04_active/02_information_gain}
\section{Active Querying for Preference-based GP Regression}
\input{04_active/03_gp}
\section{ROI Active Learning with Comparisons and Ordinal Feedback}
\input{04_active/04_roial}
\section{Active Querying for Scale Feedback}
\input{04_active/05_scale}
\section{Active Querying for Multimodal Rewards}
\input{04_active/06_rankings}
\section{Active Generation of Hierarchical Queries}
\input{04_active/07_hierarchical}
\section{Batch-mode Active Querying for Time-Efficiency}
\input{04_active/08_batch}

\section{Chapter Summary}
\input{04_active/09_summary}

\chapter{Final Words}\label{chap:conclusion}
\input{05_conclusion/00_intro}
\section{Challenges}
\input{05_conclusion/01_challenges}
\section{Closing Thoughts}
\input{05_conclusion/02_closing_thoughts}

\appendix
\chapter{Proofs}
\input{99_appendix/01_proofs}

\chapter{Derivations}
\input{99_appendix/02_derivations}
\chapter{Implementation Details}
\input{99_appendix/03_implementation_details}
\chapter{Experiment Details}
\input{99_appendix/04_experiment_details}

\chapter{Additional Results}
\input{99_appendix/05_additional_results}

\bibliographystyle{plainnat}
\bibliography{refs}
\end{document}

%% file: macros.tex
\newcommand{\argmin}{\operatornamewithlimits{arg\,min}}
\newcommand{\argmax}{\operatornamewithlimits{arg\,max}}
\DeclarePairedDelimiter\abs{\lvert}{\rvert}
\DeclarePairedDelimiter\norm{\lVert}{\rVert}
\newcommand{\asymeq}{\stackrel{\boldsymbol{\cdot}}{=}}

\newtheorem{theorem}{Theorem}
\newtheorem{proposition}{Proposition}

\newtheorem{example}{Example}

\newtheorem{remark}{Remark}
\newtheorem{definition}{Definition}

%% file: definitions.tex
\newcommand\mdp{\mathcal{M}}
\newcommand\stateSpace{\mathcal{S}}
\newcommand\actionSpace{\mathcal{A}}
\newcommand\transitionFunction{\mathcal{T}}
\newcommand\rewardFunction{r}
\newcommand\trajectoryRewardFunction{R}
\newcommand\horizon{T}
\newcommand\state{s}
\newcommand\action{a}
\newcommand\timestep{t}
\newcommand\trajectory{\xi}

\newcommand\initialStateDistribution{\pi_0}

\newcommand\trajectorySpace{\Xi}
\newcommand\weights{w}

\newcommand\trajectoryFeaturesFunction{\varPhi}
\newcommand\demonstration{\trajectory_D}
\newcommand\dataset{\mathcal{D}}
\newcommand\demonstrationDataset{\dataset_D}
\newcommand\comparisonDataset{\dataset_C}
\newcommand\rankingDataset{\dataset_R}
\newcommand\scaleDataset{\dataset_S}
\newcommand\ordinalDataset{\dataset_O}
\newcommand\hierarchicalDataset{\dataset_H}
\newcommand\query{Q}
\newcommand\queryResponse{q}
\newcommand\belief{b}
\newcommand{\rationalityCoefficient}{\beta}
\newcommand{\demonstrationRationalityCoefficient}{\rationalityCoefficient^D}
\newcommand{\comparisonRationalityCoefficient}{\rationalityCoefficient^C}
\newcommand\weightsSampleSet{\Omega}
\newcommand\costFunction{c} 
\newcommand{\numberOfFeatures}{d}
\newcommand\gpRewardFunction{f}
\newcommand\preferenceNoiseStd{\sigma_C}
\newcommand\standardNormalCdf{\Phi}
\newcommand\standardNormalPdf{\phi}
\newcommand\gpMeanVector{\bm{\upmu}}
\newcommand\gpCovarianceMatrix{\mathbf{K}}
\newcommand{\gpRbfHyperparameter}{\vartheta}
\newcommand{\gpKernelFunction}{k}
\newcommand{\gpNegativeHessianOfLoglikelihood}{\mathbf{W}}
\newcommand\gpPosteriorMean{\mu}
\newcommand\binaryEntropyFunction{h}
\newcommand\gpGFunction{g}
\newcommand\gpMFunction{m}

\newcommand\plannerFunction{\rho}
\newcommand\regretFunction{\mathcal{R}}
\newcommand\halfspace{\Lambda}
\newcommand\maxRewardGap{\delta}
\newcommand\sensitivityThreshold{\varrho}
\newcommand\sliderStepSize{\nu}
\newcommand\scaleNoiseStd{\sigma_S}
\newcommand{\scaleFeasibleSet}{\mathcal{F}}

\newcommand\numberOfModes{M}
\newcommand\modeIndex{m}
\newcommand\mixingCoefficient{\alpha}
\newcommand\rankingIndices{\iota}

\newcommand\batchSize{k} 
\newcommand\numberOfQueries{K}
\newcommand\datasetOfQueries{\mathcal{\numberOfQueries}}
\newcommand\reducedQuerySet{\mathcal{X}}
\newcommand\batch{X}
\newcommand\batchExample{A}
\newcommand\dppKernel{L} 
\newcommand\diversityParameter{\lambda}
\newcommand\scoreParameter{\gamma}
\newcommand\similarityMatrix{S}
\newcommand\similarityMatrixHyperparameter{\sigma_{\textrm{DPP}}}
\newcommand\featureDifference{\psi}
\newcommand\score{u}
\newcommand\mcrStepSize{\eta_{\textrm{MCR}}}

\newcommand\hierarchicalPriorMatrix{G}
\newcommand\modeUtilityFunction{\trajectoryRewardFunction^u}
\newcommand\utilityWeights{\theta}

\newcommand\ordinalThresholdSet{B^o}
\newcommand\ordinalCategory{\mathcal{B}^o}
\newcommand\ordinalNoiseStd{\sigma_O}
\newcommand\roialConservatism{\varepsilon}

\newcommand\aboutEqual{\Upsilon}
\newcommand\minimumPerceivableDifference{\varsigma}
\newcommand\queryCostHyperparameter{\varpi}
\newcommand\humanParametersAppendix{\kappa}
\newcommand\posteriorCovarianceAppendix{\Sigma}

\newcommand\gpModeUtilityFunction{\gpRewardFunction^{u}}

%% file: 00_thesis/01_abstract.tex
When robots enter everyday human environments, they need to understand their tasks and how they should perform those tasks. To encode these, reward functions, which specify the objective of a robot, are employed. However, designing reward functions can be extremely challenging for complex tasks and environments. A more promising approach is to learn reward functions from humans. Recently, several robot learning works embrace this approach and leverage human demonstrations to learn the reward functions. Known as inverse reinforcement learning, this approach relies on a fundamental assumption: humans can provide near-optimal demonstrations to the robot. Unfortunately, this is rarely the case -- human demonstrations to the robot are often suboptimal due to various reasons, e.g., difficulty of teleoperation, robot having high degrees of freedom, or humans' cognitive limitations.

This thesis is an attempt towards learning reward functions from human users by using other data modalities that are more reliable. Specifically, this thesis studies how reward functions can be learned using \emph{comparative feedback}, in which the human user compares multiple robot trajectories instead of (or in addition to) providing demonstrations. To this end, we first propose various forms of comparative feedback, e.g., pairwise comparisons, best-of-many choices, rankings, scaled comparisons; and describe how a robot can use these various forms of human feedback to infer a reward function, which may be parametric or non-parametric. We discuss the pros and cons of each comparative feedback modality in detail, and show how such feedback enables us to outperform standard inverse reinforcement learning that only utilizes demonstrations.

An important limitation of comparative feedback is that each comparison carries only a small amount of information: instead of observing the humans' actions at every time step of a demonstration, we only observe their comparison between trajectories. This harms data-efficiency, which is crucial in robotics due to the cost of collecting data (especially when it is coming from humans). To solve this, we propose \emph{active learning} techniques to enable the robot to ask for comparison feedback that optimizes for the expected information that will be gained from that user feedback.

While showcasing the benefits of these various techniques for learning and active querying, we also demonstrate its applicability in a wide variety of domains. Our experiment domains range from autonomous driving simulations to home robotics, from standard reinforcement learning benchmarks to lower-body exoskeletons.

%% file: 00_thesis/02_acknowledgments.tex
First and foremost, I would like to thank my advisor Dorsa Sadigh. It has been an absolute pleasure to learn from and work with her. Even though I had almost no experience in robotics when I started working with Dorsa, her passion, positivity and hard-work have always inspired and motivated me to learn more. Being her first Ph.D. student has been an honor and gave me invaluable experience. I would not have been where I am today without her guidance and support.

Many of my works in graduate school have been in close collaboration with other advisors. Great academic advice by Ramtin Pedarsani made a lot of positive impact not only in my studies but also in my life. Dylan P. Losey and Mykel J. Kochenderfer have always been role models with their dedication and time management skills. My collaborations with Nima Anari, Yisong Yue, Yanan Sui, Stephen L. Smith, and Aaron D. Ames contributed a great deal to this thesis. I would also like to thank my internship supervisors and collaborators at Google Research: Yinlam Chow, Mohammad Ghavamzadeh, Chih-wei Hsu, Alex Haig, and Craig Boutilier for giving me a perspective for preference-based learning algorithms outside robotics. It has also been an honor for me to collaborate and be co-authors with great mentors from both industry and academia: Adrien Gaidon, Guy Rosman, Judith E. Fan, Shahrouz Ryan Alimo, and Andrea Goldsmith. Although we have not collaborated on a research project yet, Scott Niekum, Emma Brunskill, and Chelsea Finn gave me a lot of guidance and support during my Ph.D. 

My first research experience was at Bilkent University during my undergraduate studies with Tolga {\c C}ukur in 2016. His continuous support to this date has been invaluable and I am thankful to him, as well as his students Efe Il\i cak, K\"ubra Keskin, L. Kerem {\c S}enel, Salman U. H. Dar. I was also lucky enough to work and be co-authors with Aykut Ko{\c c} at ASELSAN as a research engineer, and with Mohamad Dia and Jean Barbier in R\"udiger Urbanke's lab at EPFL as a research intern. Last but not least, I enjoyed working with and learning from my mentors at Bilkent: Orhan Ar\i kan, Cem Tekin, Emine \"Ulk\"u Sar\i ta{\c s}, \.Ismail Uyan\i k, and my collaborators: H. Can Baykara, Gamze G\"ul, Deniz Onural, Ahmet Safa \"Ozt\"urk, and \.Ilkay Y\i ld\i z.

I would also like to thank all my amazing co-authors who have been one of the main sources of learning and incredibly fun to work with. Daniel A. Lazar is not only an efficient collaborator with his math skills and sharpness, but also a great travel companion. Minae Kwon, who was already an intern in the lab when I started, has always been welcoming and I learned a lot from chatting with her. I also had the privilege to work with Nils Wilde, Anusha Lalitha, Amir Maleki, Mark Beliaev, Zhangjie Cao, Malayandi Palan, Nicholas C. Landolfi, Sydney M. Katz, Kejun (Amy) Li, Maegan Tucker, Ellen Novoseller, Chandrayee Basu, Erik Brockbank, Rajarshi Saha, Zhixun (Jason) He, Vivek Myers, Woodrow Z. Wang, Nicolas Huynh, Aditi Talati, Suvir Mirchandani, Kenneth Wang, Gleb Shevchuk, Zheqing (Bill) Zhu, Karan Bhasin, Jonathan Margoliash, and Allan Raventos.

Although the Ph.D. programs are infamously said to be lonely, I never felt in that way thanks to my friends. Special thanks to my labmates at ILIAD: Mengxi Li, Sidd Karamcheti, Andy Shih, Megha Srivastava, Suneel Belkhale and Zhiyang (Jerry) He for making the lab a great place to work at. I also thank Turkish Student Association, especially Burak Bartan, Serhat Arslan, Atiye Cansu Erol Arslan, S\"uleyman Kerimov, Okan Atalar, H\"useyin \.Inan for our board game nights; as well as my close friends from Turkey: Melike Ersoy, Batuhan S\"utba{\c s}, \"Omer Mert Aksoy, G\"orkem \"Unl\"u, Naz Yetimo{\u g}lu, Fatih Karao{\u g}lano{\u g}lu, \"Omer Arol for supporting me even from thousands of kilometers away and the virtual social activities we have had during the pandemic.

Finally, I would like to thank my family for all their love and support. My parents Reyhan B\i y\i k and \.Irfan B\i y\i k have always believed and put confidence in me. They brought up me in a way that made this thesis possible: they always taught and encouraged me to pursue what I enjoy, and do it with passion, responsibility and determination. And my sister, Beg\"um B\i y\i k, has been a ``best friend instead of a sibling" as she always wanted when she was younger; and my main source of good songs, which have been essential while working long hours. Overall, I am thankful to all members of my family: their perspective allowed me to understand ``Science is the most reliable guide for civilization, for life, for success in the world. Searching a guide other than the science is meaning carelessness, ignorance and heresy."

%% file: 01_introduction/00_intro.tex
In recent years, we have seen enormous effort to integrate robots and systems equipped with artificial intelligence (AI) into the society. While these agents are increasingly becoming part of our lives, most of their current interactions with the humans is one-way, e.g., a driver commands a vehicle to park autonomously, or the vehicle warns the driver about weather conditions. However, their successful integration will require them to intelligently \emph{learn}, \emph{adapt to}, and \emph{influence} the humans and other AI agents.

These two-way interactions, where agents need to learn, adapt to, and influence each other; appear in almost all real-life scenarios. Human teams that are good at collaborating are often the ones where each individual adapted themselves to the others, e.g., sports teams train together rather than trying to improve individually. However, AI agents are not yet capable of this adaptation: their inability to model others led to problems in several occasions. For example, price-setting bots tried to sell a book for 23.7 million dollars on an online retail website after blindly competing with each other and not realizing that by increasing the price, the other bots will increase the price as well, while no human would be willing to pay this price \cite{cnn_amazon_news}. Though this is an old example, we still see similar issues arise: autonomous cars fail to change lanes because they do not know the other drivers will slow down if they simply nudge in front of them \cite{dailymail_waymo_news}. The approach in this thesis to enable the robots to achieve the two-way interactions is inspired by how humans interact: we efficiently infer our partners' goals to optimize our behavior. For example, we move to one side of the sidewalk when we see a cyclist is approaching. If there is a mismatch between the inferred goal and our own goal, we try to influence our partners, e.g., if the cyclist moves to the same side, we stop for a second to imply we want to stay on this side and they should use the other side.

To achieve this human-like interaction, robots should understand the objective in the task, which encodes what they need to do and how they should do what they do. Designing these objective by hand, known as reward function, is extremely challenging. A more promising approach is to learn it from humans. Recent works dominantly focused on learning from human demonstrations of the task. However, human demonstrations are often suboptimal due to various reasons, e.g., difficulty of teleoperation, robots' high degrees of freedom, humans' cognitive limitations, etc. Therefore, many questions arise: What are some other forms of human feedback that enables robots to more reliably learn reward functions? How can robots learn from multiple data modalities? How can these methods extend to the cases where the reward function is multimodal or non-stationary? How can robots optimize for data-efficiency to mitigate the high costs of data collection? In this thesis, we attempt to answer these questions.

%% file: 01_introduction/01_thesis_approach.tex
\begin{wrapfigure}{R}{.35\linewidth}
    \centering
    \includegraphics[width=\linewidth]{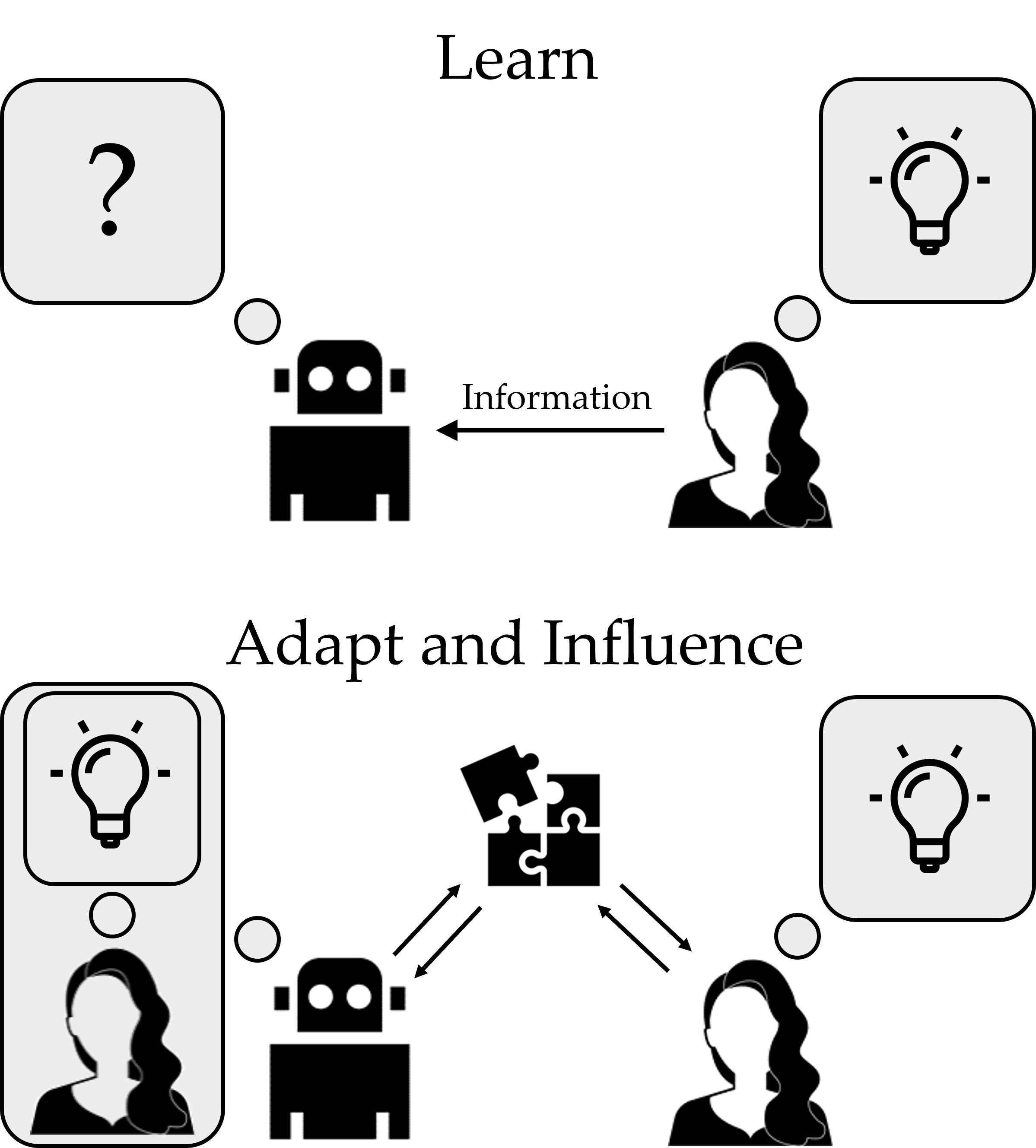}
    \caption{\textbf{(top)} The robot should first learn a model of the agent it is interacting with. \textbf{(bottom)} It will then adapt to and influence the other agent in the task.}
    \label{fig:01_01_introduction_summary}
\end{wrapfigure}
Integrating robots and systems equipped with AI into the society requires a thorough understanding of how they may learn, adapt to and influence other agents. Our approach is to divide this problem into two parts (see Figure~\ref{fig:01_01_introduction_summary}) \cite{biyik2022learning_pioneers}. First, machine learning techniques that we develop in this thesis will enable AI agents to model the behaviors and goals of the other agents by leveraging different forms of information they provide. Next, these learned behaviors and goals will enable these agents to better interact with the others to achieve online adaptation, e.g., an autonomous vehicle will adapt to both its driver and the other vehicles to better optimize its route and driving style. In this thesis, we focus on the first aspect and study how robots can learn from human feedback.

Although recent works mostly focused on learning from demonstrations, they often suffer from the suboptimality of demonstrations. In this thesis, we propose using comparative feedback to learn the objectives, where human users are asked to compare multiple trajectories of a robot based on their preferences. We develop various forms of comparative feedback, and further study how they can be collected actively to improve data-efficiency, which is crucial in robotics, especially because the data are coming from human users. For these, we bridge ideas from machine learning, information theory, human-robot interaction, optimization and control theory.

%% file: 01_introduction/02_contributions.tex
\label{sec:01_02_contributions}

This thesis makes the following contributions.

\begin{quote}
    \textit{The goal of this thesis is to develop rewards learning methods for robots that leverage comparative feedback from humans in a data-efficient way.}
\end{quote}

\subsubsection{Learning Reward Functions via Comparative Feedback}
In Chapter~\ref{chap:learning}, we study various forms of comparative feedback, how to learn reward functions using them, and how to utilize them along with user demonstrations that are possibly suboptimal. Specifically, we develop and analyze the following feedback modalities:
\begin{itemize}
    \item \textbf{Pairwise comparisons:} The user selects their preferred trajectory among two options \cite{biyik2018batch,biyik2019batch,biyik2020active,katz2021preference}.
    \item \textbf{Scale feedback:} The user uses a slider bar to indicate how much they prefer one trajectory over the other in a pairwise comparison setting \cite{wilde2021learning}.
    \item \textbf{Ordinal feedback:} In addition to pairwise comparisons, the user also labels each trajectory with an ordinal feedback. e.g., ``Bad", ``Neutral" and ``Good" \cite{li2021roial}.
    \item \textbf{Best-of-many choices:} We extend the pairwise comparisons so that the user will now choose their most preferred trajectory out of multiple options, possibly more than two \cite{biyik2022learning,biyik2019asking,biyik2019green,biyik2021incentivizing,beliaev2021incentivizing}.
    \item \textbf{Hierarchical choices:} To handle non-stationary reward functions, we develop hierarchical choice queries in which the user first responds to a standard best-of-many choice query. After their response, a new best-of-many choice query is presented such that its trajectories start from the final state of the user's preferred trajectory in the first query \cite{basu2019active}.
    \item \textbf{Rankings.} The user ranks multiple (more than two) trajectories from their most preferred to the least preferred \cite{myers2021learning}. This helps robots learn multimodal reward functions, e.g., when the data are coming from multiple people.
\end{itemize}

In addition, we study different forms of reward functions that encode how a robot or an AI agent should perform the task:
\begin{itemize}
    \item \textbf{Parametric reward functions:} Most of the thesis focuses on reward functions that are parametric \cite{biyik2018batch,biyik2019batch,katz2021preference,biyik2022learning,biyik2019asking,biyik2019green,biyik2021incentivizing,beliaev2021incentivizing}. In principle, such functions may range from linear functions to neural networks. However, as we take a Bayesian learning approach, functions with large parameter spaces are difficult to learn in practice. This restricts us to simple functional forms.
    \item \textbf{Non-parametric reward functions:} To solve this issue and be able to learn more complex rewards, we employ Gaussian processes and learn non-parametric reward functions \cite{biyik2020active,li2021roial}.
    \item \textbf{Multimodal reward functions:} If the data are coming from multiple people with different objectives, or the same person with varying objectives, unimodal reward functions fail to encode their preferences. Hence, we model the multimodal reward as a mixture of multiple parametric unimodal reward functions \cite{myers2021learning}.
    \item \textbf{Non-stationary reward functions:} As a specific case of multimodal reward functions, we study rewards that are non-stationary with some structure: the users' preferences change based on the history in the environment according to a parametric transition function \cite{basu2019active}.
\end{itemize}

\subsubsection{Active Querying for Comparative Feedback}
In Chapter~\ref{chap:active}, we address the problem of data inefficiency when learning from comparative feedback. As opposed to user demonstrations of the task, where each state in a trajectory receives an action label; comparisons contain very little information -- they only say some trajectories are better than some others. This means a robot may require enormous amounts of data to learn useful reward functions via comparative feedback, especially when there are no demonstrations to warm-start the learning. To mitigate this problem, we develop active learning techniques to actively query the users for the most informative comparative feedback. 

Specifically, we study active learning objectives that are based on volume removal \cite{sadigh2017active,biyik2018batch,biyik2019batch,biyik2019green,biyik2021incentivizing}, mutual information \cite{biyik2019asking,biyik2022learning,biyik2021incentivizing,biyik2020active,li2021roial,wilde2021learning,myers2021learning}, and max-regret \cite{wilde2020active,wilde2021learning}. While doing this, we follow a similar structure to Chapter~\ref{chap:learning}: we describe how each section of Chapter~\ref{chap:learning} can be extended with active querying. As a result of this choice, we defer all simulation and experiment results to Chapter~\ref{chap:active} where we not only investigate the learning performance but also analyze the benefits of active querying.

%% file: 01_introduction/03_organization.tex
Chapter~\ref{chap:background} is geared towards a reader who is inexperienced at robot learning: we present an introductory overview of reinforcement learning (RL) and inverse reinforcement learning (IRL) problems. While doing this, we do not focus on any particular solution -- instead we keep the problem formulations general enough so that we can present learning from comparative feedback using the same formulation. In fact, the focus of this thesis, learning reward functions from comparative feedback is closely related to the inverse reinforcement learning problem. In both of these problems, the goal is to learn a reward function that encodes the desired behavior of the robot or the AI agent.

Chapter~\ref{chap:learning} then starts with building upon an existing IRL solution, namely Bayesian inverse reinforcement learning \cite{ramachandran2007bayesian}, where the reward function is learned using human demonstrations of the task. Again taking a Bayesian approach, we study how comparative feedback can be used to learn the reward function. For this, we start with best-of-many choice queries (Section~\ref{sec:03_01_pairwise_comparisons}, \cite{biyik2019asking,palan2019learning,biyik2022learning}). We then focus on a specific version of these queries with only two options to compare, i.e., pairwise comparisons. Using this simpler query form enables us to learn non-parametric reward functions that are modeled via Gaussian processes (Section~\ref{sec:03_02_gp}, \cite{biyik2020active}), which we later improve with ordinal feedback (Section~\ref{sec:03_03_roial}, \cite{li2021roial}). We then extend the pairwise comparisons by providing the users with a slider bar to indicate how much they prefer one trajectory over the other, which would not be practical with general best-of-many choice queries (Section~\ref{sec:03_04_scale}, \cite{wilde2021learning}). After this detour, we go back to queries with more than two options, and study ranking queries which enable us to learn multimodal reward functions (Section~\ref{sec:03_05_rankings}, \cite{myers2021learning}). Finally, we look at a specific case of multimodal rewards where users transition between different modes according to the latest behavior of the robot (Section~\ref{sec:03_06_hierarchical}, \cite{basu2019active}). Section~\ref{sec:03_07_summary} summarizes Chapter~\ref{chap:learning}.

In Chapter~\ref{chap:active}, we focus on how to improve data-efficiency when we use learning from comparative feedback where information is very sparse as opposed to demonstrations. We follow a similar structure to Chapter~\ref{chap:learning}, i.e., we follow almost the same order to present the active querying techniques for each section in Chapter~\ref{chap:learning}. We first introduce volume removal (Section~\ref{sec:04_01_volume_removal}, originally proposed in \cite{sadigh2017active}) and mutual information (Section~\ref{sec:04_02_information_gain}, \cite{biyik2019asking,biyik2022learning}) based active learning objectives for best-of-many choice queries. We then proceed with mutual information based active querying when the reward is non-parametric and modeled as a Gaussian process (Section~\ref{sec:04_03_gp}, \cite{biyik2020active}), and extend it to the case where we also use ordinal feedback (Section~\ref{sec:04_04_roial}, \cite{li2021roial}). In addition to mutual information, we introduce max-regret based active querying in Section~\ref{sec:04_05_scale} where we focus on scale feedback \cite{wilde2021learning}. Following the same order as in Chapter~\ref{chap:learning}, we next present active methods for ranking queries when the reward is multimodal (Section~\ref{sec:04_06_rankings}, \cite{myers2021learning}) and for hierarchical choice queries when users transition between different reward modes (Section~\ref{sec:04_07_hierarchical}, \cite{basu2019active}). All these active querying methods require solving an optimization problem for each and every query, which might be computationally expensive and makes the querying process non-parallelizable. Hence, in Section~\ref{sec:04_08_batch}, we present various \emph{batch} active learning methods where multiple queries are optimized together in batches \cite{biyik2018batch,biyik2019batch}. Finally, Section~\ref{sec:04_09_summary} summarizes the chapter.

Chapter~\ref{chap:conclusion} is the final chapter of the thesis in which we discuss the limitations and open challenges, and conclude the ideas presented throughout the thesis. Additional material, e.g., proofs, derivations, implementation and experiment details, and additional results, are presented in the appendices.

%% file: 02_background/01_rl.tex
\label{sec:02_01_rl}

We first start with defining the reinforcement learning (RL) problem. The goal of reinforcement learning is to find how a dynamical system can be optimally controlled. For this, we will first mathematically define dynamical systems. As a running example, let's consider a robot trying to reach an object on a desk.

We use a discrete-time Markov decision process (MDP) to define a dynamical system. An MDP is a tuple $\mdp=\langle\stateSpace,\initialStateDistribution,\actionSpace,\transitionFunction,\horizon,\rewardFunction\rangle$ with the following variables. $\stateSpace$ denotes the state space. Each $\state\in\stateSpace$ fully characterizes a state of the world, e.g., the robot's and the object's poses and velocities. Each episode in the system starts with an initial state $\state_0\in\stateSpace$ drawn randomly from the initial state distribution $\initialStateDistribution$:
\begin{align}
    \state_0 \sim \initialStateDistribution(\cdot)\:.
\end{align}
$\actionSpace$ denotes the action space such that each $\action\in\actionSpace$ is an action taken in the system, e.g., the robot is given some control input. The transition distribution $\transitionFunction$ then governs how this system evolves. Based on the action $\action_\timestep$ at time step $\timestep$, the system transitions from state $\state_\timestep$ to a new state $\state_{\timestep+1}$ according to:
\begin{align}
    \state_{\timestep+1} \sim \transitionFunction(\cdot \mid \state_{\timestep}, \action_{\timestep})\:.
\end{align}
It is important that each state \emph{fully} characterizes the state of the world: knowing the current state and action is sufficient to best predict the next state, i.e.,
\begin{align}
    \transitionFunction(\cdot \mid \state_0, \action_0, \state_1, \action_1, \dots, \state_{\timestep}, \action_{\timestep}) = \transitionFunction(\cdot \mid \state_\timestep, \action_\timestep)\:.
\end{align}
This is known as the Markov property. The system evolves for $\horizon<\infty$ time steps, known as the \emph{horizon} of the MDP.\footnote{In this thesis we only consider finite-horizon MDPs. Extending to infinite-horizon MDPs requires an additional variable: discount factor. We refer to the standard text on reinforcement learning by Sutton and Barto for more details \cite{sutton2018reinforcement}.}

At each time step $\timestep$, the decision maker (the agent) receives a scalar \emph{reward}, based on the reward function $\rewardFunction$. For example, the robot may experience some positive reward for getting close to the target object, or negative reward (cost, or penalty) for colliding with some obstacles around. The goal of the agent is to select its actions to maximize the expected cumulative reward over the time steps of an episode. In literature, there are different conventions about how a reward function is defined. The three options are to define them as a function of:
\begin{itemize}[nosep]
    \item only the current state: $\rewardFunction:\stateSpace\to\mathbb{R}$,
    \item the current state and action: $\rewardFunction:\stateSpace\times\actionSpace\to\mathbb{R}$,
    \item the current state, action, and the next state: $\rewardFunction:\stateSpace\times\actionSpace\times\stateSpace\to\mathbb{R}$.
\end{itemize}
In this thesis, we will be focusing on learning from comparative feedback where multiple trajectories of a robot (or multiple episodes in an MDP) are compared. Therefore, we will use trajectory reward functions. For this, we first let a trajectory be a sequence of state-action pairs, i.e., $\trajectory = (\state_0,\action_0,\state_1,\action_1,\dots,\state_\horizon,\action_\horizon)$, and $\trajectorySpace$ denote all possible trajectories of the system. This trajectory reward function $\trajectoryRewardFunction:\trajectorySpace\to\mathbb{R}$ can then be defined for each of the reward function conventions above:
\begin{align}
    \trajectoryRewardFunction(\trajectory) &:= \sum_{\timestep=0}^{\horizon} \rewardFunction(\state_\timestep),\\
    \trajectoryRewardFunction(\trajectory) &:= \sum_{\timestep=0}^{\horizon} \rewardFunction(\state_\timestep,\action_\timestep),\textrm{ or}\\
    \trajectoryRewardFunction(\trajectory) &:= \sum_{\timestep=0}^{\horizon-1} \rewardFunction(\state_\timestep,\action_\timestep,\state_{\timestep+1})\:.
\end{align}
In fact, trajectory reward function can be defined more broadly: it does not have to be additive over time steps. Hence, it is more expressive and general. Consequently in our setup, the goal of the agent is to maximize the trajectory reward. This is the problem that reinforcement learning methods attempt to solve: how should an agent decide its actions (based on the state of the system) so that the trajectory will acquire as much reward as possible?

Reinforcement learning is still a very active area of research. Over the past decade, several methods have been successfully implemented for various versions or applications of this problem. Although the details of those methods are beyond the scope of this thesis, we include a list of widely-used methods: deep Q-networks (DQN) \cite{mnih2013playing}, deep deterministic policy gradient (DDPG) \cite{lillicrap2015continuous}, asynchronous advantage actor-critic (A3C) \cite{mnih2016asynchronous}, trust region policy optimization (TRPO) \cite{schulman2015trust}, proximal policy optimization (PPO) \cite{schulman2017proximal}, hindsight experience replay (HER) \cite{andrychowicz2017hindsight}, actor-critic using Kronecker-factored trust region method (ACKTR) \cite{wu2017scalable}, actor-critic with experience replay (ACER) \cite{wang2017sample}, twin delayed DDPG (TD3) \cite{fujimoto2018addressing}, and soft-actor critic (SAC) \cite{haarnoja2018soft}.

%% file: 02_background/02_irl.tex
\label{sec:02_02_irl}

Inverse reinforcement learning (IRL), as its name implies, tries to solve an inverse problem. In IRL, an agent who is already acting (near-)optimally in a system provides some data, i.e., they control the system. For example, an expert operator provides demonstrations of a task by teleoperating a robot. The goal in IRL is to use these expert demonstrations to identify the objective of the task, i.e., the reward function \cite{abbeel2004apprenticeship,abbeel2005exploration,ng2000algorithms,nikolaidis2015efficient,finn2016guided}.\footnote{Another interesting and very related problem is imitation learning \cite{pomerleau1988alvinn,ross2013learning,ho2016generative,stadie2017third,finn2017one,song2018multi}, where the goal is to directly learn an optimal control policy from expert demonstrations. Although the research community does not have a consensus on the scope of these terms, we use this convention: IRL tries to learn the reward function, imitation learning tries to learn the optimal policy; both from expert demonstrations.}

Formally in IRL, we are given some trajectory demonstrations $\demonstration^{(1)}, \demonstration^{(2)}, \dots \in \trajectorySpace$ (or more generally: state-action pairs, or transition tuples that also include the next state) that are known to be (near-)optimal with respect to the target task, encoded by an unknown reward function $\rewardFunction$. The goal is to learn this reward function $\rewardFunction$.

It might not be obvious why IRL is an important problem: if we already have an agent that is able to successfully control the system, why do we try to learn a reward function? The most common reason is automation. It is usually the case that the expert agent is a human, which means we need that expert human every time we need to control the system. However, if we can learn the reward function that encodes the task, then we can perform reinforcement learning in this system with the learned reward function to be able to control the system even in the absence of the expert. This is not the only use case of IRL. Another interesting application is behavior modeling \cite{liu2013understanding}. Imagine we are trying to develop an autonomous vehicle that predicts the actions of the other cars around so that it will seamlessly interact with them in traffic. To do this, understanding the objective of the other cars is crucial: if our car can infer their objective, i.e., their reward function, then it may better predict their actions. IRL has also applications in recommendation systems: Given a user's browsing history (a demonstration), the goal is to learn their preferences (reward function) so that the system can make better recommendations in future (learn a better policy).

Similar to reinforcement learning, IRL is also an active research area. Arguably the most influential methods in IRL have been apprenticeship learning \cite{abbeel2004apprenticeship}, maximum margin planning \cite{ratliff2006maximum}, Bayesian inverse reinforcement learning \cite{ramachandran2007bayesian}, and maximum entropy inverse reinforcement learning (MaxEnt-IRL) \cite{ziebart2008maximum}.

In this section, we reviewed the standard IRL problem where the goal is to learn a reward function given expert demonstrations. However in many cases, especially in robotics, expert demonstrations may not be available, or all users of the system might be providing suboptimal demonstrations \cite{gopalan2022negative}. Two common reasons for this are (1) good demonstrations might require a high level of expertise \cite{villani2018survey}, and (2) it is often too difficult to manually operate robots, especially manipulators with high degrees of freedom (DoF)~\cite{akgun2012keyframe,dragan2012formalizing,javdani2015shared,khurshid2015data}. Moreover, even when operating the high DoF of a robot is not an issue, people might have cognitive biases or habits that cause their demonstrations to not align with their actual reward functions. For example, in \cite{kwon2020when} we have shown that people tend to perform consistently risk-averse or risk-seeking actions in risky situations, depending on their potential losses or gains, even if those actions are suboptimal. As another example from the field of autonomous driving, \citet{basu2017you} suggest that people prefer their autonomous vehicles to be more timid compared to their own demonstrations. These problems show that, even though demonstrations carry an important amount of information about what the humans want, one should either try to learn from suboptimal demonstrations \cite{brown2020better,chen2020learning,cao2021learning,grollman2011donut,wu2019imitation} or go beyond demonstrations, e.g., corrections \cite{bajcsy2017learning,bajcsy2018learning,losey2018including,zhang2019learning,li2021learning}, rankings \cite{brown2019deep,brown2019extrapolating, brown2020better,chen2020learning}, critiques \cite{argall2007learning,cui2018active}, trajectory assessments \cite{shah2020interactive}, or ordinal feedback \cite{chu2005gaussian}, to properly capture the underlying reward functions. The latter approach is also the theme of this thesis: We will go beyond demonstrations and present methods that (actively) learn reward functions from comparative feedback where users compare multiple trajectories of the system. This type of feedback has been shown to be successful in several other domains such as classification \cite{chen2017near}, bandit problems \cite{busa2014survey}, and reinforcement learning \cite{wirth2017survey}.

%% file: 03_learning/00_intro.tex
Having presented the reinforcement learning (RL) and inverse reinforcement learning (IRL) problems in Chapter~\ref{chap:background}, we are now ready to start presenting our learning methods. In this chapter, we present alternative IRL solutions in which we learn reward functions using comparative feedback. Although the novelty of our methods is due to the use of comparative feedback, we still allow the use of demonstrations as in the standard IRL framework. To this end, Section~\ref{sec:03_01_pairwise_comparisons} presents how we can incorporate the information from best-of-many choices \cite{biyik2019asking,biyik2022learning} into Bayesian IRL \cite{ramachandran2007bayesian} that originally learns from demonstrations. Although we reduce the emphasis on demonstrations in later sections, the same Bayesian approach easily extends to all methods in this chapter except Section~\ref{sec:03_06_hierarchical} where we assume humans have non-stationary rewards, which makes the generation process of demonstrations ambiguous.

%% file: 03_learning/01_pairwise_comparisons.tex
\label{sec:03_01_pairwise_comparisons}

Let's start with briefly going over the other works in the literature that attempt to incorporate comparative feedback into the IRL framework.

\noindent\textbf{Learning reward functions from rankings and best-of-many choices.}
Two helpful and closely related sources of information that can be used to learn reward functions is rankings and best-of-many choices. In rankings, a human expert ranks a set of trajectories in the order of their preference~\cite{brown2019extrapolating} whereas in best-of-many choices they just pick their favorite trajectory \cite{biyik2019green}. A special case of both of these, which we also adopt in our experiments, is when these queries are pairwise \cite{akrour2012april,lepird2015bayesian,christiano2017deep,ibarz2018reward,brown2019deep,wilde2019bayesian,wirth2017survey,akrour2011preference,furnkranz2012preference,sugiyama2012preference,wilson2012bayesian}. While these works experimented their methods on some simulation environments, others leveraged pairwise comparison questions for various real-life applications, including exoskeleton gait optimization \cite{tucker2020preference}, and trajectory optimization for robots in interactive settings \cite{cakmak2011human,palan2019learning}.
 
\noindent\textbf{Learning reward functions from both demonstrations and comparisons.} \citet{ibarz2018reward} have explored combining demonstrations and comparisons, where they take a model-free approach to learn a reward function in the Atari domain. Our motivation, physical autonomous systems, differs from theirs, leading us to a structurally different method. It is difficult and expensive to obtain data from humans controlling physical robots. Hence, model-free approaches are presently impractical. In contrast, we give special attention to data-efficiency as we will discuss in detail in Sections~\ref{sec:04_01_volume_removal} and \ref{sec:04_02_information_gain}. As the resulting training process is not especially time-intensive, we efficiently learn personalized reward functions.

\subsection{Formulation}

Building on prior work, we integrate demonstrations and comparative feedback to learn the human's reward function. Here we formalize this problem setting, and introduce the two forms of human feedback that we will focus on: demonstrations and best-of-many choices. Our formulation revisits the definitions in Section~\ref{sec:02_01_rl} and extends it for comparative feedback.

\noindent \textbf{MDP.} Let us consider a fully observable dynamical system describing the evolution of the robot, which should ideally behave according to the human's preferences. We formulate this system as a discrete-time Markov Decision Process (MDP) $\mdp = \langle \stateSpace, \actionSpace, \transitionFunction, \rewardFunction, \horizon\rangle$. At time $\timestep$, $\state_\timestep \in \stateSpace$ denotes the state of the system and $\action_\timestep \in \actionSpace$ denotes the robot's action. The robot transitions to a new state according to its dynamics distribution: $\state_{\timestep+1} \sim \transitionFunction(\cdot \mid \state_\timestep, \action_\timestep)$. At every time step the robot receives reward $\rewardFunction(\state)$, and the task ends after a total of $\horizon$ time steps.

\noindent \textbf{Trajectory.} A trajectory $\trajectory \in \trajectorySpace$ is a finite sequence of state-action pairs, i.e., $\trajectory = \big((\state_\timestep, \action_\timestep)\big)_{\timestep=0}^\horizon$ over the time horizon $\horizon$.

\noindent \textbf{Reward.} The reward function captures how the human wants the robot to behave. Similar to related works \cite{abbeel2004apprenticeship,ng2000algorithms,ziebart2008maximum} where the reward is a linear combination of features, we assume the reward is a parametric function of some trajectory features. Consistent with prior work \cite{bobu2018learning,bajcsy2017learning,ziebart2008maximum}, we will assume that the trajectory features $\trajectoryFeaturesFunction(\trajectory)$ for any given trajectory $\trajectory$ are known. In practice, they can be based on the state features along that trajectory. Or more generally, the trajectory features $\trajectoryFeaturesFunction(\trajectory)$ can be defined as any function over the entire trajectory $\trajectory$ as we discussed in Section~\ref{sec:02_01_rl}. To understand what the human wants, the robot must simply learn the true parameters of the reward function which we denote with $\weights^*$. Accordingly, we denote a trajectory reward function parameterized with $\weights$ as:
\begin{align}
    \trajectoryRewardFunction_{\weights}(\trajectory) = \gpRewardFunction_{\weights}(\trajectoryFeaturesFunction(\trajectory))\:.
    \label{eq:03_01_PF1}
\end{align}

\begin{figure}[t!]
\includegraphics[width=0.5\textwidth]{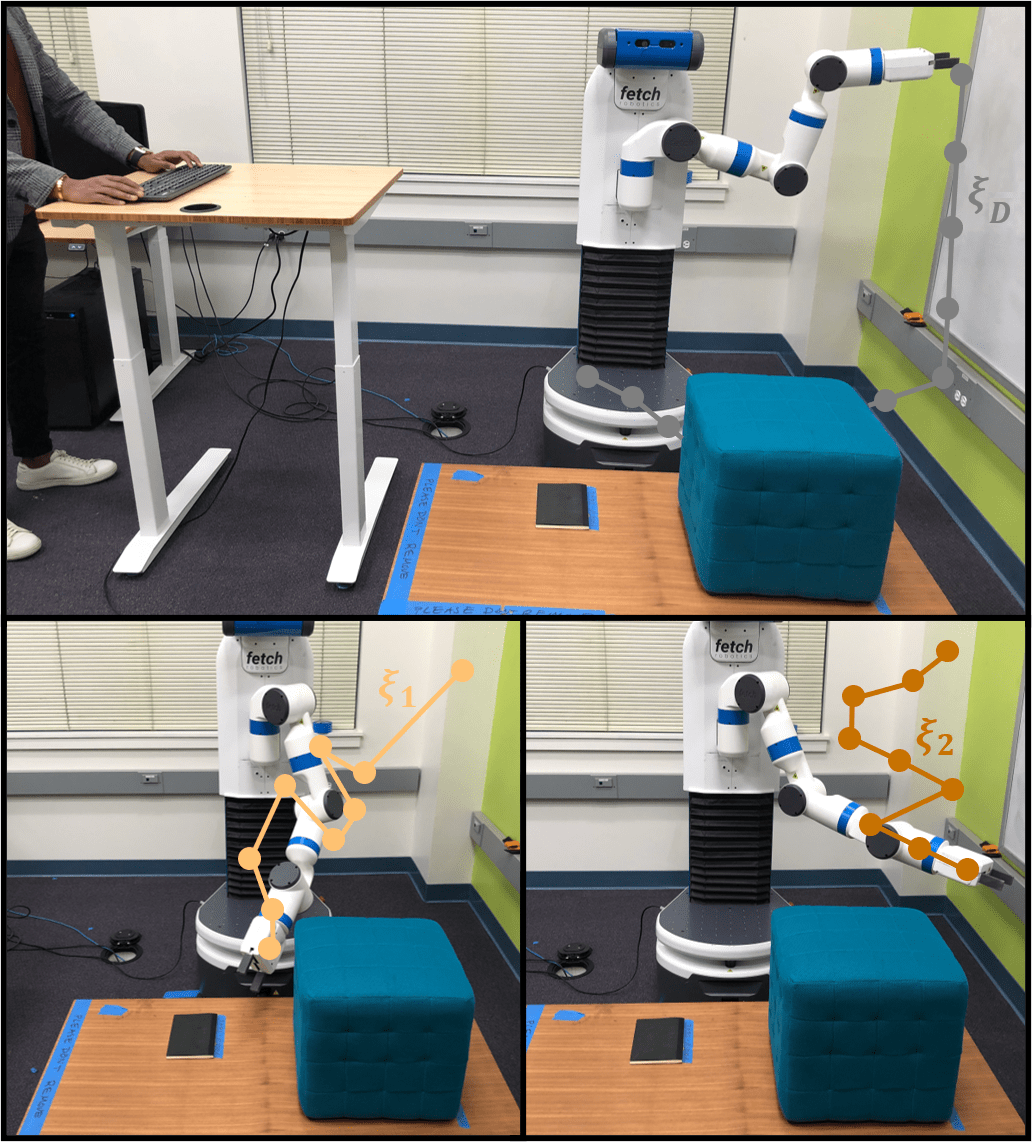}
\centering
\caption{Example of a demonstration (top) and a best-of-many choice query with $\abs{\query}=2$, i.e., a pairwise comparison query (bottom). During the demonstration the robot is \emph{passive}, and the human teleoperates the robot to produce trajectory $\trajectory_D$ from scratch. By contrast, the preference query can be \emph{active}: the robot chooses two trajectories $\trajectory_1$ and $\trajectory_2$ to show to the human, and the human answers by selecting their preferred option.}
\label{fig:03_01_examples}
\end{figure}

\noindent \textbf{Demonstrations.} One way that the human can convey their reward function parameters $\weights$ to the robot is by providing demonstrations. Each human demonstration is a trajectory $\demonstration$, and we denote a dataset of human demonstrations as $\demonstrationDataset = \{\demonstration^{(1)}, \demonstration^{(2)}, \ldots, \demonstration^{(\abs{\demonstrationDataset})}\}$. In practice, these demonstrations could be provided by kinesthetic teaching, by teleoperating the robot, or in virtual reality (see Figure~\ref{fig:03_01_examples}, top).

\noindent \textbf{Best-of-many Choices.} Another way the human can provide information is by giving feedback about the trajectories the robot shows. We define a best-of-many choice query $\query = \{\trajectory_1, \trajectory_2, \ldots, \trajectory_{\abs{\query}}\}$ as a set of $\abs{\query}$ robot trajectories. The human answers this query by picking a trajectory $\queryResponse \in \query$ that matches their personal preferences (i.e., maximizes their reward function). In practice, the robot could play $\abs{\query}$ different trajectories, and let the human choose their favorite (see Figure~\ref{fig:03_01_examples}, bottom). 

\noindent \textbf{Problem.} Our overall goal is to accurately and efficiently learn the human's reward function from multiple sources of data. In this section, we will only focus on demonstrations and best-of-many choices. Our approach should learn the reward parameters $\weights$ with a combination of demonstrations and best-of-many choice queries.

\begin{figure*}[t]
\includegraphics[width=\textwidth]{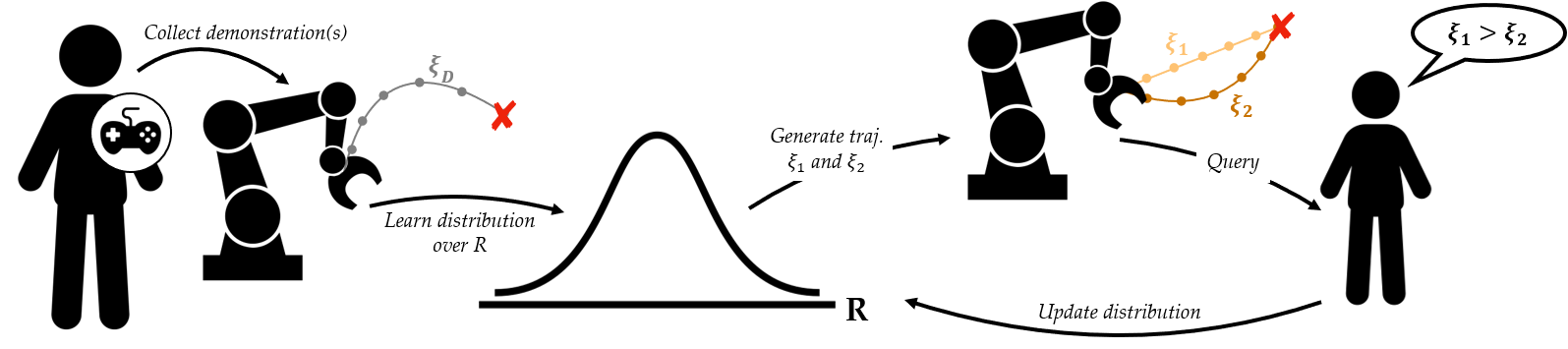}
\centering
\caption{Overview of our \textsc{DemPref} approach. The human starts by providing a set of \emph{high-level} demonstrations (left), which are used to initialize the robot's belief over the human's reward function through $\weights$. The robot then \emph{fine-tunes} this belief by asking questions (right): the robot actively generates a set of trajectories, and asks the human to choose their favorite.}
\label{fig:03_01_dempref}
\end{figure*}

\subsection{Our Approach}

We now overview our approach for integrating demonstrations and best-of-many choices to efficiently learn the human's reward function. Intuitively, demonstrations provide an informative, \emph{high-level} understanding of what behavior the human wants; however, these demonstrations are often noisy, and may fail to cover some aspects of the reward function. By contrast, preferences are \emph{fine-grained}: they isolate specific, ambiguous aspects of the human's reward, and reduce the robot's uncertainty over these regions. It therefore makes sense for the robot to start with high-level demonstrations before moving to fine-grained preferences. Indeed --- as we will show in Theorem~\ref{thm:04_02_order_of_dempref} --- starting with demonstrations and then shifting to actively collected comparative feedback is the most efficient order for gathering data. Our algorithm, which we call \textsc{DemPref} (short for demonstrations and preferences) leverages this insight to combine high-level demonstrations and low-level best-of-many choice queries (see Figure~\ref{fig:03_01_dempref}).

\subsubsection{Initializing a Belief from Offline Demonstrations}

\textsc{DemPref} starts with a set of offline trajectory demonstrations $\demonstrationDataset$. These demonstrations are collected \emph{passively}: the robot lets the human show their desired behavior, and does not interfere or probe the user. We leverage these passive human demonstrations to initialize an informative but imprecise prior over the true reward function parameters $\weights$.

\noindent \textbf{Belief.} Let the belief $\belief$ be a probability distribution over $\weights$. We initialize $\belief$ using the trajectory demonstrations, so that $\belief^0(\weights) = P(\weights \mid \demonstrationDataset)$. Applying Bayes' Theorem:
\begin{equation} \label{eq:03_01_DP1}
    \begin{split}
    \belief^0(\weights) & \propto P(\demonstrationDataset \mid \weights)P(\weights) \\
    & = P(\demonstration^{(1)}, \demonstration^{(2)}, \ldots, \demonstration^{(\abs{\demonstrationDataset})} \mid \weights)P(\weights)
    \end{split}
\end{equation}
We assume that the trajectory demonstrations are conditionally independent, i.e., the human does not consider their previous demonstrations when providing a new demonstration. Hence, Equation~\eqref{eq:03_01_DP1} becomes:
\begin{equation} \label{eq:03_01_DP2}
    \belief^0(\weights) \propto P(\weights)\prod_{\demonstration \in \demonstrationDataset} P(\demonstration \mid \weights)
\end{equation}
In order to evaluate Equation~\eqref{eq:03_01_DP2}, we need a model of $P(\demonstration \mid  \weights)$ --- in other words, how likely is the demonstrated trajectory $\demonstration$ given that the human's reward function parameters are $\weights$?

\noindent \textbf{Boltzmann Rational Model.} \textsc{DemPref} is not tied to any specific choice of the human model in Equation~\eqref{eq:03_01_DP2}, but we do want to highlight the Boltzmann rational model that is commonly used in inverse reinforcement learning \cite{ziebart2008maximum,ramachandran2007bayesian}. Under this particular model, the probability of a human demonstration is related to the reward associated with that trajectory:
\begin{align}
    P(\demonstration \mid \weights) & \propto \exp{\big(\demonstrationRationalityCoefficient \trajectoryRewardFunction_{\weights}(\demonstration)\big)}\\
    & = \exp{\big(\demonstrationRationalityCoefficient \gpRewardFunction_{\weights}(\trajectoryFeaturesFunction(\demonstration))\big)}\:.
    \label{eq:03_01_DP3}
\end{align}
Here $\demonstrationRationalityCoefficient \geq 0$ is a temperature hyperparameter, commonly referred to as the \emph{rationality coefficient}, that expresses how noisy the human demonstrations are, and we substituted Equation~\eqref{eq:03_01_PF1} for $\trajectoryRewardFunction$. Leveraging this human model, the initial belief over $\weights$ given the offline demonstrations becomes:
\begin{equation} \label{eq:03_01_DP4}
    \belief^0(\omega) \propto \exp{\Bigg(\demonstrationRationalityCoefficient\cdot \sum_{\demonstration \in \demonstrationDataset} \gpRewardFunction_{\weights}(\trajectoryFeaturesFunction(\demonstration))\Bigg)} P(\weights)
\end{equation}

\noindent \textbf{Summary.} Human demonstrations provide an informative but imprecise understanding of $\weights$. Because these demonstrations are collected passively, the robot does not have an opportunity to investigate aspects of $\weights$ that it is unsure about. We therefore leverage these demonstrations to initialize $\belief^0$, which we treat as a high-level \emph{prior} over the human's reward. Next, we will introduce how we update this belief using best-of-many choice questions to remove uncertainty and obtain a fine-grained posterior.

\subsubsection{Updating the Belief with Proactive Queries}

After initialization, \textsc{DemPref} iteratively performs two main tasks: \emph{actively} choosing the right preference query $\query$ to ask, and applying the human's answer to update $\belief$. In this section we focus on the second task: updating the robot's belief $\belief$. We will explore how robots should proactively choose the right question in Sections~\ref{sec:04_01_volume_removal} and \ref{sec:04_02_information_gain} under Chapter~\ref{chap:active}.

\noindent \textbf{Posterior.} The robot asks a new question at each iteration $i\geq 1$. Let $\query^{(i)}$ denote the $i$-th best-of-many choice query, and let $\queryResponse^{(i)}$ be the human's response to this query.\footnote{For the ease of notation, we will slightly abuse the notation and use $\queryResponse^{(i)}$ as a random variable when the user's response has not been elicited yet, and as a constant after their response is revealed. Similarly, we will use $\query^{(i)}$ as an optimization variable when we are optimizing for the next query in Chapter~\ref{chap:active}, but it will be a constant as soon as the query is made to the user.} Again applying Bayes' Theorem, the robot's posterior over $\weights$ becomes:
\begin{equation} \label{eq:03_01_DP5}
    \belief^{i}(\weights) \propto P(\queryResponse^{(1)},\ldots,\queryResponse^{(i)} \mid \query^{(1)}, \ldots, \query^{(i)}, \weights) \cdot \belief^0(\weights)\:,
\end{equation}
where $\belief^0$ is the prior initialized using human demonstrations. We assume that the human's responses $\queryResponse$ are conditionally independent, i.e., only based on the current query and reward function parameters. Equation (\ref{eq:03_01_DP5}) then simplifies to:
\begin{equation} \label{eq:03_01_DP6}
    \belief^{i}(\weights) \propto \belief^0(\weights) \cdot \prod_{i'=1}^i P(\queryResponse^{(i')} \mid \query^{(i')}, \weights)
\end{equation}
We can equivalently write the robot's posterior over $\weights$ after asking $i$ questions as:
\begin{equation} \label{eq:03_01_DP7}
    \belief^{i}(\weights) \propto P(\queryResponse^{(i)} \mid \query^{(i)}, \weights) \cdot \belief^{i-1}(\weights)
\end{equation}

\noindent \textbf{Human Model.} In Equation~\eqref{eq:03_01_DP7}, $P(\queryResponse^{(i)} \mid \query^{(i)}, \weights)$ denotes the probability that a human with reward function parameters $\weights$ will answer query $\query^{(i)}$ by selecting trajectory $\queryResponse^{(i)} \in \query^{(i)}$. Put another way, this likelihood function is a probabilistic human model, and previous work demonstrated the importance of modeling imperfect human responses \cite{kulesza2014structured}. One way to do this is to model a noisily optimal human as selecting $\queryResponse^{(i)}$ from a \emph{strict} best-of-many choice query $\query^{(i)}$ by
\begin{align}
P(\queryResponse^{(i)} = \query^{(i)}_{j} \mid \query^{(i)},\weights) &= \frac{\exp(\comparisonRationalityCoefficient\trajectoryRewardFunction_{\weights}(\query^{(i)}_j))}{\sum_{j'=1}^{\abs{\query^{(i)}}}\exp(\comparisonRationalityCoefficient\trajectoryRewardFunction_{\weights}(\query^{(i)}_{j'}))}\:,
\label{eq:03_01_noisily_optimal}
\end{align}
where $\comparisonRationalityCoefficient$ is the rationality coefficient for comparisons. We call the query strict because the human is required to select one of the trajectories. This model, backed by neuroscience and psychology \citep{daw2006cortical,luce2012individual,ben1985discrete,lucas2009rational}, is routinely used  \cite{biyik2019green,guo2010real,viappiani2010optimal,wilson2019ten}. It is also known as the multinomial logits (MNL) model \cite{chen2018nearly}.

Although we present this human choice model and use its variants in most of the subsequent sections, our \textsc{DemPref} approach is agnostic to the specific choice of $P(\queryResponse^{(i)} \mid \query^{(i)}, \weights)$ --- we test different human models in our experiments presented in Section~\ref{sec:04_02_information_gain}. For now, we simply want to highlight that this human model defines the way users respond to queries.

Having defined the learning algorithm that incorporates both demonstrations and best-of-many choices, and a human model, we now have a computational method of learning reward functions from comparative feedback in addition to demonstrations. In Sections~\ref{sec:04_01_volume_removal} and \ref{sec:04_02_information_gain}, we will boost the data-efficiency of this algorithm by developing active querying techniques. We also defer our experiments to those sections. Prior to them in the subsequent sections of this chapter, we will extend this framework to other forms of comparative feedback and relax some of the assumptions. The next section relaxes the parametric reward assumption.

%% file: 03_learning/02_gp.tex
\label{sec:03_02_gp}

In Section~\ref{sec:03_01_pairwise_comparisons}, we showed a Bayesian learning approach that incorporates demonstrations and best-of-many choice queries. An important assumption we made is that the reward function is a \emph{parametric} function of some known trajectory features. One might think that this is general enough, because we did not impose any other constraints on the functional form, so for example, deep neural networks could be good enough in most practical applications. However, computational issues often arise in practice: since we are taking a Bayesian learning approach, learning becomes infeasible as the number of parameters to learn increases. Given that deep neural networks often contain large number of learnable parameters, this means the approach is limited to simple functional forms. In fact, we will mostly focus on linear reward functions when we present our experiments as in the prior work \cite{abbeel2004apprenticeship,ng2000algorithms,ziebart2008maximum}.

Alternative approaches to Bayesian learning includes training a deep neural network using gradient-based optimization methods with a loss function that minimizes the log-likelihood where the likelihood simply comes from our belief distribution $\belief$ \cite{katz2021preference}. However, these approaches are usually limited in the sense that they only give a point-estimate of the reward function, whereas the Bayesian learning approach enables us to model the uncertainty by learning a distribution over reward functions. Motivated by this, we relax the parametric reward function assumption by explicitly learning a distribution over reward functions using Gaussian processes (GP) in this section. To do this, we focus on a special case of best-of-many choice queries with $\abs{\query}=2$, i.e., pairwise comparisons where users just compare two trajectories and choose their favorite. Our results, which we again defer until we present the corresponding active querying methods in Section~\ref{sec:04_03_gp}, show this significantly improves the expressive power of the learned reward function.

\subsection{Related Work}

\noindent \textbf{Gaussian process regression.} In the machine learning literature, \citet{gonzalez2017preferential} and \citet{chu2005preference} proposed methods for preference-based Bayesian optimization and GP regression, respectively, but they were not collecting data actively. Furthermore, \citet{gonzalez2017preferential} required to regress a GP over $2\numberOfFeatures$-dimensions to model a $\numberOfFeatures$-dimensional function, which causes a computational burden. More relevantly, \citet{houlsby2012collaborative} presented an active method for preference-based GP regression. However, similar to \cite{gonzalez2017preferential}, the regression was over a $2\numberOfFeatures$-dimensional space where the learned model predicts the outcome of a comparison rather than outputting a reward value. \citet{jensen2011pairwise} showed how to update a GP with preference data, but the active query generation was not an interest. \citet{kapoor2007active} developed an active learning approach for classification with GPs. This is a specific case of our problem, as the labels in classification are consistent, i.e., the labels assigned to the samples in the dataset, even if they are incorrect, do not change during training. In our case, the user can respond to the same pairwise comparison query inconsistently depending on their noise model. \citet{houlsby2011bayesian} and \citet{daniel2015active} proposed active GP fitting methods for classification and reward learning, respectively. While the latter focused on robotics tasks, they were not preference-based. Hence, they may be infeasible in many applications as it is difficult for humans to assign actual reward values. 

In this section, we study a method for preference-based GP regression that learns from pairwise comparisons, and extend it with active query generation in Section~\ref{sec:04_03_gp}. This method does not require the humans to assign actual reward values to the trajectories for fitting the GP.

\subsection{Formulation} \label{sec:03_02_formulation}
We again model the environment the robot is going to operate in as a discrete-time finite-horizon MDP. We use $\state_\timestep\in\stateSpace$ to denote the state and $\action_\timestep\in\actionSpace$ for the action (control inputs) at time $\timestep$. A trajectory $\trajectory\in\trajectorySpace$ within this MDP is a sequence that consists of the state and actions: $\trajectory = (\state_0,\action_0,\state_1,\action_1,\dots,\state_\horizon,\action_\horizon)$, where $\horizon$ is the finite time horizon. 

We assume a reward function over trajectories, $\trajectoryRewardFunction:\trajectorySpace\to\mathbb{R}$, that encodes the human user's preferences about how they want the system to operate.

We assume the reward function $\trajectoryRewardFunction$ can be formulated as $\trajectoryRewardFunction(\trajectory) = \gpRewardFunction(\trajectoryFeaturesFunction(\trajectory))$, where $\trajectoryFeaturesFunction:\trajectorySpace\to\mathbb{R}^\numberOfFeatures$ defines a set of trajectory features, e.g. average speed and minimum distance to the closest car in a driving trajectory. However, we emphasize that this formulation of $\trajectoryRewardFunction$ enables a more general form of functions that does not require strong assumptions -- such as linearity in the features -- which is commonly put in place when learning reward functions (as in \cite{abbeel2004apprenticeship,ng2000algorithms,ziebart2008maximum}). We use a GP to model $\gpRewardFunction$, which allows us to avoid strong assumptions about the form of $\gpRewardFunction$.\footnote{Due to computation reasons, we assume $\numberOfFeatures$ is small. Compared to previous works which assume $\trajectoryRewardFunction$ to be linear in features or a small dimensional parameter space for parametric reward functions, this is a very mild assumption.}

\begin{figure}[th]
	\includegraphics[width=0.5\textwidth]{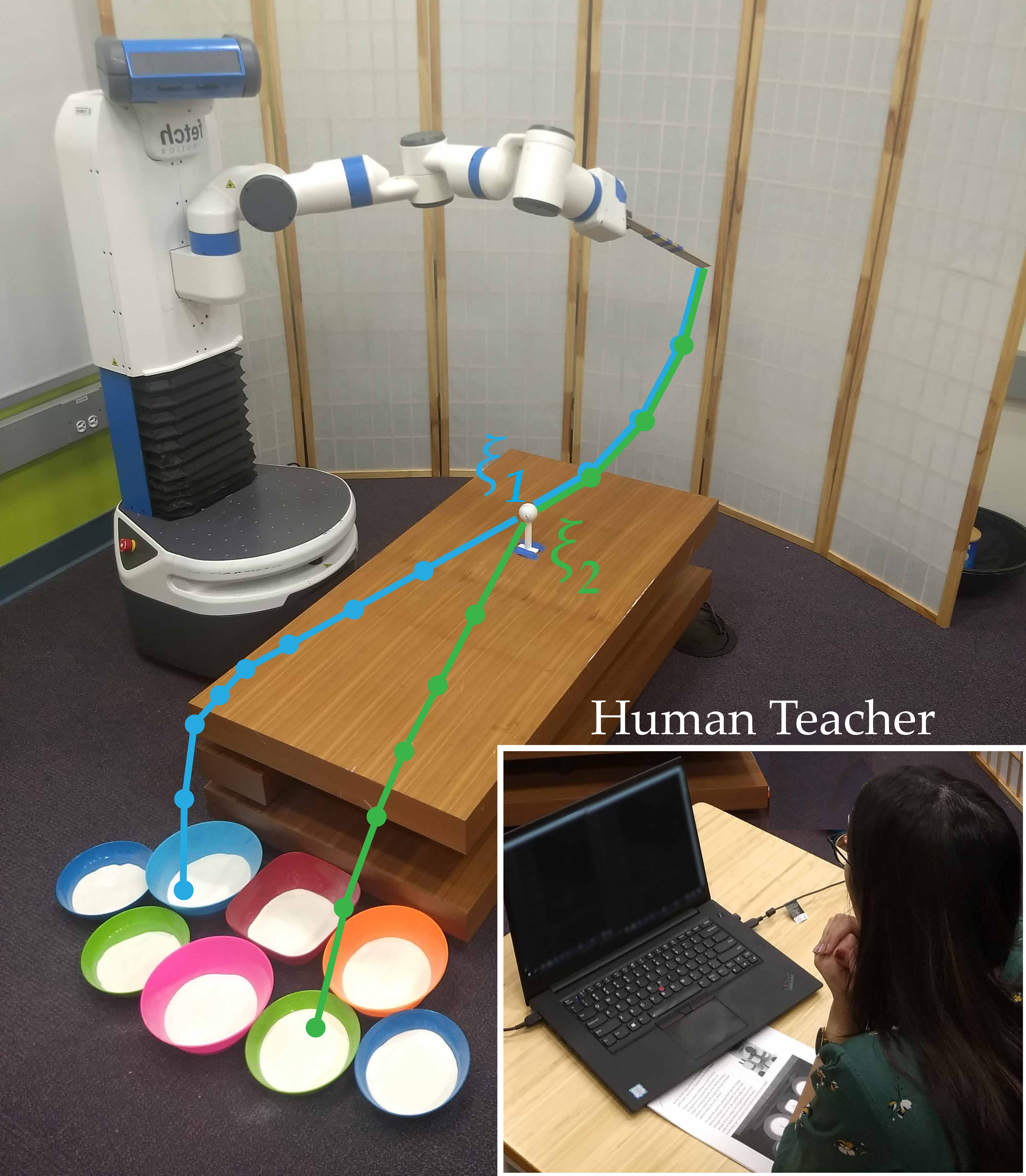}
	\centering
	\caption{The user is trying to teach the robot how to play a variant of mini-golf, where the reward differs among eight targets. In preference-based learning, instead of trying to design a reward function by hand or controlling the robot to provide demonstrations, the user simply compares two demonstrated trajectories on the robot. Here, $\trajectory_1$ and $\trajectory_2$ demonstrate two trajectories that correspond to hitting the ball towards the blue or green targets.}
	\label{fig:03_02_gp_front_fig}
\end{figure}

Our goal is to learn this more general form of reward functions using preference data in the form of pairwise comparisons. The robot will demonstrate a query $\query$ consisting of two trajectories, $\trajectory_1$ and $\trajectory_2$ as shown in Figure~\ref{fig:03_02_gp_front_fig} with blue and green curves, to the user, and will ask which trajectory they prefer. The user will respond to this query based on their preferences. The user's response provides useful information about the underlying preference reward function $\trajectoryRewardFunction$. Of course, we cannot assume human responses are perfect for every query, so we model the noise in their responses using the commonly adopted probit model (an alternative to the human choice model we presented in Equation~\eqref{eq:03_01_noisily_optimal}), which assumes humans make a binary decision between the two trajectories noisily based on the cumulative distribution function (cdf) of the difference between the two reward values:
\begin{align}
P(\queryResponse = \trajectory_1 \mid \query=\{\trajectory_1,\trajectory_2\}) = P\left(\trajectoryRewardFunction(\trajectory_1) - \trajectoryRewardFunction(\trajectory_2) > \epsilon\right),
\end{align}
where $\queryResponse\in \query$ denotes the user's choice, and $\epsilon\sim\mathcal{N}(0,2\preferenceNoiseStd^2)$ for some standard deviation $\preferenceNoiseStd\sqrt{2}$. Therefore, equivalently:
\begin{align}
P(\queryResponse=\trajectory_1 \mid \query=\{\trajectory_1,\trajectory_2\}) = \standardNormalCdf\left(\frac{\trajectoryRewardFunction(\trajectory_1)-\trajectoryRewardFunction(\trajectory_2)}{\sqrt{2}\preferenceNoiseStd}\right),
\label{eq:03_02_human_model}
\end{align}
where $\standardNormalCdf$ is the cumulative distribution function of the standard normal. This is an example of a Thurstonian model \cite{maydeu1999thurstonian}.

Having defined the problem setting, we are now ready to present our method for learning expressive reward functions using GPs.

\subsection{Our Approach} \label{subsec:03_02_approach}
In this section, we first give some background information about Gaussian Processes. We then introduce preference-based GP regression, where we show how to update a GP with the results of pairwise comparisons. We will present our approach to active preference query generation in Section~\ref{sec:04_03_gp}, where we discuss how to find the most informative query that accelerates the learning.\footnote{We make our Python code for preference-based GP regression and active query generation publicly available at \url{https://github.com/Stanford-ILIAD/active-preference-based-gpr}.} To simplify the notation, we replace $\trajectoryFeaturesFunction(\trajectory)$ with $\trajectoryFeaturesFunction$, with superscripts and subscripts when needed in this subsection and the related appendices.

\subsubsection{Gaussian Processes}
We start by introducing the necessary background on GPs for our work. We refer the readers to \cite{rasmussen2005gaussian} for other uses of GPs in machine learning.

Suppose we have a finite trajectory space $\trajectorySpace = \{\trajectoryFeaturesFunction^{(i)}\}_{i=1}^{\abs{\trajectorySpace}}$, where $\trajectoryFeaturesFunction^{(i)} \in \mathbb{R}^\numberOfFeatures$ is the features of the $i^{\textrm{th}}$ trajectory according to some arbitrary indexing in $\trajectorySpace$. We employ a probabilistic point of view for $\gpRewardFunction$ by modeling it using a GP, which enables us to model nonlinearities and uncertainties well without introducing parameters. We have:
\begin{align}
P(\mathbf{\gpRewardFunction} \mid \gpMeanVector, \gpCovarianceMatrix) = \frac{\exp\left(-\frac12 (\mathbf{\gpRewardFunction}-\gpMeanVector)^\top \gpCovarianceMatrix^{-1}(\mathbf{\gpRewardFunction}-\gpMeanVector)\right)}{(2\pi)^{\abs{\trajectorySpace}/2} \abs{\gpCovarianceMatrix}^{1/2}} ,
\label{eq:03_02_gp_prob}
\end{align}
where $\mathbf{\gpRewardFunction}=[\gpRewardFunction(\trajectoryFeaturesFunction^{(1)}), \gpRewardFunction(\trajectoryFeaturesFunction^{(2)}), \dots, \gpRewardFunction(\trajectoryFeaturesFunction^{(\abs{\trajectorySpace})})]^\top$, $\gpMeanVector$ and $\gpCovarianceMatrix$ are the mean vector and the covariance matrix of the GP distribution for the $\abs{\trajectorySpace}$ items in the dataset. Put it in another way, $\mathbf{\gpRewardFunction}$ follows a multivariate distribution. The covariance matrix depends on the used kernel. In this work, we use a variant of radial basis function (RBF) kernel with hyperparameter $\gpRbfHyperparameter$:
\begin{align*}
\gpKernelFunction(\trajectoryFeaturesFunction^{(i)},\trajectoryFeaturesFunction^{(j)}) &= \exp\left(-\gpRbfHyperparameter\norm{\trajectoryFeaturesFunction^{(i)}-\trajectoryFeaturesFunction^{(j)}}_2^2\right) - \bar{\gpKernelFunction}(\trajectoryFeaturesFunction^{(i)},\trajectoryFeaturesFunction^{(j)}),\\
\bar{\gpKernelFunction}(\trajectoryFeaturesFunction^{(i)},\trajectoryFeaturesFunction^{(j)}) &= \exp\left(-\gpRbfHyperparameter\norm{\Psi^{(i)}-\bar{\trajectoryFeaturesFunction}}_2^2 - \gpRbfHyperparameter\norm{\trajectoryFeaturesFunction^{(j)}-\bar{\trajectoryFeaturesFunction}}_2^2\right),
\end{align*}
where $\tilde{\trajectoryFeaturesFunction}\in\mathbb{R}^\numberOfFeatures$ is an arbitrary point for which we assume $\gpRewardFunction(\tilde{\trajectoryFeaturesFunction})=0$. This helps in practice because the query responses only depend on the relative difference between the two reward function values at the trajectories, i.e., $\gpRewardFunction(\trajectoryFeaturesFunction)+\epsilon$ for any $\epsilon\in\mathbb{R}$ would have the same likelihood for a dataset as $\gpRewardFunction(\trajectoryFeaturesFunction)$. By setting $\gpRewardFunction(\tilde{\trajectoryFeaturesFunction})=0$ for some arbitrary $\tilde{\trajectoryFeaturesFunction}\in\mathbb{R}^\numberOfFeatures$, we dissolve this ambiguity. It does not introduce an assumption because for any function $\gpRewardFunction'$ and for any point $\tilde{\trajectoryFeaturesFunction}$, one can define $\gpRewardFunction(\trajectoryFeaturesFunction) = \gpRewardFunction'(\trajectoryFeaturesFunction) - \gpRewardFunction'(\tilde{\trajectoryFeaturesFunction})$ without loss of generality\textemdash both $\gpRewardFunction'$ and $\gpRewardFunction$ will encode the same preferences. Finally, this variant of the RBF kernel is still positive semi-definite, because it is equivalent to the covariance kernel of a GP which is initialized with an initial data point and a standard RBF kernel prior.

\subsubsection{Preference-based GP Regression}
Although the previous subsection was needed to explain how GPs work, we only focus on preference-based learning without any demonstrations in this section. In preference-based learning with pairwise comparisons, we have a dataset $\comparisonDataset = (\query^{(i)}, \queryResponse^{(i)})_{i=1}^{\abs{\comparisonDataset}} =  ((\trajectoryFeaturesFunction^{(i)}_1,\trajectoryFeaturesFunction^{(i)}_2), \queryResponse^{(i)})_{i=1}^{\abs{\comparisonDataset}}$, consisting of pairs of trajectory features $\trajectoryFeaturesFunction^{(i)}_1, \trajectoryFeaturesFunction^{(i)}_2 \in \mathbb{R}^\numberOfFeatures$, and user responses $\mathbf{\queryResponse} = (\queryResponse^{(i)})_{i=1}^{\abs{\comparisonDataset}}$, where $\queryResponse^{(i)} \in \{\query^{(i)}_1,\query^{(i)}_2\}$ indicates whether the user preferred $\trajectoryFeaturesFunction^{(i)}_1$ or $\trajectoryFeaturesFunction^{(i)}_2$. Accordingly, $\gpCovarianceMatrix$ is now a $2\abs{\comparisonDataset} \times 2\abs{\comparisonDataset}$ matrix, whose rows and columns correspond to $\trajectoryFeaturesFunction^{(i)}_j, \forall i \in \{1,\dots,\abs{\comparisonDataset}\}, \forall j \in \{1,2\}$. Similarly, $\gpMeanVector$ is a $2\abs{\comparisonDataset}$-vector. Using a Bayesian approach to update the GP with new preference data $(\query^{(i)},\queryResponse^{(i)})$ gives:
\begin{align}
P(\gpRewardFunction \mid \gpMeanVector, \gpCovarianceMatrix, \query^{(i)}, \queryResponse^{(i)}) &\propto P(\queryResponse^{(i)} \mid \gpRewardFunction, \gpMeanVector, \gpCovarianceMatrix, \query^{(i)})P(\gpRewardFunction \mid \gpMeanVector, \gpCovarianceMatrix, \query^{(i)}) \nonumber\\
&= P(\queryResponse^{(i)} \mid \gpRewardFunction, \query^{(i)})P(\gpRewardFunction \mid \gpMeanVector, \gpCovarianceMatrix).
\label{eq:03_02_bayesian_update}
\end{align}
Here, the first term is just the probabilistic human response model given in Equation~\eqref{eq:03_02_human_model}, and the second term is Equation~\eqref{eq:03_02_gp_prob}. However, this posterior does not follow a GP distribution similar to Equation~\eqref{eq:03_02_gp_prob}, and becomes analytically intractable \cite{jensen2011pairwise}.

Prior works have shown it is possible to perform some approximation such that the posterior is another GP \cite{jensen2011pairwise,rasmussen2005gaussian}. The most common approximation techniques are: 
\begin{enumerate}
    \item Laplace approximation, where the idea is to model the posterior as a GP such that the mode of the likelihood is treated as the posterior mean, and a second-order Taylor approximation around the maximum of the log-likelihood gives the posterior covariance. This technique is computationally very fast. 
    \item Expectation Propagation (EP), where the idea is to approximate each factor of the product by a Gaussian. EP is an iterative method that processes each factor iteratively to update the distribution to minimize its KL-divergence with the true posterior. While it is more accurate than Laplace approximation, it is slower in practice \cite{nickisch2008approximations}.
\end{enumerate} 
In this section, we use the former for its computational efficiency. Hence, we now show how to compute the quantities for Laplace approximation, i.e., posterior mean and covariance.

\noindent\textbf{Finding the posterior mean.} We use the mode of the posterior as an approximation to the posterior mean:
\begin{align}
\hat{\mathbf{\gpRewardFunction}} = \argmax_{\mathbf{\gpRewardFunction}} \left(\log\left(P(\mathbf{\queryResponse} \mid \mathbf{\query}, \mathbf{\gpRewardFunction})\right) + \log\left(P(\mathbf{\gpRewardFunction}\mid \mathbf{\query},  \gpMeanVector, \gpCovarianceMatrix)\right)\right)\:,
\label{eq:03_02_posterior_mean}
\end{align}
where $\mathbf{\query}$ denotes the queries that correspond to the responses $\mathbf{\queryResponse}$. Because the preference data are conditionally independent, the first term can be written as:
\begin{align*}
\log\left(P(\mathbf{\queryResponse}\mid \mathbf{\query}, \mathbf{\gpRewardFunction})\right) &= \sum_{i=1}^{\abs{\comparisonDataset}}{\log P(\queryResponse^{(i)} \mid \query^{(i)}, \mathbf{\gpRewardFunction})}\\
&= \sum_{i=1}^{\abs{\comparisonDataset}}{\log \standardNormalCdf\left(\frac{\gpRewardFunction(\trajectoryFeaturesFunction^{(i)}_1)-\gpRewardFunction(\trajectoryFeaturesFunction^{(i)}_2)}{\sqrt{2}\preferenceNoiseStd}\right)}
\end{align*}
due to Equation~\eqref{eq:03_02_human_model}. Adopting a zero-mean prior for $f$, we can write the second term of the optimization~\eqref{eq:03_02_posterior_mean} as:
\begin{align*}
\log\left(P(\mathbf{\gpRewardFunction}\mid \bm{\trajectoryFeaturesFunction}), \gpMeanVector, \gpCovarianceMatrix\right) &= -\frac12 \log\abs{\gpCovarianceMatrix} - \abs{\comparisonDataset}\log{2\pi} - \frac12 \mathbf{\gpRewardFunction}^\top \gpCovarianceMatrix^{-1}\mathbf{\gpRewardFunction}
\end{align*}
Armed with these two expressions, we can now optimize the log-likelihood and thus find the mode of it to approximate the posterior mean.

\vspace{5pt}
\noindent\textbf{Finding the posterior covariance matrix.} Following \cite{rasmussen2005gaussian}, and omitting the derivation details for brevity, the posterior covariance matrix is $\hat{\gpCovarianceMatrix} = (\gpCovarianceMatrix^{-1} + \gpNegativeHessianOfLoglikelihood)^{-1}$ where $\gpNegativeHessianOfLoglikelihood$ is the negative Hessian of the log-likelihood:
\begin{align*}
\gpNegativeHessianOfLoglikelihood_{ij} = -\frac{\partial^2 \log\left(P(\mathbf{\queryResponse}\mid \mathbf{\query}, \mathbf{\gpRewardFunction})\right)}{\partial \gpRewardFunction^{(i)} \partial \gpRewardFunction^{(j)}}.
\end{align*}

Now, we know how to approximate the posterior mean and the posterior covariance for the Laplace approximation of Equation~\eqref{eq:03_02_bayesian_update}. This allows us to model and update the reward with preference data using a GP.

We also want to perform inference from this approximated GP. Inference is not only useful for active query generation as we will show in Section~\ref{sec:04_03_gp}, but it also enables us to compute the reward expectations and variances given a trajectory.

\noindent\textbf{Inference.} Given two points $\trajectoryFeaturesFunction_1^*, \trajectoryFeaturesFunction_2^* \in \mathbb{R}^\numberOfFeatures$, we want to be able to compute the expected mean rewards $\gpMeanVector^*$ and also the covariance matrix between those two points $\gpCovarianceMatrix^*$, both of which will be useful for active query generation in Section~\ref{sec:04_03_gp}. These are given by:
\begin{align}
\gpMeanVector^* = \mathbb{E}\left[\mathbf{\gpRewardFunction}^* \mid \mathbf{\query}, \mathbf{\queryResponse}, \trajectoryFeaturesFunction_1^*, \trajectoryFeaturesFunction_2^*\right] = {\gpKernelFunction^*}^\top \gpCovarianceMatrix^{-1}\mathbf{\gpRewardFunction},
\label{eq:03_02_inference_mean}
\end{align}
where ${\gpKernelFunction^*}$ is a $2\times 2\abs{\comparisonDataset}$ matrix whose $i^{\textrm{th}}$ row consists of $\gpKernelFunction(\trajectoryFeaturesFunction^*_i, \trajectoryFeaturesFunction_1^{(j)})$ and $\gpKernelFunction(\trajectoryFeaturesFunction^*_i, \trajectoryFeaturesFunction^{(j)}_2)$ values for $j\in\{1,\ldots,\abs{\comparisonDataset}\}$, and
\begin{align}
\gpCovarianceMatrix_* = \gpCovarianceMatrix_{\bm{0}} - \gpKernelFunction_*\left(\bm{I}_{2\abs{\comparisonDataset}} + \gpNegativeHessianOfLoglikelihood\gpCovarianceMatrix\right)^{-1}\gpNegativeHessianOfLoglikelihood \gpKernelFunction_*^\top,
\label{eq:03_02_inference_cov}
\end{align}
where ${\gpCovarianceMatrix_{\bm{0}}}_{ij} = \gpKernelFunction\left(\trajectoryFeaturesFunction^*_i,\trajectoryFeaturesFunction^*_j\right)$ is a $2\times 2$ matrix, $\bm{I}_{2\abs{\comparisonDataset}}$ is the $2\abs{\comparisonDataset}\times 2\abs{\comparisonDataset}$ identity matrix.

Having a way to find the posterior mean and covariance as well as to perform inference means we now know how to learn a reward function modeled using a GP. In practice, the posterior mean can be used as a point estimate of the reward function, and the posterior covariance is useful for modeling the uncertainty over rewards. In the next section, we incorporate ordinal feedback on top of pairwise comparisons (as in \cite{chu2005gaussian}), which also enables us to define a \emph{region of avoidance} for safety-critical applications.

%% file: 03_learning/03_roial.tex
\label{sec:03_03_roial}

\begin{figure}[tb]
 \centering
 \includegraphics[width=0.5\linewidth]{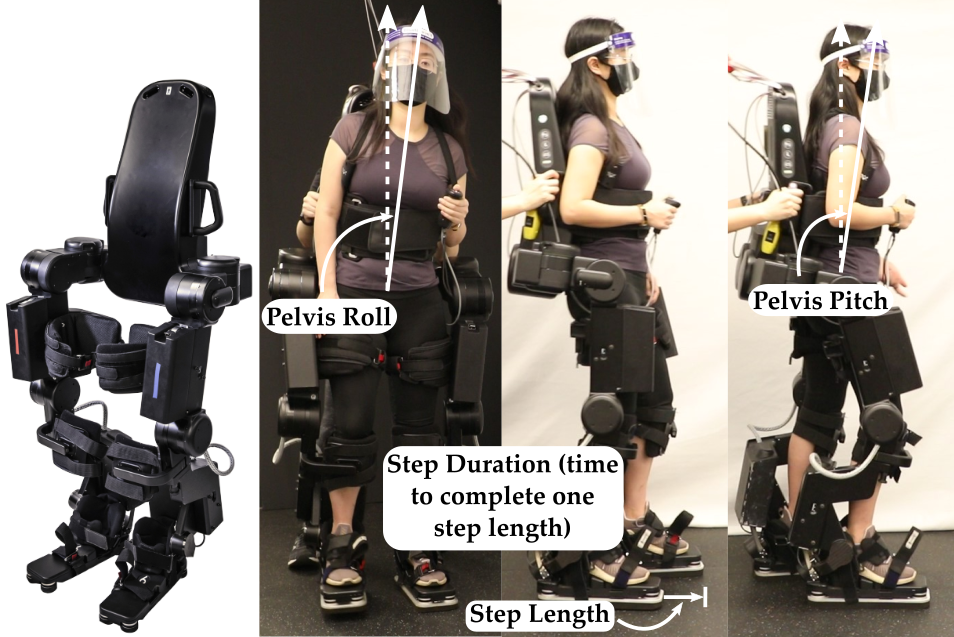}
 \caption{The Atalante exoskeleton, designed by Wandercraft, has 12 actuated joints, 6 on each leg. The experiments explore four gait parameters: step length, step duration, pelvis roll, and pelvis pitch.}
 \label{fig:03_03_Atalante}
\end{figure} 

Learning from comparative feedback naturally involves demonstrating some suboptimal trajectories to the human expert. In some cases, this might be problematic. For example, suppose we are trying to learn optimal gait parameters of a lower body exoskeleton (see Figure~\ref{fig:03_03_Atalante}) where each gait corresponds to a trajectory. The human user who will give the comparison feedback will wear this exoskeleton and make comparisons about how comfortable they feel. Asking them comparison questions that involve highly suboptimal trajectories will cause them to feel uncomfortable and/or unsafe. In practice, we need to avoid this as much as possible.

Although we only focus on learning from offline datasets in this chapter and defer active comparison data collection until Chapter~\ref{chap:active}, we now present how we can define such regions in the trajectory space $\trajectorySpace$ to avoid by utilizing ordinal feedback (in addition to pairwise comparisons) from humans.

More formally in this section, we denote this region of undesirable trajectories as the ``Region of Avoidance'' (ROA) and the region of remaining trajectories as the ``Region of Interest'' (ROI). In prior work on the highly-related area of safe exploration  \cite{sui2015safe, schreiter2015safe,berkenkamp2016safe,sui2018stagewise}, unsafe trajectories (or actions, or parameters) are considered to be catastrophically bad and therefore must be avoided completely.  However, the resulting algorithms can be overly conservative in settings such as ours, where occasionally sampling from bad regions is tolerable.

This section, together with Section~\ref{sec:04_04_roial}, proposes the Region of Interest Active Learning (ROIAL) algorithm, a novel active learning framework which queries the user for qualitative (ordinal) or preference (comparative) feedback to locate the ROI and estimate the reward function as accurately as possible over the ROI. In this section, we describe the learning algorithm, and Section~\ref{sec:04_04_roial} focuses on active querying.

The vast majority of prior work on preference learning obtains at most $1$ bit of information per pairwise comparison query \cite{houlsby2011bayesian,tucker2020preference, tucker2020linecospar,xu2010fast,furnkranz2010preference, thatte2018method,wilde2020active,qian2015learning,sui2017multi,bengs2021preference}. ROIAL additionally learns from ordinal labels \cite{chu2005gaussian}, which assign trajectories to discrete ordered categories such as ``bad,'' ``neutral,'' and ``good.'' Ordinal feedback enables ROIAL to both: 1) locate the ROI by learning the boundary between the least-preferred category (ROA) and remaining trajectories (ROI), and 2) estimate the reward function more efficiently within the ROI. Compared to the $1$ bit of information obtained per pairwise comparison query, each ordinal query yields up to $\log_2(\abs{\ordinalCategory})$ bits of information where $\ordinalCategory$ is the set of ordinal labels. Since ordinal feedback is identical for trajectories within each ordinal category, pairwise comparisons provide finer-grained information about the reward function's shape within the categories.

We describe the learning algorithm that performs GP regression using both pairwise comparisons and ordinal feedback, and learns the ROA in the subsequent subsections. In Section~\ref{sec:04_04_roial}, we extend it with active query generation to complete the description of the ROIAL algorithm, and validate it both in simulation and experimentally using the aforementioned lower-body exoskeleton task.

\subsection{Formulation}
We again consider a learning problem over a finite (but potentially-large) trajectory space $\trajectorySpace$, with a trajectory feature function $\trajectoryFeaturesFunction: \trajectorySpace\to \mathbb{R}^\numberOfFeatures$. Each trajectory $\trajectory \in \trajectorySpace$ is assumed to have an underlying reward to the user, $\trajectoryRewardFunction(\trajectory)=\gpRewardFunction(\trajectoryFeaturesFunction(\trajectory))$. The algorithm aims to learn the unknown reward function $\trajectoryRewardFunction: \trajectorySpace \to \mathbb{R}$. The trajectories' rewards can be written in the vectorized form $\bm{\gpRewardFunction} := [\gpRewardFunction(\trajectoryFeaturesFunction(\trajectory^{(1)})),\gpRewardFunction(\trajectoryFeaturesFunction(\trajectory^{(2)})),...,\gpRewardFunction(\trajectoryFeaturesFunction(\trajectory^{(\abs{\trajectorySpace})}))]^\top$, where $\{\trajectory^{(j)} \mid j = 1, \ldots, \abs{\trajectorySpace}\}$ are the trajectories in $\trajectorySpace$.

Specific to this section, we slightly change the comparison dataset structure: instead of having pairwise comparisons between two arbitrary trajectories, we assume the human user experiences (or watches) trajectories one by one and compares each trajectory to the previous one. This choice is made as it is more natural and time-efficient for the lower-body exoskeleton task we mentioned. However mathematically, this does not change anything: the learning algorithm could still handle pairwise comparison datasets with arbitrary trajectories as in the previous sections.

Accordingly, we let $\trajectory^{(i)} \in \trajectorySpace$ be the trajectory selected in trial $i$. We receive qualitative information about $\gpRewardFunction$ after each trial $i$, consisting of an ordinal label $\queryResponse_o^{(i)}$ and a comparison between $\trajectory^{(i)}$ and $\trajectory^{(i-1)}$ for $i \geq 2$. We use $\trajectory^{(i1)} \succ \trajectory^{(i2)}$ to denote a preference for trajectory $\trajectory^{(i1)}$ over $\trajectory^{(i2)}$, and following each trial $i$, collect these pairwise comparisons into a dataset $\comparisonDataset^{(i)} = \{\trajectory^{(i1)} \succ \trajectory^{(i2)} \mid i = 1,2, ...,\abs{\comparisonDataset} \}$. The ordinal labels are similarly collected into $\ordinalDataset^{(i)} = \{(\trajectory^{(i)},\queryResponse^{(i)}_o) \mid i = 1,2,...,\abs{\ordinalDataset} \}$. Again assuming no expert demonstrations in this section, the full user feedback dataset after iteration $i$ is defined as $\dataset^{(i)} := \comparisonDataset^{(i)} \cup \ordinalDataset^{(i)}$.

Ordinal feedback assigns one of pre-determined ordered labels to each sampled action. These (possibly noisy) labels are assumed to reflect ground truth ordinal categories (e.g., ``bad,'' ``neutral,'' ``good,'' etc.), which partition $\trajectorySpace$ into the sets that correspond to each ordinal label. We define the region of avoidance (ROA) as the trajectories that would fall into the set of lowest ordinal label. For instance, in the lower-body exoskeleton setting, it consists of gaits that make the user feel unsafe or uncomfortable. Similarly, the ROA could be defined as the union of multiple sets that correspond to the bottom ordinal labels. We define the region of interest (ROI) as the complement of the ROA, i.e., $\trajectorySpace \setminus \textrm{ROA}$. 

\subsection{Learning Algorithm}
This subsection describes the learning algorithm, which leverages qualitative human feedback to estimate the ROI and reward function (code available at \url{https://github.com/kli58/ROIAL}). We first discuss Bayesian modeling of the reward function, and then explain the procedure for rendering it tractable in high dimensions. We then detail the process for estimating the ROI.

\subsubsection{Bayesian Posterior Inference}
To simplify notation, this section omits the iteration $i$ from all quantities. Given the feedback dataset $\dataset = \comparisonDataset \cup \ordinalDataset$, the utilities $\bm{\gpRewardFunction}$ have posterior:
\begin{align} \label{eq:03_03_posterior}
    P(\bm{\gpRewardFunction} \mid \comparisonDataset,\ordinalDataset) \propto P(\comparisonDataset \mid \bm{\gpRewardFunction}) P(\ordinalDataset \mid \bm{\gpRewardFunction})P(\bm{\gpRewardFunction}),
\end{align}
where $P(\bm{\gpRewardFunction})$ is a Gaussian prior over the utilities $\bm{\gpRewardFunction}$:
\begin{align*}
    P(\bm{\gpRewardFunction}) = \frac{1}{(2\pi)^{\abs{\trajectorySpace}/2} \abs{\gpCovarianceMatrix}^{1/2}} \exp\left(-\frac{1}{2}\bm{\gpRewardFunction}^\top\gpCovarianceMatrix^{-1}\bm{\gpRewardFunction}\right),
\end{align*}
in which $\gpCovarianceMatrix \in \mathbb{R}^{\abs{\trajectorySpace} \times \abs{\trajectorySpace}}$, $\gpCovarianceMatrix_{jj'} = \gpKernelFunction(\trajectoryFeaturesFunction(\trajectory^{(j)}),\trajectoryFeaturesFunction(\trajectory^{(j')}))$, and $\gpKernelFunction$ is a kernel of choice. This section and Section~\ref{sec:04_04_roial} use the squared exponential kernel. 

\noindent\textbf{Comparison feedback.} We assume that the users' preferences are corrupted by noise as in \cite{chu2005preference}, such that:
\begin{align}
   & P(\trajectory^{(1)} \succ \trajectory^{(2)} \mid \bm{\gpRewardFunction}) = g_{C}\left(\frac{\gpRewardFunction(\trajectoryFeaturesFunction(\trajectory^{(1)}))-\gpRewardFunction(\trajectoryFeaturesFunction(\trajectory^{(2)}))}{\preferenceNoiseStd}\right), 
\end{align}
where $g_{C}: \mathbb{R} \to (0,1)$ is a monotonically-increasing link function, and $\preferenceNoiseStd > 0$ quantifies noisiness in the comparisons. Note that this is a generalized version of the Thurstonian model we used in Equation~\eqref{eq:03_02_human_model}.

\noindent\textbf{Ordinal feedback.} We define set of thresholds $\ordinalThresholdSet$ such that $-\infty = \ordinalThresholdSet_0 < \ordinalThresholdSet_1 < \ordinalThresholdSet_2 < \ldots < \ordinalThresholdSet_{\abs{\ordinalCategory}} = \infty$. These thresholds partition the trajectory space into $\abs{\ordinalCategory}$ ordinal categories $\ordinalCategory_1, \ordinalCategory_2, \dots, \ordinalCategory_{\abs{\ordinalCategory}}$. For any $\trajectory \in \trajectorySpace$, if $\gpRewardFunction(\trajectoryFeaturesFunction(\trajectory)) < \ordinalThresholdSet_1$, then $\trajectory \in \ordinalCategory_1$, and $\trajectory$ has an ordinal label of 1. Similarly, if $\ordinalThresholdSet_j \leq \gpRewardFunction(\trajectoryFeaturesFunction(\trajectory)) < \ordinalThresholdSet_{j+1}$, then $\trajectory \in \ordinalCategory_{j+1}$, and $\trajectory$ corresponds to an ordinal label of $j+1$.  We assume that the users' ordinal labels are corrupted by noise as in \cite{chu2005gaussian}, such that:
\begin{align}
    P(\queryResponse_o \mid \bm{\gpRewardFunction}, \trajectory) = g_{O}\left( \frac{\ordinalThresholdSet_{\queryResponse_o}-\gpRewardFunction(\trajectoryFeaturesFunction(\trajectory))}{\ordinalNoiseStd}\right) - g_{O}\left(\frac{\ordinalThresholdSet_{\queryResponse_o-1}-\gpRewardFunction(\trajectoryFeaturesFunction(\trajectory))}{\ordinalNoiseStd}\right),
\end{align}
where $g_{O}: \mathbb{R} \to (0,1)$ is a monotonically-increasing link function, and $\ordinalNoiseStd > 0$ quantifies the ordinal noise.

Assuming conditional independence of queries, the likelihoods $P(\comparisonDataset\mid\bm{\gpRewardFunction})$ and $P(\ordinalDataset\mid\bm{\gpRewardFunction})$ are:
\begin{align*}
    P(\comparisonDataset \mid \bm{\gpRewardFunction}) = & \prod_{i=1}^{\abs{\comparisonDataset}} g_{C}\left(\frac{\gpRewardFunction(\trajectoryFeaturesFunction(\trajectory^{(i1)}))-\gpRewardFunction(\trajectoryFeaturesFunction(\trajectory^{(i2)}))}{\preferenceNoiseStd}\right), \\
    P(\ordinalDataset \mid \bm{\gpRewardFunction}) = & \prod_{i=1}^{\abs{\ordinalDataset}} \left[g_{O} \left( \frac{\ordinalThresholdSet_{\queryResponse_o^{(i)}}-\gpRewardFunction(\trajectoryFeaturesFunction(\trajectory^{(i)}))}{\ordinalNoiseStd}\right) - g_{O}\left(\frac{\ordinalThresholdSet_{\queryResponse_O^{(i-1)}}-\gpRewardFunction(\trajectoryFeaturesFunction(\trajectory^{(i)}))}{\ordinalNoiseStd}\right)\right].
\end{align*}
Our simulations and experiments in Section~\ref{subsec:04_04_experiments} fix the hyperparameters $\preferenceNoiseStd$, $\ordinalNoiseStd$, and $\{\ordinalThresholdSet_j \mid j = 1, \ldots, \abs{\ordinalCategory} - 1\}$ in advance. One could also estimate them during learning using strategies such as evidence maximization, but this can be computationally very expensive, especially with large trajectory spaces.

Common choices of link function ($g_C$ and $g_O$) include the Gaussian cumulative distribution function \cite{chu2005preference,chu2005gaussian} and the sigmoid function, $g(x) = (1 + e^{-x})^{-1}$ \cite{tucker2020linecospar}. We model feedback via the sigmoid link function because empirical results suggest that a heavier-tailed noise distribution improves performance. We use the Laplace approximation to approximate the posterior as Gaussian as in Section~\ref{sec:03_02_gp}: $P(\bm{\gpRewardFunction} \mid \dataset^{(i)}) \approx \mathcal{N}(\bm{\hat{\gpRewardFunction}}^{(i)}, \hat{\gpCovarianceMatrix}^{(i)})$ \cite{williams2006gaussian}.

\subsubsection{High-Dimensional Tractability}
Calculating the model posterior is the algorithm's most computationally expensive step, and is intractable for large trajectory spaces. Most existing work in high-dimensional Gaussian process learning requires quantitative feedback \cite{kandasamy2015high, wang2013bayesian}. Previous work in preference-based high-dimensional Gaussian process learning \cite{tucker2020linecospar} restricts posterior inference to one-dimensional subspaces. However, the approach in \cite{tucker2020linecospar} is more amenable to the regret minimization problem because each one-dimensional subspace is biased toward regions of high posterior reward. Instead, to increase the online computing speed over high-dimensional spaces, in each iteration $i$ we select a subset $\trajectorySpace_S^{(i)} \subset \trajectorySpace$ of trajectories uniformly at random, and evaluate the posterior only over $\trajectorySpace_S^{(i)}$.

\subsubsection{Estimating the Region of Interest}
Since we lack prior knowledge about the ROI, it must be estimated during the learning process. In each iteration $i$, we model the ROI as the following set of trajectories: $\{\trajectory\in\trajectorySpace \mid \hat{\gpRewardFunction}^{(i-1)}(\trajectoryFeaturesFunction(\trajectory)) + \roialConservatism\hat{\sigma}^{(i-1)}(\trajectory) > \ordinalThresholdSet_1\}$, where $\hat{\sigma}^{(i-1)}(\trajectory)$ is the posterior standard deviation associated with $\trajectory$. The variable $\roialConservatism$ is a user-defined hyperparameter that determines the algorithm's conservatism in estimating the ROI; positive $\roialConservatism$'s are optimistic, while negative $\roialConservatism$'s are more conservative in avoiding the ROA. In practice, we evaluate trajectories in the randomly-selected subset $\trajectorySpace_S^{(i)}$ and define $\trajectorySpace_{\textrm{ROI}}^{(i)} = \{ \trajectory \in \trajectorySpace_S^{(i)} \mid \hat{\gpRewardFunction}^{(i-1)}(\trajectoryFeaturesFunction(\trajectory)) + \roialConservatism\hat{\sigma}^{(i-1)}(\trajectory) > \ordinalThresholdSet_1 \}$ in each iteration 
$i$.  Note that this definition is optimistic, whereas safe exploration approaches use pessimistic definitions \cite{sui2015safe,schreiter2015safe,berkenkamp2016safe,sui2018stagewise}.

\subsubsection{Summary}
In Section~\ref{sec:03_02_gp}, we studied a GP regression method that uses pairwise comparisons. In this section, we extended it with ordinal feedback, and defined region of avoidance (ROA) and region of interest (ROI) based on the ordinal categories. Together, these two sections present a computational method of learning non-parametric reward functions from pairwise comparisons and ordinal feedback. We will extend these methods with active querying in Chapter~\ref{chap:active} and present experiment results that include the lower-body exoskeleton task we mentioned in the beginning of this section. The subsequent sections in this chapter goes back to parametric reward functions and focuses on reward learning algorithms that make use of other forms of comparative feedback. Since the Bayesian learning approach is maintained, they can be easily extended to non-parametric reward functions with GPs via Laplace approximation as long as the reward function is stationary and unimodal.\footnote{In fact, best-of-many choice queries as presented in Section~\ref{sec:03_01_pairwise_comparisons} could also be used for GP regression. However, we intentionally focused on pairwise comparisons as we adopt them later when we introduce active query generation in Sections~\ref{sec:04_03_gp} and \ref{sec:04_04_roial}.}

%% file: 03_learning/04_scale.tex
\label{sec:03_04_scale}

In Section~\ref{sec:03_01_pairwise_comparisons}, we introduced best-of-many choice queries and presented a Bayesian approach for learning parametric reward functions using them along with expert demonstrations. In Sections~\ref{sec:03_02_gp} and \ref{sec:03_03_roial} we focused on a special case of best-of-many choice queries where the user is presented with only $2$ options, making it a pairwise comparison question between the options. Using this special case, we showed how we can learn non-parametric reward functions by modeling them as Gaussian processes, optionally along with ordinal feedback.

\begin{figure}[!t]
	\centering
    \includegraphics[width=\textwidth]{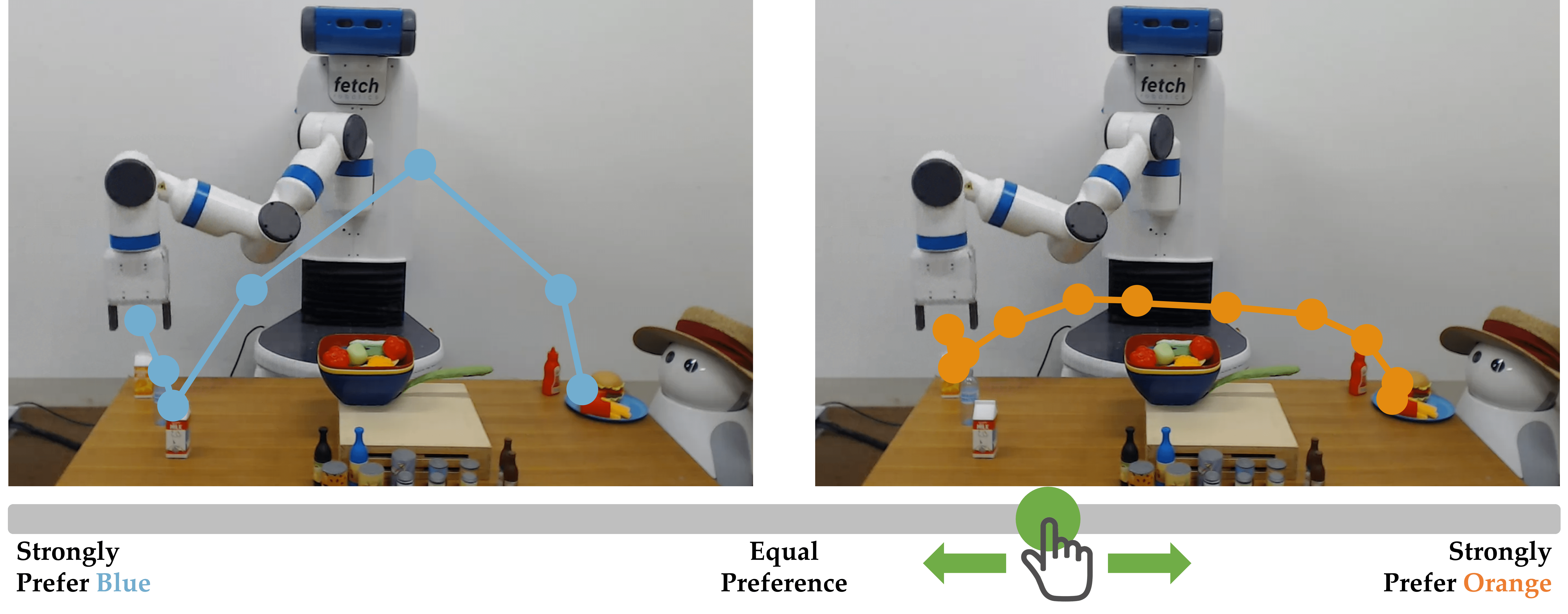}
	\caption{Scale feedback allows users to provide finely detailed comparisons between different options.}
	\label{fig:03_04_frontfig}
\end{figure}

Pairwise comparisons, or best-of-many choices in general, although simple to collect, are limiting in a number of ways. Consider the example shown in Figure~\ref{fig:03_04_frontfig}, where a robot is tasked to serve a drink to a customer. The customer might have different preferences over the type of drink to have (milk, orange juice, or water), or the specifics of the trajectory the robot takes (e.g., if it goes over the stove or around it which can affect the temperature of the drink or the likelihood of the robot accidentally hitting the pan handle). A strict pairwise comparison between two trajectories, although minimizing interface complexity and mental effort for the user, does not really capture these intricacies of human preferences. In addition, when the user is indifferent towards both options, learning becomes difficult since users may become noisier in their responses. We thus need to have a more expressive way of collecting data from humans. 

Several works, e.g., \citet{holladay2016active} and \citet{basu2018learning}, investigate modifications of learning from pairwise comparisons where users can also answer \emph{About Equal} (which we also experiment with in Section~\ref{sec:04_02_information_gain}). These two forms of pairwise comparison feedback are usually referred to as \emph{strict} and \emph{weak} pairwise comparisons. When the user chooses the neutral answer, the robot learns to assign about equal reward to the presented trajectories.

In the proposed scale feedback framework in this section, we take the weak pairwise comparisons approach one step further: Instead of three discrete values for feedback (prefer A, prefer B, neutral) users give quasi-continuous feedback. Our key insight is that allowing users to provide a scaled approach on a slider (as shown in Figure~\ref{fig:03_04_frontfig}) can provide a more expressive medium for learning from humans and capture nuances in their preferences. This allows the user to indicate \emph{how much} they prefer one option over the other.

Slider bars have been used in robotics for tuning parameters \cite{racca2020interactive}. More related to our work, \citet{cabi2020scaling} proposed using them for \emph{reward sketching}. Instead of assigning a numerical preference between presented options, users continuously indicate the robot's progress towards some goal. However, this requires users to assign scores to different parts of trajectories. Developing the scale feedback for preference-based learning, we retain the ease of comparing trajectories.

To this end, we propose \emph{scale feedback} as a new mode of interaction: Instead of a strict question on which of the two proposed trajectories the user prefers, we allow for more nuanced feedback using a slider bar. We design a Gaussian model for how users provide scale feedback, and learn a reward function capturing human preferences. Similar to Section~\ref{sec:03_01_pairwise_comparisons} and prior work in robotics, we assume this reward is a parametric function of a set of trajectory features~\cite{abbeel2004apprenticeship,wilde2020improving,palan2019learning,holladay2016active}, where the main task of learning from scale feedback is to recover the parameters of this reward function.

We demonstrate the performance benefit of scale feedback over pairwise comparisons in a driving simulation. Further, we investigate its practicality in two user studies with the real robot experiment shown in Figure~\ref{fig:03_04_frontfig}. Our results suggest scale feedback leads to significant improvements in learning performance. We present these simulation and experiment results in Chapter~\ref{chap:active} after we develop the active querying methods for scale feedback in Section~\ref{sec:04_05_scale}.

\subsection{Formulation}\label{subsec:03_04_formulation}

We now introduce the notation we use in this section and formulate the learning from scale feedback problem.

\noindent\textbf{Reward function.} We again consider the scenario where a robot needs to learn a reward function from a user, for example for customizing its behavior to the preferences of the user.
We assume the user evaluates robot trajectories $\trajectory\in \trajectorySpace$ from a potentially infinite trajectory space $\trajectorySpace$ based on a vector of features $\trajectoryFeaturesFunction(\trajectory)\in\mathbb{R}^{\numberOfFeatures}$. Similar to Section~\ref{sec:03_01_pairwise_comparisons} and prior works in robotics \cite{abbeel2004apprenticeship,wilde2020improving,palan2019learning,holladay2016active}, we define a parametric reward function $\trajectoryRewardFunction$ that assigns a numerical value to any trajectory $\trajectory$:
\begin{align}
    \trajectoryRewardFunction_{\weights}(\trajectory) = \gpRewardFunction_{\weights}(\trajectoryFeaturesFunction(\trajectory))\:.
\end{align}
These features are usually provided by a domain expert incorporating the core factors that the reward needs to capture, e.g., collision with other objects, or distance to the goal.

Further, we assume in this section the robot has access to a motion planner that finds an optimal trajectory given reward function parameters, i.e., the planner is a (deterministic) function $\plannerFunction$ where $\plannerFunction(\weights) = \argmax_{\trajectory\in\trajectorySpace}\trajectoryRewardFunction_{\weights}(\trajectory)$.

\noindent\textbf{Regret.}
Similar to \cite{wilde2020active}, we define the \emph{regret} between any two parameter sets $(\weights,\weights')$ as the difference in the reward $\weights'$ assigns to the trajectories $\plannerFunction(\weights)$ and $\plannerFunction(\weights')$:
\begin{align}
    \regretFunction(\weights, \weights') = \trajectoryRewardFunction_{\weights'}(\plannerFunction(\weights'))  - \trajectoryRewardFunction_{\weights'}(\plannerFunction(\weights))\:,
    \label{eq:03_04_regret}
\end{align}
which quantifies the suboptimality when the true weights are $\weights'$, but the trajectory is optimized using $\weights$.

\noindent\textbf{Learning.}
Let $\weights^*$ denote the true weights for the reward function. These weights are not known to the robot; the only information initially available is a prior distribution $\belief^0 = P(\weights=\weights^*)$, which might be initialized using offline demonstrations as in Section~\ref{sec:03_01_pairwise_comparisons}. The robot learns $\weights^*$ by iteratively presenting the user with two trajectories $\query^{(i)} = \left(\trajectory_1^{(i)}, \trajectory_2^{(i)}\right)$ for iterations $i\in\{1,2,\ldots\}$. We extend the learning from pairwise comparisons framework, where users simply indicate the trajectory they prefer, to a setting where they instead provide a more finely detailed \emph{scale feedback}.

\noindent\textbf{Scale Feedback.} Presented with two trajectories $\trajectory_1$ and $\trajectory_2$, the user returns numerical feedback  $\queryResponse\in[-1,1]$. If $\queryResponse=0$, this means the user has no preference between the trajectories, $\queryResponse=1$ equals a strong preference for trajectory $\trajectory_1$ and $\queryResponse=-1$ a strong preference for trajectory $\trajectory_2$.

From an interface design and expressiveness perspective, it is undesirable to have users give a numerical value for $\queryResponse$. Instead, they can express such a feedback with a slider bar with a more fine-grained set of options. An example is illustrated in Figure~\ref{fig:03_04_frontfig}. We let $\scaleDataset=\{(\query^{(i)},\queryResponse^{(i)}\}_{i=1}^{\abs{\scaleDataset}}=\{(\trajectory^{(i)}_1,\trajectory^{(i)}_2,\queryResponse^{(i)})\}_{i=1}^{\abs{\scaleDataset}}$ be the set of recorded scale feedback from the user.

\noindent\textbf{Performance Measures.} Let $\hat{\weights}$ be the robot's estimate of $\weights^*$. We consider two performance metrics. One is \emph{alignment} of parameters \cite{sadigh2017active,biyik2019asking}, $\mathtt{Alignment} = \frac{\hat{\weights}\cdot \weights^*}{\norm{\hat{\weights}}\cdot \norm{\weights^*}}$, measuring the cosine similarity of vectors $\hat{\weights}$ and $\weights^*$, i.e., how well the parameters of the user's reward function are learned. Alternatively, \citet{wilde2020active} proposed the relative error in \emph{cost}. We adapt this as the $\mathtt{Relative\_Reward}=\frac{\trajectoryRewardFunction_{\weights^*}(\plannerFunction(\hat{\weights}))}{\trajectoryRewardFunction_{\weights^*}(\plannerFunction(\weights^*))}$, measuring how much the user likes the trajectory optimized for $\hat{\weights}$ compared to the one optimized for $\weights^*$. 

\noindent\textbf{Problem Statement.} Given a robot motion planner $\plannerFunction$ and a user whose preferences come from the prior $\belief^0 = P(\weights=\weights^*)$, our goal in this section is to develop a learning model that maximizes either of the performance measures by performing inference of reward function parameters from scale feedback. Later in Section~\ref{sec:04_05_scale}, we will develop an adaptive querying policy for querying the user with maximally informative scale feedback questions for some number of rounds.

\subsection{Our Approach}
We now briefly review learning from pairwise comparisons from a new perspective, and then extend the framework to scale feedback.

\subsubsection{Pairwise Comparisons Feedback}
\begin{figure}[th]
    \centering
    \includegraphics[width=0.7\textwidth]{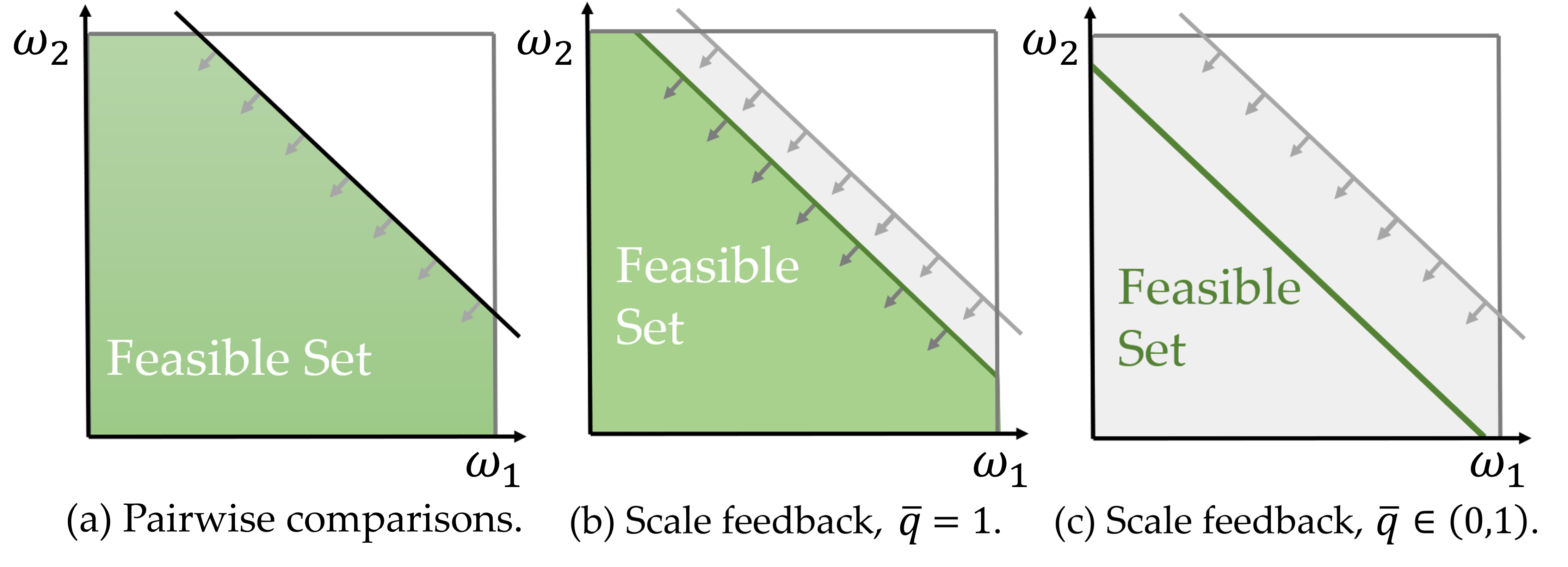}
    \caption{Different feasible sets learned from pairwise comparison and scale feedback under the linear reward model. Shown is the updated weight space (green) after observing user feedback for one $(\trajectory_1,\trajectory_2)$ pair. If $\bar{\queryResponse}=1$, scale feedback enables us to learn a tighter half-space; when $\bar{\queryResponse}\in(0,1)$, scale feedback enables us to learn an equality, i.e., a hyperplane.}
    \label{fig:03_04_feasible_sets}
\end{figure}

When presented with two trajectories $\trajectory_1$ and $\trajectory_2$, a user returns an ordering $\trajectory_1\succeq \trajectory_2$ ($\trajectory_1$ is preferred) or $\trajectory_1\preceq \trajectory_2$ ($\trajectory_2$ is preferred). In a noiseless setting, we have
\begin{align}
\trajectoryRewardFunction_{\weights^*}(\trajectory_1) - \trajectoryRewardFunction_{\weights^*}(\trajectory_2)\geq 0  \iff \trajectory_1\succeq \trajectory_2 \:.
\label{eq:03_04_learn_from_choice}
\end{align}
That is, the trajectory $\trajectory_1$ has a reward that is at least as high as that of $\trajectory_2$ with respect to the hidden true user weights $\weights^*$. Equation~\eqref{eq:03_04_learn_from_choice} already contains an observation model: If the user chooses trajectory $\trajectory_1$, the robot can infer that $\trajectory_1$ has a higher reward with respect to $\weights^*$.
This inequality defines a subset in the parameter space: $\halfspace(\trajectory_1,\trajectory_2)=\{\weights \mid (\trajectoryRewardFunction_{\weights}(\trajectory_1)-\trajectoryRewardFunction_{\weights}(\trajectory_2)) \geq 0\}$ containing all weights that are \emph{feasible} given the observed user choice. Over $\abs{\comparisonDataset}$ iterations, we can intersect the subsets $\halfspace(\trajectory_1^{(1)},\trajectory_2^{(1)}), \dots,\halfspace(\trajectory_1^{(\abs{\comparisonDataset})},\trajectory_2^{(\abs{\comparisonDataset})})$ to obtain the \emph{feasible set} $\scaleFeasibleSet^{(\abs{\comparisonDataset})}$. An example is shown in Figure~\ref{fig:03_04_feasible_sets}a for a linear reward function.

\subsubsection{Scale Feedback}

Scale feedback allows the robot to gain more information: the robot can also infer by \emph{how much} the user prefers $\trajectory_1$, allowing for learning tighter feasible sets.
We extend the model in \eqref{eq:03_04_learn_from_choice} and show how a noiseless user would provide scale feedback and then study how a robot can learn from it.

\begin{definition}[Maximum Reward Gap]
Given true parameters $\weights^*$ for a user, the maximum reward gap is 
\begin{align}
\maxRewardGap^* = \max_{\trajectory_1,\trajectory_2\in \trajectorySpace}\left(\trajectoryRewardFunction_{\weights^*}(\trajectory_1) - \trajectoryRewardFunction_{\weights^*}(\trajectory_2)\right)\:.
\label{eq:03_04_max_reward_gap}
\end{align}
\end{definition}
We notice that the maximum reward gap cannot be computed, since $\weights^*$ is unknown to the robot. Nevertheless, we can formulate the user choice model and then derive an observation model.

\noindent\textbf{User model.}
The maximum reward gap helps to define when a noiseless user would indicate a strong preference. We assume this occurs if and only if the difference in reward of $\trajectory_1$ and $\trajectory_2$ with respect to $\weights^*$ is at least $\sensitivityThreshold^*\maxRewardGap^*$ for some $0<\sensitivityThreshold^*\leq1$. Here $\sensitivityThreshold^*$ is a saturation parameter which governs at what reward difference (w.r.t. to the maximum gap) the user's feedback gets saturated to a strong preference. For any other $(\trajectory_1,\trajectory_2)$ where $\abs{\trajectoryRewardFunction_{\weights^*}(\trajectory_1)-\trajectoryRewardFunction_{\weights^*}(\trajectory_2)}\in [0,\sensitivityThreshold^*\maxRewardGap^*)$, we assume the user to linearly scale the noiseless response $\bar{\queryResponse}$ between $-1$ and $1$, which leads to the following model.

\begin{definition}[Noiseless User Model]
Presented with two trajectories $\trajectory_1$ and $\trajectory_2$, a noiseless user with the saturation parameter $\sensitivityThreshold^*\in(0,1]$ will always provide the following feedback:
\begin{equation}
\begin{aligned}
   \bar{\queryResponse} = 
   \begin{cases}
   1
   \quad\text{ if } \trajectoryRewardFunction_{\weights^*}(\trajectory_1) - \trajectoryRewardFunction_{\weights^*}(\trajectory_2)
   \geq \sensitivityThreshold^*\maxRewardGap^*,\\
   -1
   \quad\text{ if } 
   \trajectoryRewardFunction_{\weights^*}(\trajectory_2)-\trajectoryRewardFunction_{\weights^*}(\trajectory_1)\geq \sensitivityThreshold^{\weights^*}{\maxRewardGap^{\weights^*}},\\
   \frac{\trajectoryRewardFunction_{\weights^*}(\trajectory_1) - \trajectoryRewardFunction_{\weights^*}(\trajectory_2)}{\sensitivityThreshold^*\maxRewardGap^*}
   \quad\text{ otherwise }.
   \end{cases}
\end{aligned}
\label{eq:03_04_noisefree_model_saturated}
\end{equation}
\end{definition}

\begin{figure}[!t]
		\centering
		\centering
		\begin{subfigure}[t]{0.43\textwidth}
        \includegraphics[width=.99\textwidth]{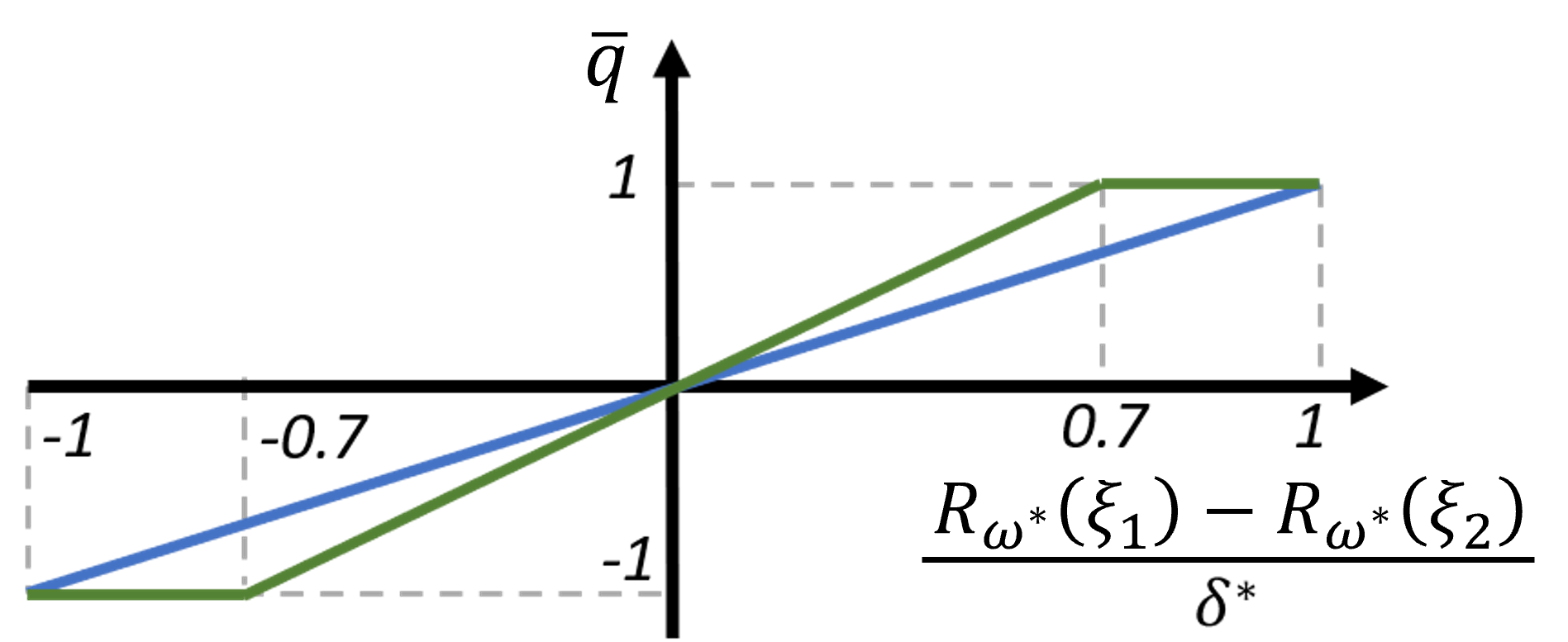}
		\caption{User model for providing scale feedback with $\sensitivityThreshold^*=1$ (blue) and $\sensitivityThreshold^*=0.7$ (green).}
		\label{fig:03_04_saturated_model_alpha}
		\end{subfigure}
		\hfill
		\begin{subfigure}[t]{0.56\textwidth}
		\centering
        \includegraphics[width=.99\textwidth]{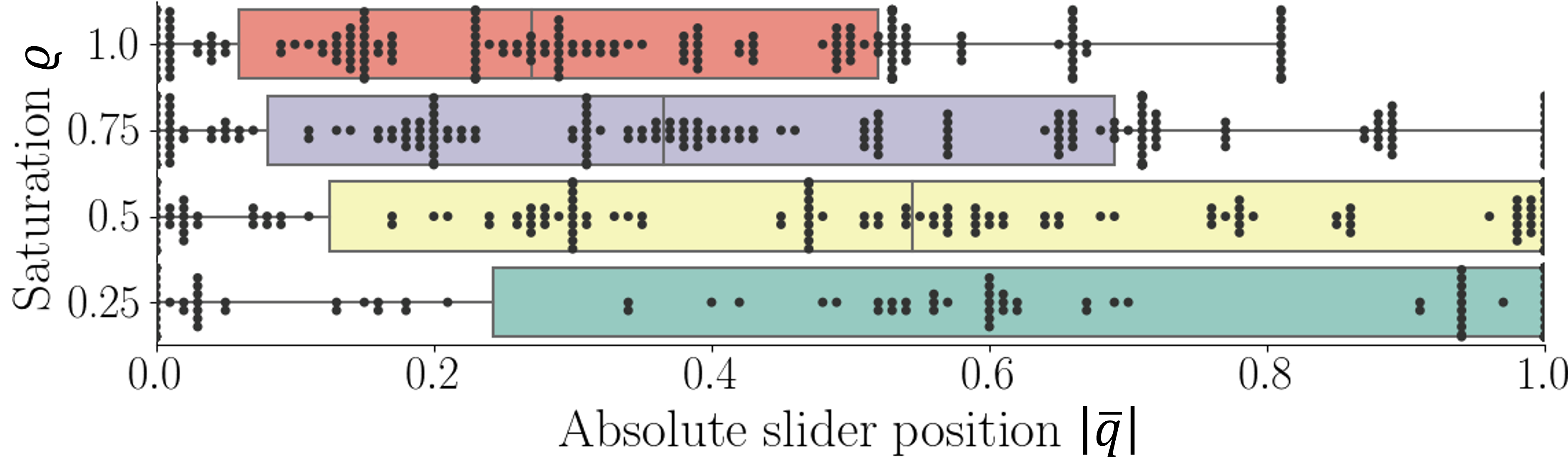}
		\caption{Example slider feedback for different $\sensitivityThreshold$. The boxplots indicate the four quartiles of the absolute slider values.}
		\label{fig:03_04_saturated_model_examples}
		\end{subfigure}
		\caption{Noiseless user model.}
		\label{fig:03_04_sat_model}
\end{figure}

We illustrate the noiseless user model in Figure~\ref{fig:03_04_saturated_model_alpha} under different saturation parameters $\sensitivityThreshold^*$. In Figure~\ref{fig:03_04_saturated_model_examples}, we show a simulated example: for a fixed $\weights^*$ we simulate how users with different values for $\sensitivityThreshold^*$ would provide scale feedback to the same $20$ queries. For larger $\sensitivityThreshold^*$, they position the slider closer to the neutral position. Finally, we derive an observation model for the noiseless user:
\begin{equation}
\begin{aligned}
\bar{\queryResponse}=-1 &\implies \trajectoryRewardFunction_{\weights^*}(\trajectory_1) - \trajectoryRewardFunction_{\weights^*}(\trajectory_2)\leq \bar{\queryResponse}\sensitivityThreshold^*\maxRewardGap^*\\
\bar{\queryResponse}\in(-1,1)  &  \implies \trajectoryRewardFunction_{\weights^*}(\trajectory_1) - \trajectoryRewardFunction_{\weights^*}(\trajectory_2)= \bar{\queryResponse}\sensitivityThreshold^*\maxRewardGap^*,\\
\bar{\queryResponse}=1 &\implies \trajectoryRewardFunction_{\weights^*}(\trajectory_1) - \trajectoryRewardFunction_{\weights^*}(\trajectory_2)\geq \bar{\queryResponse}\sensitivityThreshold^*\maxRewardGap^*.
\end{aligned}
\label{eq:03_04_obs_model_det_saturated}
\end{equation}

Figures \ref{fig:03_04_feasible_sets}b and \ref{fig:03_04_feasible_sets}c illustrate the resulting feasible sets from \eqref{eq:03_04_obs_model_det_saturated} for a linear reward model. Moreover, we notice the user-specific and unknown parameters $\sensitivityThreshold^*$ and $\maxRewardGap^*$ always appear as a product.
Thus, this product can be seen as a single additional parameter, and the notion of feasible set $\scaleFeasibleSet$ can be readily extended to this augmented parameter space.

\subsubsection{Probabilistic User Feedback}
\label{sec:prob_model}

In practice, users are often noisy; they might consider additional or slightly different features than the robot, not follow the parametric reward function, or simply be uncertain in some answers. Since we cannot expect users to always provide slider feedback following \eqref{eq:03_04_noisefree_model_saturated}, we introduce a probabilistic model where we add uncertainty to the placement of the slider.

Another practical limitation is the fact that we cannot collect truly continuous feedback from the users. Instead, the slider bar has a step size $\sliderStepSize \in (0,1]$ such that the user provides feedback of the form $n\sliderStepSize$ for $n\in\mathbb{Z}$ and $-\sliderStepSize^{-1}\leq n\leq\sliderStepSize^{-1}$. Note that $\sliderStepSize\to0$ retains the continuous scale feedback, whereas $\sliderStepSize=1$ gives the standard weak pairwise comparison model where the feedback is always in $\{-1,0,1\}$.

\begin{definition}[Probabilistic User Model]
Given a user $\weights^*$ and a query $(\trajectory_1,\trajectory_2)$, let $\bar{\queryResponse}$ be the user feedback defined in the noiseless user model in \eqref{eq:03_04_noisefree_model_saturated}.
A probabilistic user using a slider bar with a step size of $\sliderStepSize$ then provides feedback
\begin{align}
   \queryResponse = \mathrm{round}\!\left(\bar{\queryResponse} + \epsilon, \sliderStepSize \right)
   \label{eq:03_04_noisy_model}
\end{align}
where $\epsilon$ is a zero-mean Gaussian noise, i.e., $\epsilon\sim \mathcal{N}(0, \scaleNoiseStd^2)$ with standard deviation $\scaleNoiseStd$, and $\mathrm{round}\!\left(x,\sliderStepSize\right)$ outputs $n\sliderStepSize$ closest to $x$ such that $n\in\mathbb{Z}\cap [-\sliderStepSize^{-1},\sliderStepSize^{-1}]$.
\label{def:03_04_probabilistic_user_model}
\end{definition}

\noindent\textbf{Probabilistic Observation Model.}
Given the probabilistic user model, we now show how a robot can infer about $\weights^*$ from scale feedback.
In the noiseless case, user feedback defines a feasible set. For the probabilistic case, we instead derive a distribution over $\weights$ and $\sensitivityThreshold$.
Let $\maxRewardGap(\weights) = \max_{\trajectory_1,\trajectory_2\in\trajectorySpace}(\trajectoryRewardFunction_{\weights}(\trajectory_1)-\trajectoryRewardFunction_{\weights}(\trajectory_2))$, similar to \eqref{eq:03_04_max_reward_gap}. Then for $0 < \sensitivityThreshold \leq 1$, the belief is defined
\begin{align}
P(\weights, \sensitivityThreshold \mid \bar{\queryResponse}, \trajectory_1, \trajectory_2) =
\begin{cases}
\tilde{P}(\weights, \sensitivityThreshold \mid \bar{\queryResponse}, \trajectory_1, \trajectory_2) & \textrm{if } \bar{\queryResponse} \in(-1,1),\\
P^+(\weights, \sensitivityThreshold \mid \bar{\queryResponse}, \trajectory_1, \trajectory_2) & \textrm{if } \bar{\queryResponse}=1,\\
P^-(\weights, \sensitivityThreshold \mid \bar{\queryResponse}, \trajectory_1, \trajectory_2) & \textrm{if } \bar{\queryResponse}=-1,
\end{cases}
\label{eq:03_04_user_observation_noisy}
\end{align}
where
\begin{align}
\tilde{P}(\weights, \sensitivityThreshold \mid \bar{\queryResponse}, \trajectory_1, \trajectory_2) &\propto
\begin{cases}
1 \text{ if } \trajectoryRewardFunction_{\weights}(\trajectory_1)-\trajectoryRewardFunction_{\weights}(\trajectory_2)=\bar{\queryResponse}\sensitivityThreshold\maxRewardGap(\weights),\\
0\text{ otherwise }.
\end{cases}\\
P^+(\weights, \sensitivityThreshold \mid \bar{\queryResponse}, \trajectory_1, \trajectory_2) &\propto
\begin{cases}
1 \text{ if } \trajectoryRewardFunction_{\weights}(\trajectory_1)-\trajectoryRewardFunction_{\weights}(\trajectory_2) \geq \bar{\queryResponse}\sensitivityThreshold\maxRewardGap(\weights),\\
0\text{ otherwise }.
\end{cases}\\
P^-(\weights, \sensitivityThreshold \mid \bar{\queryResponse}, \trajectory_1, \trajectory_2) &\propto
\begin{cases}
1 \text{ if } \trajectoryRewardFunction_{\weights}(\trajectory_1)-\trajectoryRewardFunction_{\weights}(\trajectory_2) \leq \bar{\queryResponse}\sensitivityThreshold\maxRewardGap(\weights),\\
0\textrm{ otherwise}.
\end{cases}
\label{eq:03_04_prob_obs_beta}
\end{align}

Given noisy user feedback $\queryResponse$ as in \eqref{eq:03_04_noisy_model}, we can define a probabilistic density function $P(\bar{\queryResponse} \mid \queryResponse)$.
Together with \eqref{eq:03_04_user_observation_noisy} we derive a compound probability distribution
\begin{align}
P(\weights, \sensitivityThreshold \mid \queryResponse, \trajectory_1, \trajectory_2)
=
\int_{-1}^1 P(\weights, \sensitivityThreshold \mid \bar{\queryResponse}, \trajectory_1, \trajectory_2)P(\bar{\queryResponse} \mid \queryResponse) d\bar{\queryResponse}.
\label{eq:03_04_compound}
\end{align}
where we can write $P(\bar{\queryResponse} \mid \queryResponse)$ for $\bar{\queryResponse}\in[-1,1]$ as
\begin{align}
P(\bar{\queryResponse} \mid \queryResponse) \propto \begin{cases}
\standardNormalCdf\left(\frac{\queryResponse-\bar{\queryResponse}+\sliderStepSize/2}{\scaleNoiseStd}\right) & \textrm{if }\queryResponse=-1,\\
\standardNormalCdf\left(\frac{\bar{\queryResponse}-\queryResponse+\sliderStepSize/2}{\scaleNoiseStd}\right) - \standardNormalCdf\left(\frac{\bar{\queryResponse}-\queryResponse-\sliderStepSize/2}{\scaleNoiseStd}\right) & \textrm{if }\queryResponse\in(-1,1),\\
\standardNormalCdf\left(\frac{\bar{\queryResponse}-\queryResponse+\sliderStepSize/2}{\scaleNoiseStd}\right) & \textrm{if }\queryResponse=1,\\
\end{cases}
\label{eq:03_04_slider_from_noisy}
\end{align}
and $P(\bar{\queryResponse} \mid \queryResponse)=0$ for $\bar{\queryResponse}\not\in[-1,1]$. Here, $\standardNormalCdf$ denotes the cdf of the standard normal distribution. Finally, given a dataset $\scaleDataset=\{(\trajectory_1^{(i)},\trajectory_2^{(i)},\queryResponse^{(i)})\}_{i=1}^{\abs{\scaleDataset}}$ and some prior $\belief^0(\weights,\sensitivityThreshold)$, the joint posterior is
\begin{align}
\belief^{\abs{\scaleDataset}}(\weights, \sensitivityThreshold) \propto \belief^0(\weights, \sensitivityThreshold)
\prod_{i=1}^{\abs{\scaleDataset}} P(\weights, \sensitivityThreshold \mid \queryResponse^{(i)}, \trajectory_1^{(i)}, \trajectory_2^{(i)}).
\label{eq:03_04_joint_post}
\end{align}
Here, we can factor $\belief^0(\weights, \sensitivityThreshold)$ as $P(\weights)P(\sensitivityThreshold)$ by assuming $\weights$ and $\sensitivityThreshold$ are independent and we also have a prior for $\sensitivityThreshold^*$. We can then take the expectation of the posterior $P(\weights, \sensitivityThreshold \mid \scaleDataset)$ as a point estimate of the learned user model.

\subsection{Algorithm Design}
\label{sec:algorithm}
We now outline the learning algorithm.
Over $\abs{\scaleDataset}$ iterations:
(i) the robot generates a query $(\trajectory_1^{(i)}, \trajectory_2^{(i)})$ (active query generation for scale feedback will be presented in Section~\ref{sec:04_05_scale}), (ii) the user provides scale feedback to the query in the form of the slider value $\queryResponse^{(i)}$ (in the noiseless case, $\queryResponse^{(i)}=\bar{\queryResponse}^{(i)}$), and (iii) the robot updates its dataset and posterior using Equation~\eqref{eq:03_04_joint_post}.
After iteration $\abs{\scaleDataset}$, the algorithm returns the expected parameters $\hat{\weights}=\mathbb{E}\left[\weights \mid \scaleDataset\right]$.

\subsubsection{Worst Case Error Bound}

To compare scale feedback to pairwise comparisons, we establish a worst case bound on the performance measures for both frameworks.
We introduce the \emph{worst-case error} as the maximum negative performance measure, $1-\mathtt{Alignment}(\weights, \weights^*)$ (the bounds also generalize to $1-\mathtt{Relative\_Reward}(\weights, \weights^*)$, but we use only $\mathtt{Alignment}$ for brevity). The constant in front ensures a positive value, which we then discount with the posterior belief, given observations $\scaleDataset$ (or $\comparisonDataset$ in the case of pairwise comparisons):
\begin{align}
\mathtt{Err}^{\max}(\weights^*, \scaleDataset)=\max_{\weights} b^{\abs{\scaleDataset}}(\weights) (1-\mathtt{Alignment}(\weights, \weights^*))\:
\label{eq:03_04_upper_bound_true_error}
\end{align}
where $b^{\abs{\scaleDataset}}(\weights)$ is obtained by marginalizing the posterior $b^{\abs{\scaleDataset}}(\weights,\sensitivityThreshold)$ over $\sensitivityThreshold$. This describes the worst $\weights$ the robot could pick, discounted by the posterior distribution learned from data $\scaleDataset$.
In the noiseless setting, this simplifies to $\max_{\weights\in\halfspace}1-\mathtt{Alignment}(\weights, \weights^*)$ where $\halfspace$ is the feasible set.

\begin{proposition}[Upper error bound]
\label{prop:03_04_upper_bound}
Let $\scaleDataset$ denote the dataset of scale feedback and $\comparisonDataset$ be the dataset of pairwise comparisons for the same set of queries (trajectory pairs). For any user weights $\weights^*$, it holds in the noiseless setting that $\mathtt{Err}^{\max}(\weights^*, \scaleDataset)\leq \mathtt{Err}^{\max}(\weights^*, \comparisonDataset)$.
\end{proposition}

The proof follows from the observation $\halfspace^{\mathtt{Scale}}\subseteq\halfspace^{\mathtt{Choice}}$, i.e., scale feedback removes more volume from the parameter space. Hence, the worst choice of an estimate $\hat{\weights}$ given observations is guaranteed to have a smaller worst-case error when using scale feedback. The full proof is in Appendix~\ref{app:03_04_proposition_proof}.

We defer the simulation and user study results to Section~\ref{sec:04_05_scale} where we will also present an active querying approach for scale feedback. In the remainder of this chapter, we move to more complex settings where the reward functions to be learned are either multimodal or non-stationary.

%% file: 03_learning/05_rankings.tex
\label{sec:03_05_rankings}

Up to this point in the thesis, we focused on learning a \emph{unimodal} reward function that models human preferences on a target task. However, this unimodality assumption does not always hold: human preferences are usually more complex and need to be captured via a multimodal representation. Further, even if the preferences of a human are truly unimodal, we often use a mixture of data from multiple humans, which can be difficult to disentangle, leading to multimodality.

\begin{figure}[!t]
		\centering
		\centering
		\begin{subfigure}[t]{0.35\textwidth}
        \includegraphics[width=.99\textwidth]{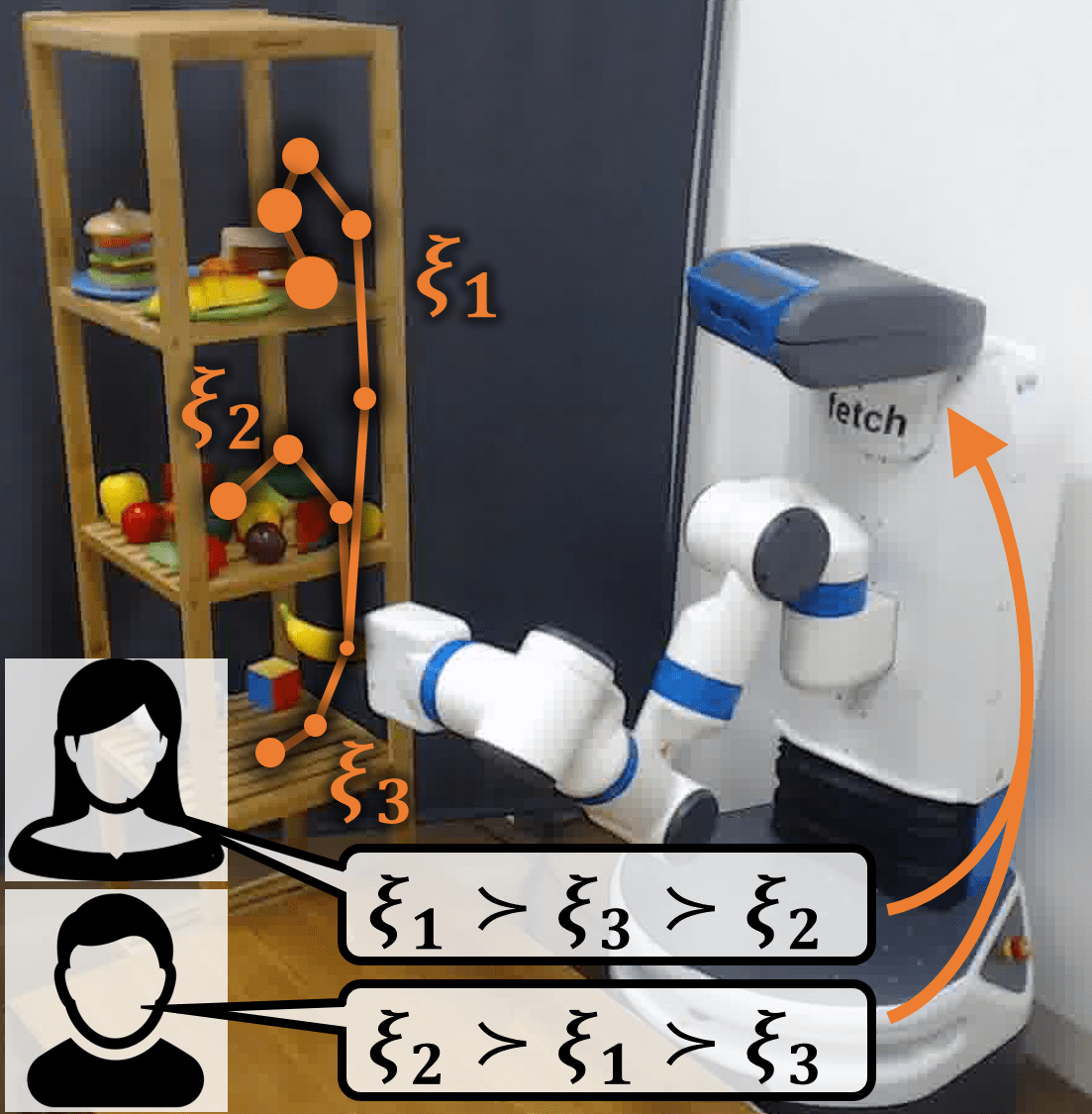}
		\caption{Fetch robot putting a banana on one of the three shelves. The two users have different preferences, and so they provide different rankings to the robot. The robot needs to be able to model multimodal reward functions for successfully achieving the task.}
		\label{fig:03_05_robot_experiment_visual}
		\end{subfigure}
		\hfill
		\begin{subfigure}[t]{0.59\textwidth}
		\centering
        \includegraphics[width=.99\textwidth]{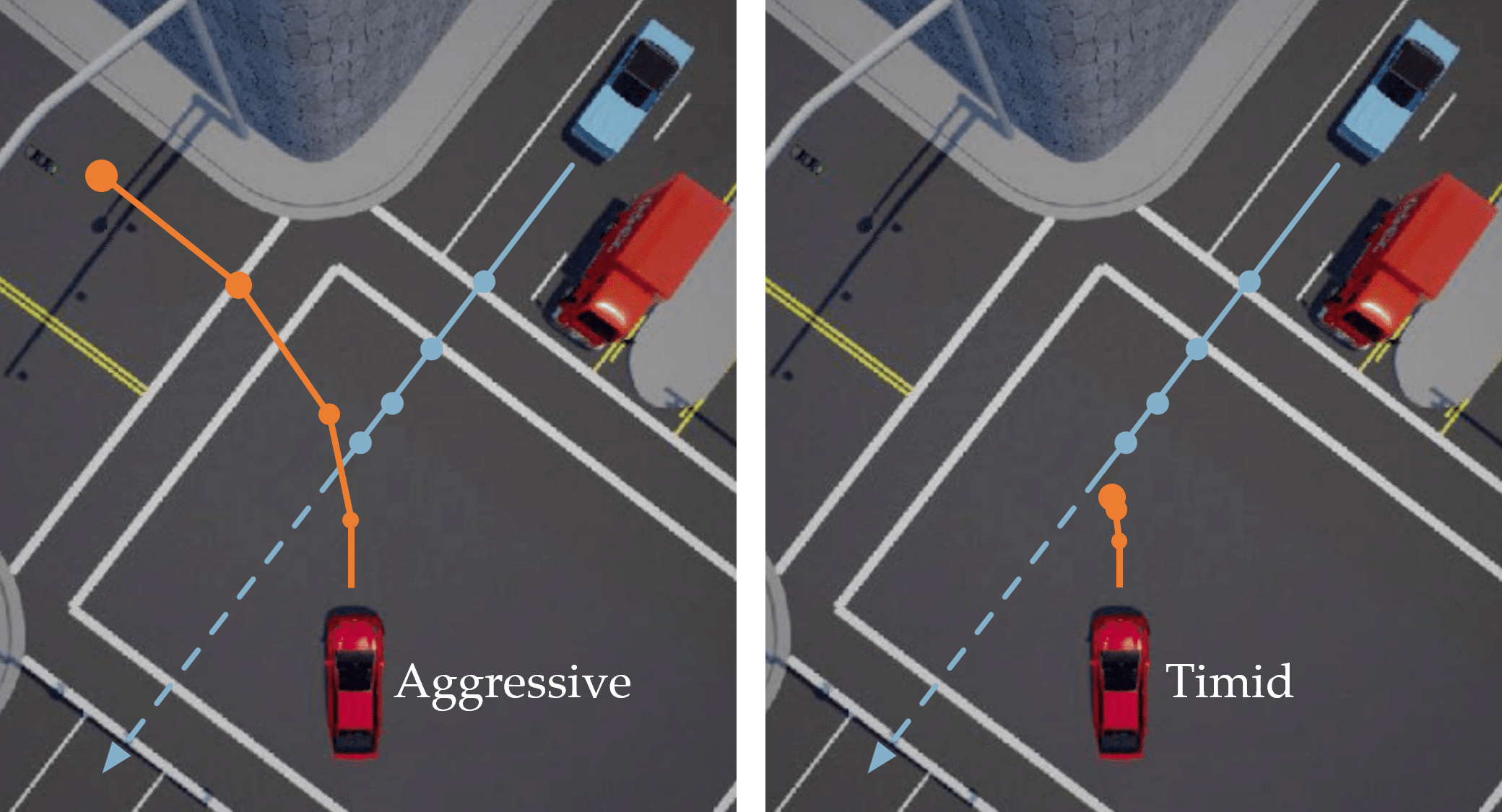}
		\caption{The red car does not observe the blue car due to the occluding truck until it comes to the intersection. It is possible to avoid accident by \textbf{(left)} completing the turn aggressively or \textbf{(right)} making a hard-brake. An autonomous vehicle trying to learn from such mixture data must be able to model multimodal reward functions for safe and efficient driving.}
		\label{fig:03_05_driving_example}
		\end{subfigure}
		\caption{Examples of why multimodal reward functions might be needed.}
\end{figure}

As an example, consider a robot placing a banana on one of the three shelves (see Figure~\ref{fig:03_05_robot_experiment_visual}). The middle shelf is often used for fruits, but it has no room left and if the robot tries to put the banana there, it may cause other fruits to fall. The top shelf has some space but it has been used for cooked meals. The bottom shelf has a lot of free space, but is usually used only for toys. In such a scenario, people may have very different preferences about what the robot should do. If we try to learn a unimodal reward using data collected from multiple people, the robot is likely to fail in the task, because the data will include inconsistent preferences.

As another example, consider the driving scenario presented in \cite{cao2020reinforcement} (see Figure~\ref{fig:03_05_driving_example}), where the red car attempts to make an unprotected left turn, but fails to observe the blue car occluded behind a truck. An aggressive driver might accelerate and avoid the accident by completing the turn before the blue car reaches the collision point. Similarly, a timid driver would move slowly and brake sharply the moment it sees the blue car, which also prevents the accident. Even though both modes can avoid the accident, a driving policy learned by a mixture of this data is likely to fail while trying to comply with both modes. In fact, \citet{cao2020reinforcement} demonstrated simply applying imitation learning using such a mixture data fails in this case, and a separation of different modes is needed.

One solution is of course to label the different modes in the data. For example, one could separate the data based on the preferred shelf in the banana placing example, and learn different reward functions for each shelf. However, this separation is not always straightforward. For example in the driving example, it is unclear what should be labeled as aggressive or timid driving. Clustering the data based on the human who provided the data is also not viable as it will introduce data-inefficiency issues, and perhaps more importantly, humans are not always unimodal: a usually timid driver can drive more aggressively when in a hurry.

These examples motivate us to develop methods that can learn \emph{multimodal} reward functions using datasets that are not specifically labeled with the modes. To this end, previous work proposed learning from demonstrations to learn multimodal policies \cite{hausman2018multi,fei2020triple} or reward functions with multiple possible intentions \cite{babes2011apprenticeship,ramponi2020truly,ho2016generative}. However, learning from expert demonstrations is often extremely challenging in robotics as we discussed in earlier sections, e.g., providing demonstrations on a robot with high degrees of freedom is non-trivial \cite{akgun2012keyframe}, and humans have difficulty giving demonstrations that align with their preferences due to their cognitive biases \cite{basu2017you,kwon2020when}. Thus, it is desirable to have methods that learn from other more reliable sources of data, e.g., pairwise comparisons of trajectories \cite{brown2019extrapolating,basu2018learning}.

While learning from pairwise comparisons provides a rich source of data for learning reward functions, the theoretical results by \citet{zhao2016learning} imply that extending the pairwise-comparison-based reward learning techniques to multimodal reward functions is not possible, i.e. failure cases can be constructed, where pairwise comparisons are not sufficient for identifying different modes of a multimodal function. Our insight is that it is possible to learn a multimodal reward function by going beyond pairwise comparisons and instead using \emph{rankings}.

We want to emphasize this is a different problem from learning nonlinear rewards. Nonlinear reward functions allow users to have multiple sets of desired behavior: a user may prefer both aggressive and timid driving over a behavior that is in between which can cause an accident. However, the existing methods that handle nonlinearities assume unimodal human feedback, which means the user must have a consistent preference towards either of the modes. Therefore, either (\emph{aggressive} $\succ$ \emph{timid} $\succ$ \emph{accident}) or (\emph{timid} $\succ$ \emph{aggressive} $\succ$ \emph{accident}) is expected. In this work, we relax this assumption and learn multimodal reward functions without requiring consistent rankings in the dataset. Our framework allows both (\emph{aggressive} $\succ$ \emph{timid} $\succ$ \emph{accident}) and (\emph{timid} $\succ$ \emph{aggressive} $\succ$ \emph{accident}) to be in the dataset, and we recover both modes of the reward function.

To achieve this, we formulate multimodal reward learning as a mixture learning problem in this section, and use rankings from humans to learn the mixture.

\subsubsection{Computational Models for Rankings}

Before we move into formulation, we briefly review computational models for human rankings. In the previous sections, we introduced different examples of such models for best-of-many choice queries (Equation~\eqref{eq:03_01_noisily_optimal}), pairwise comparisons (Equations~\eqref{eq:03_01_noisily_optimal} and \eqref{eq:03_02_human_model}), and scale queries (Equation~\eqref{eq:03_04_noisefree_model_saturated}). In fact, our best-of-many choice model is known as the multinomial logits (MNL) model \cite{chen2018nearly}, and it has been widely used for human preferences in many fields \cite{ben2018discrete,wilson2019ten,biyik2021incentivizing,beliaev2021incentivizing} including robotics \cite{sadigh2017active,biyik2018batch,basu2018learning}. Similarly, the model is known as the Bradley-Terry model for the special case of pairwise comparisons \cite{chen2013pairwise}.

To extend these models to rankings, Plackett-Luce \cite{maystre2015fast,archambeau2012plackett} and Mallows models \cite{lu2011learning,vitelli2018probabilistic,lu2014effective,busa2014preference} are commonly employed. In this section, we use the Plackett-Luce model as it is a natural extension of MNL, which we already used in Section~\ref{sec:03_01_pairwise_comparisons}. We formalize this model in Section~\ref{subsec:03_05_formulation}.

\subsubsection{Learning Mixture Models from Rankings.}
Our approach to learning multimodal reward functions is through mixture models, where we assume the data come from different individual models with some unknown probabilities. Relatedly, previous works considered mixtures of MNLs \cite{de2010bayesian,chierichetti2018learning}, Plackett-Luce models \cite{zhao2016learning}, and Mallows models \cite{liu2018efficiently}. Other works adopt different methods to model multimodality, such as by assuming latent state dynamics that transition between different modes \cite{morton2017simultaneous} or by learning the different modes from labeled datasets \cite{cao2020reinforcement,qureshi2019composing}. To avoid these modeling assumptions, we focus on directly learning the mixture model.

While \citet{zhao2016learning} have theoretically studied the mixture of Plackett-Luce choice models, which also informs our algorithm in terms of the query sizes, they only focus on learning the rewards of a discrete set of items. In this section, we deal with a continuous hypothesis space under a mixture of Plackett-Luce models.

\subsection{Formulation} 
\label{subsec:03_05_formulation}

\noindent\textbf{Setup.} We again consider a fully-observable dynamical system. A trajectory $\trajectory$ in this system is a series of states and actions, i.e., $\trajectory=(\state_0,\action_0,\ldots, \state_\horizon, \action_\horizon)$. The set of feasible trajectories is $\trajectorySpace$.

We assume there is a set of $\numberOfModes$ individual reward functions that are possibly different, each of which encodes some preference between the trajectories in $\trajectorySpace$. For the rest of the formalism, we refer to each individual reward function as an \emph{expert} for the clarity of the presentation.

Following the common assumption in reward learning \cite{ziebart2008maximum,biyik2022learning,wilde2020active}, we assume each preference can be modeled as a parametric reward function over a known fixed feature space, so the reward associated with a trajectory $\trajectory$ with respect to the $\modeIndex^{\textrm{th}}$ expert is $\trajectoryRewardFunction_{\weights_{\modeIndex}}(\trajectory) = \gpRewardFunction_{\weights_\modeIndex}( \trajectoryFeaturesFunction(\trajectory))$, where $\weights_\modeIndex$ is the unknown vector of parameters. Across the expert population, there exists some unknown distribution, corresponding to the ratio of the data provided by the experts. We represent this distribution with mixing coefficients $\mixingCoefficient_\modeIndex$ such that $\sum_{\modeIndex=1}^{\numberOfModes} \mixingCoefficient_\modeIndex=1$.
We will then learn both the unknown reward functions $\{\weights_\modeIndex\}_{\modeIndex=1}^\numberOfModes$ and the mixing coefficients $\{\mixingCoefficient_\modeIndex\}_{\modeIndex=1}^\numberOfModes$, using ranking queries made to the $\numberOfModes$ experts.

\noindent\textbf{Ranking Model.}
We define a \emph{ranking query} to be a set of the form $\query = \{\trajectory_1,\ldots,\trajectory_{\abs{\query}}\}$ for a fixed query size $\abs{\query}$. The response to a ranking query is a ranking over the items contained therein, of the form $\queryResponse = (\trajectory_{\rankingIndices_1},\ldots, \trajectory_{\rankingIndices_{\abs{\query}}})$, where $\rankingIndices_1$ is the index of the expert's top choice, $\rankingIndices_2$ is the second top choice, and so on. While it is not known which expert provided the response to the query, we know the prior that a response comes from expert $\modeIndex$ with some unknown probability $\mixingCoefficient_\modeIndex$, i.e., $P(\trajectoryRewardFunction=\trajectoryRewardFunction_{\weights_{\modeIndex}})=\mixingCoefficient_\modeIndex$. Going back to our banana placing example, a ranking query of $\abs{\query}$ robot trajectories is generated by the algorithm, and a user---whose identity is unknown to the algorithm---responds to this query.

We then capture how human experts respond to these ranking queries by modeling a ranking distribution through an iterative process using Luce's choice axiom \cite{luce2012individual}. In this process, the experts repeatedly select their top choice $\rankingIndices_1$ with a probability distribution generated with the softmax rule to generate a ranking from the order items were selected:
\begin{align*}
    P\left(\queryResponse_1=\trajectory_{\rankingIndices_1}\mid \trajectoryRewardFunction=\trajectoryRewardFunction_{\weights_{\modeIndex}}, \weights_\modeIndex\right)
     &= \frac
    {\exp\left(\comparisonRationalityCoefficient\trajectoryRewardFunction_{\weights_{\modeIndex}}(\trajectory_{\rankingIndices_1})\right)}
    {\sum_{j=1}^{\abs{\query}}  \exp\left(\comparisonRationalityCoefficient\trajectoryRewardFunction_{\weights_{\modeIndex}}(\trajectory_{\rankingIndices_j})\right)}\:.
\end{align*}
where $\comparisonRationalityCoefficient$ is the rationality coefficient for comparative feedback. In the following iterations, the experts select their top choice among the remaining trajectories:
\begin{align}
    P\left(\queryResponse_j=\trajectory_{\rankingIndices_j}\mid \queryResponse_1,\ldots, \queryResponse_{j-1},\trajectoryRewardFunction=\trajectoryRewardFunction_{\weights_{\modeIndex}},\weights_\modeIndex\right)
     &= \frac
    {\exp\left(\comparisonRationalityCoefficient\trajectoryRewardFunction_{\weights_{\modeIndex}}(\trajectory_{\rankingIndices_j})\right)}
    {\sum_{j'=j}^{\abs{\query}}  \exp\left(\comparisonRationalityCoefficient\trajectoryRewardFunction_{\weights_{\modeIndex}}(\trajectory_{\rankingIndices_{j'}})\right)}\:.
    \label{eq:03_05_conditional}
\end{align}
This is known as the Plackett-Luce ranking model \cite{maystre2015fast,archambeau2012plackett}. Together with the prior over experts $\mixingCoefficient_\modeIndex$, the resulting distribution over rankings $\queryResponse$ is a mixture of {Plackett-Luce} models with mixing coefficients $\mixingCoefficient_\modeIndex$ and weights proportional to $\exp{\left(\trajectoryRewardFunction_{\weights_{\modeIndex}}(\trajectory)\right)}$.

Hence, the ranking distribution first selects the reward function $\trajectoryRewardFunction_{\weights_{\modeIndex}}$ with probability $\mixingCoefficient_\modeIndex$, and then selects trajectories from $\query$ sequentially with probability proportional to the exponent of their reward, i.e., $\exp{\left(\trajectoryRewardFunction_{\weights_{\modeIndex}}\right)}$, among the remaining trajectories until none is left, generating a ranking of the trajectories.

So given knowledge of the true reward function weights $\weights_\modeIndex$ and mixing coefficients $\mixingCoefficient_\modeIndex$, we have the following joint mass over responses $\queryResponse$ from a query $\query$:
\begin{align}
    P(\queryResponse\mid \query, \weights, \mixingCoefficient) = \sum_{\modeIndex=1}^{\numberOfModes}\mixingCoefficient_\modeIndex\prod_{j=1}^{\abs{\query}} \frac{\exp{\left(\comparisonRationalityCoefficient\trajectoryRewardFunction_{\weights_{\modeIndex}}(\trajectory_{\rankingIndices_j})\right)}}
    {\sum_{j'=j}^{\abs{\query}}  \exp{\left(\comparisonRationalityCoefficient\trajectoryRewardFunction_{\weights_{\modeIndex}}(\trajectory_{\rankingIndices_{j'}})\right)}}\:.
    \label{eq:03_05_pmf}
\end{align}

\noindent\textbf{Objective.} Our goal is to design a series of adaptive queries $\query^{(i)}$ to optimally learn the reward function parameters $\weights_\modeIndex$ and corresponding mixing coefficients $\mixingCoefficient_\modeIndex$ upon observing the query responses $\queryResponse^{(i)}$. We constrain all queries to consist of a fixed number of trajectories $\abs{\query}$.

In the subsequent section, we present our learning framework. We defer the active query generation algorithm that makes the queries adaptive to Section~\ref{sec:04_06_rankings} along with simulation and experiment results.

\subsection{Our Approach}
\label{subsec:03_05_approach}

To learn the reward parameters $\weights_\modeIndex$ and mixing coefficients $\mixingCoefficient_\modeIndex$, we again adopt a Bayesian learning approach. For this, we maintain a posterior over the parameters $\weights_\modeIndex$ and $\mixingCoefficient_\modeIndex$. Given a dataset of rankings $\rankingDataset$, this posterior can be written as
\begin{align}
    P(\weights, \mixingCoefficient \mid \rankingDataset) &= P(\weights, \mixingCoefficient \mid \query^{(1)}, \queryResponse^{(1)}, \dots, \query^{(\abs{\rankingDataset})}, \queryResponse^{(\abs{\rankingDataset})})\nonumber\\ &\propto P(\weights, \mixingCoefficient)P(\query^{(1)}, \queryResponse^{(1)}, \dots, \query^{(\abs{\rankingDataset})}, \queryResponse^{(\abs{\rankingDataset})} \mid \weights, \mixingCoefficient)\nonumber\\
    &=P(\weights, \mixingCoefficient)\prod_{i=1}^{\abs{\rankingDataset}}P(\queryResponse^{(i)}, \query^{(i)} \mid \weights, \mixingCoefficient, \query^{(1)}, \queryResponse^{(1)}, \dots, \query^{(i-1)}, \queryResponse^{(i-1)}) \nonumber\\
    &\propto P(\weights,\mixingCoefficient)\prod_{i=1}^{\abs{\rankingDataset}}P(\queryResponse^{(i)} \mid \weights, \mixingCoefficient, \query^{(i)})\:,
    \label{eq:03_05_bayesian_learning}
\end{align}
where we use the conditional independence of rankings $\queryResponse^{(i)}$ given $\weights, \mixingCoefficient$ and the conditional independence of the $\query^{(i)}$ on $\weights, \mixingCoefficient$ given $\query^{(1)}, \queryResponse^{(1)}, \dots, \query^{(i-1)}, \queryResponse^{(i-1)}$ in the last equation. To be able to compute this posterior, we assume some prior distribution over the reward parameters and the mixing coefficients, which is system-dependent and may come from domain knowledge, and use Equation~\eqref{eq:03_05_pmf} to calculate the likelihood terms. For example, in our simulations and user studies in Section~\ref{sec:04_06_rankings}, we adopted a Gaussian prior $\weights_\modeIndex\sim\mathcal{N}(0,I)$ and a uniform prior over the unit $\numberOfModes-1$ simplex for $\mixingCoefficient$. Learning this posterior distribution in Equation~\eqref{eq:03_05_bayesian_learning}, one can compute a maximum likelihood estimate (MLE) or expectation as the predicted reward parameters and mixing coefficients.

Having presented the learning framework for multimodal rewards from rankings, we are now ready to proceed to the last section of Chapter~\ref{chap:learning}, where we will relax the stationarity assumption of the reward function, assume a specific mode transition model, and develop a new query type (namely, hierarchical choice queries) for learning these nonstationary rewards. 

%% file: 03_learning/06_hierarchical.tex
\label{sec:03_06_hierarchical}

In the previous sections, we assumed a stationary reward function, which is not expressive enough to match human preferences in all environments. Real world is often non-stationary due to environment complexity or changes in objectives in the environment. Surrounding agents continuously change their behavior which in turn requires the robot to adapt to these changes. For example in driving, people continuously adapt their reward functions in response to traffic complexity and behavior of other drivers. It is quite common for us to get impatient behind a slow driver and make drastic maneuvers different from our usual driving style. Here, we may weigh efficiency more than collision avoidance than we usually do.

As an important class of non-stationary environments, human-robot and robot-robot adaptation have recently attracted much attention, where the aim is to ensure robots adapt to their changing environments and other agents \cite{nikolaidis2017human,al2017continuous}. In contrast, our goal in this section is to learn the reward functions that dynamically change depending on the interactions between the agents and the environment. We augment learning from comparative feedback to recover such multimodal reward functions. 

Prior works theoretically investigated how to perform preference-based learning for multimodal reward functions \cite{zhao2016learning,liu2018efficiently,chierichetti2018learning}, which we also discussed in Section~\ref{sec:03_05_rankings}. In this section, we relax this problem by assuming there is a structure between the modes, i.e., the mode from which the next comparison data will come can be estimated based on the current comparative feedback. Although we assume and explicitly model these transitions between modes, the problem is not necessarily easier, because we are also interested in learning how users' preferences transition between the modes.

Modeling behaviors in such environments is a well-studied problem especially for driving. For example, \citet{dong2016characterizing} characterize driving styles based on sensor data using deep learning. In a more related paper, \citet{morton2017simultaneous} modeled the drivers with a latent state space which can affect their driving behavior. While they stated these latent states might change over time, both of these works made the assumption that latent states remain unchanged over the trajectories of interest, so they did not address changing behaviors. \citet{berndt2008continuous} modeled the latent states of the drivers using Hidden Markov Models (HMM) where they also allow adaptation. However, they did not specifically learn reward functions, and they focused on identifying the maneuvers the drivers will perform from a predefined database. With a similar objective, \citet{kulic2007affective} used HMM for latent state estimation for human-robot interaction. We recently studied how trust of humans in a robot, a specific latent state, changes based on the performance in a task \cite{brockbank2022how}.

In this section, we propose to learn an expressive representation of preferences in non-stationary scenarios, where interactions and adaptations better reflect the real-world conditions. We assume that the non-stationary scenarios arise from changing behaviors of other agents interacting with our system, which in turn affect human preferences. We formalize the reward dynamics which encodes not only different human preferences but also \emph{how} they change.
\begin{quote}
  \emph{Our insight is that a dynamic reward model may match human preferences more accurately in a wide range of scenarios than a stationary reward function.} 
\end{quote}

We tailor best-of-many choice queries to capture longer term interactions between the robot and the surrounding agents, and develop a probabilistic model of user responses for any number of stationary reward functions and the transitions between them.

In this section, we make the following contributions:

\noindent\textbf{Reward dynamics.} User preferences may change based on the behaviors of other agents in the environment. We encode the momentary human preference by a static reward function and assume at any point of time the human has an internal \emph{preference mode} (mood) which dictates what reward function the human will optimize next. We introduce the notion of \emph{reward dynamics} as a tuple of reward functions and parameters governing transitions between those.

\noindent\textbf{Hierarchical choice queries.} We formalize the \emph{hierarchical choice queries} as a sequence of best-of-many choice queries, each of which we call a \emph{sub-query}.
The sub-queries sequentially follow each other so that the user moods are reflected into their choices.

\subsection{Formulation}
Let us consider a robot that should match human preferences in an environment of interest that includes other agents (e.g. driving scenarios). These other agents can act differently at different times. For example, in the case of driving, some cars aggressively swerve through the traffic and others may follow a more cooperative strategy of allowing other cars to merge smoothly. Prior works on autonomous driving \cite{sadigh2016planning,sadigh2016information,sadigh2018planning,sadigh2019verifying,bai2015intention,sezer2015towards,fisac2018hierarchical,stefansson2019human} assume the robot should follow the same reward function over time in both of the above scenarios. We argue user preferences may vary in response to the changing behaviors of the environment agents in both driving and potentially other multi-agent environments. Our goal is to learn an expressive reward function corresponding to these dynamic preferences.   

We model the environment as a fully-observable dynamical system. For driving, the continuous state of the system $\state \in \stateSpace$ includes the positions and the velocities of the robot and the other agents. The state of the system changes based on actions of all the agents through the transition function $\transitionFunction$, which we can now write as
\begin{align}
\state_{\timestep+1} \sim \transitionFunction(\cdot \mid \state_\timestep, {\action_{\textrm{robot}}}_\timestep, {\action_{\textrm{others}}}_\timestep)
\end{align}
where ${\action_{\textrm{others}}}_\timestep$ are the actions of the other agents at time step $\timestep$, which affect the reward function and in turn the actions of the robot.\footnote{More generally, the other agents in the environment can be thought of as part of the environment. All information regarding them, including their trajectories, can be encoded into the states. This will reduce the setup to the setup we used in earlier sections, e.g., Section~\ref{sec:03_01_pairwise_comparisons}} We define a finite trajectory $\trajectory \in \trajectorySpace$ as a sequence of continuous state-action pairs $\trajectory = (\state_0, {\action_{\textrm{robot}}}_0, {\action_{\textrm{others}}}_0, \dots, \state_{\horizon}, {\action_{\textrm{robot}}}_\horizon, {\action_{\textrm{others}}}_\horizon)$ over a finite horizon $\horizon$, and $\trajectorySpace$ is the set of all feasible trajectories that satisfy the dynamics of the system.
\begin{figure*}[t]
    \centering
    \includegraphics[width=\textwidth]{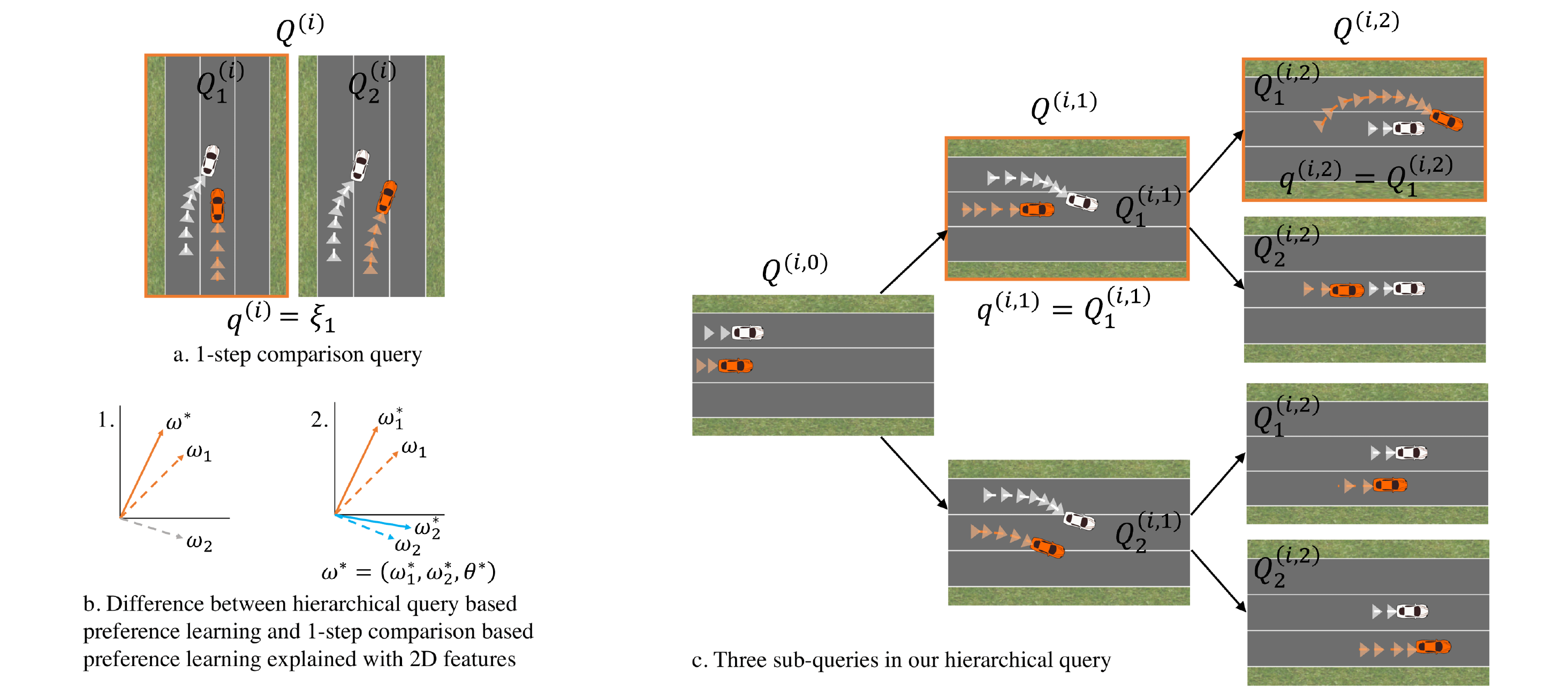}
    \caption{(a) 1-step comparison query. In any two iterations a user with bimodal preference may pick the trajectories optimal with respect to two different true weights $\weights_{1}^*$ and $\weights_{2}^*$. (b) This ambiguity shows up as noise in 1-step comparison based learning where the goal is to learn a single reward function $\weights^*$ (on the left). In reality the true preference function of the user changes between $\weights_{1}^*$ and $\weights_{2}^*$ depending on the environment, $\utilityWeights^*$ governs the transition. Our algorithm learns such a bimodal preference: $\weights_1$ close to $\weights_{1}^*$ and a $\weights_2$ close to $\weights_{2}^*$ (on the right). (c) Our proposed hierarchical query consists of 3 sub-queries. In iteration $i$ of querying, $\query^{(i,0)}$ is a context sub-query, $\query^{(i,1)}$ is a comparison between two trajectories, each a different continuation of $\query^{(i,0)}$, and $\query^{(i,2)}$ continues the preferred trajectory from $\query^{(i,1)}$.}
    \label{fig:03_06_query_comparison}
\end{figure*}

Our goal is to learn human preferences for how the robot should behave in the presence of different environment agents. We learn this reward function by making hierarchical comparison queries to the users.

\subsection{Hierarchical Comparison Queries}
Prior works learned static reward functions by asking people to compare between two different trajectories of robots. There, each query is a pair of short videos that demonstrate two trajectories of the system \cite{sadigh2017active,christiano2017deep}. Such short trajectories do not capture the nuances of interaction in a non-stationary multi-agent system. As an example, in a 1-step comparison query in Figure~\ref{fig:03_06_query_comparison}a, an environment agent (white car) aggressively merged in front of the ego agent (orange car). One option for the ego agent is to slow down (optimize a cooperative reward function). This sudden slow down may have frustrated the user, causing a mode change. So in a similar situation later (in another query) the user prefers a trajectory optimal with respect to a competitive reward function and prevents the other agents in the environment from merging in front. This change in preference manifests as noise in 1-step preference-based learning approaches (see Figure~\ref{fig:03_06_query_comparison}b). However, we would like to learn a composite reward function that not only captures both of these preferences but also \emph{how} they have changed in such a non-stationary environment.

To do so, we allow the users to change their preferences within the same query. We present each query $\query^{(i)}$ as a sequence of several sub-queries. Each sub-query $\query^{(i,j)}$ in the sequence is a continuation from the final state of the preferred trajectory of the previous sub-query $\query^{(i,j-1)}$. This allows us to learn how the behavior of other interacting agents in one sub-query affects user preference in the next sub-query. We assume that the users' next immediate \emph{preference mode} depends only on their current experience. We, therefore, reset their preference mode at the beginning of each query with a demonstration, which we denote as $\query^{(i,0)}$. After $\query^{(i,0)}$, each sub-query is a best-of-many choice between trajectories from $\Xi$: $\query_1^{(i,1)}$, $\query_2^{(i,1)}$, $\dots$, $\query_{\abs{\query^{(i,1)}}}^{(i,1)}$ are all continuations of $\query^{(i,0)}$, where $\abs{\query^{(i,1)}}$ denotes the number of trajectories (options) in sub-query $\query^{(i,1)}$. In general, for the rest of the sub-queries, $\query_{j'}^{(i,j)}$ for $j'=1,2,\dots,\abs{\query^{(i,j)}}$ are continuations from $\queryResponse^{(i,j-1)}$, which is the answer for the $(j-1)^{\textrm{th}}$ sub-query, as shown in Figure~\ref{fig:03_06_query_comparison}c.

\subsection{Reward Dynamics Model}
\label{subsec:03_06_reward_dynamics}
\subsubsection{Preliminaries}
Throughout this section, we will use $[n]$ to denote the integer set $\{1,2,\dots,n\}$ for $n\in\mathbb{Z}_{>0}$.

We denote the $j^\textrm{th}$ sub-query in query $i$ as $\query^{(i,j)}$. $\abs{\query^{(i)}}$ denotes the number of sub-queries in query $i$ (as opposed to number of options in a question as in other sections where queries were not hierarchical), and $\abs{\query^{(i,j)}}$ denotes the number of options in the sub-query.

We assume there is a finite set of modes $[\numberOfModes]$ where $\numberOfModes$ is the number of modes. We also assume the mode of the user is stable during a sub-query. Slightly abusing the notation, we denote the mode in the $j^\textrm{th}$ sub-query of query $i$ as $\numberOfModes(i,j) \in [\numberOfModes]$.

Each sub-query $\query^{(i,j)}$ consists of $\abs{\query^{(i,j)}}$ trajectories: $\query^{(i,j)}_1,\query^{(i,j)}_2,\dots,\query^{(i,j)}_{\abs{\query^{(i,j)}}} \in \trajectorySpace$. When $j=0$, $\abs{\query^{(i,j)}}=1$, as the $0^\textrm{th}$ sub-query is only for setting the initial mode of the user. The user selects one of the trajectories in a sub-query as their response to that sub-query. The user's response to $\query^{(i,j)}$ is $\queryResponse^{(i,j)}\in\query^{(i,j)}$.

In addition, we assume a trajectory features function $\trajectoryFeaturesFunction:\trajectorySpace\to \mathbb{R}^\numberOfFeatures$ that maps every trajectory to a $\numberOfFeatures$-dimensional feature space. This function may depend on both the robot and the other agents in the environment. We assume $\trajectoryFeaturesFunction$ is known. For example, some representative features for driving are distance to the closest environment car, distance to the road boundaries, the speed and the heading angle of the ego vehicle.

\subsubsection{Human Preference Model}
\noindent \textbf{Reward functions under known modes.} We define a user-specific reward function parameterized by the mode of the user, for example, two different reward functions representing calm and rushed driving: $\trajectoryRewardFunction_{\weights_\modeIndex}:\trajectorySpace\to\mathbb{R}$ for $\modeIndex\in [\numberOfModes]$. With a parametric assumption, it is defined as: $\trajectoryRewardFunction_{\weights_\modeIndex}(\trajectory) = \gpRewardFunction_{\weights_{\modeIndex}}\left(\trajectoryFeaturesFunction(\trajectory)\right)$ where $\trajectory\in\trajectorySpace$ and $\weights$ is a user-specific reward parameter matrix, and $\weights_\modeIndex$ is the $\modeIndex^{\textrm{th}}$ column of $\weights$, with each column corresponding to a particular mode for a user. Then, the user response to a sub-query $\query^{(i,j)}$ is probabilistic based on the softmax model \cite{luce2012individual,holladay2016active}, as we also used before:
\begin{align}
	P(\queryResponse^{(i,j)} \mid \query^{(i,j)}, \numberOfModes(i,j)=\modeIndex, \weights) &=  \frac{\exp(\comparisonRationalityCoefficient\trajectoryRewardFunction_{\modeIndex}(\queryResponse^{(i,j)}))}{\sum_{\trajectory \in \query^{(i,j)}} \exp(\comparisonRationalityCoefficient\trajectoryRewardFunction_{\modeIndex}(\trajectory))}
\end{align}
for any $\queryResponse^{(i,j)}\in\query^{(i,j)}$. This models the probability of the human making a choice given a sub-query, the humans' mode during that sub-query, and the user-specific preferences.

\noindent \textbf{Prior on mode transitions.}
We also learn how people change modes. For example, how likely a person is to transition from aggressive driving to defensive driving or vice versa. We assume that a prior $\hierarchicalPriorMatrix\in\mathbb{R}^{\numberOfModes\times \numberOfModes}$ over the mode transitions is given by the designer. The matrix $\hierarchicalPriorMatrix$ alone represents the natural propensity to transition between different modes and is independent of the sub-queries and the current state of the learning algorithm. For example, some mode transitions are naturally more likely than the others: If we have three modes that correspond to defensive, neutral and offensive moods, then it would be more likely for a defensive user to switch to the neutral mode than to the offensive mode. $\hierarchicalPriorMatrix$ captures this prior. Hence, it is constrained to be a proper Markov chain matrix. We note that Markov chains are employed similarly for mood changes by psychiatrists, e.g. \cite{holmes2016applications}. We explain the formulation of the mode transitions next.

\noindent \textbf{Mode transition model.}
The users change their mode based on what they experienced in the previous sub-query and their previous mode. We define a \emph{mode-utility} function to capture this effect of the sub-queries. Specifically, we model the mode transitions as follows: The user has an underlying mode-utility function that quantifies the previous trajectories. If the user thinks they would have higher utility with mode $\modeIndex$, then they transition to $\modeIndex$. As an example, imagine you are driving in a very calm mood. If someone suddenly cuts in front of you, you would think ``if I were aggressive, I could keep a shorter headway with the car in front and the other car would not have been able to cut in front of me", and you also switch to an aggressive mood. It is of course also possible that you keep calm. Hence, the transition should be stochastic.

We model the mode-utility as a function of trajectories: $\modeUtilityFunction_{\utilityWeights_\modeIndex}:\trajectorySpace\to\mathbb{R}$ for $\modeIndex\in[\numberOfModes]$. Again with the assumption that it is a parametric function, it is defined as: $\modeUtilityFunction_{\utilityWeights_\modeIndex}(\trajectory) = \gpModeUtilityFunction_{\utilityWeights_{\modeIndex}}\left(\trajectoryFeaturesFunction(\trajectory)\right)$ where $\utilityWeights$ is another user-specific parameter matrix and $\utilityWeights_\modeIndex$ is the $\modeIndex^{\textrm{th}}$ column of $\utilityWeights$.

The probability of transitioning from any mode $\modeIndex$ in sub-query $\query^{(i,j)}$ to any mode $\modeIndex'$ in the next sub-query $\query^{(i,j+1)}$ is given by multiplying the prior $\hierarchicalPriorMatrix$ with the likelihood computed using the mode-utility function:
\begin{align}
    P_{\modeIndex \modeIndex'}(\query^{(i,j)},\queryResponse^{(i,j)},\utilityWeights) &:= P(\numberOfModes(i,j+1) = \modeIndex' \mid \numberOfModes(i,j)=\modeIndex, \query^{(i,j)}, \queryResponse^{(i,j)}, \utilityWeights) \nonumber \\
	&= \frac1{Z} \frac{\exp(\modeUtilityFunction_{\modeIndex'}(\queryResponse^{(i,j)}))}{\sum_{\modeIndex''\in [\numberOfModes]}\exp(\modeUtilityFunction_{\modeIndex''}(\queryResponse^{(i,j)}))} \hierarchicalPriorMatrix_{{\modeIndex}{\modeIndex'}}
\end{align}
where $Z$ is the normalization constant. In a completely ``neutral case", when the likelihood (softmax) gives equal values for each mode, the transition is solely defined by the prior $\hierarchicalPriorMatrix$. Some examples of $\hierarchicalPriorMatrix$ are:
\begin{itemize}
	\item $\hierarchicalPriorMatrix_{\modeIndex\modeIndex'}=1/\numberOfModes$ for $\forall (\modeIndex,\modeIndex)\in[\numberOfModes]^2$ means that the user may change from any mode to any other mode just based on the previous sub-query with a uniform prior. This is suitable when the modes are categorical, not sequential.
	\item $\hierarchicalPriorMatrix=I$ means the user will not ever change her mode and will remain in her initial mode. Note that the initial mode will also be modeled in a probabilistic way.
	\item If $\hierarchicalPriorMatrix$ is a band matrix, then the user can only change between the modes that are ``close". This is suitable for sequential modes.
\end{itemize}

While our learning model is valid for any feasible $\hierarchicalPriorMatrix$, we will do simplifying assumptions to actively select the hierarchical queries for sample-efficient learning in Section~\ref{sec:04_07_hierarchical}.

\begin{definition}
\emph{Reward dynamics} of a user is a tuple of $(\weights, \utilityWeights)$, which governs both the user preferences and how they transition between modes with the interactions the user is involved in.
\end{definition}
Therefore, our aim is to learn the \emph{reward dynamics} rather than a static unimodal reward function.

\noindent \textbf{Initial Mode.}
We do not know the initial mode $\numberOfModes(i,0)$ of the user, which is the active mode during $\query^{(i,0)}$. One simple way is to assume uniform distribution over all modes. However, imagine $\hierarchicalPriorMatrix$ is such that transitioning to a mode $\modeIndex$ is very unlikely from any mode. Then, the uniform assumption will not hold, because the user is unlikely to be in mode $\modeIndex$. Then a better model is the following:
\begin{align}
	P_\modeIndex := P(\numberOfModes(i,j) = \modeIndex) = \bar{\hierarchicalPriorMatrix}_\modeIndex
\end{align}
where $\bar{\hierarchicalPriorMatrix}_\modeIndex$ denotes the probability of mode $\modeIndex$ in the stationary distribution of the Markov chain $\hierarchicalPriorMatrix$. If there exist several stationary distributions, the designer should pick one of them using domain knowledge.
	
\subsection{Learning Reward Dynamics}\label{subsec:rewards dynamics}
To make the learning of reward dynamics effective and efficient, we should restrict the continuous space of reward dynamics. For that, we make assumptions on the norms of the columns of $\weights$ and $\utilityWeights$ similar to \cite{sadigh2017active,biyik2018batch}, i.e., we assume those norms are not larger than $1$.

There is also the problem of \emph{label switching}. That is, all the probabilities will remain the same if we switch the order of modes both in $\weights$ and $\utilityWeights$. Since this can completely disable the learning, we enforce another constraint on the ordering, as mentioned by \cite{zhao2016learning}, such that $\utilityWeights_{1,1} > \utilityWeights_{2,1} > \dots > \utilityWeights_{\numberOfModes,1}$ where $\utilityWeights_{\modeIndex,1}$ is the first element of $\modeIndex^{\textrm{th}}$ column of $\utilityWeights$.

Our goal is to learn a distribution over the \emph{reward dynamics} by using hierarchical choice queries. We start with a uniform prior over the space of all feasible $(\weights,\utilityWeights)$. After receiving all the responses to a query $\query{(i)}$, ${(\queryResponse^{(i,1)}, \queryResponse^{(i,2)}, \dots, \queryResponse^{(i,\abs{\query^{(i)}})})}$, we perform a Bayesian update:
\begin{align}
P(\weights,\utilityWeights &\mid \queryResponse^{(i,1)}, \queryResponse^{(i,2)}, \dots, \queryResponse^{(i,\abs{\query^{(i)}})}, \query^{(i,1)}, \query^{(i,2)}, \dots, \query^{(i,\abs{\query^{(i)}})}) \nonumber \\
&\propto P(\queryResponse^{(i,1)}, \queryResponse^{(i,2)}, \dots, \queryResponse^{(i,\abs{\query^{(i)}})} \mid \weights,\utilityWeights, \query^{(i,1)}, \query^{(i,2)}, \dots, \query^{(i,\abs{\query^{(i)}})}) P(\weights,\utilityWeights)
\label{eq:03_06_bayesian_update}
\end{align}

Next we derive the expression for the update function, i.e., the first term in the right-hand side of Equation~\eqref{eq:03_06_bayesian_update}, and present some simplifications that we adopted for our implementation.

\subsection{Derivation and Simplifications}
In this section, we present how we compute the update function for the prior $p(\weights,\utilityWeights)$.
We note $\query^{(i,0)}$ does not receive any response. For the simplicity of notation, we let $\queryResponse^{(i,0)}$ denote the only trajectory in $\query^{(i,0)}$, so that $P_{\modeIndex \modeIndex'}(\query^{(i,j)},\queryResponse^{(i,j)},\utilityWeights)$ is well-defined for $\forall(\modeIndex,\modeIndex')\in[\numberOfModes]^2$ when $j=0$. We then derive
\begin{align}
&P(\queryResponse^{(i,1)}, \queryResponse^{(i,2)}, \dots, \queryResponse^{(i,\abs{\query^{(i)}})} \mid \weights,\utilityWeights, \query^{(i,1)}, \query^{(i,2)}, \dots, \query^{(i,\abs{\query^{(i)}})}) \nonumber \\
&=\sum_{(\modeIndex_0,\dots,\modeIndex_{\abs{\query^{(i)}}})\in[\numberOfModes]^{\abs{\query^{(i)}}+1}} P_{\modeIndex_0}P_{\modeIndex_0\modeIndex_1}(\query^{(i,0)},\queryResponse^{(i,0)},\utilityWeights)\dots P_{\modeIndex_{\abs{\query^{(i)}}-1}\modeIndex_{\abs{\query^{(i)}}}}(\query^{(i,\abs{\query^{i}}-1)},\queryResponse_{\abs{\query^{(i)}}\!-1},\utilityWeights) \nonumber \\
&\qquad\prod_{j=1}^{\abs{\query^{i}}} P(\queryResponse^{(i,j)} | \weights, \query^{(i,j)}, \numberOfModes(i,j) = \modeIndex_j)
\end{align}

In our implementation, we restrict ourselves to the cases where $\abs{\query^{(i)}}=3$ for all iterations $i=1,2,\dots$. Then, the above equation is simplified as
\begin{align}
P&(\queryResponse^{(i,1)},\queryResponse^{(i,2)}\mid \weights,\utilityWeights,\query^{(i,0)},\query^{(i,1)},\query^{i,2})\nonumber \\
&= \sum_{\modeIndex_0\in[\numberOfModes]}\sum_{\modeIndex_1\in[\numberOfModes]}\sum_{\modeIndex_2\in[\numberOfModes]}P_{\modeIndex_0}P_{\modeIndex_0\modeIndex_1}(\query^{(i,0)},\queryResponse^{(i,0)},\utilityWeights)P_{\modeIndex_1\modeIndex_2}(\query^{(i,1)},\queryResponse^{(i,1)},\utilityWeights)\nonumber \\
&\qquad P(\queryResponse^{(i,1)} \mid \weights,\query^{(i,1)},\numberOfModes(i,1)=\modeIndex_1)P(\queryResponse^{(i,2)} \mid \weights,\query^{(i,2)},\numberOfModes(i,2)=\modeIndex_2)
\end{align}

To eliminate the normalization $Z$ from the equation, we assume $\hierarchicalPriorMatrix_{\modeIndex\modeIndex'}\in\{0,1/c_{\modeIndex}\}$ for $\forall (\modeIndex,\modeIndex')\in[\numberOfModes]^2$ where $c_{\modeIndex}$ is an appropriate constant. That is, we assume the model designer will just decide on whether or not it is possible to move between any two modes and will not assign specific prior probabilities. Then,
\begin{align}
P_{\modeIndex \modeIndex'}(\query^{(i,0)},\queryResponse^{(i,0)},\utilityWeights) &= \frac{\exp(\modeUtilityFunction_{\utilityWeights_{\modeIndex'}}(\queryResponse^{(i,0)}))}{\sum_{\modeIndex''\in [\numberOfModes]:\hierarchicalPriorMatrix_{\modeIndex \modeIndex''}=1/{c_{\modeIndex}}}\exp(\modeUtilityFunction_{\utilityWeights_{\modeIndex''}}(\queryResponse^{(i,0)}))}
\end{align}

If we further assume $\numberOfModes=2$ and $\hierarchicalPriorMatrix_{\modeIndex \modeIndex'}=1/2$ for $\forall(\modeIndex,\modeIndex')\in[\numberOfModes]^2$, such as the case of cooperative and competitive modes, we also have $P_{\modeIndex}=\frac12$, so we can write:
\begin{align}
&P(\queryResponse^{(i,1)},\queryResponse^{(i,2)} \mid \weights,\utilityWeights,\query^{(i,0)},\query^{(i,1)},\query^{(i,2)})\nonumber \\
&= \sum_{(\modeIndex_1,\modeIndex_2)\in\{1,2\}^2}\prod_{j\in\{1,2\}}\frac{\exp(\trajectoryRewardFunction_{\weights_{\modeIndex_j}}(\queryResponse^{(i,j)}))}{\sum_{\trajectory\in \query^{(i,j)}}\exp(\trajectoryRewardFunction_{\weights_{\modeIndex_j}}(\trajectory))} \frac{\exp(\modeUtilityFunction_{\utilityWeights_{\modeIndex_j}}(\queryResponse^{(i,j-1)}))}{\exp(\modeUtilityFunction_{\utilityWeights_1}(\queryResponse^{(i,j-1)}))+\exp(\modeUtilityFunction_{\utilityWeights_2}(\queryResponse^{(i,j-1)}))}
\label{eq:03_06_simplified_response_model}
\end{align}
This formulation enables us to update $P(\weights,\utilityWeights)$. We will present our active hierarchical choice query selection algorithm that improves data-efficiency in Section~\ref{sec:04_07_hierarchical}.

%% file: 03_learning/07_summary.tex
\label{sec:03_07_summary}

In this chapter, we developed reward learning techniques that use comparative feedback from humans instead of or in addition to expert demonstrations. This problem setup is closely related to inverse reinforcement learning as we discussed in Section~\ref{sec:02_02_irl}. However, it brings an important advantage in practice: we do not need to collect human demonstrations of the task that are (almost) optimal, which is infeasible or extremely challenging in many domains, e.g., regular users of a lower-body exoskeleton (see Section~\ref{sec:03_03_roial}) are people with paralysis (a group with nearly 5.4 million people in the US alone \cite{armour2016prevalence}), who cannot possibly give demonstrations to these systems.

On the other hand, comparative feedback is easy to collect and does not require expertise on controlling the system. In most cases, a human user who has the knowledge of the target task can easily compare two (or more) trajectories of a robot in terms of their task performance. Motivated by this, we developed and investigated various techniques for reward learning using comparative feedback. We studied different query types and functional forms for the reward function (as we outlined in Section~\ref{sec:01_02_contributions}).

An important limitation of comparative feedback, especially when we are working on trajectory level, i.e., human users comparing trajectories rather than individual states or actions, is the fact that each comparison query carries a small amount of information compared to expert demonstrations. Although we showed in Section~\ref{sec:03_01_pairwise_comparisons} that demonstrations can still be used together with comparative feedback, this limitation hurts the practicality and scalability of the methods we developed. Therefore, techniques that actively query the users for the highest information gain are crucial. In the next chapter, we will focus on such approaches that are based on active learning optimizations. We will also present the simulation and experiment results that we deferred in Chapter~\ref{chap:learning}.

%% file: 04_active/00_intro.tex
Even though having humans provide comparative feedback does not suffer from similar problems to collecting demonstrations, each comparison question is much less informative than an expert demonstration. For example, each pairwise comparison query can provide at most $1$ bit of information. A promising approach to tackle this problem is to actively generate the queries for comparative feedback \cite{sadigh2017active,katz2019learning,wilde2020active}.

In Chapter~\ref{chap:learning}, we covered how the robot can update its understanding of the reward function parameters $\weights$ given the human's comparative feedback; but how does the robot choose the right questions in the first place? Active query generation deals with this problem. Unlike demonstrations --- where the robot is passive --- here the robot is \emph{active}, and purposely probes the human to get fine-grained information about specific parts of $\weights$ that are unclear. By actively querying the user, the robot attempts to get as much information as possible, mitigating the data-inefficiency issue of learning from comparative feedback. At the same time, the robot needs to remember that a human is answering these questions, and so the options need to be easy and intuitive for the human to respond to. Proactively choosing intuitive queries is arguably the most challenging part of learning from comparative feedback. Accordingly, we will explore methods for actively generating queries $\query$ in the subsequent sections.

\noindent \textbf{Greedy Robot.} Ideally, the robot must find the best \emph{adaptive sequence} of queries to clarify the human's reward. Unfortunately, reasoning about a sequence of queries is --- in general --- NP-hard \cite{ailon2012active}. We therefore proceed in a \emph{greedy} fashion throughout this chapter: at every iteration $i$, the robot chooses $\query^{(i)}$ while thinking only about the next posterior belief $\belief^{i}$ (e.g., see Equation~\eqref{eq:03_01_DP7}).

In Section~\ref{sec:04_01_volume_removal}, we start with describing the maximum volume removal based active querying method \cite{sadigh2017active}, which has been the dominant approach for active query generation to maximize data-efficiency \cite{palan2019learning}. However, in Section~\ref{sec:04_02_information_gain}, we will show optimizing mutual information is a better approach for data-efficiency, and also helps with generating easier questions for the human users. We will then use this technique to extend the majority of the methods we presented in Chapter~\ref{chap:learning} with active querying in Sections~\ref{sec:04_03_gp} through \ref{sec:04_07_hierarchical}.  Finally, Section~\ref{sec:04_08_batch} will introduce various batch-active querying techniques that enable actively generating queries in batches for all the methods presented until that point. All sections in this chapter will also present simulation and experiment results both for the active querying techniques and the corresponding learning methods from Chapter~\ref{chap:learning}.

%% file: 04_active/01_volume_removal.tex
\label{sec:04_01_volume_removal}

In this section, we introduce an active querying method that is called maximum volume removal. The objective in this method is to maximize the amount of space that will be removed from the hypothesis space of reward function parameters $\weights$, and the overall optimization problem is submodular, allowing this approach to enjoy some theoretical guarantees as shown by \citet{sadigh2017active}.

Although the volume removal objective can be used with any of the query types we introduced in Chapter~\ref{chap:learning}, we will use the DemPref method we introduced in Section~\ref{sec:03_01_pairwise_comparisons} for its simplicity. This will also enable us to show in the next section that active comparative feedback must be collected after the demonstrations are incorporated into the belief distribution via Equation~\eqref{eq:03_01_DP4}. However, we would like to note once again that demonstrations can be used along with any, possibly actively collected, comparative feedback type, as long as the reward function is stationary and unimodal.

\subsection{Maximum Volume Removal Optimization}

Maximizing volume removal is a widely used strategy for selecting queries. The method attempts to generate the most-informative queries by finding the $\query^{(i)}$ that maximizes the expected difference between the prior and \emph{unnormalized} posterior \cite{sadigh2017active, palan2019learning}. Formally, the method generates a query of $\abs{\query^{(i)}} \geq 2$ trajectories at iteration $i$ by solving:
\begin{align}
\argmax_{\query^{(i)}= \{\trajectory_1,\dots,\trajectory_{\abs{\query^{(i)}}}\}} \mathbb{E}_{\queryResponse^{(i)}} \left[\int \left(\belief^{i-1}(\weights) - \belief^{i-1}(\weights)P(\queryResponse^{(i)} \mid \query^{(i)},\weights)\right)d\weights \right]
\end{align}
where the prior is on the left and the unnormalized posterior from Equation~\eqref{eq:03_01_DP7} is on the right. The integration is over the all possible values of $\weights$. This optimization problem can equivalently be written as:
\begin{align} \label{eq:04_01_VR1}
\query^{(i)}_* = \argmax_{\query^{(i)} = \{\trajectory_1,\dots,\trajectory_{\abs{\query^{(i)}}}\}} \mathbb{E}_{\queryResponse^{(i)}} \mathbb{E}_{\weights\sim\belief^{i-1}} \left[1 - P(\queryResponse^{(i)} \mid \query^{(i)},\weights)\right]\:,
\end{align}
or in the special case of pairwise comparison queries, i.e., $\abs{\query^{(i)}}=2$, as we show in Appendix~\ref{app:99_01_ijrr_dpp_worst},
\begin{align} \label{eq:04_01_VR1_worst_case}
\query^{(i)}_* = \argmax_{\query^{(i)} = \{\trajectory_1,\trajectory_2\}} \min_{\queryResponse^{(i)}} \mathbb{E}_{\weights\sim\belief^{i-1}} \left[1 - P(\queryResponse^{(i)} \mid \query^{(i)},\weights)\right]\:.
\end{align}

The distribution $\belief^{i-1}$ can get very complex and thus --- to tractably compute the expectations in Equation~\eqref{eq:04_01_VR1} --- we are forced to leverage sampling. Letting $\weightsSampleSet$ denote a set of samples drawn from the prior $\belief^{i-1}$, and $\asymeq$ denote asymptotic equality as the number of samples $\abs{\weightsSampleSet}\to\infty$, the optimization problem in Equation~\eqref{eq:04_01_VR1} becomes:
\begin{align} \label{eq:04_01_VR2}
\query^{(i)}_* \asymeq \argmin_{\query^{(i)} = \{\trajectory_1,\dots,\trajectory_{\abs{\query^{(i)}}}\}}\sum_{\queryResponse^{(i)}\in \query^{(i)}}\left(\sum_{{\weights}\in\weightsSampleSet}P(\queryResponse^{(i)}\mid \query^{(i)},{\weights})\right)^2
\end{align}
In practice, we can use, for example, Metropolis-Hastings \cite{chib1995understanding} for sampling from the prior belief $\belief^{i-1}$.

\noindent \textbf{Intuition.} When solving Equation~\eqref{eq:04_01_VR2}, the robot looks for queries $\query^{(i)}$ where each answer $\queryResponse^{(i)} \in \query^{(i)}$ is equally likely given the current belief over $\weights$. These questions appear useful because the robot is maximally uncertain about which trajectory the human will prefer.

\noindent \textbf{When Does This Fail?} Although prior works have shown that volume removal can work in practice, we here identify two key shortcomings. First, we point out a failure case: the robot may solve for questions where the answers are equally likely but \emph{uninformative} about the human's reward. Second, the robot does not consider the human's ability to answer when choosing questions --- and this leads to \emph{challenging}, indistinguishable queries that are hard to answer!

\subsubsection{Uninformative Queries}
The optimization problem used to identify maximum volume removal queries fails to capture our original goal of generating informative queries. Consider a trivial query where all options are identical: $\query^{(i)} = \{\trajectory_1, \trajectory_1, \ldots, \trajectory_1\}$. Regardless of which answer $\queryResponse^{(i)}$ the human chooses, here the robot gets no information about the right reward function; put another way, $\belief^i = \belief^{i-1}$. Asking a trivial query is a waste of the human's time --- but we find that this uninformative question is actually a best-case solution to Equation~\eqref{eq:04_01_VR1}.
\begin{theorem}
	The trivial query $\query = \{\trajectory_1, \trajectory_1, \ldots, \trajectory_1\}$ (for any $\trajectory_1\in\trajectorySpace$) is a global solution to Equation~\eqref{eq:04_01_VR1}.
	\begin{proof}
	For a given $\query^{(i)}$ and $\weights$, $\sum_{\queryResponse^{(i)}} P(\queryResponse^{(i)}\mid \query^{(i)},\weights) = 1$. Thus, we can upper bound the objective in Equation~\eqref{eq:04_01_VR1} as follows:
	\begin{align}
		\mathbb{E}_{\queryResponse^{(i)}\mid\query^{(i)},\belief^{i-1}} \mathbb{E}_{\weights\sim\belief^{i-1}} &[1 - P(\queryResponse^{(i)}\mid \query^{(i)},\weights)]\\
		&= 1 - \mathbb{E}_{\weights\sim\belief^{i-1}}\left[\sum_{\queryResponse^{(i)}\in \query^{(i)}}P(\queryResponse^{(i)}\mid \query^{(i)},\weights)^2\right] \leq 1-1/\abs{\query^{(i)}}\:,
	\end{align}
	recalling that $\abs{\query^{(i)}}$ is the total number of options in $\query^{(i)}$. For the trivial query $\query = \{\trajectory_1, \trajectory_1, \ldots, \trajectory_1\}$, the objective in Equation~\eqref{eq:04_01_VR1} has value $\mathbb{E}_{\queryResponse^{(i)}\mid \query^{(i)}, \belief^{i-1}} \mathbb{E}_{\weights\sim\belief^{i-1}} \left[1 - P(\queryResponse^{(i)}\mid \query^{(i)},\weights)\right]\ = 1 - 1/\abs{\query^{(i)}}$. This is equal to the upper bound on the objective, and thus the trivial, uninformative query of identical options is a global solution to Equation~\eqref{eq:04_01_VR1}.
	\end{proof}
\label{thm:04_01_volume_removal_failure}
\end{theorem}

\subsubsection{Challenging Queries}
Volume removal prioritizes questions where each answer is equally likely. Even when the options are not identical (as in a trivial query), the questions may still be very challenging for the user to answer. We explain this issue through a concrete example (also see Figure~\ref{fig:04_01_corl19_domains}):

\begin{figure}[t]
\includegraphics[width=0.8\textwidth]{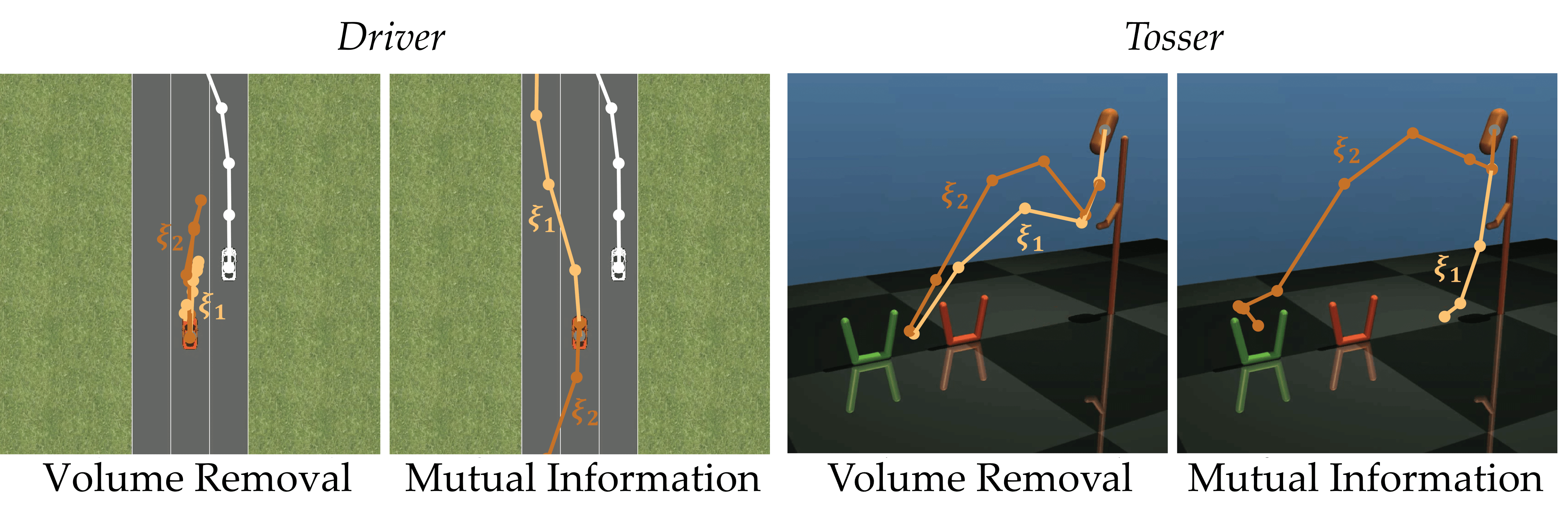}
\centering
\caption{Sample queries generated with the volume removal and information gain methods on \emph{Driver} and \emph{Tosser} tasks. Volume removal generates queries that are difficult, because the options are almost equally good or equally bad.}
\label{fig:04_01_corl19_domains}
\end{figure}

\begin{figure*}[t]
\includegraphics[width=\textwidth]{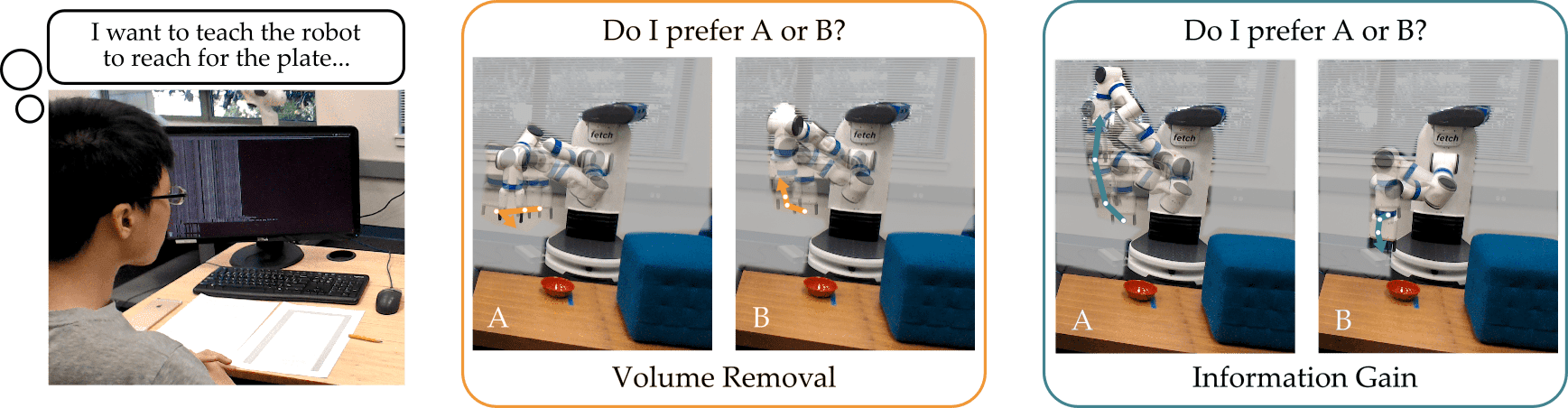}
\centering
\caption{Comparing preference queries that do not account for the human's ability to answer to queries generated using our information gain approach. Here the robot is attempting to learn the user’s reward function, and demonstrates two possible trajectories. The user should select the trajectory that better aligns with their own preferences. While the trajectories produced by the state-of-the-art volume removal method are almost indistinguishable, our information theoretic approach results in questions that are easy to answer, which eventually increase the robot’s overall learning efficiency.}
\label{fig:04_01_frontfig}
\end{figure*}

\begin{example}\label{ex:04_01_example_question}
	\normalfont Let the robot query the human while providing $\abs{\query}=2$ different answer options, i.e., with a pairwise comparison query, $\trajectory_1$ and $\trajectory_2$.
	
	\textit{Question A.} Here the robot asks a question where both options are equally good choices. Consider query $Q_A$ such that $P(q = \trajectory_{A,1} \mid \query_A, \weights) = P(\queryResponse = \trajectory_{A,2} \mid \query_A, \weights) \quad \forall \weights\in\weightsSampleSet$. Responding to $\query_A$ is difficult for the human, since both options $\trajectory_{A,1}$ and $\trajectory_{A,2}$ equally capture their reward function. 
	
	\textit{Question B.} Alternatively, this robot asks a question where only one option matches the human's true reward. Consider a query $\query_B$ such that:
	\begin{align}
	    & P(\queryResponse = \trajectory_{B,1} \mid \query_B, \weights) \approx  1 \quad \forall \weights \in\weightsSampleSet^{(1)} \\
	    & P(\queryResponse = \trajectory_{B,2} \mid \query_B, \weights) \approx  1 \quad \forall \weights \in\weightsSampleSet^{(2)} \\
        & \weightsSampleSet^{(1)} \cup \weightsSampleSet^{(2)} = \weightsSampleSet, \quad \abs{\weightsSampleSet^{(1)}}=\abs{\weightsSampleSet^{(2)}}
	\end{align}
	If the human's weights $\weights$ lie in $\weightsSampleSet^{(1)}$, the human will always answer with $\trajectory_{B,1}$, and --- conversely --- if the true $\weights$ lies in $\weightsSampleSet^{(2)}$, the human will always select $\trajectory_{B,2}$. Intuitively, this query is easy for the human: regardless of what they want, one option stands out when answering the question.
\end{example}	

\noindent \textbf{Incorporating the Human.} Looking at Example~\ref{ex:04_01_example_question}, it seems clear that the robot should ask question $\query_B$. Not only does $\query_A$ fail to provide any information about the human's reward (because their response could be equally well explained by any $\weights$), but it is also hard for the human to answer (since both options seem equally viable). Unfortunately, when maximizing volume removal the robot thinks $\query_A$ is \emph{just as good} as $\query_B$: they are both global solutions to its optimization problem! Here volume removal gets it wrong because it fails to take the human into consideration. Asking questions based only on how uncertain the robot is about the human's answer can naturally lead to confusing, uninformative queries. Figure~\ref{fig:04_01_corl19_domains} demonstrates some of these hard queries generated by the volume removal formulation.

Theorem~\ref{thm:04_01_volume_removal_failure} and Example~\ref{ex:04_01_example_question} make it clear that volume removal is not the true objective we should be optimizing for. It works in practice not despite the local optima, but thanks to them! Therefore, in the next section, we will introduce a new active querying approach that is based on maximizing the mutual information. We will show this approach does not suffer from similar issues. It will further enable us to show demonstrations must be incorporated into the belief \emph{before} actively querying the user for comparative feedback. Before we move into this new objective, we present our experiment results with the volume removal optimization.

\subsection{Experiments} \label{subsec:04_01_experiments}

We conduct two sets of experiments to assess the performance of DemPref with volume removal maximization under various metrics. In all experiments, we assume a reward function that is linear in trajectory features, i.e., $\trajectoryRewardFunction_{\weights}(\trajectory) = \weights^{\top}\trajectoryFeaturesFunction(\trajectory)$ for any trajectory $\trajectory\in\trajectorySpace$ with $\norm{\weights}\leq1$.\footnote{In this section, unless otherwise noted, we adopt $\demonstrationRationalityCoefficient=0.02$, $\comparisonRationalityCoefficient=1$, and assume a uniform prior over reward parameters $\weights$, i.e., $P(\weights)$ is constant for any $\norm{\weights}_2\leq1$. We use Metropolis-Hastings algorithm \cite{chib1995understanding} for sampling the set $\weightsSampleSet$ from belief distribution over $\weights$.} We start by describing the simulation domains and the user study environment. Each subsequent subsection presents a set of experiments and tests the relevant hypotheses.

\begin{figure}[t]
\includegraphics[width=0.7\textwidth]{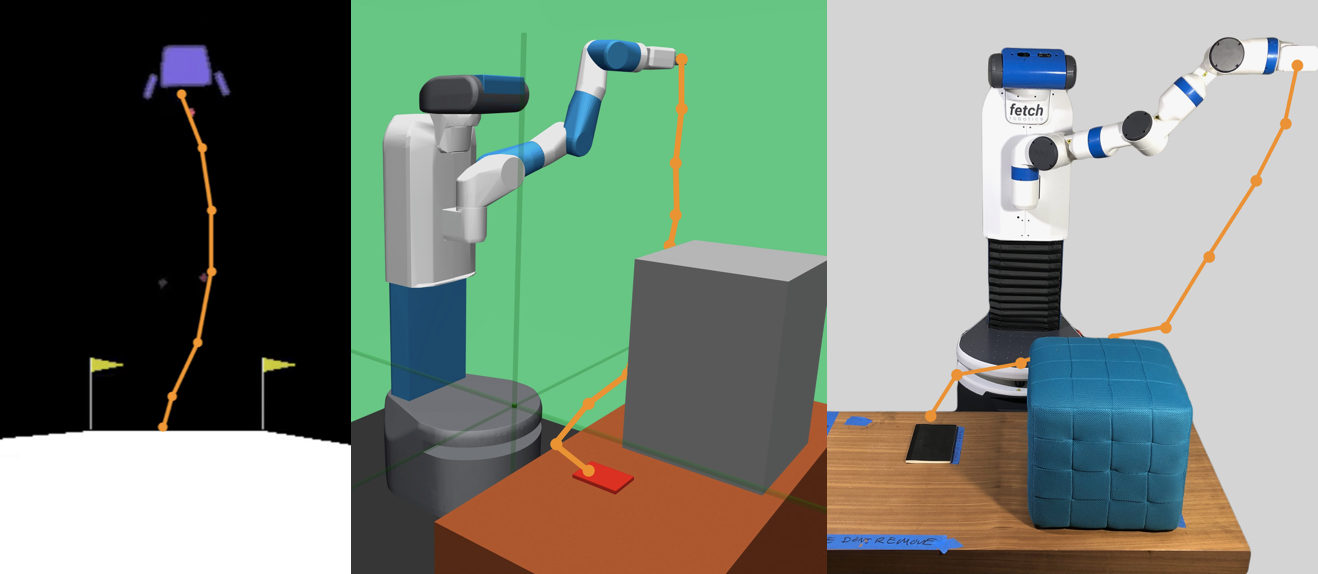}
\centering
\caption{Views from simulation domains, with a demonstration in orange: \textbf{(a)} \emph{LunarLander}, \textbf{(b)} \emph{FetchReach} (simulated), \textbf{(c)} \emph{FetchReach} (physical). }
\label{fig:04_01_rss19_domains}
\end{figure}

\subsubsection{Simulation Domains}
In each experiment, we use a subset of the following domains, shown in Figures~\ref{fig:04_01_corl19_domains} and \ref{fig:04_01_rss19_domains}, as well as a linear dynamical system:

\noindent\emph{LunarLander.} We use the continuous ``LunarLander" environment from OpenAI Gym \cite{brockman2016openai}, where the lander has to safely reach the landing pad. The trajectory features correspond to the lander's average distance from the landing pad, its angle, its velocity, and its final distance to the landing pad. 

\noindent\emph{FetchReach.} Inspired by \cite{bajcsy2017learning}, we use a modification of the ``FetchReach" environment from OpenAI Gym (built on top of MuJoCo), where the robot has to reach a goal with its arm, while keeping its arm as low-down as possible (see Figure~\ref{fig:04_01_rss19_domains}). The trajectory features correspond to the robot gripper's average distance to the goal, its average height from the table, and its average distance to a box obstacle in the domain. See Appendix~\ref{app:99_04_ijrr_features} for the formal feature definitions.

For our user studies, we employ a version of the \emph{FetchReach} environment with the physical Fetch robot (see Figure~\ref{fig:04_01_rss19_domains}) \cite{wise2016fetch}.

\subsubsection{Evaluation Metric}
To judge convergence of the inferred reward function parameters to true parameters in simulations, we adopt the \emph{alignment metric} from \cite{sadigh2017active}: 
\begin{align}
\texttt{Alignment} = \frac{1}{\abs{\weightsSampleSet}}\sum_{\bar{\weights}\in\weightsSampleSet}\frac{\weights^*\cdot\bar{\weights}}{\norm{\weights^*}_2\norm{\bar{\weights}}_2}\:,
\label{eq:04_01_alignment}
\end{align}
where $\weights^*$ is the true reward function parameters.

We are now ready to present our two sets of experiments each of which demonstrates a different aspect of the proposed DemPref framework:
\begin{enumerate}[nosep]
    \item The utility of initializing with demonstrations,
    \item The advantages comparative feedback provide over using only demonstrations,
\end{enumerate}

\subsubsection{Initializing with Demonstrations}
We first investigate whether initializing the learning framework with user demonstrations is helpful. Specifically, we test the following hypotheses:

\noindent\textbf{H1.} \emph{DemPref accelerates learning by initializing the prior belief $\belief^0$ using user demonstrations.}

\noindent\textbf{H2.} \emph{The convergence of DemPref improves with the number of demonstrations used to initialize the algorithm.}

To test these two claims, we perform simulation experiments in \emph{Driver}, \emph{LunarLander} and \emph{FetchReach} environments. For each environment, we simulate a human user with hand-tuned reward function parameters $\weights$, which gives reasonable performance. We generate demonstrations by applying model predictive control (MPC) to solve: $\max_\trajectory \trajectoryRewardFunction_{\weights^*}(\trajectory)$. After initializing the belief with varying number of such demonstrations ($\abs{\demonstrationDataset}\in\{0,1,3\}$), the simulated users in each environment respond to $25$ pairwise comparison queries ($\abs{\query}=2$) according to Equation~\eqref{eq:03_01_noisily_optimal}, each of which is actively synthesized with the volume removal optimization.\footnote{The environments we use are deterministic, i.e., state transitions are not stochastic. Hence, we fix the initial state and simply optimize over the sequence of actions to solve the volume removal maximization problem.} We repeat the same procedure for $8$ times to obtain confidence bounds.

\begin{figure*}[ht]
\includegraphics[width=0.85\textwidth]{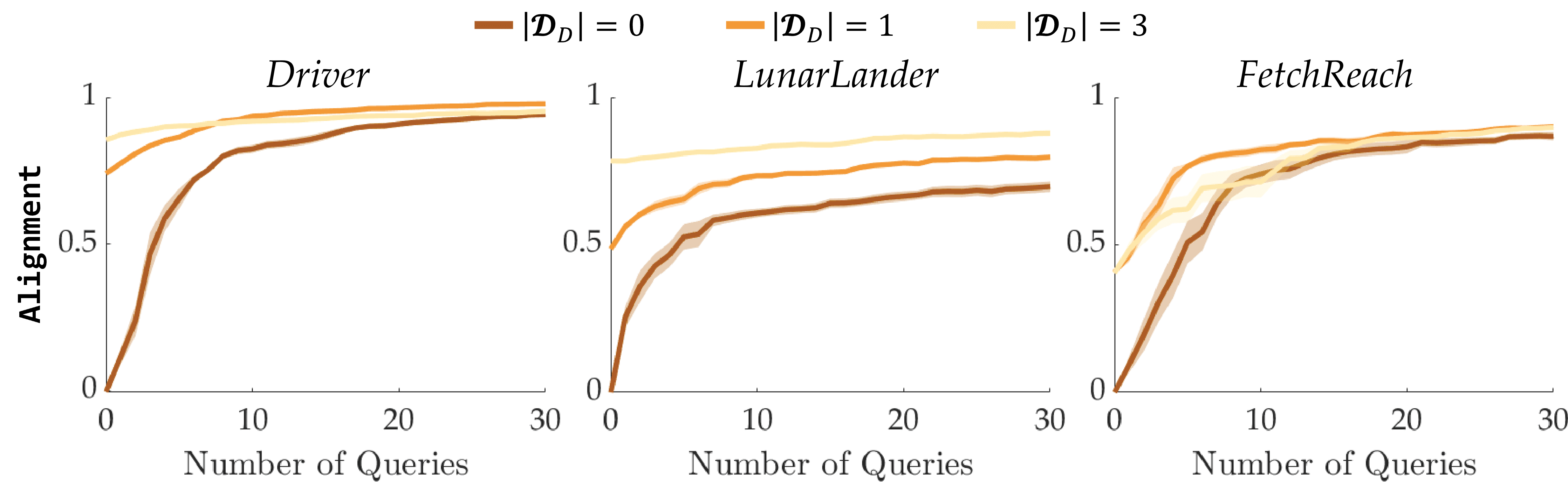}
\centering
\caption{The results of our first experiment, investigating whether initializing with demonstrations improves the learning rate of the algorithm, on three domains. On the \emph{Driver}, \emph{LunarLander}, and \emph{FetchReach} (simulated) environments, initializing with one demonstration improved the rate of convergence significantly.}
\label{fig:04_01_varying_D}
\end{figure*}

The results of the experiment are presented in Figure~\ref{fig:04_01_varying_D}. On all three environments, initializing with demonstrations improves the convergence rate of the preference-based algorithm significantly; to match the \texttt{Alignment} value attained by DemPref with only one demonstration in $10$ pairwise comparison queries, it takes the pure preference-based algorithm, i.e., without any demonstrations, $30$ pairwise comparisons on \emph{Driver}, $35$ on \emph{LunarLander}, and $20$ on \emph{FetchReach}. These results provide strong evidence in favor of \textbf{H1}.

The results regarding \textbf{H2} are more complicated. Initializing with three instead of one demonstration improves convergence significantly only on the \emph{Driver} and \emph{LunarLander} domains. (The improvement on \emph{Driver} is only at the early stages of the algorithm, when fewer than $10$ pairwise comparisons are used.) However, on the \emph{FetchReach} domain, initializing with three instead of one demonstration hurts the performance of the algorithm. (Although, we do note that the results from using three demonstrations are still an improvement over the results from not initializing with demonstrations). This is unsurprising. It is much harder to provide good demonstrations on the \emph{FetchReach} environment than on the \emph{Driver} or \emph{LunarLander} environments, and therefore the demonstrations are of lower quality. Using more demonstrations when they are of lower quality leads to the prior being more concentrated further away from the true reward function, and can cause the the learning algorithm to slow down.

In practice, we find that using a single demonstration to initialize the algorithm leads to reliable improvements in convergence, regardless of the complexity of the domain.

\subsubsection{DemPref vs. IRL}
Next, we analyze if preference elicitation improves learning performance. To do that, we conduct a within-subjects user study where we compare our DemPref algorithm with Bayesian IRL \cite{ramachandran2007bayesian}. The hypotheses we are testing are:

\noindent\textbf{H3.} \emph{The robot which uses the reward function learned by DemPref will be more successful at the task (as evaluated by the users) than the IRL counterpart.}

\noindent\textbf{H4.} \emph{Participants will prefer to use the DemPref framework as opposed to the IRL framework.}

For these evaluations, we use the \emph{FetchReach} domain with the physical Fetch robot. Participants were told that their goal was to get the robot's end-effector as close as possible to the goal, while (1) avoiding collisions with the block obstacle and (2) keeping the robot's end-effector low to the ground (so as to avoid, for example, knocking over objects around it). Participants provided demonstrations via teleoperation (using end-effector control) on a keyboard interface; each user was given some time to familiarize themselves with the teleoperation system before beginning the experiment.

Participants trained the robot using two different systems. (1) IRL: Bayesian IRL with $5$ demonstrations. (2) DemPref: our DemPref framework (with the volume removal optimization) with $1$ demonstration and $15$ proactive pairwise comparison queries\footnote{The number of demonstrations and pairwise comparisons used in each system were chosen such that a simulated agent achieves similar convergence to the true reward on both systems.}. We counter-balanced across which system was used first, to minimize the impact of familiarity bias with our teleoperation system.

\begin{figure*}[t]
\includegraphics[width=\textwidth]{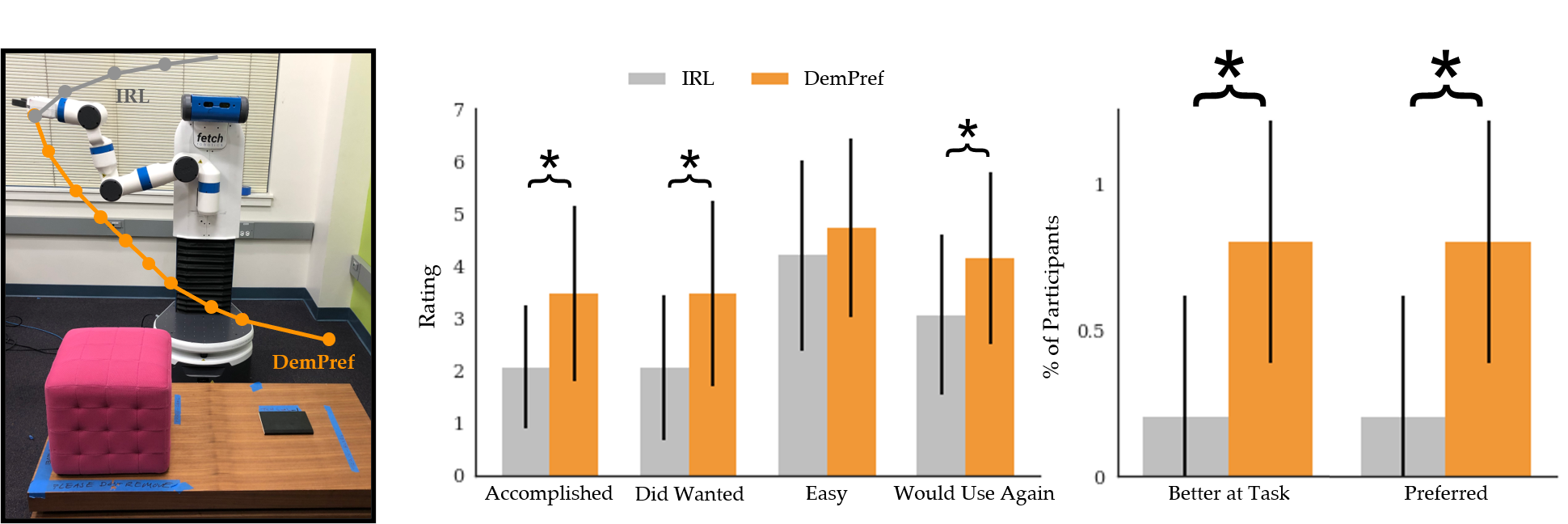}
\centering
\caption{(Left) Our testing domain, with two trajectories generated according to the reward functions learned by IRL and DemPref from a specific user in our study. (Right) The results of our usability study -- the error bars correspond to standard deviation and significant results are marked with an asterisk. We find that users rated the robot trained with DemPref as significantly better at accomplishing the task and preferred to use our method for training the robot significantly more than they did IRL. However, we did not find evidence to suggest that users found our method easier to use than standard IRL.}
\label{fig:04_01_rss_user_study}
\end{figure*}

After learning from human feedback, the robot was trained in simulation using Proximal Policy Optimization (PPO) with the reward function learned from each system \cite{schulman2017proximal}. To ensure that the robot was not simply overfitting to the training domain, we used different variants of the domain for training and testing the robot. We used two different test domains (and counter-balanced across them) to increase the robustness of our results against the specific testing domain. Figure~\ref{fig:04_01_rss_user_study}~(left) illustrates one of our testing domains.
We rolled out three trajectories in the test domains for each algorithm on the physical Fetch robot. After observing each set of trajectories, the users were asked to rate the following statements on a 7-point rating scale:
\begin{enumerate}[nosep]
    \item The robot accomplished the task well. (Accomplished)
    \item The robot did what I wanted. (Did Wanted)
    \item It was easy to train the robot with this system. (Easy)
    \item I would want to use this system to train a robot in the future. (Would Use Again)
\end{enumerate}
They were also asked two comparison questions:
\begin{enumerate}[nosep]
    \item Which robot accomplished the task better? (Better at Task)
    \item Which system would you prefer to use if you had to train a robot to accomplish a similar task? (Preferred)
\end{enumerate}
They were finally asked for general comments.

For this user study, we recruited $15$ participants (11 male, 4 female), six of whom had prior experience in robotics but none of whom had any prior exposure to our system.

We present our results in Figure~\ref{fig:04_01_rss_user_study}~(right). When asked which robot accomplished the task better, users preferred the DemPref system by a significant margin ($p < 0.05$, Wilcoxon paired signed-rank test); similarly, when asked which system they would prefer to use in the future if they had to train the robot, users preferred the DemPref system by a significant margin ($p < 0.05$). This provides strong evidence in favor of both \textbf{H3} and \textbf{H4}. 

As expected, many users struggled to teleoperate the robot. Several users made explicit note of this fact in their comments: ``I had a hard time controlling the robot", ``I found the [IRL system] difficult as someone who [is not] kinetically gifted!", ``Would be nice if the controller for the [robot] was easier to use." Given that the robot that employs IRL was only trained on these demonstrations, it is perhaps unsurprising that DemPref outperforms IRL on the task.

We were however surprised by the extent to which the IRL-powered robot fared poorly: in many cases, it did not even attempt to reach for the goal. Upon further investigation, we discovered that IRL was prone to, in essence, ``overfitting" to the training domain. In several cases, IRL had overweighted the users' preference for obstacle avoidance. This proved to be an issue in one of our test domains where the obstacle is closer to the robot than in the training domain. Here, the robot does not even try to reach for the goal since the loss in value (as measured by the learned reward function) from going near the obstacle is greater than the gain in value from reaching for the goal. Figure~\ref{fig:04_01_rss_user_study}~(left) shows this test domain and illustrates, for a specific user, a trajectory generated according to reward function learned by each of IRL and DemPref.

While we expect that IRL would overcome these issues with more careful feature engineering and increased diversity of the training domains, it is worth noting DemPref was affected much less by these issues. These results suggest learning from comparative feedback methods may be more robust to poor feature engineering and a lack of training diversity than IRL; however, a rigorous evaluation of these claims is beyond the scope of this thesis.

It is interesting that despite the challenges that users faced with teleoperating the robot, they did not rate the DemPref system as being ``easier" to use than the IRL system ($p = 0.297$). Several users specifically referred to the time it took to generate each query ($\sim$45 seconds) as negatively impacting their experience with the DemPref system: ``I wish it was faster to generate the preference [queries]", ``The [DemPref system] will be even better if time cost is less." Additionally, one user expressed difficulty in evaluating the preference queries themselves, commenting ``It was tricky to understand/infer what the preferences were [asking]. Would be nice to communicate that somehow to the user (e.g. which [trajectory] avoids collision better)!", which highlights the fact that volume removal formulation may generate queries that are extremely difficult for the humans. Hence, we analyze in the next section how mutual information objective improves the experience for the users.

%% file: 04_active/02_information_gain.tex
\label{sec:04_02_information_gain}

As we just showed both mathematically and empirically, maximizing volume removal sometimes fails to generate \emph{informative} queries, and also does not consider the ease and intuitiveness of every query for the human-in-the-loop. This can lead to queries that are \emph{difficult} for the human to answer, e.g., two queries that are equally good (or bad) from the human's perspective.

We \emph{resolve} this issue with a second active querying method, mutual information: here the robot balances (a) how much information it will get from a correct answer against (b) the human’s ability to answer that question confidently. We also present a \emph{set of tools} that can be used to enhance the user's experience, including an optimal condition for determining when the robot should stop asking questions.

\subsection{Maximum Mutual Information Optimization}
At each iteration, we find the query $\query^{(i)}$ that maximizes the mutual information about $\weights$. We do so by solving the following optimization problem:
\begin{align}\label{eq:04_02_IG1}
\query^{(i)}_* &= \argmax_{\query^{(i)}} I(\weights; \queryResponse^{(i)} \mid \query^{(i)}, \belief^{i-1}) \nonumber\\
&= \argmax_{\query^{(i)}} H(\weights \mid \query^{(i)}, \belief^{i-1}) - \mathbb{E}_{\queryResponse^{(i)}\mid\query^{(i)},\belief^{i-1}} H(\weights \mid \queryResponse^{(i)}, \query^{(i)}, \belief^{i-1}),
\end{align}
where $I$ is the mutual information and $H$ is Shannon's information entropy \cite{cover2012elements}. Approximating the expectations via sampling, we re-write this optimization problem below (see Appendix~\ref{app:99_02_mutual_information_derivation} for the full derivation):
\begin{align}\label{eq:04_02_IG2}
\query^{(i)}_* \asymeq \argmax_{\query^{(i)} = \{\trajectory_1,\dots,\trajectory_{\abs{\query^{(i)}}}\}} &\frac{1}{\abs{\weightsSampleSet}} \sum_{\queryResponse^{(i)}\in \query^{(i)}}\sum_{{\weights}\in\weightsSampleSet}\Bigg(P(\queryResponse^{(i)} \mid \query^{(i)},{\weights}) \log_2{ \left( \frac{\abs{\weightsSampleSet} \cdot P(\queryResponse^{(i)} \mid  \query^{(i)},{\weights})}{\sum_{\weights'\in\weightsSampleSet} P(\queryResponse^{(i)} \mid \query^{(i)},\weights')} \right) \Bigg)}\:,
\end{align}
where $\weightsSampleSet$ again denotes the samples from the prior belief $\belief^{i-1}$.

\noindent\textbf{Intuition.} To see why accounting for the human is naturally part of the mutual information solution, re-write Equation~\eqref{eq:04_02_IG1}:
\begin{equation} \label{eq:04_02_IG3}
    \query^{(i)}_* = \argmax_{\query^{(i)}} ~ H(\queryResponse^{(i)} \mid \query^{(i)}, \belief^{i-1}) - \mathbb{E}_{\weights\sim\belief^{i-1}} H(\queryResponse^{(i)} \mid \weights, \query^{(i)})
\end{equation}
Here the first term in Equation~\eqref{eq:04_02_IG3} is the \emph{robot's uncertainty} over the human's response: given a query $\query^{(i)}$ and the robot's understanding of $\weights$, how confidently can the robot predict the human's answer? The second entropy term captures the \emph{human's uncertainty} when answering: given a query and their true reward, how confidently will they choose option $\queryResponse^{(i)}$? Optimizing for mutual information with Equations~\eqref{eq:04_02_IG1} or \eqref{eq:04_02_IG2} naturally considers both robot and human uncertainty, and favors questions where (a) the robot is unsure how the human will answer but (b) the human can answer easily. We contrast this to maximum volume removal, where the robot purely focused on questions where the human's answer was unpredictable.

\noindent \textbf{Why Does This Work?} To highlight the advantages of this method, let us revisit the shortcomings of volume removal. Below we show how mutual information optimization successfully addresses the problems described in Theorem~\ref{thm:04_01_volume_removal_failure} and Example~\ref{ex:04_01_example_question}. Further, we emphasize that the computational complexity of computing objective (\ref{eq:04_02_IG2}) is equivalent --- in order --- to the volume removal objective from Equation~\eqref{eq:04_01_VR2}. Thus, the mutual information based method avoids the previous failures while being at least as computationally tractable.

\subsubsection{Uninformative Queries}
Recall from Theorem~\ref{thm:04_01_volume_removal_failure} that any trivial query $\query = \{\trajectory_1, \ldots, \trajectory_1\}$ is a global solution for volume removal. In reality, we know that this query is a worst-case choice: no matter how the human answers, the robot will gain no insight into $\weights$. Mutual information ensures that the robot will not ask trivial queries: under Equation~\eqref{eq:04_02_IG1}, $\query = \{\trajectory_1,\ldots, \trajectory_1\}$ is actually the \emph{global minimum}!

\subsubsection{Challenging Questions}
Revisiting Example~\ref{ex:04_01_example_question}, we remember that $\query_B$ was a much easier question for the human to answer, but volume removal values $\query_A$ as highly as $\query_B$. Under mutual information, the \emph{robot} is equally uncertain about how the human will answer $\query_A$ and $\query_B$, and so the first term in Equation~\eqref{eq:04_02_IG3} is the same for both. But the robot using mutual information additionally recognizes that the \emph{human} is very uncertain when answering $\query_A$: here $\query_A$ attains the global maximum of the second term while $\query_B$ attains the global minimum! Thus, the overall value of $\query_B$ is higher and --- as desired --- the robot recognizes that $\query_B$ is a better question.

\subsection{Additional Tools and Analysis}
We introduced how robots can generate proactive questions to maximize mutual information. Below we highlight some additional tools that designers can leverage to improve the computational performance and applicability of these methods. In particular, we draw the reader's attention to an optimal stopping condition, which tells the DemPref robot when to stop asking the human questions.

\subsubsection{Optimal Stopping}
We propose a novel extension --- specifically for mutual information --- that tells the robot when to stop asking questions. Intuitively, the DemPref querying process should end when the robot's questions become more costly to the human than informative to the robot. 

Let each query $\query$ have an associated cost $\costFunction(\query) \in \mathbb{R}_{\geq 0}$. This function captures the \emph{cost} of a question: e.g., the amount of time it takes for the human to answer, the number of similar questions that the human has already seen, or even the interpretability of the question itself. We subtract this cost from our mutual information objective in Equation~\eqref{eq:04_02_IG1}, so that the robot (greedily) maximizes mutual information while biasing its search towards low-cost questions:
\begin{align} \label{eq:04_02_EX1}
\max_{\query=\{\trajectory_1,\ldots,\trajectory_{\abs{\query}}\}} I(\weights; \queryResponse \mid \query, \belief^{i-1}) - \costFunction(\query)
\end{align}
Now that we have introduced a cost into the query selection problem, the robot can reason about when its questions are becoming prohibitively expensive or redundant. We find that the best time to stop asking questions in expectation is when their cost exceeds their value:
\begin{theorem}\label{thm:04_02_optimal_stopping}
	A robot using mutual information to perform active preference-based learning should stop asking questions if and only if the global solution to Equation~\eqref{eq:04_02_EX1} is negative at the current iteration.
\end{theorem}
See the Appendix~\ref{app:99_01_ijrr_active_optimal_stopping} for our proof. We emphasize that this result is valid only for mutual information, and adapting Theorem~\ref{thm:04_02_optimal_stopping} to volume removal is not trivial.

The decision to terminate our DemPref algorithm is now fairly straightforward. At each iteration $i$, we search for the question $\query^{(i)}_*$ that maximizes the trade-off between mutual information and cost. If the value of Equation~\eqref{eq:04_02_EX1} is non-negative, the robot shows this query to the human and elicits their response; if not, the robot cannot find any sufficiently important questions to ask, and the process ends. This automatic stopping procedure makes the active learning process more user-friendly by ensuring that the user does not have to respond to unnecessary or redundant queries.

\subsubsection{Why Demonstrations First?}
Now that we have a user-friendly strategy for generating queries and stopping, we want to determine \emph{in what order} the robot should leverage the demonstrations and the comparisons.

Recall that demonstrations provide coarse, high-level information, while comparison queries hone-in on isolated aspects of the human's reward function. Intuitively, it seems like we should start with high-level demonstrations before probing low-level preferences: but is this really the right order of collecting data? What about the alternative --- a robot that instead waits to utilize the demonstrations dataset $\demonstrationDataset$ until after asking questions? 

When leveraging mutual information maximization to generate queries, we here prove that the robot will gain \emph{at least as much} information about the human's preferences as any other order of demonstrations and queries. Put another way, starting with demonstrations \emph{in the worst case} is just as good as any other order; and \emph{in the best case} we obtain more information.

\begin{theorem}
	Under the Boltzmann rational human model for demonstrations presented in Equation~\eqref{eq:03_01_DP3}, our DemPref approach --- where best-of-many choice queries are actively generated after collecting demonstrations --- results in at least as much information about the human's preferences as would be obtained by reversing the order of queries and demonstrations.
	\begin{proof}
	Let $\query^{(i)}_*$ be the (greedily) optimal query with respect to the mutual information optimization \emph{after collecting demonstrations}. From Equation~\eqref{eq:04_02_EX1}, $\query^{(i)}_* = \argmax_{\query^{(i)}} \left(I(\weights; \queryResponse^{(i)} \mid \query^{(i)}, \belief^{i-1}) - \costFunction(\query^{(i)})\right)$. We let $\queryResponse^{(i)}_*$ denote the human's response to query $\query^{(i)}_*$. Similarly, let ${\tilde\query}^{(i)}$ be the mutual information query \emph{before incorporating the demonstrations into the belief}, so that ${\tilde\query}^{(i)} = \argmax_{\query^{(i)}} \left(I(\weights; \queryResponse^{(i)} \mid \query^{(i)}, ({\tilde\query}^{(j)}, {\tilde\queryResponse}^{(j)})_{j=0}^{i-1}) - \costFunction(\query^{(i)})\right)$. Again, we let ${\tilde\queryResponse}^{(i)}$ denote the human's response to query ${\tilde\query}^{(i)}$. Noting the total cost of queries will not change (and hence, the theorem and the proof extend to the case where $\costFunction(\query)=0$ for all queries $\query$), we can compare the overall mutual information for each order of questions and demonstrations:
	\begin{align}
	    I\Big(&\weights;\demonstrationDataset,(\queryResponse^{(1)}_*,\queryResponse^{(2)}_*,\ldots) \mid (\query^{(1)}_*,\query^{(2)}_*,\ldots)\Big) \nonumber\\
	    &= I(\weights;\demonstrationDataset) + I\left(\weights;(\queryResponse^{(1)}_*,\queryResponse^{(2)}_*,\ldots) \mid (\belief^0, \query^{(1)}_*,\query^{(2)}_*,\ldots)\right) \nonumber\\
	    &\geq I(\weights;\demonstrationDataset) + I\left(\weights;({\tilde\queryResponse}^{(1)},{\tilde\queryResponse}^{(2)},\ldots) \mid (\belief^0, {\tilde\query}^{(1)},{\tilde\query}^{(2)},\ldots)\right) \nonumber\\
	    &= I\left(\weights;({\tilde\queryResponse}^{(1)},{\tilde\queryResponse}^{(2)},\ldots, \demonstrationDataset) \mid ({\tilde\query}^{(1)},{\tilde\query}^{(2)},\ldots)\right)
	\end{align}
	\end{proof}
\label{thm:04_02_order_of_dempref}
\end{theorem}

\noindent \textbf{Intuition.} We can explain Theorem~\ref{thm:04_02_order_of_dempref} through two main insights. First, the mutual information from a passively collected demonstration dataset is the same regardless of when that demonstration is incorporated as long as it is conditioned on the same variables. Second, proactively generating questions based on a prior leads to more incisive queries than choosing questions from scratch. In fact, Theorem~\ref{thm:04_02_order_of_dempref} can be generalized to show that active information resources should be utilized after passive resources.

\subsubsection{Bounded Regret}
At the start of this chapter we mentioned that --- instead of looking for the optimal sequence of future questions --- our techniques will greedily choose the best query at the current iteration. Prior work has shown that this greedy approach is reasonable for volume removal, where it is guaranteed to have bounded suboptimality in terms of the volume removed~\cite{sadigh2017active}. However, this volume is defined in terms of the unnormalized distribution, and so this result does not say much about the learning performance. Unfortunately, the mutual information also does not provide theoretical guarantees, as it is only submodular, but not \emph{adaptive} submodular.

\subsection{Algorithm}

\begin{algorithm}[t]
  \caption{DemPref with a Human-in-the-Loop}
  \label{alg:04_02_DemPref}
  \begin{algorithmic}[1]
    \State Collect human demonstrations: $\demonstrationDataset = \{\demonstration^{(1)}, \demonstration^{(2)}, \ldots, \demonstration^{(\abs{\demonstrationDataset})} \}$
    \State Initialize belief over the human's reward weights $\weights$: $\belief^0(\weights) \propto \exp{\left(\demonstrationRationalityCoefficient \weights \cdot \sum_{\demonstration \in \demonstrationDataset} \trajectoryFeaturesFunction(\demonstration)\right)} P(\weights)$
    \For{$i \gets 1, 2, \ldots$}
        \State Choose proactive question $\query^{(i)}$: $\query^{(i)}_* \gets \argmax_{\query^{(i)}}~I(\weights; \queryResponse^{(i)} \mid \query^{(i)}, \belief^{i-1}) - \costFunction(\query^{(i)})$
        \If{$I(\weights; \queryResponse^{(i)} \mid \query^{(i)}, \belief^{i-1}) - \costFunction(\query^{(i)}) < 0$}
            \State \Return $\belief^{i-1}$
        \EndIf
        \State Elicit human's answer $\queryResponse^{(i)}$ to query $\query^{(i)}$
        \State Update belief over $\weights$ given query and response: $\belief^{i}(\weights) \propto P(\queryResponse^{(i)} \mid \query^{(i)}, \weights) \belief^{i-1}(\weights)$
    \EndFor
  \end{algorithmic}
\end{algorithm}

We present the complete DemPref pseudocode with active querying in Algorithm~\ref{alg:04_02_DemPref}. This algorithm involves two main steps: first, the robot uses the human's offline trajectory demonstrations $\demonstrationDataset$ to initialize a high-level understanding of the human's preferred reward. Next, the robot actively generates user-friendly questions $\query$ to fine-tune its belief $\belief$ over $\weights$. These questions can be selected using volume removal or mutual information objectives (we highlight the mutual information approach in Algorithm~\ref{alg:04_02_DemPref}). As the robot asks questions and obtains a precise understanding of what the human wants, the informative value of new queries decreases: eventually, asking new questions becomes suboptimal, and the DemPref algorithm terminates.

\noindent \textbf{Advantages.} Before moving to the simulation and experiment results, we conclude our presentation of DemPref with mutual information maximization by summarizing its two main contributions:
\begin{enumerate}[nosep]
    \item The robot learns the human's reward by synthesizing two types of information: high-level demonstrations and fine-grained best-of-many choice queries.
    \item The robot generates questions while accounting for the human's ability to respond, naturally leading to user-friendly and informative queries.
\end{enumerate}

\subsection{Experiments}\label{subsec:04_02_experiments}
We conduct three sets of experiments to evaluate the performance of DemPref with mutual information maximization. Similar to Section~\ref{sec:04_01_volume_removal}, we assume a reward function that is linear in trajectory features in all experiments, i.e., $\trajectoryRewardFunction_{\weights}(\trajectory) = \weights^{\top}\trajectoryFeaturesFunction(\trajectory)$ for any trajectory $\trajectory\in\trajectorySpace$.\footnote{Unless otherwise noted, we adopt $\demonstrationRationalityCoefficient=0.02$, $\comparisonRationalityCoefficient=1$, constant $\costFunction(\query)$ for $\forall \query$, and assume a uniform prior over reward parameters $\weights$, i.e., $P(\weights)$ is constant for any $\norm{\weights}_2\leq1$. We use Metropolis-Hastings algorithm \cite{chib1995understanding} for sampling the set $\weightsSampleSet$ from belief distribution over $\weights$.}

\subsubsection{Simulation Domains}
In each experiment in addition to the experiment domains presented in Section~\ref{subsec:04_01_experiments}, we use a subset of the following domains. These domains are shown in Figure~\ref{fig:04_01_corl19_domains} (and see Figure~\ref{fig:04_01_rss19_domains} for the domains adopted from Section~\ref{subsec:04_01_experiments}).

\noindent\emph{Linear Dynamical System (LDS).} We use a linear dynamical system with six dimensional state and three dimensional action spaces. State values are directly used as state features without any transformation.

\noindent\emph{Driver.} We use a 2D driving simulator \cite{sadigh2016planning}, where the agent has to safely drive down a highway. The trajectory features correspond to the distance of the agent from the center of the lane, its speed, heading angle, and minimum distance to other vehicles during the trajectory (white in Figure~\ref{fig:04_01_corl19_domains}~(top)). See Appendix~\ref{app:99_04_ijrr_features} for the formal feature definitions.

\noindent\emph{Tosser.} We use a ``Tosser" robot simulation built in MuJoCo \cite{todorov2012mujoco} that tosses a capsule-shaped object into baskets. The trajectory features are the maximum horizontal distance forward traveled by the object, the maximum altitude of the object, the number of flips the object does, and the object's final distance to the closest basket. See Appendix~\ref{app:99_04_ijrr_features} for the formal feature definitions.

For our user studies, we again employ the same version of the Fetch environment as in Section~\ref{subsec:04_01_experiments} with the physical Fetch robot (see Figure~\ref{fig:04_01_rss19_domains}) \cite{wise2016fetch}.

\subsubsection{Human Choice Models}
As we described in Section~\ref{sec:03_01_pairwise_comparisons}, we require a probabilistic model for the human's response $\queryResponse^{(i)}$ in a query $\query^{(i)}$ conditioned on their reward function parameters $\weights$. In the results we presented in this section, we use two specific models. One is the softmax model we introduced in Equation~\eqref{eq:03_01_noisily_optimal}, re-stating:
\begin{align}
    P(\queryResponse^{(i)} = \query^{(i)}_{j} \mid \query^{(i)},\weights) &= \frac{\exp(\comparisonRationalityCoefficient\trajectoryRewardFunction_{\weights}(\query^{(i)}_j))}{\sum_{j'=1}^{\abs{\query^{(i)}}}\exp(\comparisonRationalityCoefficient\trajectoryRewardFunction_{\weights}(\query^{(i)}_{j'}))}\:.
\end{align}

As the second model, we generalize pairwise comparison queries and this preference model to include an ``About Equal" option. This is similar to scale queries we introduced in Section~\ref{sec:03_04_scale}. However, we are using a different, simpler choice model as we are not allowing any choice other than ``About Equal" and the trajectory choices. We denote this new ``About Equal" option by $\aboutEqual$ and define a \emph{weak pairwise comparison query} $\query^+ := \query\cup\{\aboutEqual\}$ when $\abs{\query}=2$.

Building on prior work by \citet{krishnan1977incorporating}, we incorporate the information from the ``About Equal" option by introducing a minimum perceivable difference parameter $\minimumPerceivableDifference \geq 0$, and defining:
\begin{align}
P&(\queryResponse^{(i)}=\aboutEqual \mid {\query^{(i)}}^+,\weights) = \left(\exp(2\minimumPerceivableDifference) - 1\right)P(\queryResponse^{(i)}=\query^{(i)}_1\mid {\query^{(i)}}^+,\weights)P(\queryResponse^{(i)}=\query^{(i)}_2 \mid  {\query^{(i)}}^+,\weights)\:, \nonumber\\
P&(\queryResponse^{(i)}=\query^{(i)}_j \mid {\query^{(i)}}^+,\weights) = \frac{1}{1 + \exp(\minimumPerceivableDifference + \trajectoryRewardFunction_{\weights}(\query^{(i)}_{j'}) - \trajectoryRewardFunction_{\weights}(\query^{(i)}_j))}, \quad \{\query^{(i)}_j,\query^{(i)}_{j'}\} = {\query^{(i)}}^+ \setminus \{\aboutEqual\}\:.
\label{eq:04_01_weak_noisily_optimal}
\end{align}
Notice that Equation~\eqref{eq:04_01_weak_noisily_optimal} reduces to Equation~\eqref{eq:03_01_noisily_optimal} when $\minimumPerceivableDifference=0$; in which case we model the human as always perceiving the difference in options. All derivations in Sections~\ref{sec:03_01_pairwise_comparisons}, \ref{sec:04_01_volume_removal} and \ref{sec:04_02_information_gain} hold with weak pairwise comparison queries. In particular, we include a discussion of extending our formulation to the case where $\minimumPerceivableDifference$ is user-specific and unknown in Appendices~\ref{app:99_02_ijrr_active_unknown_parameter} and \ref{app:99_05_ijrr_active_unknown_delta}. The additional parameter causes no trouble in practice. For all our experiments in this section, we set $\abs{\query}=2$, and $\minimumPerceivableDifference=1$ (whenever relevant).

We note that there are alternative choice models compatible with our framework for weak pairwise comparisons (e.g., \cite{holladay2016active} and our scale feedback model in Section~\ref{sec:03_04_scale}). Additionally, one may generalize the weak pairwise comparison queries to $\abs{\query}>2$, though it complicates the choice model as the user must specify which of the trajectories create uncertainty.

We are now ready to present our three sets of experiments each of which demonstrates a different aspect of the proposed active DemPref framework:
\begin{enumerate}[nosep]
    \item The advantages of mutual information formulation over volume removal,
    \item The order of demonstrations and preferences, and
    \item Optimal stopping condition under the mutual information objective.
\end{enumerate}
We again use the \texttt{Alignment} metric to quantitatively evaluate the performance (see Equation~\eqref{eq:04_01_alignment}).

\subsubsection{Mutual Information vs. Volume Removal}
To investigate the performance and user-friendliness of the mutual information and volume removal methods for learning from comparative feedback, we conduct experiments with simulated users in \emph{LDS}, \emph{Driver}, \emph{Tosser} and \emph{FetchReach} environments; and real user studies in \emph{Driver}, \emph{Tosser} and \emph{FetchReach} (with the physical robot). We are particularly interested in the following three hypotheses:

\noindent\textbf{H5.} \emph{Mutual information formulation outperforms volume removal in terms of data-efficiency.}

\noindent\textbf{H6.} \emph{Mutual information queries are easier and more intuitive for the human than those from volume removal.}

\noindent\textbf{H7.} \emph{A user's preference aligns best with reward parameters learned via mutual information maximization.} 

To enable faster computation, we discretized the search space of the optimization problems by generating $500,\!000$ random pairwise comparison queries and precomputing their trajectory features. Each call to an optimization problem then performs a loop over this discrete set.

\begin{figure*}[t]
	\centering
	\includegraphics[width=\textwidth]{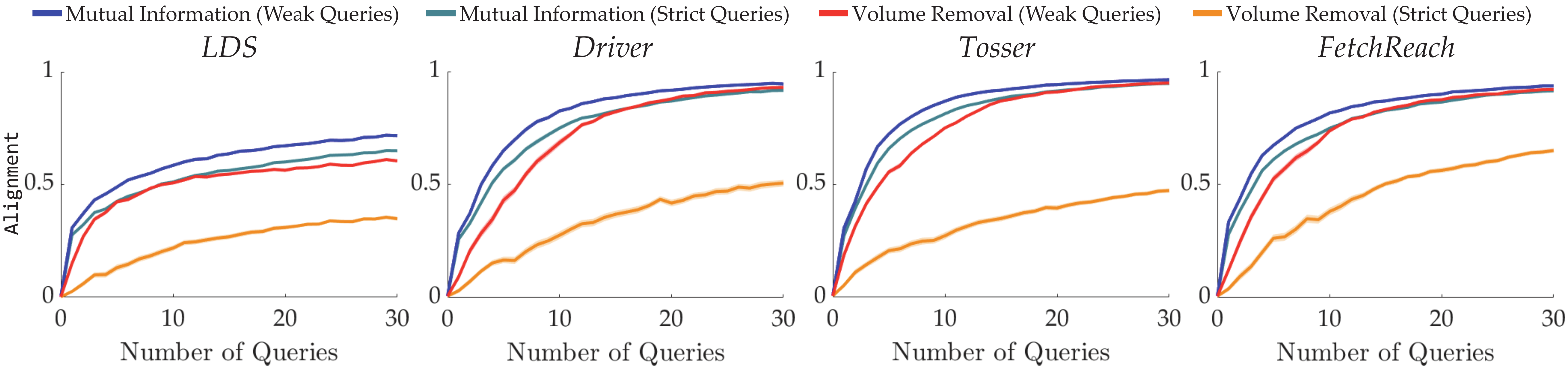}
	\caption{Alignment values are plotted (mean $\pm$ standard error) to compare mutual information and volume removal formulations. Standard errors are so small that they are mostly invisible in the plots. Dashed lines show the weak pairwise comparison query variants. Mutual information provides a significant increase in learning rate in all cases. While weak pairwise comparison queries lead to a large amount of improvement under volume removal, mutual information formulation is still superior in terms of the convergence rate.}
	\label{fig:04_02_simulation_results_m}
\end{figure*}

In simulation experiments, we learn the randomly generated reward functions via both strict and weak pairwise comparison queries where the ``About Equal" option is absent and present, respectively. We repeat each experiment $100$ times to obtain confidence bounds. Figure~\ref{fig:04_02_simulation_results_m} shows the \texttt{Alignment} value against query number for the $4$ different tasks. Even though the ``About Equal" option improves the performance of volume removal by preventing the trivial query, $\query=\{\trajectory_1,\trajectory_1,\dots\}$, from being a global optimum, mutual information gives a significant improvement on the learning rate both with and without the ``About Equal" option in all environments.\footnote{See Appendix~\ref{app:99_05_ijrr_active_no_discretization} for results without query space discretization.} These results strongly support \textbf{H5}.

\begin{figure*}[t]
	\centering
	\includegraphics[width=\textwidth]{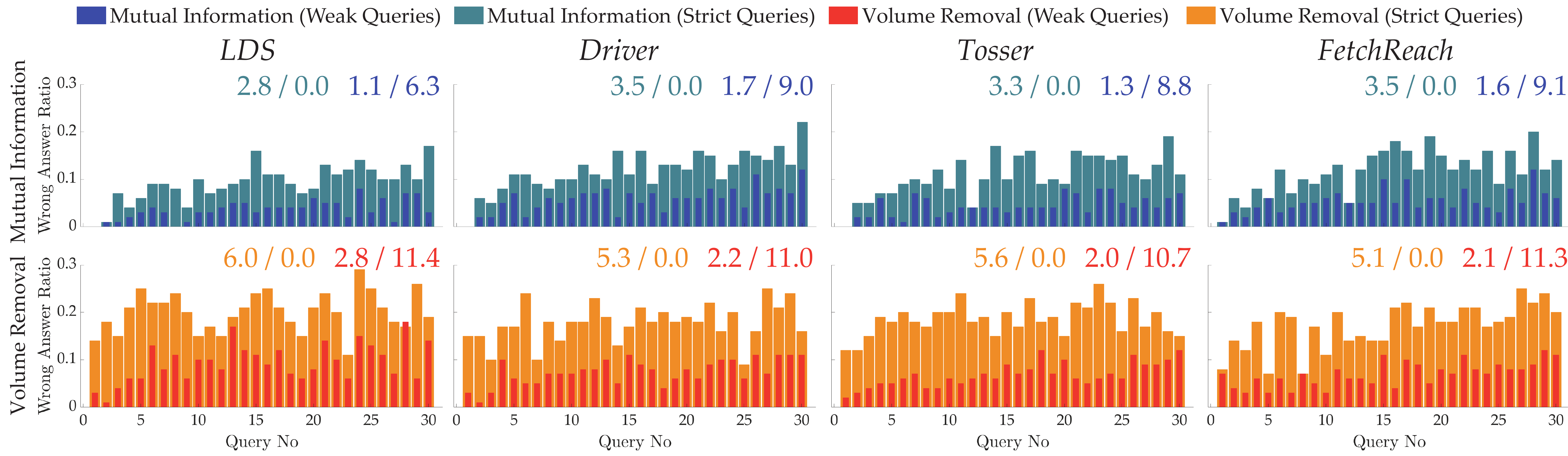}
	\caption{Wrong answer ratios on different queries are shown. The numbers at top show the average number of wrong responses and ``About Equal" choices, respectively, for both strict and weak queries. Mutual information formulation yields smaller numbers of wrong and ``About Equal" answers, especially in the early stages.}
	\label{fig:04_02_simulation_results_wrong_responses}
\end{figure*}

The numbers given within Figure~\ref{fig:04_02_simulation_results_wrong_responses} count the wrong answers and ``About Equal" choices made by the simulated users. The mutual information formulation significantly improves over volume removal. Moreover, weak pairwise comparison queries consistently decrease the number of wrong answers, which can be one reason why it performs better than strict queries.\footnote{Another possible explanation is the information acquired by the ``About Equal" responses. We analyze this in Appendix~\ref{app:99_05_ijrr_active_value_of_idk} by comparing the results with what would happen if this information was discarded.} Figure~\ref{fig:04_02_simulation_results_wrong_responses} also shows when the wrong responses are given. While wrong answer ratios are higher with volume removal formulation, it can be seen that mutual information reduces wrong answers especially in early queries, which leads to faster learning. These results support \textbf{H6}.

In the user studies for this part, we used \emph{Driver} and \emph{Tosser} environments in simulation and the \emph{FetchReach} environment with the physical robot. We began by asking participants to rank a set of features (described in plain language) to encourage each user to be consistent in their preferences. Subsequently, we queried each participant with a sequence of $30$ questions generated actively; $15$ from volume removal and $15$ via mutual information. We prevent bias by randomizing the sequence of questions for each user and experiment: the user does not know which algorithm generates a question.

Participants responded to a 7-point rating scale survey after each question:
\begin{itemize}[nosep]
    \item It was easy to choose between the trajectories that the robot showed me.
\end{itemize}
They were also asked the Yes/No question:
\begin{itemize}[nosep]
    \item Can you tell the difference between the options presented?
\end{itemize}

In concluding the \emph{Tosser} and \emph{Driver} experiments, we showed participants two trajectories: one optimized using reward function parameters from mutual information (trajectory A) and one optimized using reward parameters from volume removal (trajectory B)\footnote{We excluded \emph{FetchReach} for this question to avoid prohibitive trajectory optimization (due to large action space).}.
Participants responded to a 7-point rating scale survey: \begin{itemize}[nosep]
    \item Trajectory A better aligns with my preferences than trajectory B.
\end{itemize}

We recruited $15$ participants ($8$ female, $7$ male) for the simulations (\emph{Driver} and \emph{Tosser}) and $12$ for the \emph{FetchReach} ($6$ female, $6$ male). We used strict pairwise comparison queries. A video demonstration of these user studies is available at \url{http://youtu.be/JIs43cO\_g18}.

\begin{figure*}[t]
	\centering
	\includegraphics[width=0.85\textwidth]{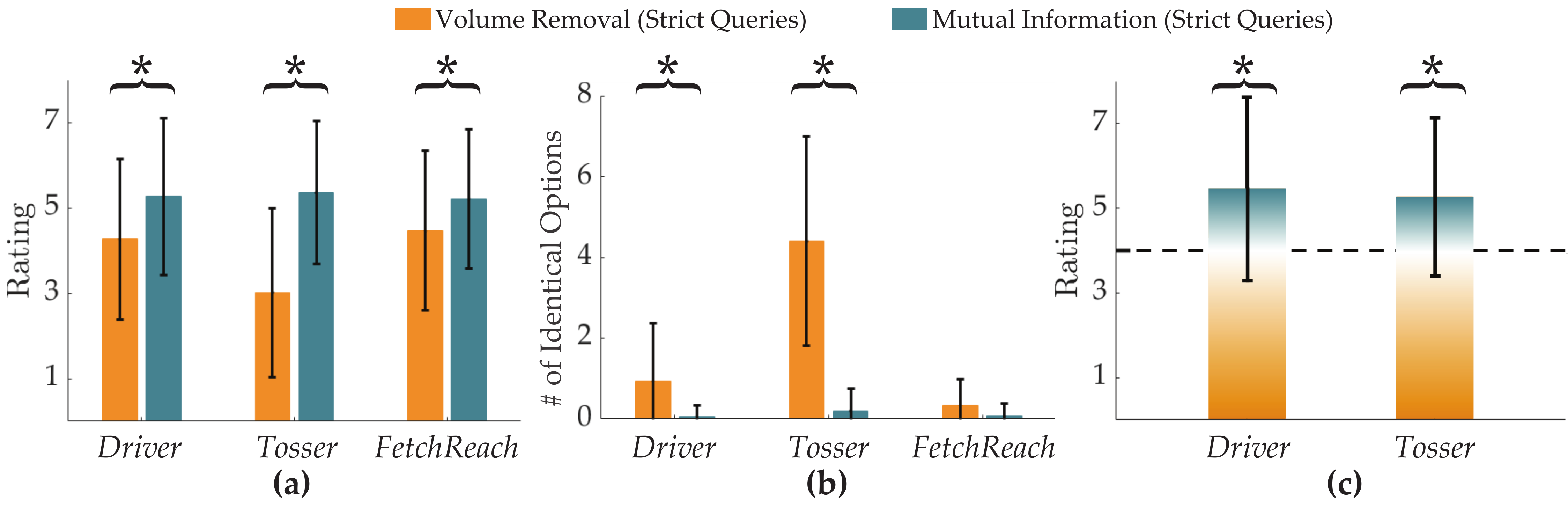}
	\caption{User study results. Error bars show std. Asterisks show statistical significance. \textbf{(a)} Easiness survey results averaged over all queries and users. Queries generated using the mutual information maximization method are rated significantly easier by the users than the volume removal queries. \textbf{(b)} The number of identical options in the experiments averaged over all users. In Driver and Tosser, users indicated significantly less indistinguishable queries with mutual information maximization compared to volume removal maximization.  \textbf{(c)} Final preferences averaged over the users. $7$ means the user strongly prefers the optimized trajectory w.r.t. the learned reward by the mutual information formulation, and $1$ is the volume removal. Dashed line represents indifference between two methods.  Users significantly prefer the robot who learned using the mutual information maximization method for active query generation.}
	\label{fig:04_02_corl_user_studies}
\end{figure*}

Figure~\ref{fig:04_02_corl_user_studies}a shows the results of the easiness surveys. In all environments, users found mutual information queries easier: the results are statistically significant ($p<0.005$, two-sample $t$-test). Figure~\ref{fig:04_02_corl_user_studies}b shows the average number of times the users stated they cannot distinguish the options presented. The volume removal formulation yields several queries that are indistinguishable to the users while the mutual information formulation avoids this issue. The difference is significant for \emph{Driver} ($p<0.05$, paired-sample $t$-test) and \emph{Tosser} ($p<0.005$). Taken together, these results support \textbf{H6}.

Figure~\ref{fig:04_02_corl_user_studies}c shows the results of the survey the participants completed at the end of experiment. Users significantly preferred the mutual information trajectory over that of volume removal in both environments ($p<0.05$, one-sample $t$-test), supporting \textbf{H7}.

\subsubsection{The Order of Information Sources}
Having seen mutual information maximization provides a significant boost in the learning rate, we checked whether the passively collected demonstrations or the actively queried preferences should be given to the model first. Specifically, we tested:

\noindent\textbf{H8.} \emph{If passively collected demonstrations are used before the actively collected comparison query responses, the learning becomes faster.}

While Theorem~\ref{thm:04_02_order_of_dempref} asserts that we should first initialize DemPref via demonstrations, we performed simulation experiments to check this notion in practice. Using \emph{LDS}, \emph{Driver}, \emph{Tosser} and \emph{FetchReach}, we ran three sets of experiments where we adopted weak pairwise comparison queries: (i) We initialize the belief with a single demonstration and then query the simulated user with $15$ pairwise comparison questions, (ii) We first query the simulated user with $15$ pairwise comparison questions and we add the demonstration to the belief independently after each question, and (iii) We completely ignore the demonstration and use only $15$ pairwise comparison queries. The reason why we chose to have only a single demonstration is because having more more demonstrations tends to increase the alignment value for both (i) and (ii), thereby making the difference between the methods' performances very small. We ran each set of experiment $100$ times with different, randomly sampled, true reward functions. We again used the same dataset of $500,\!000$ queries for query generation. We also used the trajectory that gives the highest reward to the simulated user out of this dataset as the demonstration in the first two sets of experiments. Since the demonstration is not subject to noises or biases due to the control of human users, we set $\demonstrationRationalityCoefficient=0.2$.

\begin{figure*}[th]
	\centering
	\includegraphics[width=\textwidth]{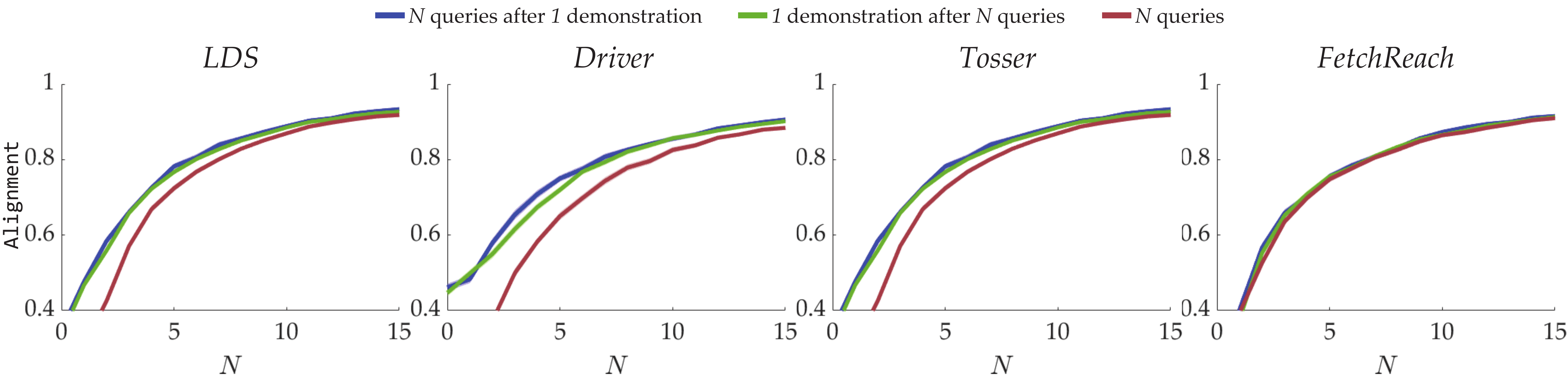}
	\caption{Simulation results for the order of demonstrations and preference queries. \texttt{Alignment} values are plotted (mean$\pm$s.e.). It is consistently better to first utilize the passively collected demonstrations rather than actively generated preference queries. The differences in the \texttt{Alignment} value is especially small in the \emph{FetchReach} simulations, which might be due to the fact that it is a simpler environment in terms of the number of trajectory features.}
	\label{fig:04_02_demfirst_vs_demsecond}
\end{figure*}

Figure~\ref{fig:04_02_demfirst_vs_demsecond} shows the \texttt{Alignment} value against the number of queries. The last set of experiments has significantly lower \texttt{Alignment} values than the first two sets especially when the number of pairwise comparison queries is small. This indicates the demonstration has carried an important amount of information. Comparing the first two sets of experiments, the differences in the \texttt{Alignment} values are small. However, the values are consistently higher when the demonstrations are used to initialize the belief distribution. This supports \textbf{H8} and numerically validates Theorem~\ref{thm:04_02_order_of_dempref}.

\subsubsection{Optimal Stopping}
Finally, we experimented our optimal stopping extension for mutual information based active querying algorithm in \emph{LDS}, \emph{Driver}, \emph{Tosser} and \emph{FetchReach} environments with simulated users. Again adopting query discretization, we tested:

\noindent\textbf{H9.} \emph{Optimal stopping enables cost-efficient reward learning under various costs.}

\begin{figure*}[th]
	\centering
	\includegraphics[width=\textwidth]{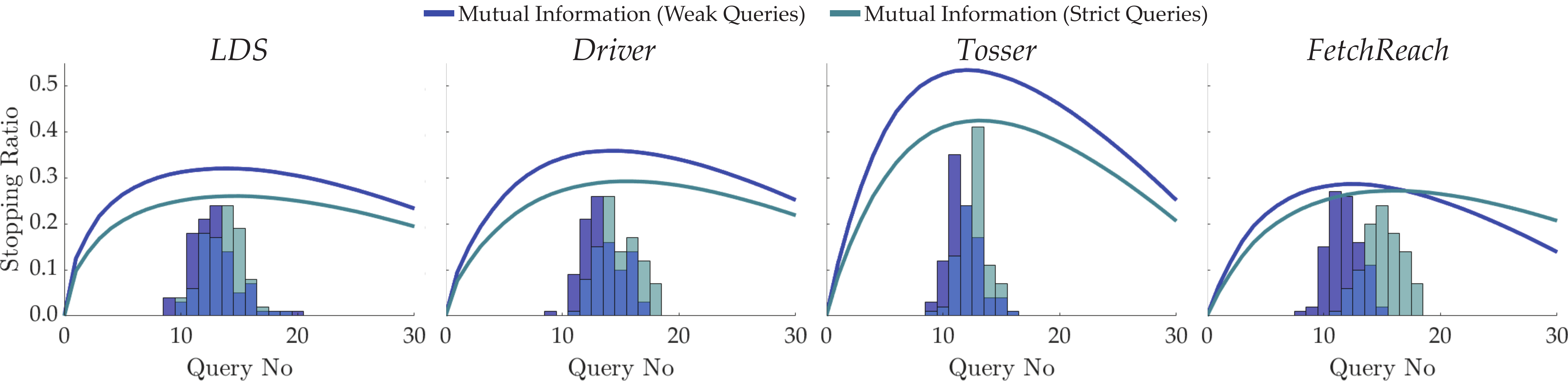}
	\caption{Simulation results for optimal stopping. Line plots show cumulative active learning rewards (cumulative difference between the mutual information values and the query costs), averaged over $100$ test runs and scaled for visualization. Histograms show when optimal stopping condition is satisfied, which aligns with the desired cumulative rewards.}
	\label{fig:04_02_optimal_stopping_query_dependent}
\end{figure*}

As the query cost, we employed a cost function to improve interpretability of queries, which may have the associated benefit of making learning more efficient \cite{bajcsy2018learning}. We defined a cost function:
\begin{align*}
\costFunction(\query) = \queryCostHyperparameter-\abs{\featureDifference_{j^*}} + \max_{j'\in\{1,\dots\}\setminus\{j^*\}}\abs{\featureDifference_{j'}},\: j^*=\argmax_j{\abs{\featureDifference_j}},
\end{align*}
where $\featureDifference=\trajectoryFeaturesFunction(\query_1)-\trajectoryFeaturesFunction(\query_2)$. This cost favors queries in which the difference in one feature is larger than that between all other features. Such a query may prove more interpretable. We first simulate $100$ random users and tune $\queryCostHyperparameter$ accordingly: For each simulated user, we record the $\queryCostHyperparameter$ value that makes the objective zero in the $i^{\textrm{th}}$ query (for smallest $i$) such that $\texttt{Alignment}_i, \texttt{Alignment}_{i-1}, \texttt{Alignment}_{i-2} \in [x,x+0.02]$ for some $x$. We then use the average of these $\queryCostHyperparameter$ values for our tests with $100$ different random users. Figure~\ref{fig:04_02_optimal_stopping_query_dependent} shows the results.\footnote{We found similar results with query-independent costs minimizing the number of queries. See Appendix~\ref{app:99_05_ijrr_active_query_independent_stopping}.} Optimal stopping rule enables terminating the process with near-optimal cumulative active learning rewards (the cumulative difference between the mutual information and the cost as in Equation~\eqref{eq:04_02_EX1}) in all environments, which supports \textbf{H9}.

Following the same order as in Chapter~\ref{chap:learning}, we now proceed with preference-based GP regression where a Gaussian process model is trained using pairwise comparisons. We extend this learning framework with mutual information based active querying to improve data-efficiency.

%% file: 04_active/03_gp.tex
\label{sec:04_03_gp}

While we know how to learn a non-parametric reward function $f$ using only pairwise comparisons from Section~\ref{sec:03_02_gp}, this endeavor can require tremendous amount of data, because each query will give at most $1$ bit of information. Furthermore, we can expect a decreasing trend in the information gain as we learn the reward function. Therefore, it is important to select the queries for the human such that each query gives as much information as possible. For parametric reward models, Section~\ref{sec:04_02_information_gain} has shown that this can be done by maximizing the mutual information, which also makes the queries easy for the user. Extending this formulation to the reward functions modeled with a GP is not trivial, because one needs to sample from the GP many times for each trajectory, whereas a parametric reward form allows the reward prediction after sampling the parameters only once.

Hence, in this section, our goal is to perform mutual information maximization with GPs.

\subsection{Formulation}
Formally, we want to solve the following problem, using the same notation as in Section~\ref{sec:03_02_gp}:
\begin{align*}
\query_* = (\trajectoryFeaturesFunction_*^{(1)}, \trajectoryFeaturesFunction_*^{(2)}) = \argmax_{\trajectoryFeaturesFunction^{(1)}, \trajectoryFeaturesFunction^{(2)}} I(\gpRewardFunction ; \queryResponse \mid \trajectoryFeaturesFunction^{(1)}, \trajectoryFeaturesFunction^{(2)}, \mathbf{\query}, \mathbf{\queryResponse}),
\end{align*}
where $I$ is the mutual information and $\queryResponse$ is the response to the query $\query=(\trajectoryFeaturesFunction^{(1)}, \trajectoryFeaturesFunction^{(2)})$. This optimization is equivalent to
\begin{align}
\argmax_{\trajectoryFeaturesFunction^{(1)}, \trajectoryFeaturesFunction^{(2)}} \left(H(\queryResponse \mid \trajectoryFeaturesFunction^{(1)}, \trajectoryFeaturesFunction^{(2)}, \mathbf{\query}, \mathbf{\queryResponse}) - \mathbb{E}_{\gpRewardFunction \mid \mathbf{\query}, \mathbf{\queryResponse}}\left[H(\queryResponse \mid \trajectoryFeaturesFunction^{(1)}, \trajectoryFeaturesFunction^{(2)}, \gpRewardFunction)\right]\right),
\label{eq:04_03_problem}
\end{align}
where $H$ is the information entropy.

This optimization can be interpreted as follows:
On one hand, maximizing the first entropy term $H(\queryResponse \mid \trajectoryFeaturesFunction^{(1)}, \trajectoryFeaturesFunction^{(2)}, \mathbf{\query}, \mathbf{\queryResponse})$ encourages fast convergence by maximizing the uncertainty of the outcome of every query for the learned GP model. On the other hand, minimizing the second entropy term $H(\queryResponse \mid \trajectoryFeaturesFunction^{(1)}, \trajectoryFeaturesFunction^{(2)}, \gpRewardFunction)$ encourages the ease of responding to the queries by the user meaning the user should be certain about their choices.

We defer the full derivation of \eqref{eq:04_03_problem} to Appendix~\ref{app:99_02_rss20_gp_derivation}, but here we give an easy-to-implement formulation of the optimization. Denoting the posterior mean of $\gpRewardFunction(\trajectoryFeaturesFunction^{(i)})$, which is obtained using Equation~\eqref{eq:03_02_inference_mean}, with $\gpPosteriorMean^{(i)}$, the objective function can be written as
\begin{align}
\binaryEntropyFunction\left(\Phi\left(\frac{\gpPosteriorMean^{(1)}-\gpPosteriorMean^{(2)}}{\sqrt{2\preferenceNoiseStd^2 + \gpGFunction(\trajectoryFeaturesFunction^{(1)},\trajectoryFeaturesFunction^{(2)})}}\right)\right) - \gpMFunction\left(\trajectoryFeaturesFunction^{(1)},\trajectoryFeaturesFunction^{(2)}\right)
\label{eq:04_03_active_query_generation}
\end{align}
where $\preferenceNoiseStd$ is the noise parameter of the human response model we introduced in Equation~\eqref{eq:03_02_human_model},
\begin{align*}
\gpGFunction(\trajectoryFeaturesFunction^{(1)},\trajectoryFeaturesFunction^{(2)})=&\mathrm{Var}\left(\gpRewardFunction(\trajectoryFeaturesFunction^{(1)})\right) + \mathrm{Var}\left(\gpRewardFunction(\trajectoryFeaturesFunction^{(2)})\right) - 2 \: \mathrm{Cov}\left(\gpRewardFunction(\trajectoryFeaturesFunction^{(1)}),\gpRewardFunction(\trajectoryFeaturesFunction^{(2)})\right),
\end{align*}
whose terms can be computed using Equation~\eqref{eq:03_02_inference_cov};
$\binaryEntropyFunction$ is the binary entropy function, i.e., 
\begin{equation*}
    \binaryEntropyFunction(p) = -p\log_2(p) - (1-p)\log_2(1-p),
\end{equation*}
and
\begin{align*}
\gpMFunction\left(\trajectoryFeaturesFunction^{(1)},\trajectoryFeaturesFunction^{(2)}\right) = \frac{\sqrt{\pi\ln(2)\preferenceNoiseStd^2}\exp\left(-\frac{(\gpPosteriorMean^{(1)} - \gpPosteriorMean^{(2)})^2}{\pi\ln(2)\preferenceNoiseStd^2 + 2\gpGFunction(\Psi^{(1)},\Psi^{(2)})}\right)}{\sqrt{\pi\ln(2)\preferenceNoiseStd^2 + 2\gpGFunction(\trajectoryFeaturesFunction^{(1)},\trajectoryFeaturesFunction^{(2)})}}.
\end{align*}

Synthesizing queries that maximize this objective will give us very informative data points for preference-based GP regression and improve data-efficiency.

Previously, we have shown in Section~\ref{sec:04_02_information_gain} for the parametric reward models that using a mutual information based formulation accelerates the learning whereas volume removal based methods (as in Section~\ref{sec:04_01_volume_removal}) rely on local optima and can produce trivial queries that compare the exact same trajectory and so gives no information. In the following, we show our formulation in this section also does not suffer from this trivial query problem.

\begin{remark}
The trivial query $\query = \{\trajectoryFeaturesFunction^{(1)},\trajectoryFeaturesFunction^{(1)}\}$ does not maximize our acquisition function given in \eqref{eq:04_03_active_query_generation}, and is in fact a global minimizer.
\end{remark}
\begin{proof}
For the query $\query = \{\trajectoryFeaturesFunction^{(1)}, \trajectoryFeaturesFunction^{(1)}\}$, we re-write \eqref{eq:04_03_active_query_generation} as 
\begin{align*}
    \binaryEntropyFunction\left(\standardNormalCdf\left(\frac{\gpPosteriorMean^{(1)}-\gpPosteriorMean^{(1)}}{\sqrt{2\preferenceNoiseStd^2 + \gpGFunction(\trajectoryFeaturesFunction^{(1)},\trajectoryFeaturesFunction^{(1)})}}\right)\right) - \gpMFunction\left(\trajectoryFeaturesFunction\right) &= 1 - \gpMFunction(\trajectoryFeaturesFunction^{(1)}, \trajectoryFeaturesFunction^{(1)})
\end{align*}
where $\mathrm{Var}\left(\gpRewardFunction(\trajectoryFeaturesFunction^{(1)})\right)=\mathrm{Cov}\left(\gpRewardFunction(\trajectoryFeaturesFunction^{(1)}),\gpRewardFunction(\trajectoryFeaturesFunction^{(1)})\right)$, and so $\gpGFunction(\trajectoryFeaturesFunction^{(1)},\trajectoryFeaturesFunction^{(1)})=0$, and
\begin{align*}
    \gpMFunction\left(\trajectoryFeaturesFunction\right) = \frac{\sqrt{\pi\ln(2)\preferenceNoiseStd^2}\exp\left(-\frac{(\gpPosteriorMean^{(1)} - \gpPosteriorMean^{(1)})^2}{\pi\ln(2)\preferenceNoiseStd^2 + 2\gpGFunction(\trajectoryFeaturesFunction^{(1)},\trajectoryFeaturesFunction^{(1)})}\right)}{\sqrt{\pi\ln(2)\preferenceNoiseStd^2 + 2\gpGFunction(\trajectoryFeaturesFunction^{(1)},\trajectoryFeaturesFunction^{(1)})}}
    = 1
\end{align*}
which makes the objective value $0$. Since the mutual information has to be nonnegative, this completes the proof that the trivial query is a global minimizer of the objective.
\end{proof}

We now proceed with our simulations and experiments on GP regression using actively collected pairwise comparison feedback.

\subsection{Experiments} \label{subsec:04_03_gp_experiments}

\subsubsection{Simulation Experiments}
\begin{figure}[t]
	\includegraphics[width=0.45\textwidth]{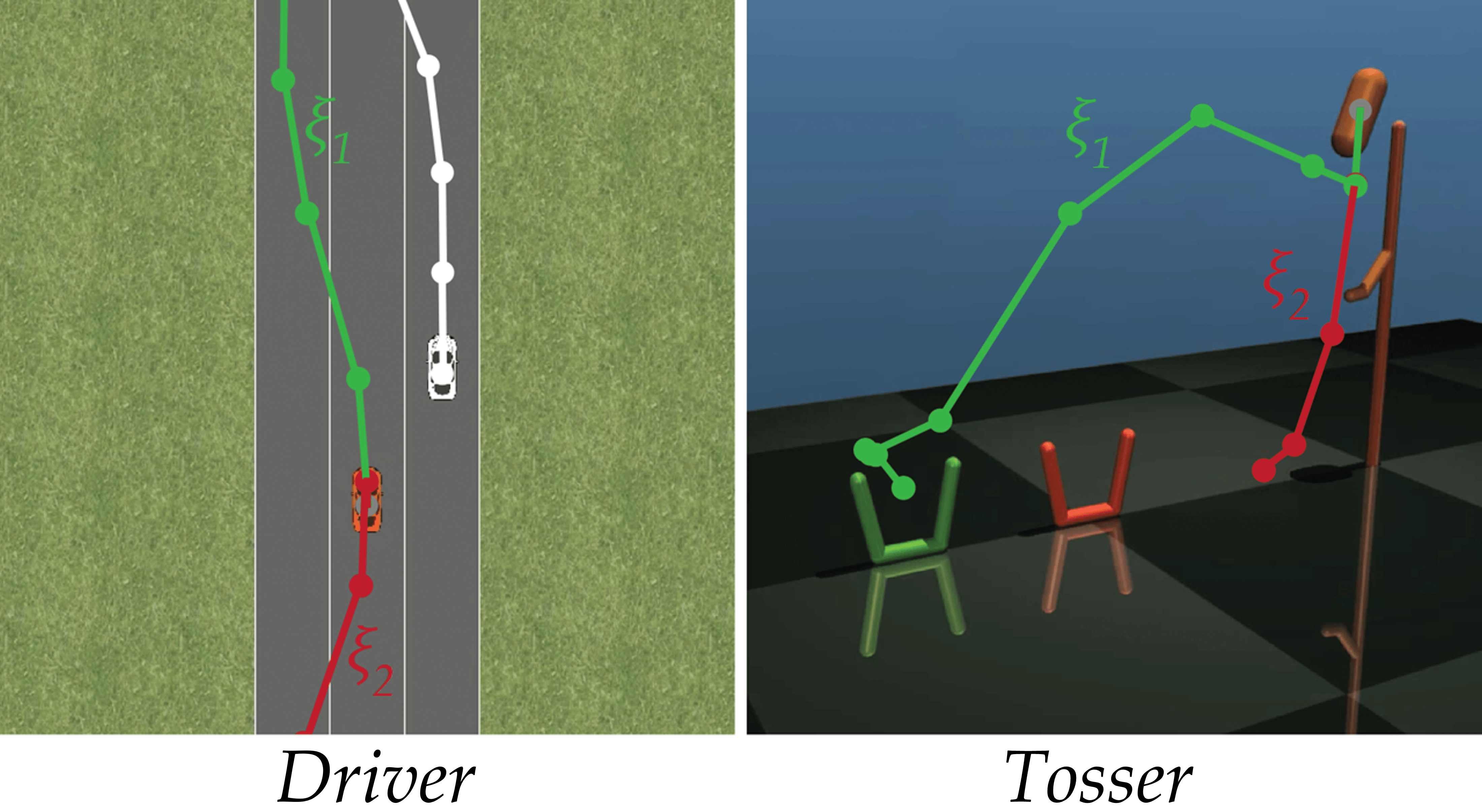}
	\centering
	\caption{Sample trajectories are shown for the two simulation environments. In \emph{Driver}, another car is cutting in front of the ego vehicle. In \emph{Tosser}, the robot must hit the dropping capsule such that it will fall into one of the baskets.
	}
	\label{fig:04_03_rss20_simulation_visuals}
\end{figure}
In this subsection, we present our experiments in two simulation domains to demonstrate how (i) GP rewards improve expressiveness over linear reward functions (which is often assumed as we did in Sections~\ref{subsec:04_01_experiments} and \ref{subsec:04_02_experiments}, also see \cite{abbeel2004apprenticeship,ng2000algorithms,ziebart2008maximum}), and (ii) active query generation improves data-efficiency over random querying.

\noindent \textbf{Environments.} To validate our framework on robotics tasks, we used two simulation environments from Section~\ref{subsec:04_02_experiments} with slight modifications: a 2D \emph{Driver} simulation \cite{sadigh2016planning} and a MuJoCo \cite{todorov2012mujoco} environment to simulate a \emph{Tosser} robot that tries to throw an object into a basket. For reader's convenience, we again show visuals from these environments with sample trajectories in Figure~\ref{fig:04_03_rss20_simulation_visuals}. For example in \emph{Driver}, the user is asked whether they would move forward or backward in the given scenario. While the users would have a common response to this query, some questions may differ among the users. For instance in \emph{Tosser}, the query asks the user whether to throw the ball into the green basket or to drop it instead. Depending on the users' preferences about the green basket, different users may have different responses.

\begin{figure*}[ht]
	\includegraphics[width=\textwidth]{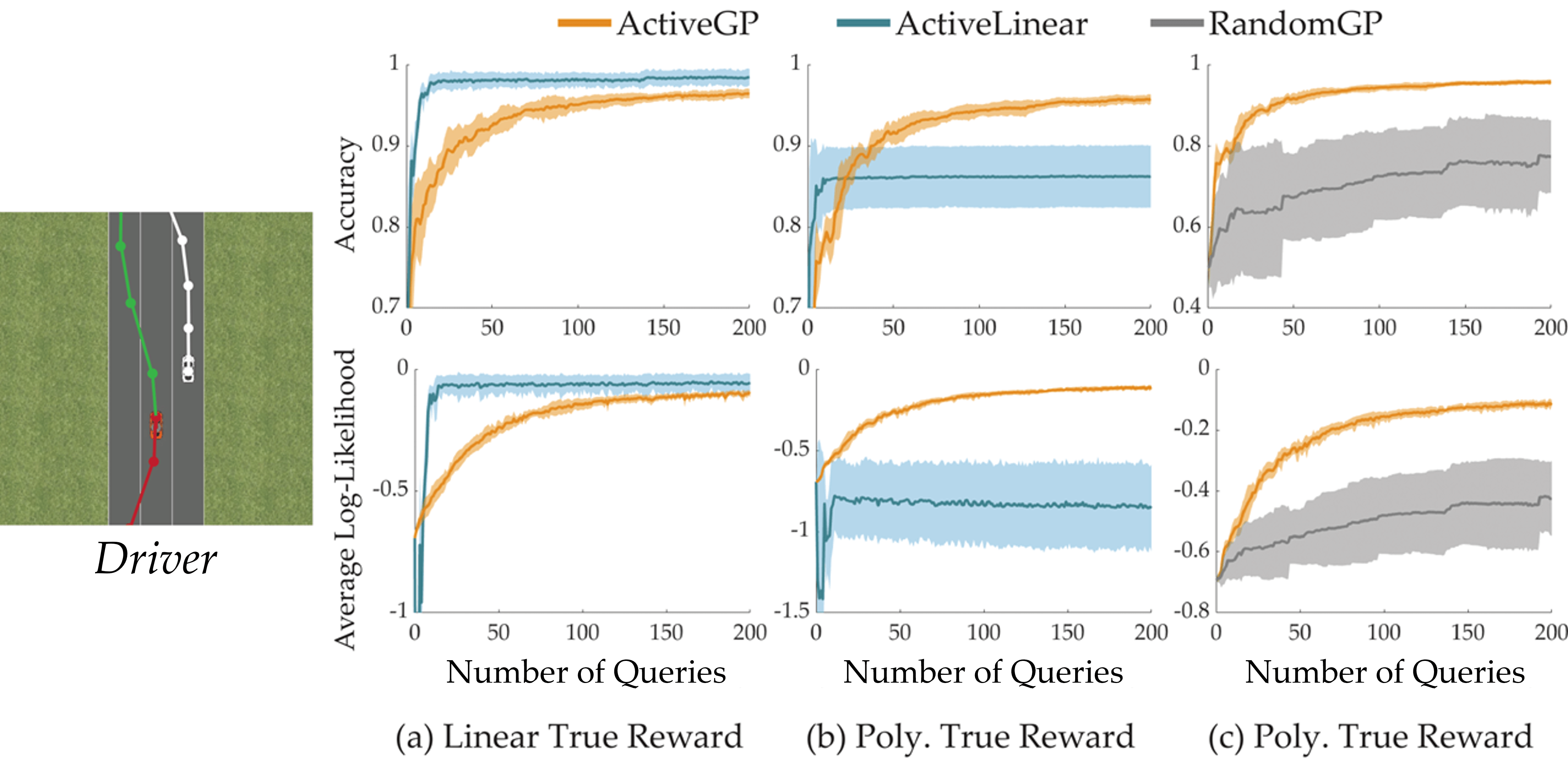}
	\centering
	\caption{Accuracies and average log-likelihoods for test set queries are shown for the \emph{Driver} environment (mean$\pm$std over $5$ runs). \textbf{(a)} Expressiveness results when the true underlying reward function is linear. \textbf{(b)} Expressiveness results when the true underlying reward function is a degree-of-two polynomial. \textbf{(c)} Data-efficiency results that compare \textsc{ActiveGP} with \textsc{RandomGP}. Accuracies and average log-likelihoods for test set queries are shown (mean$\pm$std). Active query generation improves data-efficiency over random querying in both tasks. This can be seen through both accuracy and log-likelihood.}
	\label{fig:04_03_gp_driver_results}
\end{figure*}

In these two environments, we use the following simple features for the function $\trajectoryFeaturesFunction$:
\begin{itemize}[nosep]
	\item \emph{Driver}: Distance to the other car, speed, heading angle, distance to the closest lane center.
	\item \emph{Tosser}: The maximum horizontal range, and the number of capsule flips.
\end{itemize}
In contrast to the other sections and the prior work, here we do not need to fine-tune the feature hyperparameters to learn the reward functions because GPs can effectively capture nonlinearities.

\noindent \textbf{Simulated Human Model.} We simulated human responses with an underlying true reward function $\trajectoryRewardFunction$ with some Gaussian noise, in accordance with the probit model presented in Equation~\eqref{eq:03_02_human_model}. We modeled the true $\trajectoryRewardFunction$ as either a degree-of-two polynomial or a linear function. In both cases, we selected the parameters of true $\trajectoryRewardFunction$ as i.i.d. random samples from the standard normal distribution. We repeated each simulation experiment $5$ times with varying underlying true reward functions.

\noindent \textbf{Baselines.} For our analyses, we compared three methods:
\begin{itemize}
	\item \textsc{RandomGP}: The reward is modeled using a Gaussian process. The two distinct trajectories selected in each training query are sampled from a training dataset uniformly at random.
	\item \textsc{ActiveLinear}: The reward is modeled as a linear combination of features, and the active query generation method of Section~\ref{sec:04_02_information_gain} selects the most informative comparison queries at every step of training.
	\item \textsc{ActiveGP}: The reward is modeled as a Gaussian process. We will use our active query generation method to generate the most informative comparison queries to efficiently learn the reward function.
\end{itemize}

We generated a training dataset of trajectories with uniformly randomly selected actions, as in Section~\ref{subsec:04_02_experiments}. At every iteration of \textsc{ActiveGP} and \textsc{ActiveLinear}, we computed the mutual information of each possible query from this dataset to select the most informative query. This approach decreases the computation time compared to solving a continuous optimization over all possible trajectories as it was done in Section~\ref{subsec:04_01_experiments} and by \cite{sadigh2017active,palan2019learning}.

\noindent\textbf{Evaluation.} We compare GP reward with linear reward in terms of \emph{expressiveness} (\textsc{ActiveGP} vs. \textsc{ActiveLinear}), and compare active query generation with random querying baseline in terms of \emph{data-efficiency} (\textsc{ActiveGP} vs. \textsc{RandomGP}).

\noindent\textbf{Test Set Generation.}
For both analyses on the expressiveness and data-efficiency, we also generated test sets of trajectories from the same distribution as the training set. However, it would not be fair to use the test set as is. Obtained with uniformly random action sequences, the majority of the training set is uninteresting trajectories, e.g. the ego car moves slightly forward and backward (similar to a random walk) in \emph{Driver}, or the robot does not hit the capsule in \emph{Tosser}. Using the test set without further modifications would mean we give more importance to these uninteresting behaviors as they form the majority in the datasets. Obviously, this is not the case. We want to learn the reward function everywhere in the dynamically feasible region with equal importance.

\begin{figure}[ht]
	\includegraphics[width=0.42\textwidth]{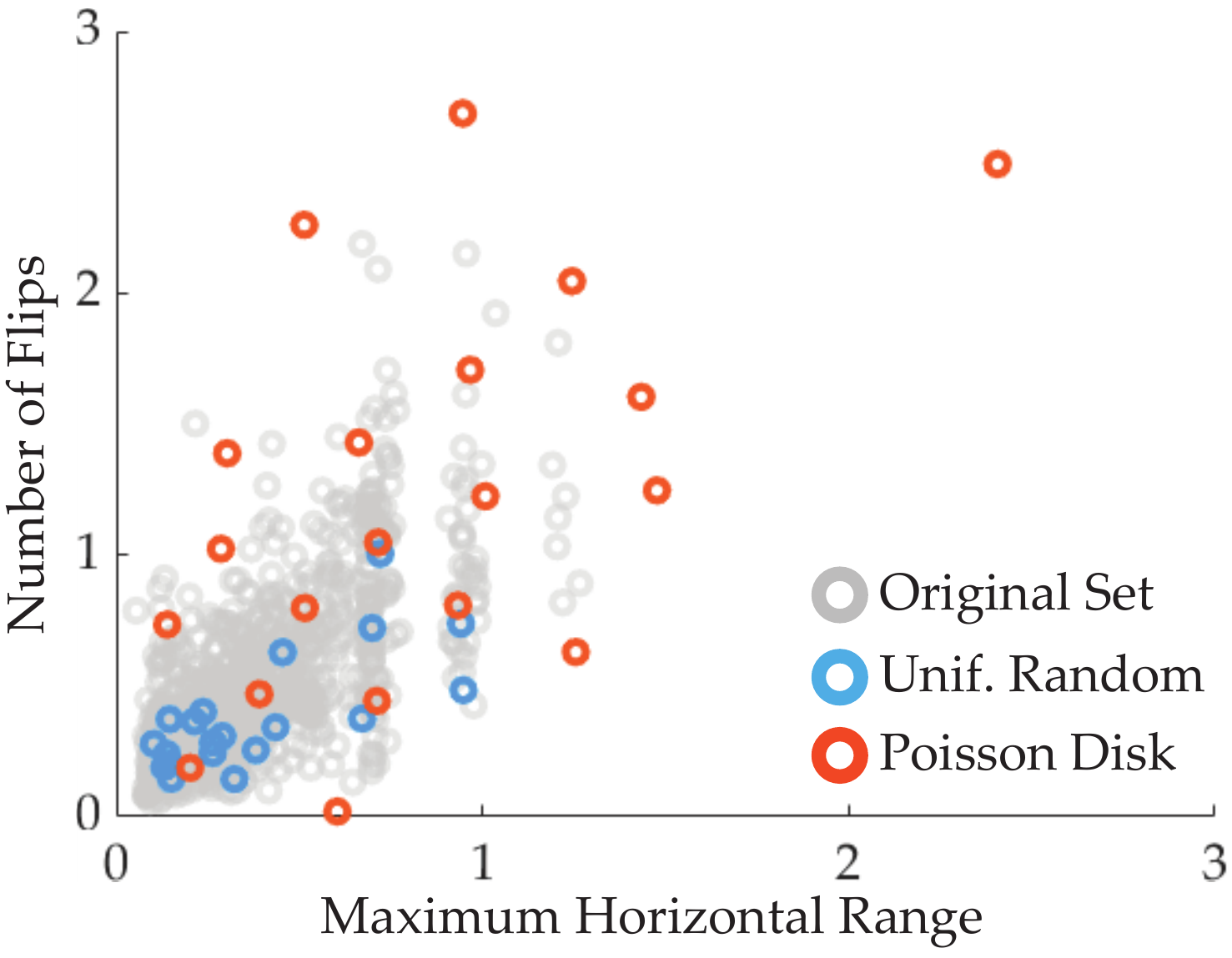}
	\centering
	\caption{Features of $1000$ \emph{Tosser} trajectories are visualized in two-dimensional plane (gray). Poisson disk sampling allows us to obtain a diverse set of $20$ samples (orange), whereas sampling uniformly at random yields mostly uninteresting trajectories (blue).}
	\label{fig:04_03_gp_poisson}
\end{figure}

Hence, we adopted Poisson disk sampling \cite{bridson2007fast} to get a diverse set of trajectories from the test set. Poisson disk sampling makes sure the difference between trajectories\footnote{We used $L_2$ distance between the feature vectors.} is above some threshold by rejecting the samples that violate this constraint. A small example set of samples is compared to uniformly random samples in Figure~\ref{fig:04_03_gp_poisson} for the \emph{Tosser} environment.

After obtaining the diverse test set, we stored the true (noiseless) response of the simulated user for each possible query in this set. For the analysis on expressiveness, we computed the accuracy and the log-likelihood of the true responses under the reward functions that are learned with $\abs{\comparisonDataset}$ actively chosen queries (up to $\abs{\comparisonDataset}=200$). For data-efficiency analysis, we again used the true human responses to the queries in the diverse test set (only from the polynomial reward functions) to calculate the accuracy and the log-likelihood under the learned reward functions.

\begin{figure*}[ht]
	\includegraphics[width=\textwidth]{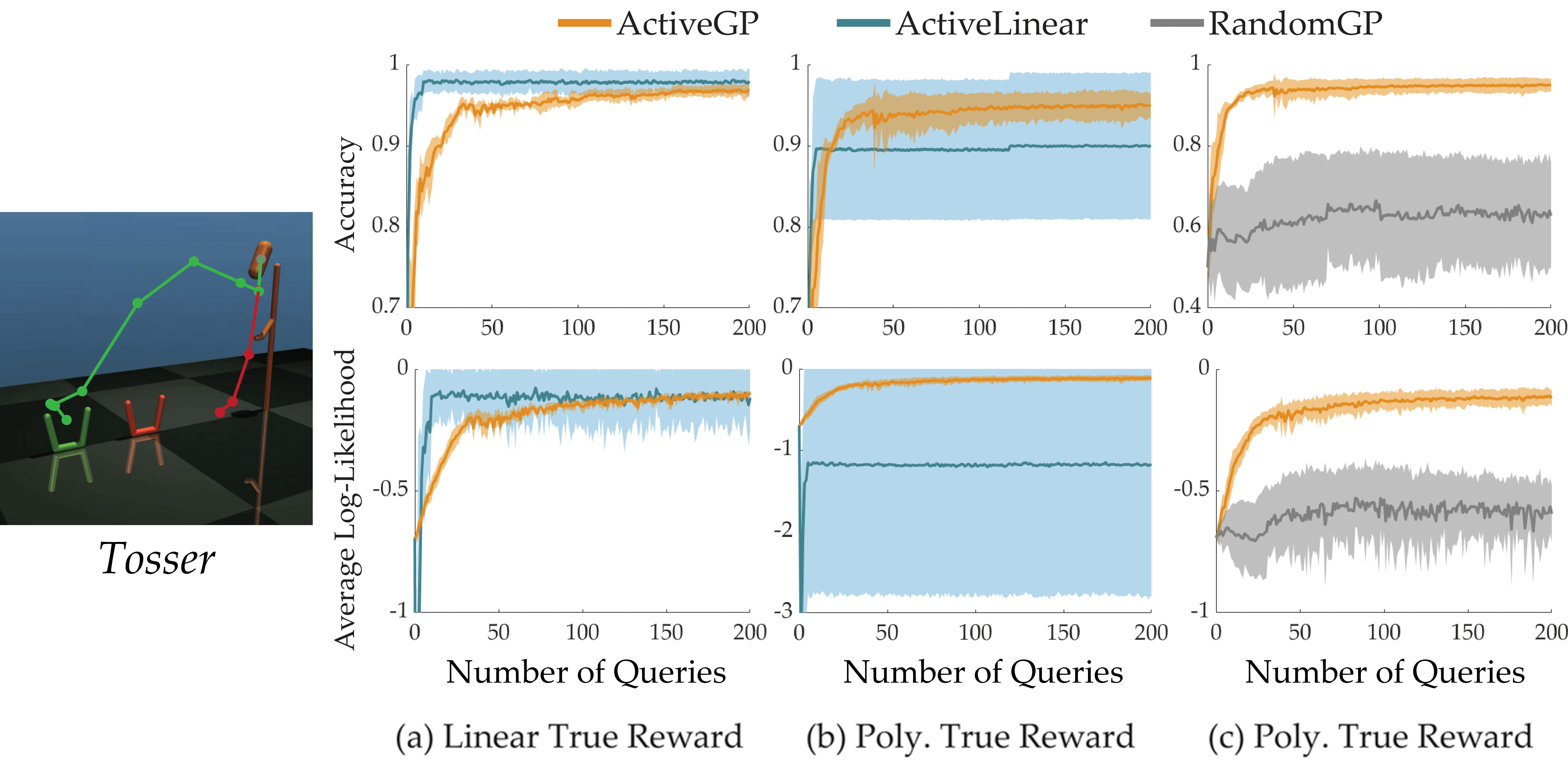}
	\centering
	\caption{Accuracies and average log-likelihoods for test set queries are shown for the \emph{Tosser} environment (mean$\pm$std over $5$ runs). \textbf{(a)} Expressiveness results when the true underlying reward function is linear. \textbf{(b)} Expressiveness results when the true underlying reward function is a degree-of-two polynomial. \textbf{(c)} Data-efficiency results that compare \textsc{ActiveGP} with \textsc{RandomGP}. Accuracies and average log-likelihoods for test set queries are shown (mean$\pm$std). Active query generation improves data-efficiency over random querying in both tasks. This can be seen through both accuracy and log-likelihood.}
	\label{fig:04_03_gp_tosser_results}
\end{figure*}

\noindent\textbf{Expressiveness.} Figures~\ref{fig:04_03_gp_driver_results}a, \ref{fig:04_03_gp_driver_results}b, \ref{fig:04_03_gp_tosser_results}a, and \ref{fig:04_03_gp_tosser_results}b show the results of expressiveness simulations. When the true reward is polynomial, the linear model results in very high variance in both accuracy and likelihood, because its performance relies on how good a linear function can explain the true nonlinear reward. In this case, the GP model captures nonlinearities better than the linear model and provides better learning (Figures~\ref{fig:04_03_gp_driver_results}b and \ref{fig:04_03_gp_tosser_results}b). When the true reward function is linear in features, a linear model naturally learns faster. However, as shown in Figures~\ref{fig:04_03_gp_driver_results}a and \ref{fig:04_03_gp_tosser_results}a, even in that case, GP model can achieve linear model's performance. To further improve the reward model, one can consider an approach to combine the linear and GP models by keeping a belief distribution over whether the true reward is linear or not, and actively querying the user according to this belief. We leave this extension as future work.

\noindent\textbf{Data-Efficiency.} We then evaluated how our active query generation helps with data-efficiency. Figures~\ref{fig:04_03_gp_driver_results}c and \ref{fig:04_03_gp_tosser_results}c compare \textsc{ActiveGP} and \textsc{RandomGP} for the simulation environments. It can be seen that active querying significantly accelerates learning over random querying. It should be noted that the number of samples taken via Poisson disk sampling matters: While choosing a very small number will increase the variance in the results, choosing a very large number will make random querying seem like it performs comparable to (or even better than) the active querying as the test set will mostly consist of uninteresting trajectories, which are also abundant in the training set, as we stated earlier.

\subsubsection{User Studies}
\begin{figure}[t]
	\includegraphics[width=0.45\textwidth]{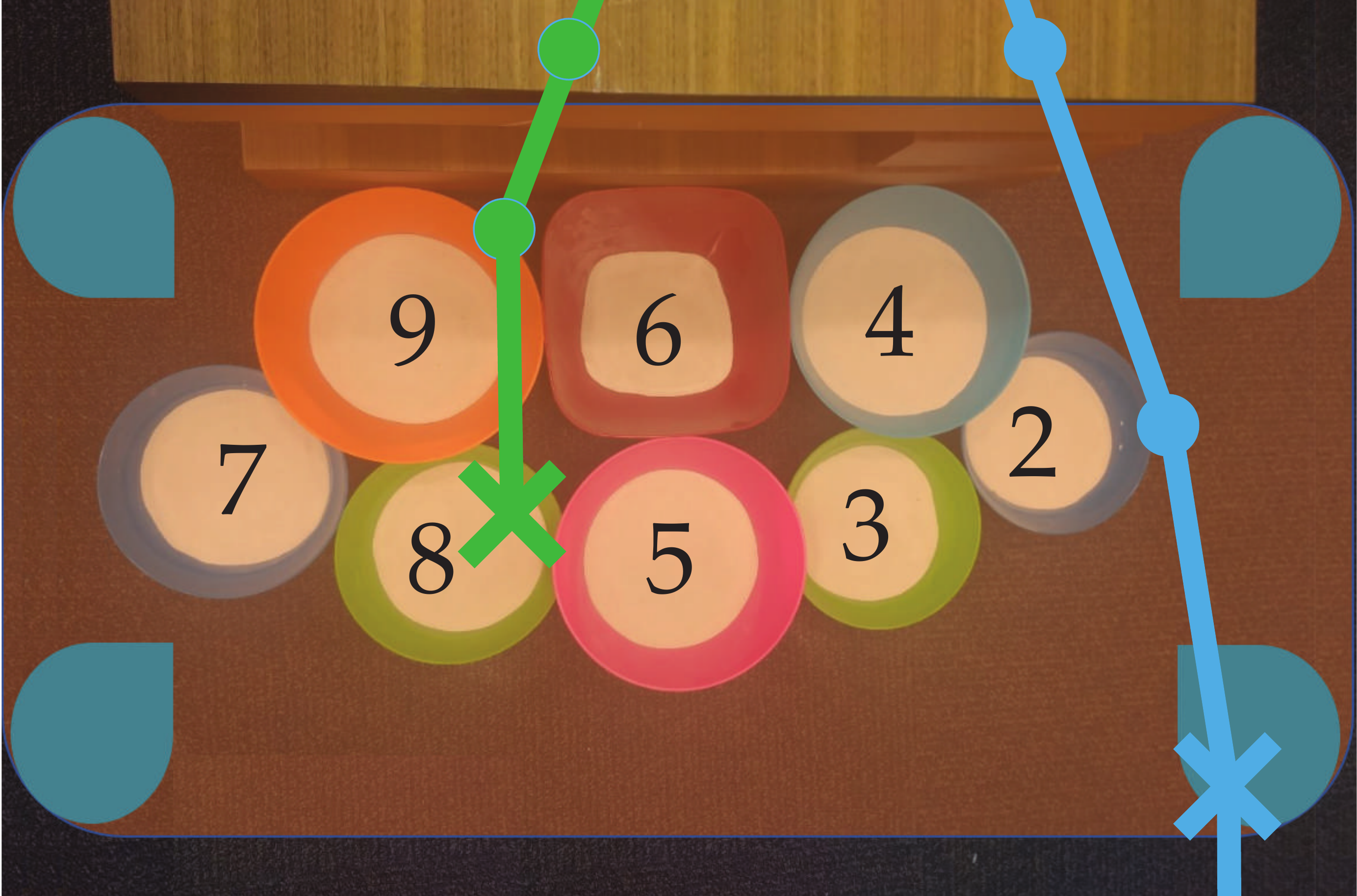}
	\centering
	\caption{Top view of the eight targets in the variant of mini-golf user study. The users assign distinct scores from $2$ to $9$ to the targets. The figure shows an example of this ranking. While the robot is capable of hitting the ball into the entire shaded region, the maximizers of a linear reward always lie near the corners of the shaded region in blue. Therefore, while the GP reward model can query the user with better trajectories (e.g. the green trajectory), the linear model only explores the boundaries (e.g. the blue trajectory that throws the ball outside of this region). Crosses show where the ball hits the ground.}
	\label{fig:04_03_gp_targets}
\end{figure}

\noindent\textbf{Experiment Setup}. We also compare our method \textsc{ActiveGP} with \textsc{ActiveLinear} and \textsc{RandomGP} on a user study with a Fetch mobile manipulator robot \cite{wise2016fetch}. In this study, the human subjects teach the Fetch robot how to play a variant of mini golf where the robot can achieve different scores by hitting the ball to different targets (see Figures~\ref{fig:03_02_gp_front_fig} and \ref{fig:04_03_gp_targets} for the setup). However, these scores are only known to the human. In fact, the robot does not even know the locations of the targets, and it tries to learn the reward as a function of its control inputs. Fixing some of the joints, we let the robot vary only its shot speed and angle, which are also the features of the reward function.

This experiment setting is interesting because a linear reward function can only encode whether the robot must hit the ball to the right or to the left, or whether it must hit with high or low speed. It cannot particularly encourage (or discourage) hitting with a modest angle and/or speed. Therefore, as we show in Figure~\ref{fig:04_03_gp_targets}, the targets that are around the middle region cannot be the maximizers of a linear reward function.

\noindent\textbf{Subjects and Procedure.} We recruited $10$ users ($6$ male, $4$ female) with an age range from $19$ to $28$. Each user first assigned their distinct scores (from $2$ to $9$) to the eight targets. The robot then queried them with $50$ pairwise comparison questions: $15$ for \textsc{ActiveGP}, $15$ for \textsc{ActiveLinear}, $15$ for \textsc{RandomGP} and $5$ queries generated uniformly at random to create a test set. We shuffled the order of queries to avoid any bias. We used the reward models, each of which is learned with $15$ queries, to predict the user responses in the test set. The prediction score on the test set provides an accuracy metric.

In addition to the accuracy, we assessed whether the robot could successfully learn how to perform a good shot. For this, after the subjects responded to $50$ queries, the robot demonstrated $3$ more trajectories each of which corresponds to the optimal trajectory of one method, the trajectory that maximizes the learned reward function. Again, the order of these trajectories was shuffled. After watching each demonstration, the subjects assigned a score to the shot from a 9-point rating scale ($1$-very bad, $9$-very good).

\noindent\textbf{Results and Discussion}
We provide a video that gives an overview of user studies and their results at \url{https://youtu.be/SLSO2lBj9Mw}.

\begin{figure*}[th]
	\includegraphics[width=\textwidth]{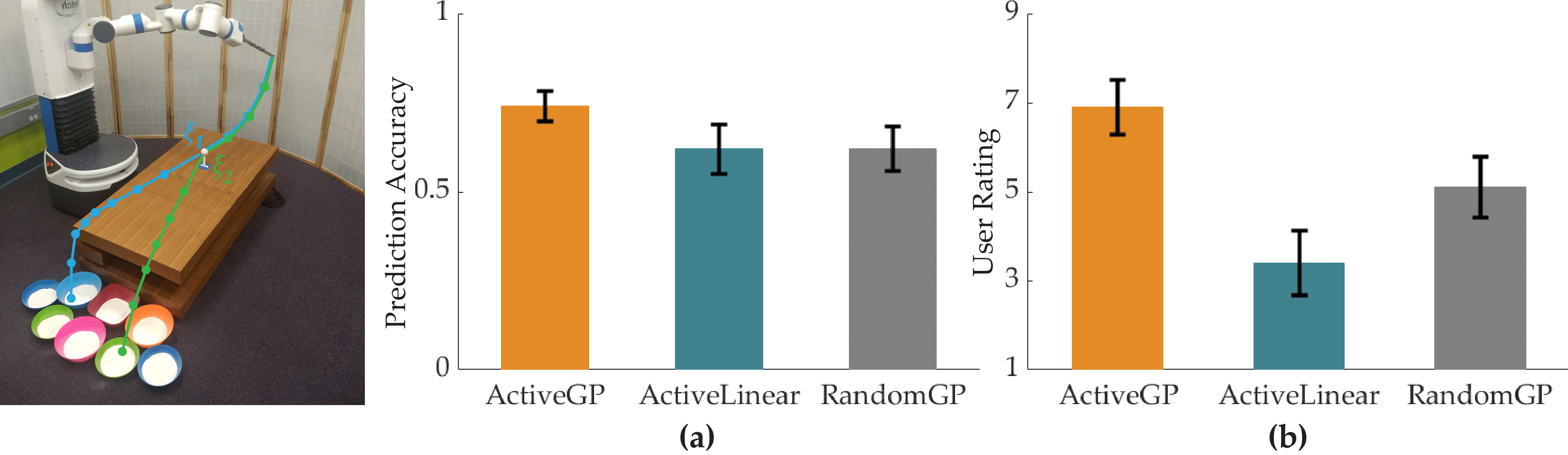}
	\centering
	\caption{\textbf{(a)} Prediction accuracy results (mean$\pm$se). Each trained with $15$ queries, \textsc{ActiveGP} achieves significantly higher prediction accuracy than both \textsc{ActiveLinear} and \textsc{RandomGP} ($p<0.05$). \textbf{(b)} User ratings on the final robot performance (mean$\pm$se). \textsc{ActiveGP} accomplishes the task significantly better than both \textsc{ActiveLinear} and \textsc{RandomGP} ($p<0.05$).}
	\label{fig:04_03_gp_fetch_results}
\end{figure*}

Figure~\ref{fig:04_03_gp_fetch_results}a shows the prediction accuracy values on the test sets collected from the subjects (averaged over the subjects). By modeling the reward using a GP and querying the users with the most informative questions, \textsc{ActiveGP} achieves significantly higher prediction accuracy ($0.74\pm0.04$, mean$\pm$se) compared to both \textsc{ActiveLinear} ($0.62\pm0.07$) and \textsc{RandomGP} ($0.62\pm0.06$) with $p<0.05$ (Wilcoxon signed rank test). The results from this user study are aligned with our simulation studies.

In reward learning, it is crucial to validate whether the learned reward function can encode the desired behavior or not. Figure~\ref{fig:04_03_gp_fetch_results}b shows the user ratings to the trajectories that the robot showed after learning the user preferences via $3$ different methods. \textsc{ActiveGP} obtains significantly higher scores ($6.9\pm0.6$) than both \textsc{ActiveLinear} ($3.4\pm0.7$) and \textsc{RandomGP} ($5.1\pm0.7$) with $p<0.05$. While \textsc{ActiveLinear} occasionally achieves high scores when the users' preferred target is near the edge, it generally fails to produce the desired behavior due to its low expressive power.

The next section will present more simulation and experiment results with active preference-based GP regression, after discussing active querying in the presence of ordinal feedback.

%% file: 04_active/04_roial.tex
\label{sec:04_04_roial}

Having seen the success of mutual information maximization based active querying for non-parametric reward functions in Section~\ref{subsec:04_03_gp_experiments}, we want to extend it in the same way as we did in Chapter~\ref{chap:learning}: we want to be able to utilize ordinal feedback in addition to pairwise comparisons. Furthermore, in Section~\ref{sec:03_03_roial}, we established that it is sometimes desired to define a ``region of avoidance" (ROA) in the trajectory space $\trajectorySpace$ such that we should not query the user with the trajectories in that space. This is especially relevant when we are actively querying the user: we should constrain our active querying optimization to avoid ROA.

Therefore in this section, we extend the GP regression model we presented in Section~\ref{sec:03_03_roial} with active querying. We again use mutual information maximization. However, differently from Section~\ref{sec:04_03_gp}, we now select the trajectories such that:
\begin{enumerate}[nosep]
    \item Each trajectory is compared with the previous trajectory in the querying sequence (as opposed to optimizing for a couple of trajectories for pairwise comparisons),
    \item We want to maximize the information from not only the comparative feedback, but also the ordinal feedback,
    \item We try to avoid querying the user with trajectories from ROA.
\end{enumerate}
We call our algorithm ROIAL, short for region of interest active learning.

At the end of this section, we will demonstrate in simulation that ROIAL estimates both the region of interest (ROI) and the reward function within the ROI with high accuracy. We experimentally demonstrate ROIAL on the lower-body exoskeleton Atalante (Figure~\ref{fig:03_03_Atalante}) to learn the reward functions of three non-disabled users over four gait parameters. The obtained landscapes highlight both agreement and disagreement in preferences among the users. Previous algorithms for exoskeleton gait optimization were incapable of drawing such conclusions; thus, this work represents progress towards establishing a better understanding of the science of walking with respect to exoskeleton gait design.

\subsection{Formulation}
ROIAL selects samples by modeling a Bayesian posterior over the reward function using Gaussian processes and maximizing the mutual information (over the ROI) with respect to this posterior. In this section, we continue to use the notation in Section~\ref{sec:03_03_roial}, as we are now making it active.

Defining $\bm{\hat{\gpRewardFunction}}^{(i)} := [\hat{\gpRewardFunction}^{(i)}(\trajectoryFeaturesFunction(\trajectory^{(1)})), \dots, \hat{\gpRewardFunction}^{(i)}(\trajectoryFeaturesFunction(\trajectory^{(\abs{\trajectorySpace})}))]^\top$ as the maximum a posteriori (MAP) estimate of the rewards $\bm{\gpRewardFunction}$ given $\dataset^{(i)}$, we aim to adaptively select the next trajectories $\trajectory^{(1)}, \trajectory^{(2)}, \ldots \in \trajectorySpace$ that minimize the error in estimating $\bm{\gpRewardFunction}$ over the ROI. We model the error as $\text{Error}(i) := \sum_{\trajectory \in \text{ROI}} \lvert \bm{\gpRewardFunction}-\bm{\hat{\gpRewardFunction}}^{(i)} \rvert$, where the absolute value is taken element-wise.

\subsubsection{Trajectory Selection via Mutual Information Maximization}
To learn the reward function in as few trials as possible, we select trajectories to maximize the mutual information between the reward function and the comparison-based and ordinal human feedback. We again adopt the greedy approach as in the previous sections to solve the following optimization in each iteration $i$:
\begin{align}
    \max_{\trajectory^{(i)}\in \textrm{ROI}^{(i)}} & \:I(\bm{\gpRewardFunction} ; \queryResponse_o^{(i)},\queryResponse^{(i)} \mid \dataset^{(i-1)}, \trajectory^{(i)}),\label{eq:04_04_infogain}
\end{align}
where $\queryResponse^{(i)}$ denotes the user's response to a pairwise comparison query between $\trajectory^{(i)}$ and $\trajectory^{(i-1)}$, and $\queryResponse^{(i)}_o$ denotes the ordinal feedback for trajectory $\trajectory^{(i)}$. One can re-write \eqref{eq:04_04_infogain} in terms of information entropy:
\begin{align*}
    \max_{\trajectory^{(i)}} H(\queryResponse_o^{(i)},\queryResponse^{(i)} \mid \dataset^{(i - 1)}, \trajectory^{(i)}) - \mathbb{E}_{\bm{\gpRewardFunction} \mid \dataset^{(i - 1)}} \left[H(\queryResponse_o^{(i)},\queryResponse^{(i)} \mid \dataset^{(i - 1)}, \trajectory^{(i)}, \bm{\gpRewardFunction})\right].
\end{align*}
Again, we can interpret the first term as the uncertainty about trajectory $\trajectory^{(i)}$'s ordinal label and preference relative to $\trajectory^{(i-1)}$. We aim to maximize this term, because queries with high model uncertainty could potentially yield significant information. The second term is conditioned on $\bm{\gpRewardFunction}$, and so represents the user's expected uncertainty. If the user is very uncertain about their feedback, then the action $\trajectory^{(i)}$ gives only a small amount of information. Hence, we aim to minimize this second term. In this way, mutual information maximization produces queries that are both informative and easy for users.

The second term is estimated via sampling from the Laplace-approximated Gaussian posterior $P(\bm{\gpRewardFunction}\mid\dataset^{(i - 1)})$. Computing the first term requires the probability $P(\queryResponse_o^{(i)},\queryResponse^{(i)} \mid \dataset^{(i-1)},\trajectory^{(i)})$. We derive it as:
\begin{align}
    P(\queryResponse_o^{(i)},\queryResponse^{(i)} \mid \dataset^{(i-1)},\trajectory^{(i)}) &= \int_{\mathbb{R}^{\abs{\trajectorySpace}}} P(\bm{\gpRewardFunction} \mid \dataset^{(i-1)},\trajectory^{(i)})P(\queryResponse_o^{(i)}, \queryResponse^{(i)} \mid \dataset^{(i-1)},\trajectory^{(i)}, \bm{\gpRewardFunction})d\bm{\gpRewardFunction} \noindent\\
    &= \mathbb{E}_{\bm{\gpRewardFunction} \mid \dataset^{(i-1)}}\left[P(\queryResponse_o^{(i)}, \queryResponse^{(i)} \mid \dataset^{(i-1)},\trajectory^{(i)},\bm{\gpRewardFunction})\right],
\end{align}
which we approximate with samples from $P(\bm{\gpRewardFunction} \mid \dataset^{(i-1)})$.

\subsubsection{ROIAL Algorithm}
Algorithm~\ref{alg:04_04_ROIAL} presents the pseudocode for the ROIAL algorithm we develop. Line 8 solves the mutual information maximization problem, whereas the procedures for lines 5-7, which include learning and estimating ROI, were presented in Section~\ref{sec:03_03_roial}.

\begin{algorithm}[tb]
\caption{ROIAL Algorithm}
\begin{algorithmic}[1]
\Require{Reward prior parameters; ordinal thresholds $\ordinalThresholdSet_1, \ldots, \ordinalThresholdSet_{\abs{\ordinalThresholdSet}}$; subset size $\abs{\trajectorySpace_S^{(i)}}$ for $\forall i$; confidence parameter $\roialConservatism$}
\State $\dataset^{(0)}= \emptyset$, \Comment{$\dataset^{(i)}$: user feedback dataset including iteration $i$}
\State Select a trajectory $\trajectory^{(1)}$ at random
\State Add ordinal feedback to data to obtain $\dataset^{(1)}$
\For{i = 2,\dots}{}
 \State Update the model posterior $P(\bm{\gpRewardFunction} \mid \dataset^{(i-1)})$ \Comment{Equation~\eqref{eq:03_03_posterior}}
 \State Determine $\trajectorySpace_S^{(i)}$ by randomly selecting $\abs{\trajectorySpace_S^{(i)}}$ actions
 \State Determine $\textrm{ROI}^{(i)} \subset \trajectorySpace_S^{(i)}$
 \State $\trajectory^{(i)} \gets \argmax_{\trajectory \in \textrm{ROI}^{(i)}}  I(\bm{\gpRewardFunction} ; \queryResponse_o^{(i)}, \queryResponse^{(i)} \mid \dataset^{(i-1)},\trajectory)$
 \State Add preference and ordinal feedback to data to obtain $\dataset^{(i)}$
 \EndFor
\end{algorithmic}
\label{alg:04_04_ROIAL}
\end{algorithm}

\subsection{Simulations and Experiments}\label{subsec:04_04_experiments}
\subsubsection{Simulation Results}
We evaluate ROIAL's performance on the Hartmann3 function---which is a standard benchmark for learning non-convex, smooth functions---and on 3-dimensional synthetic functions, sampled from  a  Gaussian  process  prior  over  a $20 \times 20 \times 20$ grid. As evaluation metrics, we use the algorithm's errors in pairwise comparison and ordinal label prediction; these allow us to quantify performance when the true reward function is unknown. The average ordinal prediction error is defined as $\overline{\text{Error}}(i) := \frac{1}{i}\sum_{i' = 1}^i \abs{\hat{\queryResponse}_o^{(i')} - {\queryResponse_o^{(i')}}^*}$, and all simulations use 5 ordinal categories.\footnote{Unless otherwise stated, hyperparameters are held constant across simulations and experiments, and their values can be found in \url{https://github.com/kli58/ROIAL}.}

\begin{figure*}[tb]
     \centering
     \begin{subfigure}[b]{0.24\textwidth}
         \centering
         \includegraphics[width=\textwidth]{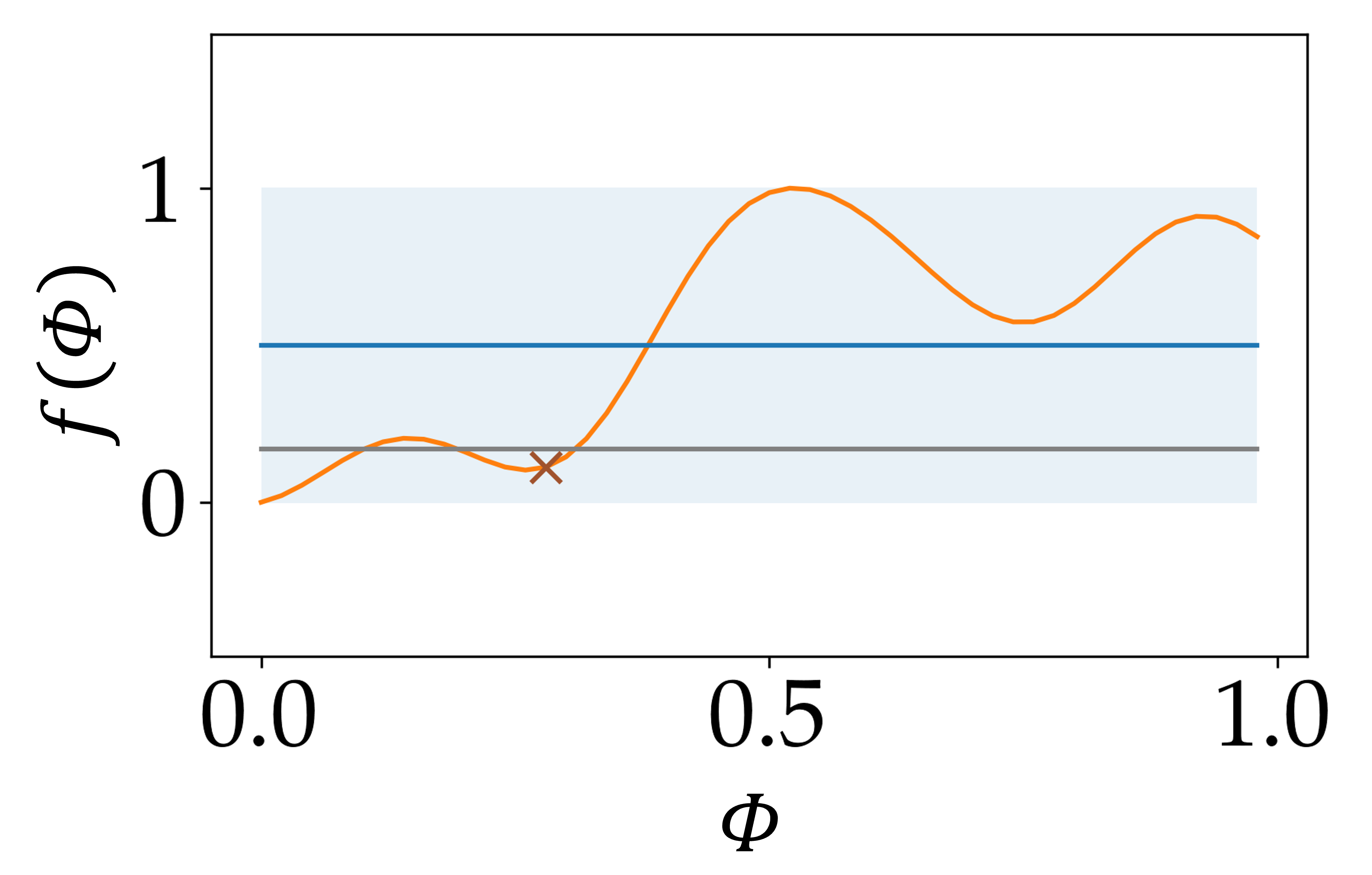}
         \caption{Iteration 1}
          \label{fig:04_04_it_1}
     \end{subfigure}
     \begin{subfigure}[b]{0.24\textwidth}
         \centering
         \includegraphics[width=\textwidth]{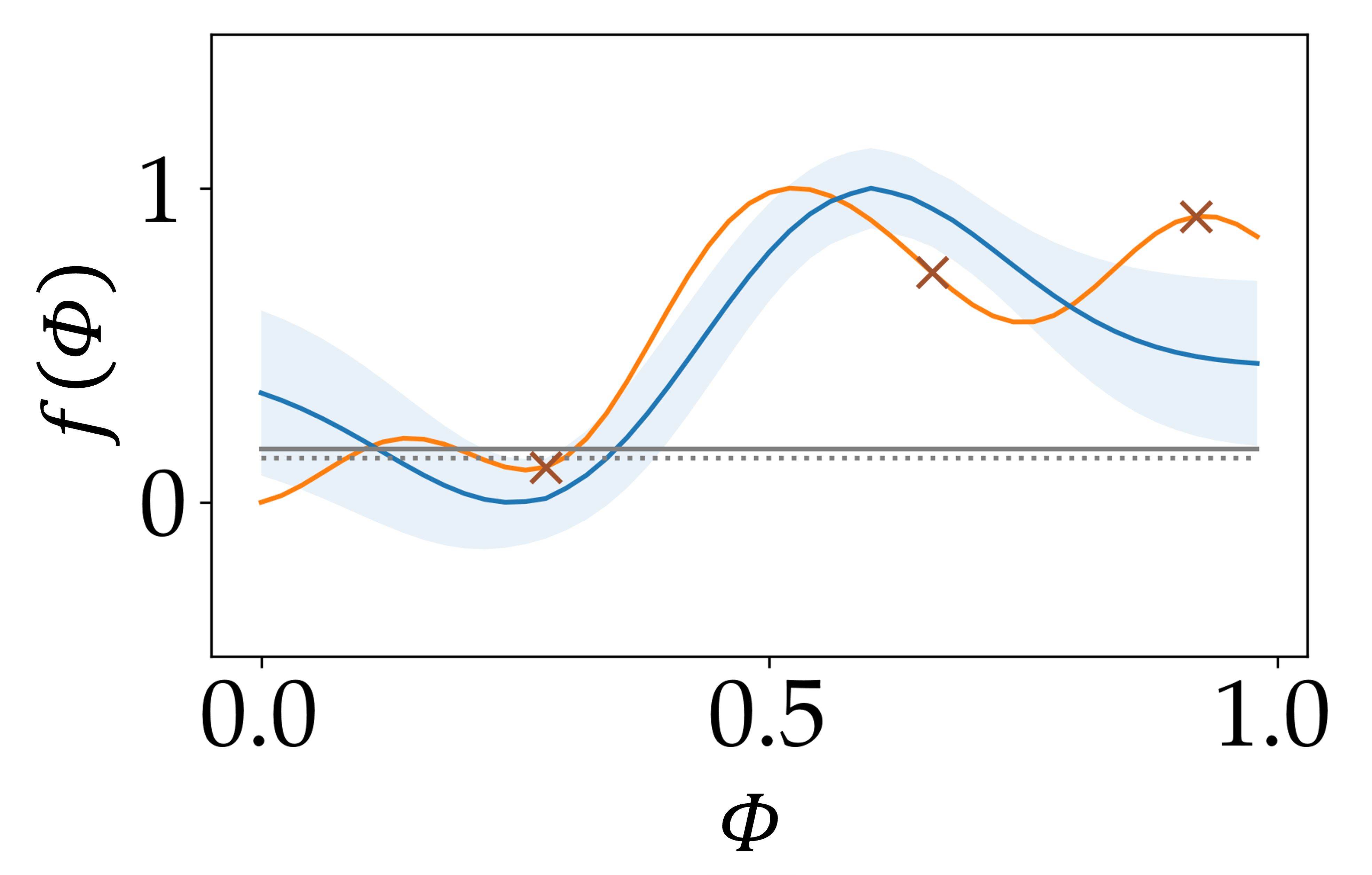}
         \caption{Iteration 3}
          \label{fig:04_04_it_3}
     \end{subfigure}
     \begin{subfigure}[b]{0.24\textwidth}
         \centering
         \includegraphics[width=\textwidth]{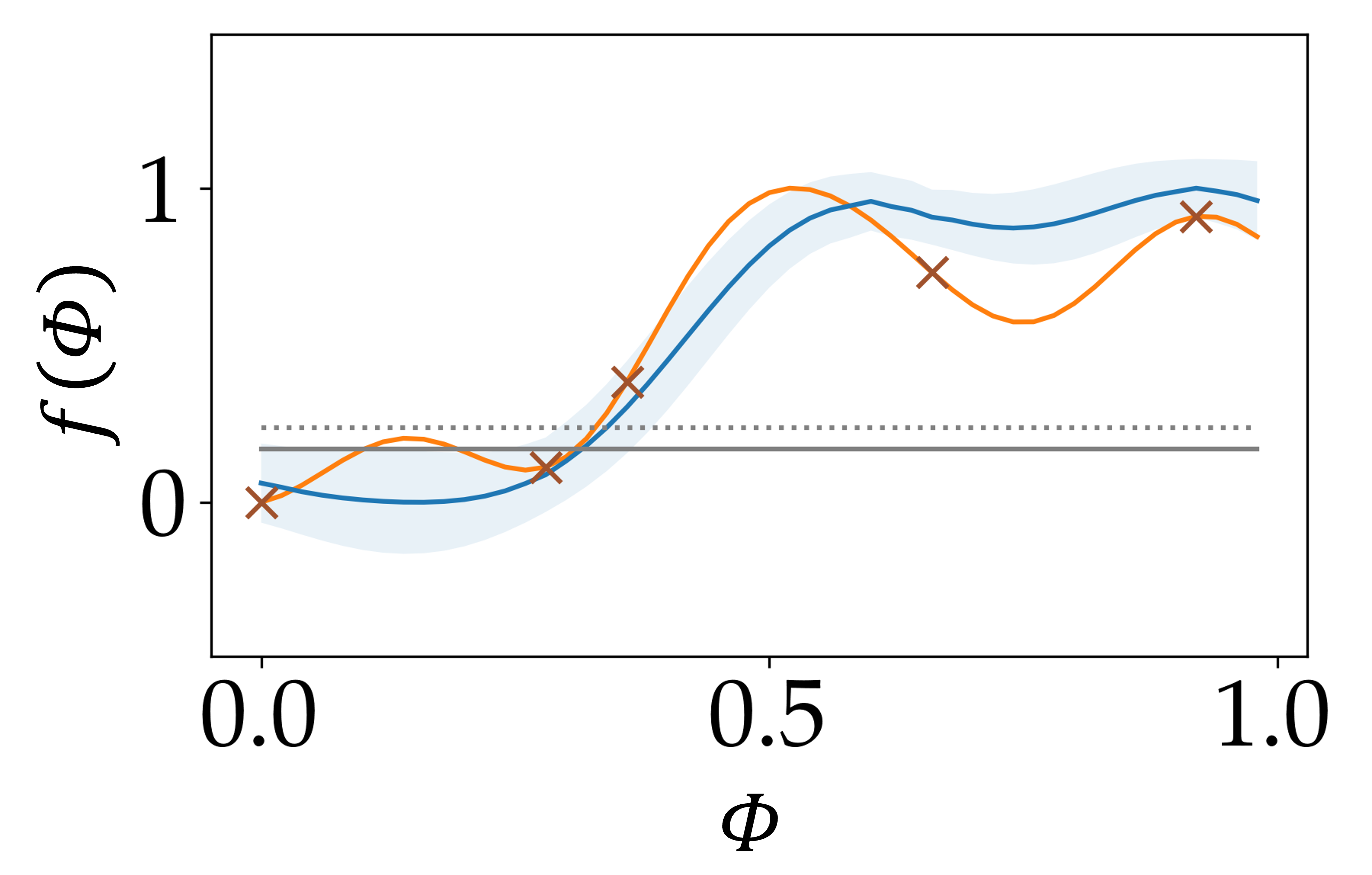}
         \caption{Iteration 5}
          \label{fig:04_04_it_5}
     \end{subfigure}
     \begin{subfigure}[b]{0.24\textwidth}
         \centering
         \includegraphics[width=\textwidth]{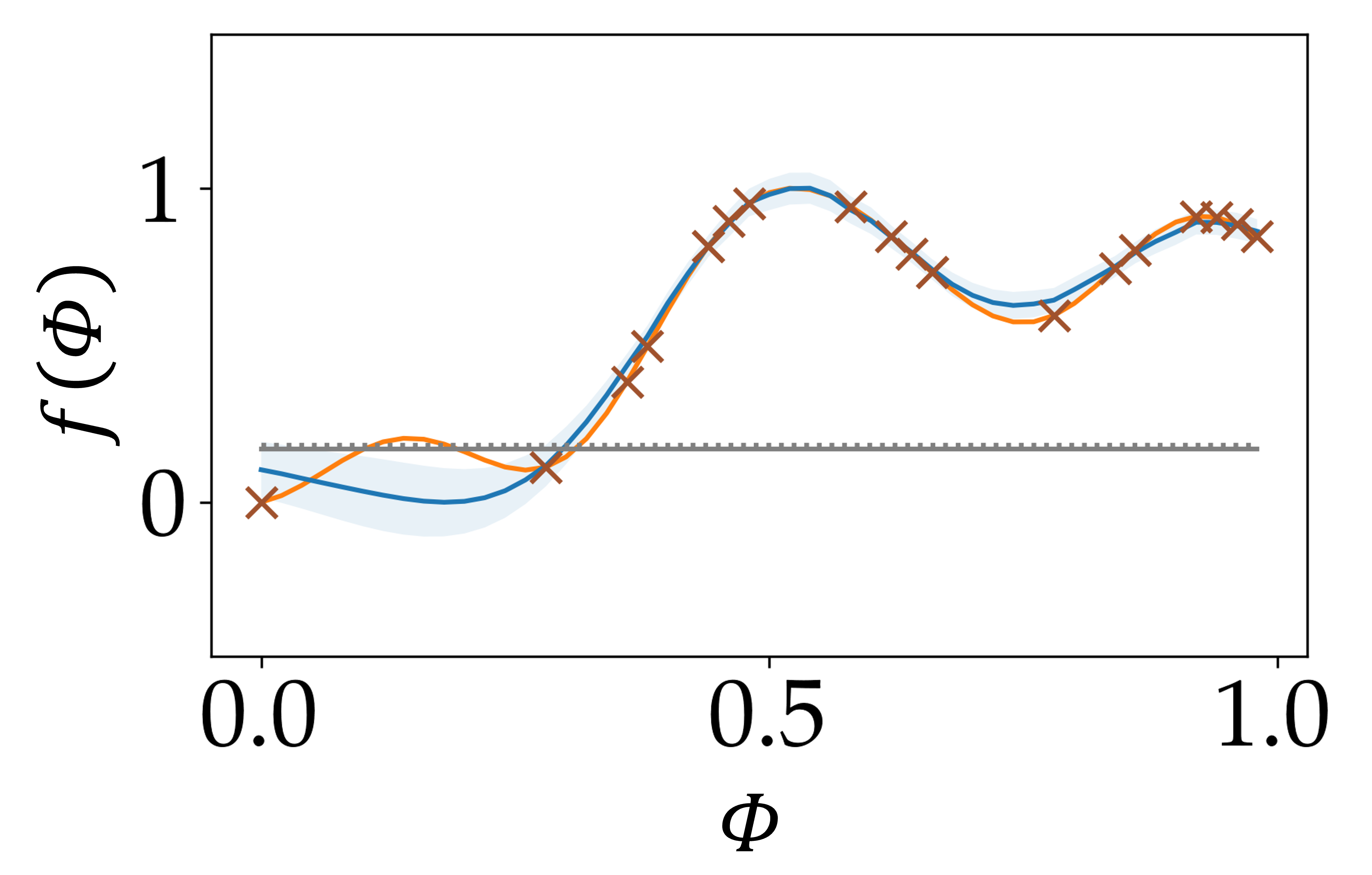}
         \caption{Iteration 20}
         \label{fig:04_04_it_20}
     \end{subfigure}
        \caption{1D posterior illustration. The true objective function is shown in orange, and the algorithm's posterior mean is blue. Blue shading indicates the confidence region for $\roialConservatism = 0.5$. The solid grey line indicates the true ordinal threshold $\ordinalThresholdSet_1$: the ROI is above this threshold, while the ROA is below it. The dotted grey line is the algorithm's $\ordinalThresholdSet_1$ hyperparameter. The actions queried so far are indicated with ``x"s. Utilities are normalized in each plot so that the posterior mean spans the range from 0 to 1.}
        \label{fig:04_04_1dposterior}
\end{figure*}

\noindent\textbf{1D illustration of ROIAL.} Figure~\ref{fig:04_04_1dposterior} illustrates the algorithm for a 1D objective (reward) function. Initially, ROIAL samples widely across the space (Figure~\ref{fig:04_04_it_1}-\ref{fig:04_04_it_5}). As seen by comparing iterations 5 and 20 (Figure~\ref{fig:04_04_it_5}-\ref{fig:04_04_it_20}), the algorithm stops querying points in the ROA (points in $\ordinalCategory_1$) because the upper confidence bound (top of the blue shaded region) there falls below the hyperparameter $\ordinalThresholdSet_1$ (dotted gray line).

\begin{figure*}
\centering
\tabskip=0pt
\valign{#\cr
  \hbox{%
    \begin{subfigure}[b]{.21\textwidth}
    \hspace{-5mm}
    \includegraphics[width=\textwidth]{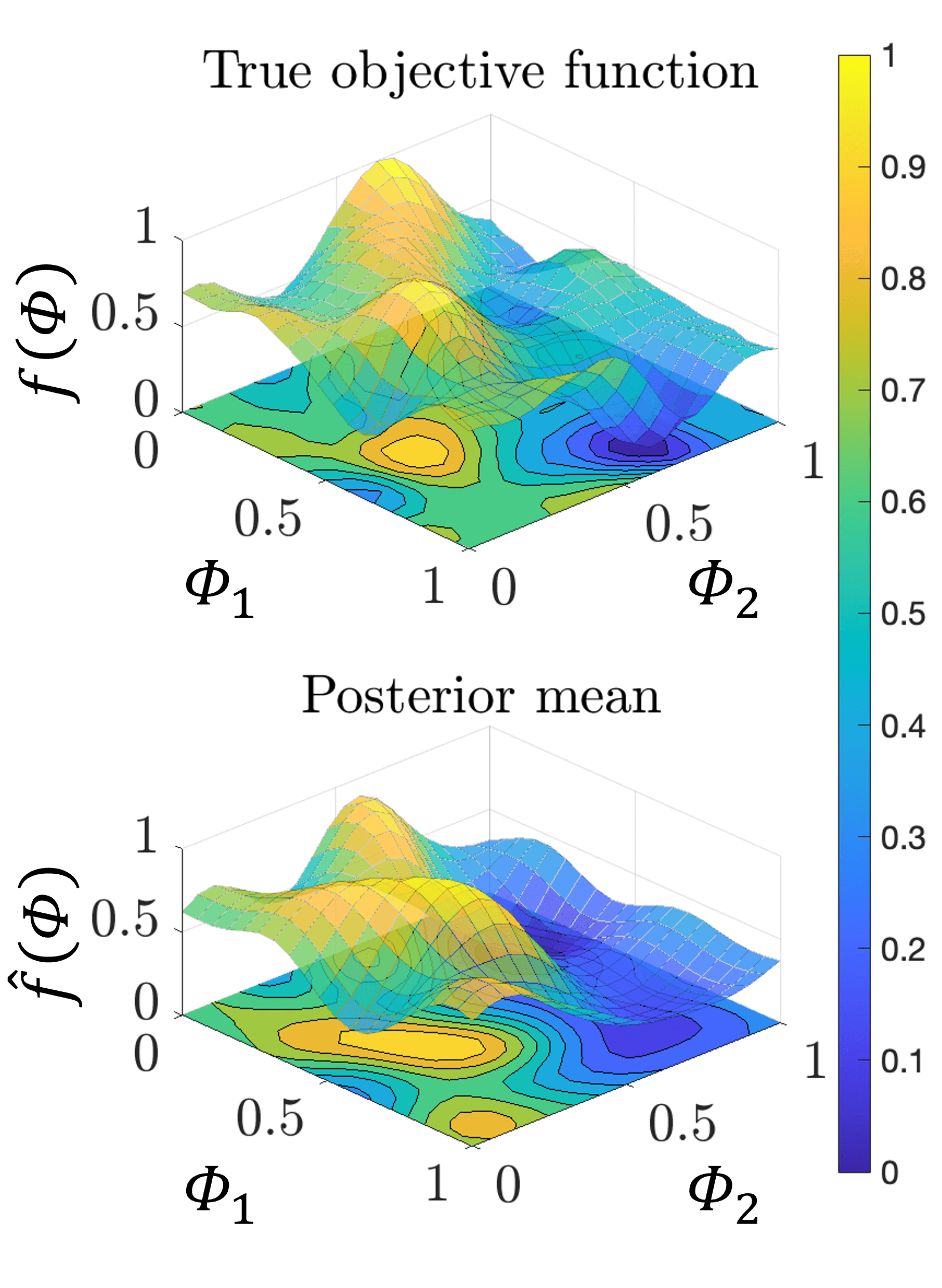}
    \caption{Synthetic function posterior}
    \label{fig:04_04_simulation_posterior}
    \end{subfigure}%
  }\cr
  \hbox{%
    \begin{subfigure}{.5\textwidth}
    \hspace{22mm}
    \includegraphics[width=\textwidth]{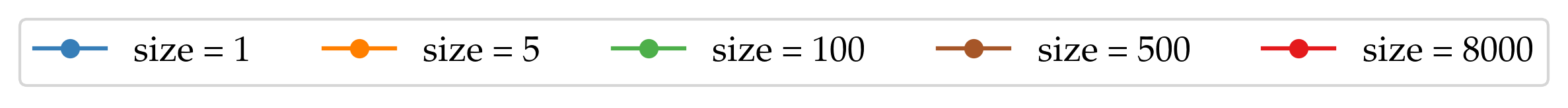}
    \end{subfigure}%
  }
  \hbox{%
    \hspace{1mm}\begin{subfigure}{.35\textwidth}
    \centering
    \includegraphics[width=\textwidth]{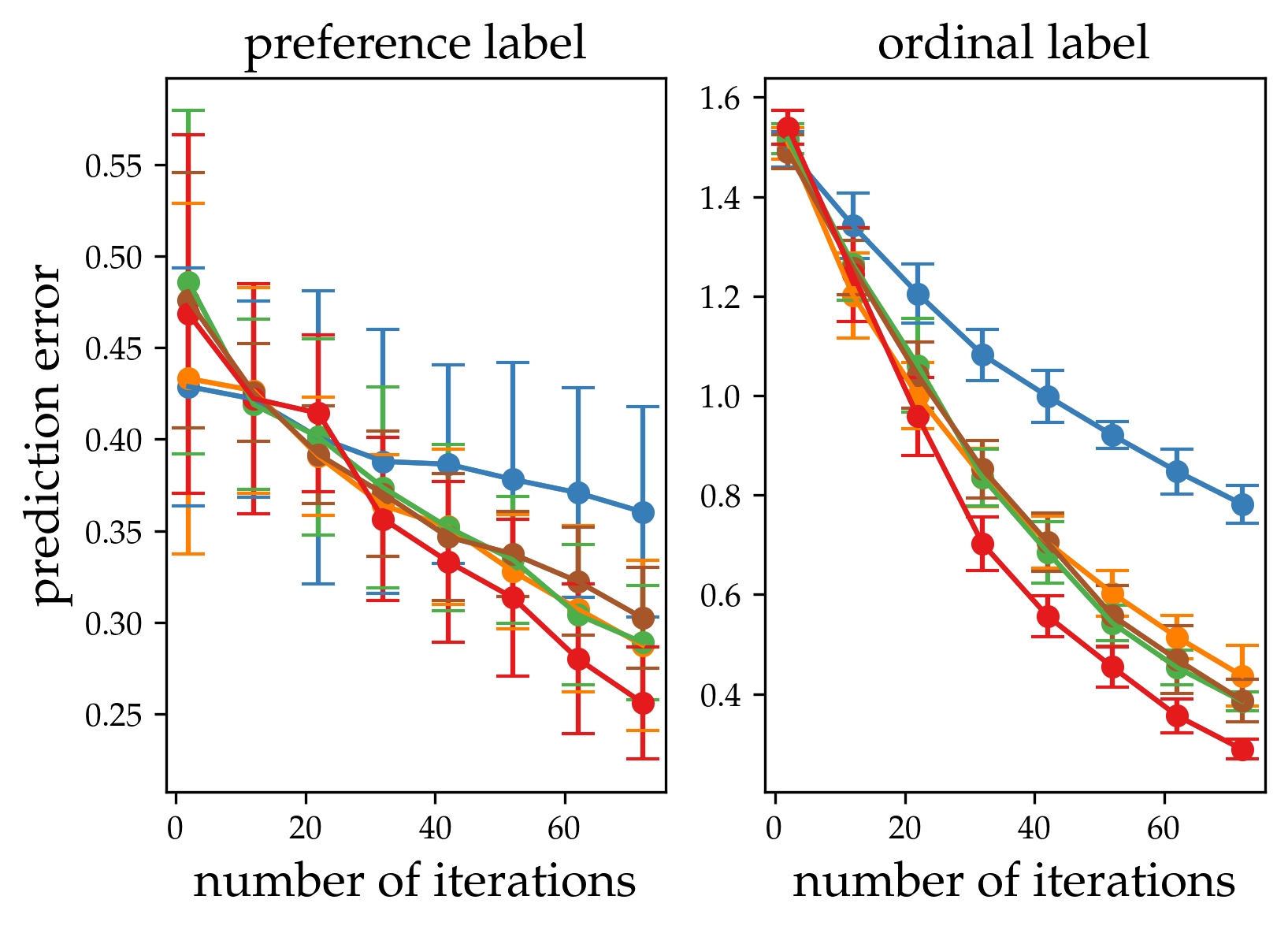}
    \caption{Hartmann3 prediction error}
    \label{fig:04_04_subset_label_hartmann3}
    \end{subfigure}%
\hspace{4mm}
  \begin{subfigure}{.35\textwidth}
    \centering
    \includegraphics[width=\textwidth]{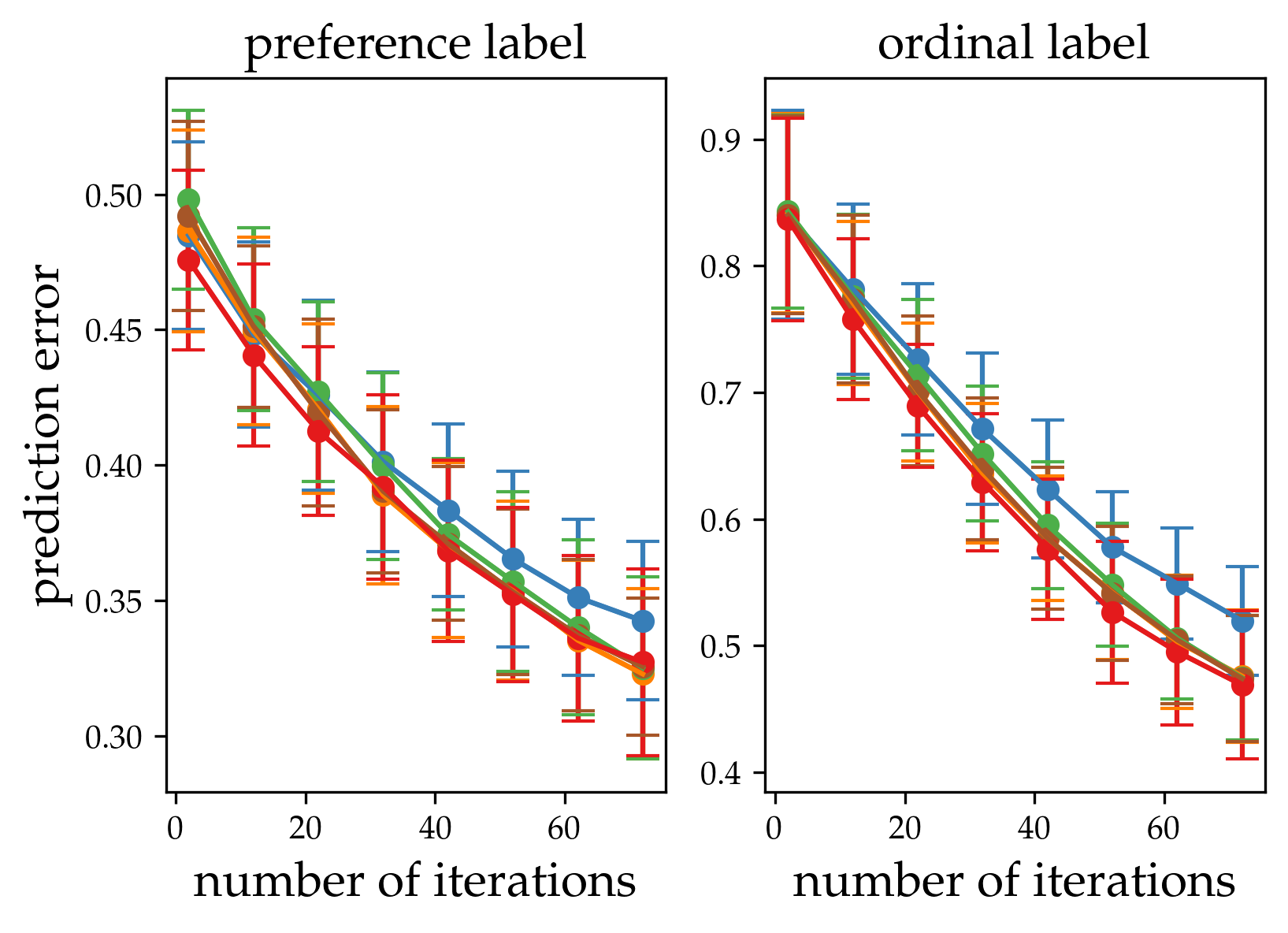}
      \caption{Synthetic function prediction error}
    \label{fig:04_04_subset_label_synthetic}
    \end{subfigure}%
  }\cr
}
\caption{Impact of random subset size on algorithm performance. a) Example 3D synthetic objective function and posterior learned by ROIAL with subset size = 500 after 80 iterations. Values are averaged over the 3$^{\textrm{rd}}$ dimension and normalized to range from 0 to 1. b-c) Algorithm's error in predicting preferences and ordinal labels (mean $\pm$ std). Each simulation evaluated performance at $1000$ randomly- selected points; the model posterior was used to predict preferences between consecutive pairs of points and ordinal labels at each point.}
  \label{fig:04_04_subset}
\end{figure*}

\noindent\textbf{Extending to higher dimensions.}
To characterize the impact of the random subset size on algorithmic performance, we compare performance of different sizes in simulation for both the Hartmann3 and synthetic reward functions. We calculate the posterior over the entire space only every 10 steps to reduce computation time, and then use this posterior to evaluate the algorithm's error in predicting preference and ordinal labels. Figure~\ref{fig:04_04_simulation_posterior} provides an example of a 3D posterior, Figure~\ref{fig:04_04_subset_label_hartmann3} depicts the average performance for Hartmann3 over 10 simulation repetitions, and Figure~\ref{fig:04_04_subset_label_synthetic} shows the average performance over a set of 50 unique synthetic functions. We find that a subset size of at least 5 yields performance close to using all points.

\begin{figure*}[tb]
\begin{center}
    \begin{subfigure}[b]{0.57\textwidth}
    \includegraphics[width=\textwidth]{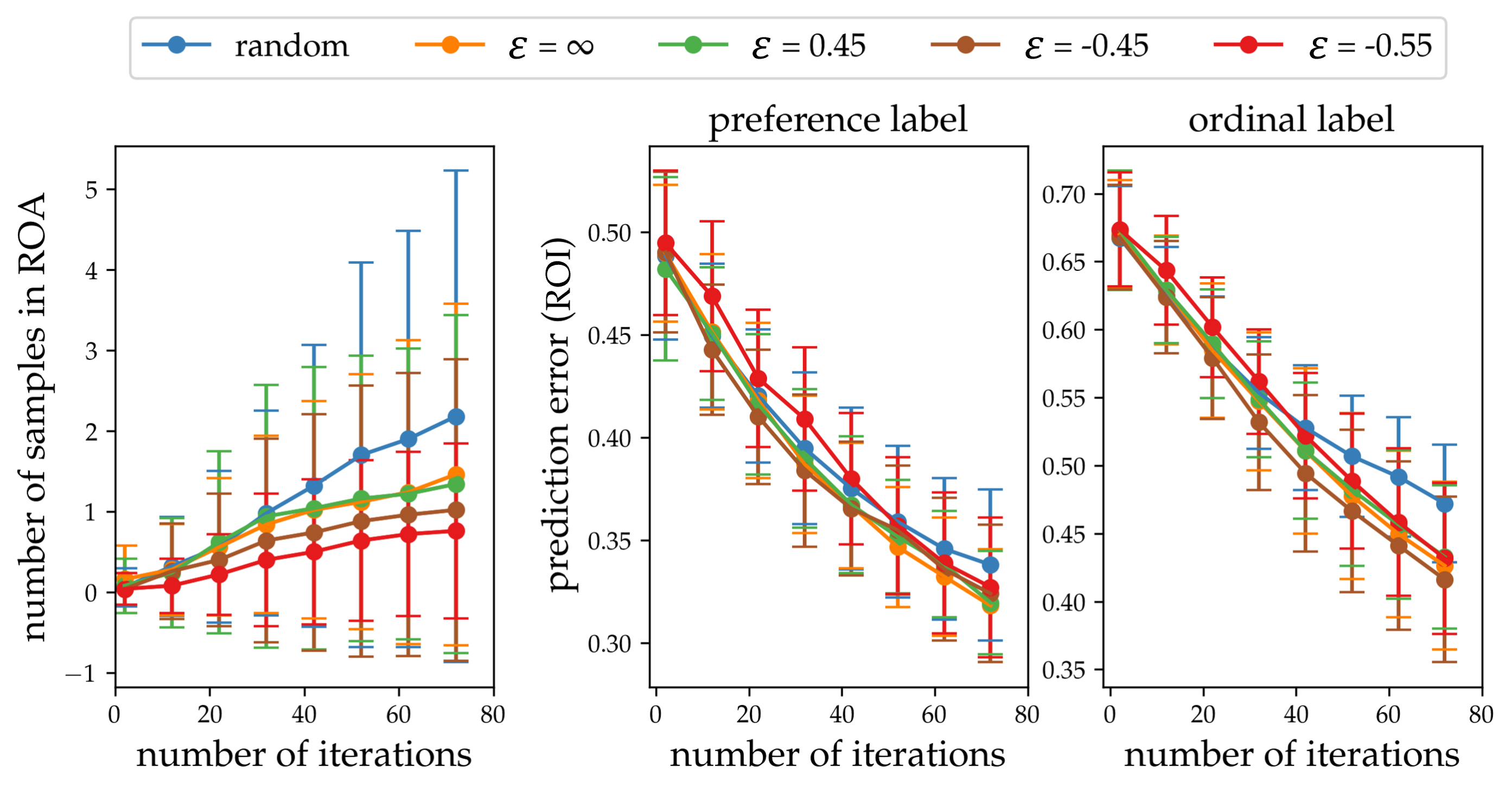}
    \caption{Number of samples in the ROA and prediction error in the ROI}
    \label{fig:04_04_ucb_idx_label}
  \end{subfigure}
  \begin{subfigure}[b]{0.4185\textwidth}
    \includegraphics[width=\textwidth]{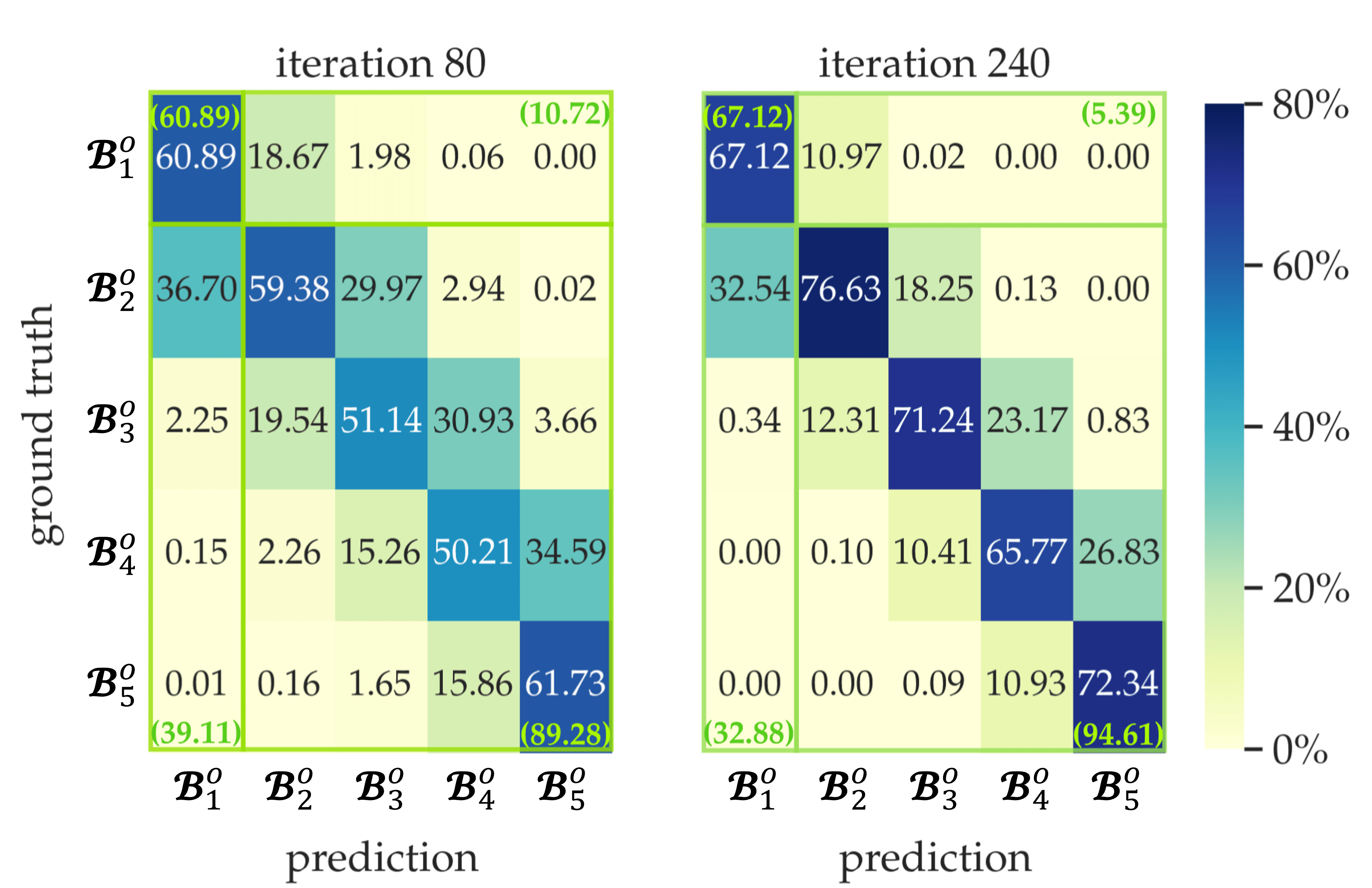}
    \caption{Confusion matrices}
    \label{fig:04_04_confusion_matrix}
  \end{subfigure}
  
  \caption{Effect of the confidence interval. All simulations are run over 50 reward synthetic functions with a random subset size of 500. a) Left: cumulative number of points in the ROA ($\ordinalCategory_1$) queried at each iteration (mean $\pm$ std).  Note that as $\roialConservatism$ increases, more samples are required for the confidence interval to fall below the ROA threshold, at which point ROIAL starts avoiding the ROA. Middle and right: error in predicting comparison and ordinal labels for different values of $\roialConservatism$; predictions are over 1,\!000 random actions (mean $\pm$ std). b) Confusion matrices (column-normalized) of ordinal label prediction  over the entire action space at iterations 80 and 240 with $\roialConservatism$ = -0.45. The $2 \times 2$ confusion matrices for ROI prediction accuracy are outlined in green. Prediction accuracy increases with the number of iterations.}
  \label{fig:04_04_ucb}
\end{center}
\end{figure*}

\noindent\textbf{Estimating the region of interest.}
We demonstrate the effect of the confidence parameter $\roialConservatism$ on the number of points sampled from the ROA and on prediction error in the ROI. Figure~\ref{fig:04_04_ucb_idx_label} demonstrates that across various values of $\roialConservatism$, visits to the ROA decrease as $\roialConservatism$ decreases. To confirm that restricting queries to the estimated ROI does not harm performance, we also compare label prediction error in the ROI across values of $\roialConservatism$. When $\roialConservatism = -0.45$, ROIAL achieves similar preference prediction accuracy and slightly-improved ordinal label prediction within the ROI compared to $\roialConservatism = \infty$, which permits sampling over the entire space (Figure~\ref{fig:04_04_ucb_idx_label}). Additionally, the confusion matrix (Figure~\ref{fig:04_04_confusion_matrix}) shows that the algorithm usually predicts either the correct ordinal label or an adjacent ordinal category. The ROI prediction accuracy (green text in Figure~\ref{fig:04_04_confusion_matrix}) indicates that ROIAL predicts whether points belong to the ROI with relatively-high accuracy.

\begin{figure}[tb]
\begin{center}
    \includegraphics[width=0.455\textwidth]{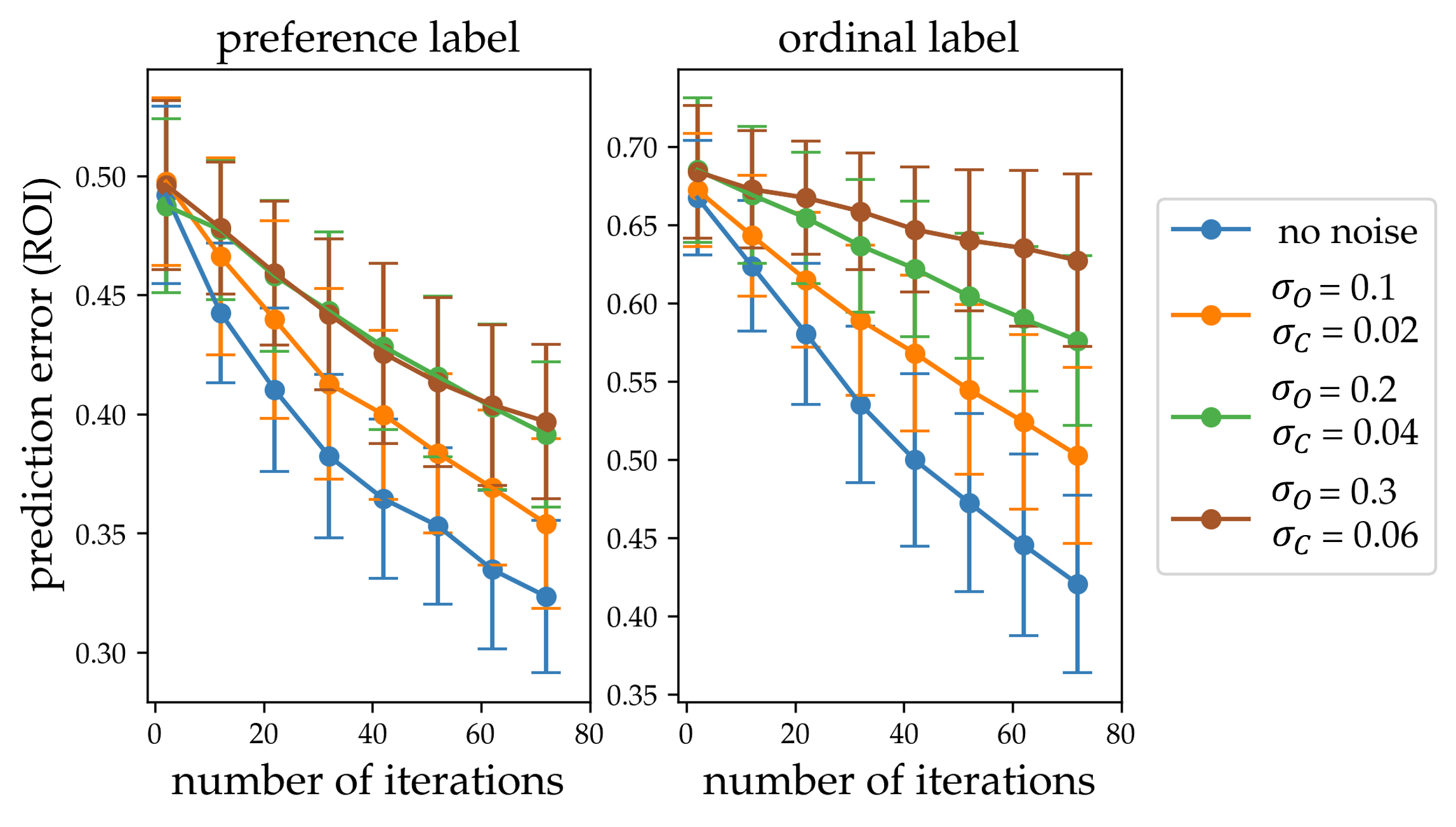}
  \caption{Effect of noisy feedback. The ordinal and pairwise comparison noise parameters, $\ordinalNoiseStd$ and $\preferenceNoiseStd$, range from $0.1$ to $0.3$ and $0.02$ to $0.06$, respectively. All cases use a random subset size of 500 and $\roialConservatism = -0.45$, and each simulation uses 1,\!000 random points to evaluate label prediction. Plots show means $\pm$ standard deviation.}
  \label{fig:04_04_ord_noise}
\end{center}
\end{figure}

\noindent\textbf{Robustness to noisy feedback.}
Since user feedback is expected to be noisy, we evaluate the algorithm's robustness to noisy feedback generated from the distributions $P(\queryResponse_o \mid \bm{\gpRewardFunction}, \trajectory) = g_{O}\left( \frac{\ordinalThresholdSet_{\queryResponse_o}-\gpRewardFunction(\trajectoryFeaturesFunction(\trajectory))}{\ordinalNoiseStd}\right) - g_{O}\left(\frac{\ordinalThresholdSet_{\queryResponse_o-1}-\gpRewardFunction(\trajectoryFeaturesFunction(\trajectory))}{\ordinalNoiseStd}\right)$ and $P(\trajectory^{(1)} \succ \trajectory^{(2)} \mid \bm{\gpRewardFunction}) = g_{C}\left(\frac{\gpRewardFunction(\trajectoryFeaturesFunction(\trajectory^{(1)}))-\gpRewardFunction(\trajectoryFeaturesFunction(\trajectory^{(2)}))}{\preferenceNoiseStd}\right)$ for ordinal and pairwise comparison feedback, respectively, with true ordinal thresholds $\{ \ordinalThresholdSet_j \mid j = 0, \dots, \abs{\ordinalCategory}\}$ and simulated noise parameters $\preferenceNoiseStd$ and $\ordinalNoiseStd$. We set $\ordinalNoiseStd > \preferenceNoiseStd$ because we expect ordinal labels to be noisier than pairwise comparisons, as they require users to recall all past experience to give consistent feedback, whereas a pairwise comparison only involves the previous and current points (or trajectories). The algorithm learns more slowly with noisier feedback (Figure~\ref{fig:04_04_ord_noise}).

\subsubsection{Exoskeleton Experiments}
After demonstrating ROIAL's performance in simulation, we experimentally deployed it on the lower-body exoskeleton Atalante, developed by Wandercraft (video: \url{https://youtu.be/04lMJmKmZrQ}, ROIAL hyperparameters: \url{https://github.com/kli58/ROIAL}). Atalante, shown in Figure~\ref{fig:03_03_Atalante},  is an 18 degree of freedom robot designed to restore assisted mobility to patients with motor complete paraplegia through the control of 12 actuated joints: 3 joints at each hip, 1 joint at each knee, and 2 degrees of actuation in each ankle. For more details on Atalante, refer to \cite{agrawal2017first,harib2018feedback,gurriet2018towards}.

\begin{figure}[tb]
    \centering
    \includegraphics[width=0.36\textwidth]{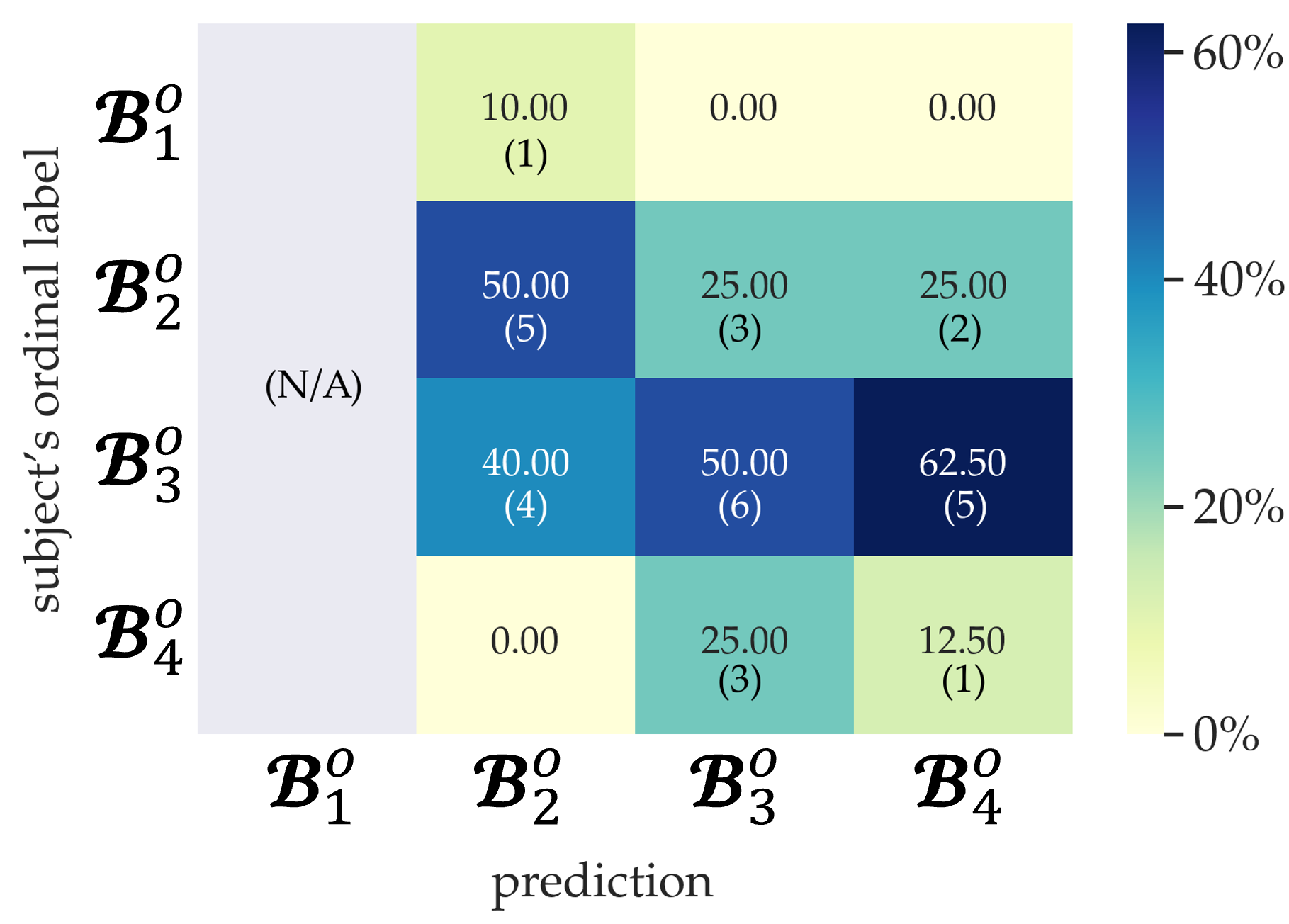}
    \caption{Confusion matrix of the validation phase results for all three subjects. The first column is gray because trajectories in the ROA ($\ordinalCategory_1$) were purposefully avoided to prevent subject discomfort. Percentages are normalized across columns. Parentheses show the numbers of gait trials in each case.}
  \label{fig:04_04_exo_confusion}
\end{figure}

\begin{figure*}
    \centering
    \includegraphics[width=\textwidth]{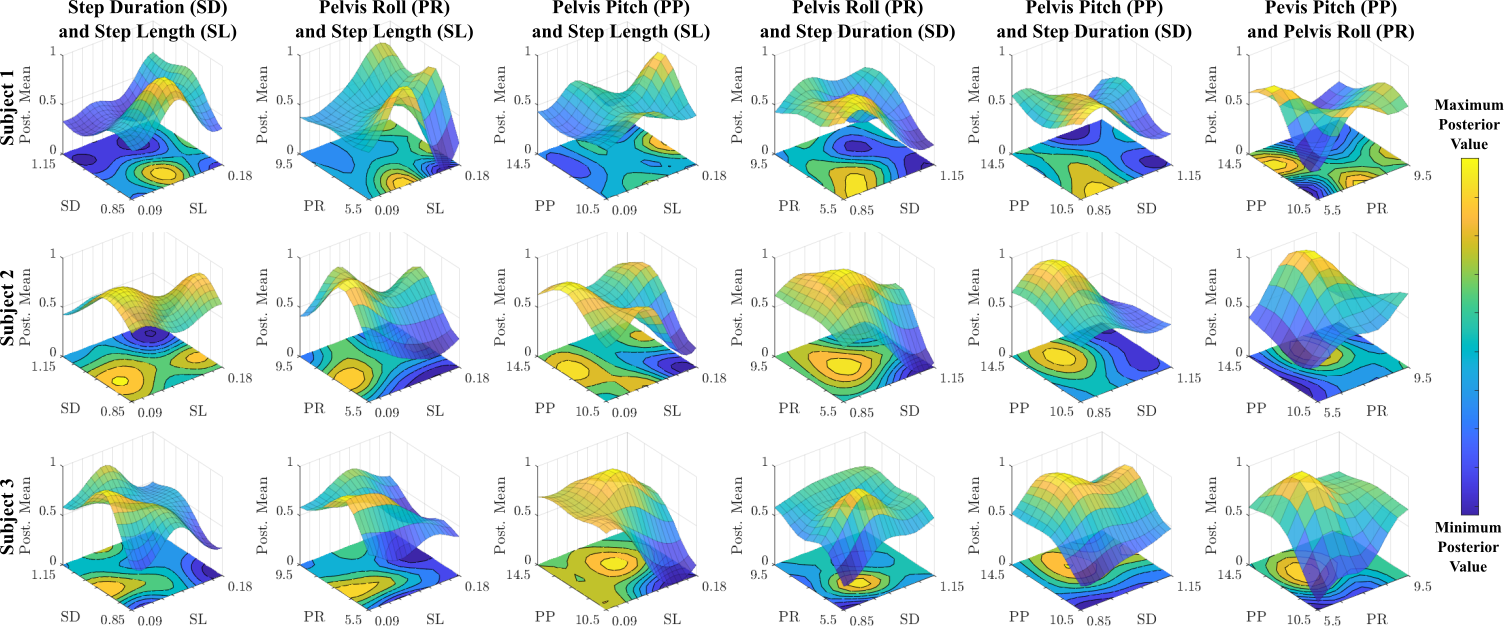}
    \caption{4D posterior mean reward across exoskeleton gaits. Rewards are plotted over each pair of gait space parameters, with the values averaged over the remaining 2 parameters in each plot. Each row corresponds to a subject: Subject 1 is the most experienced exoskeleton user, Subject 2 is the second-most experienced user, and Subject 3 never used the exoskeleton prior to the experiment.}
    \label{fig:04_04_posterior}
\end{figure*}

Dynamically stable crutch-less exoskeleton walking gaits are generated through nonconvex optimization techniques (see Section II of \citet{tucker2020preference}), based on the theory of hybrid zero dynamics (HZD) introduced by \citet{ames2014human} and the HZD-based optimization method presented in \cite{hereid2018dynamic}. 
These periodic gaits are parameterized by various features, and this studies focuses on four: step length (SL) in meters, step duration (SD) in seconds, maximum pelvis roll (PR) in degrees, and maximum pelvis pitch (PP) in degrees (Figure~\ref{fig:03_03_Atalante}). These parameters were selected because exoskeleton users frequently suggested modifications to SL, SD, and PR in prior work (see \url{https://sites.google.com/view/roial-icra2021}), and we wanted to further study the relationship between PR and PP. We discretized these parameters into bins of sizes 10, 7, 5, and 5, respectively, resulting in 1,750 trajectories within a 4D gait (or feature) space. ROIAL randomly selected 500 trajectories in each iteration and used $\roialConservatism = 0.45$ to estimate the ROI. 

The experimental procedure was conducted for three non-disabled subjects and consisted of 40 trials divided into a \emph{training phase} (30 trials) and a \emph{validation phase} (10 trials). Subjects were not informed of when the validation phase began. Subjects provided ordinal labels for all 40 gaits, and optional pairwise comparisons between the current and previous gaits for all but the first trial.\footnote{The users were given chance to not give pairwise comparisons, in which case we simply update the posterior only with the ordinal feedback.} Four ordinal categories were considered and described to the users as:
\begin{enumerate}[nosep]
    \item \textbf{Very Bad} ($\ordinalCategory_1$): User feels unsafe or uncomfortable to the point that the user never wants to repeat the gait.
    \item \textbf{Bad} ($\ordinalCategory_2$): User dislikes the gait but does not feel unsafe or uncomfortable.
    \item \textbf{Neutral} ($\ordinalCategory_3$): User neither dislikes nor likes the gait and would be willing to try the gait again.
    \item \textbf{Good} ($\ordinalCategory_4$): User likes the gait and would be willing to continue walking with it for a long period of time. 
\end{enumerate}
While including additional ordinal categories could increase the potential information gain from each query, it also increases the cognitive burden for the users and thus makes the labels less reliable. Validation gaits were selected so that at least two samples were predicted to belong to $\ordinalCategory_2, \ordinalCategory_3$, and $\ordinalCategory_4$, with the remaining four validation gaits sampled at random. Gaits predicted to belong in $\ordinalCategory_1$ were excluded because they are likely to make the user feel uncomfortable or unsafe, and gaits sampled during the training phase were explicitly excluded from the validation trials.

\noindent \textbf{Experimental results.} Figure~\ref{fig:04_04_exo_confusion} depicts the results of the validation phase for all three subjects. These results show a reliable correlation between the predicted categories and the users' reported ordinal labels, in which the majority of the predicted ordinal labels are within one category of the true ordinal labels. Since less than $2\%$ of the gait space was explored during the experiment, we expect that the prediction accuracy would increase with additional exoskeleton trials as observed in simulation (Figure~\ref{fig:04_04_confusion_matrix}). Overall, these results suggest that ROIAL can yield reliable preference landscapes within a moderate number of samples.

Figure~\ref{fig:04_04_posterior} depicts the final posterior mean for each of the subjects. These reward functions highlight both regions of agreement and disagreement among the subjects. For example, all subjects strongly dislike gaits at the lower bound of PP and lower bound of PR. However, all subjects disagree in their reward landscapes across SL and SD. This type of insight could not be derived from direct gait optimization, which mostly obtains information near the optimum.

We also evaluated the effect of each gait parameter on the posterior rewards using the permutation feature importance metric. The results of this test for each respective subject across the four gait parameters (SL, SD, PR, PP) are: (0.20, 0.30, 0.33, 0.27), (0.26, 0.36, 0.38, 0.29), and (0.23, 0.16, 0.21, 0.45). These values suggest that the preferences of more experienced users (Subjects 1 and 2) may be most influenced by SD and PR, while the least-experienced user's feedback may be most weighted by PP (Subject 3). The code for this test is available on GitHub: \url{https://github.com/kli58/ROIAL}. These results demonstrate that ROIAL is capable of obtaining preference landscapes within relatively-few exoskeleton trials while avoiding gaits that make users feel unsafe or uncomfortable.

%% file: 04_active/05_scale.tex
\label{sec:04_05_scale}

After presenting how to actively learn a non-parametric reward function using GPs and mutual information maximization, we go back to parametric reward functions and continue our presentation with other forms of comparative feedback, just like we did in Chapter~\ref{chap:learning}. Following the same structure, we proceed with scale feedback. In this section, we again use the mutual information based active querying method, but we also introduce the max regret method (originally used by \citet{wilde2020active}) to generate the scale queries. Afterwards, we present our simulation and experiment results with these two acquisition functions in Section~\ref{subsec:04_05_corl21_scale_experiments}.

\subsection{Two Acquisition Functions for Active Scale Feedback}

To learn the true reward parameters $\weights^*$ efficiently, the robot actively chooses the query $\query^{(i)}$ it presents to the user at every iteration $i$. Two approaches for learning from pairwise comparisons are mutual information maximization (Sections~\ref{sec:04_02_information_gain} through \ref{sec:04_04_roial}) and max regret optimization \cite{wilde2020active}.

Mutual information based active querying method seeks to reduce the robot's uncertainty over $\weights$ while choosing queries that are easy to answer for the user. Max regret optimization, on the other hand, minimizes the maximum regret (as defined in Equation~\eqref{eq:03_04_regret}) by showing mutual worst case trajectories, which also results in easy queries. We leverage both of these methods for our active query generation in scale feedback. 

\subsubsection{Mutual Information Maximization}
We start with the mutual information. Letting $H$ denote Shannon's information entropy \cite{wasserman2010all}, a greedy step takes the expectation over the user's response $\queryResponse^{(i)}$ to the query $\query^{(i)}$ being optimized to actively learn both the reward parameters $\weights$ and the sensitivity threshold $\sensitivityThreshold$ (see Definition~\eqref{def:03_04_probabilistic_user_model}):
\begin{align}
\query^{(i)} = \underset{\query^{(i)}}{\argmax}\;
H(\weights,\sensitivityThreshold \mid \query^{(i)}, \belief^i) 
- \mathbb{E}_{\queryResponse^{(i)}\mid \query^{(i)},\belief^i}\big[H(\weights, \sensitivityThreshold \mid \queryResponse^{(i)},\query^{(i)}, \belief^i)\big].
\label{eq:04_05_entropy_greedy}
\end{align}
Noting the belief distribution $\belief^i$ was defined over both $\weights$ and $\sensitivityThreshold$ when we use scale feedback, we approximate the computation of entropies by summing over a set $\weightsSampleSet$ of samples of $(\weights,\sensitivityThreshold)\sim b^i$. Thus, following the derivation in Section~\ref{sec:04_02_information_gain} (thereby Appendix~\ref{app:99_02_mutual_information_derivation}), the new query $\query^{(i)}_*$ solves
\begin{align}
\query^{(i)}_* = \underset{\query^{(i)}}{\argmax}
\sum_{\queryResponse^{(i)}}\sum_{(\weights,\sensitivityThreshold)\in\weightsSampleSet}
\frac{P(\queryResponse^{(i)} \mid \query^{(i)},\weights,\sensitivityThreshold)}{\abs{\weightsSampleSet}}
\log_2\left(
\frac{\abs{\weightsSampleSet} \cdot P(\queryResponse^{(i)} \mid \query^{(i)},\weights,\sensitivityThreshold)}
{\sum_{(\weights',\sensitivityThreshold')\in\weightsSampleSet} P(\queryResponse^{(i)} \mid \query^{(i)},\weights',\sensitivityThreshold')}
\right).
\label{eq:04_05_entropy_greedy_extended}
\end{align}

\subsubsection{Max Regret Optimization}
The max regret policy generates queries $\query^{(i)} = (\query^{(i)}_1,\query^{(i)}_2)$ such that if the robot learned $\query^{(i)}_1$ as the optimal trajectory but the user optimal solution would be $\query^{(i)}_2$ is a worst case. With a symmetric perspective over $\query^{(i)}_1$ and $\query^{(i)}_2$, we solve
\begin{align}
\underset{\weights,\sensitivityThreshold,\weights',\sensitivityThreshold'}{\max}
\belief^i(\weights,\sensitivityThreshold)\belief^i(\weights', \sensitivityThreshold')
\bigg(
\regretFunction(\weights,\weights') + \regretFunction(\weights',\weights)
\bigg),
\label{eq:04_05_maxregret_greedy}
\end{align}
where $\regretFunction(\cdot,\cdot)$ is the reward difference (regret) defined in Equation~\eqref{eq:03_04_regret}. The optimal query with respect to max regret optimization is then $\query^{(i)} = (\query^{(i)}_1,\query^{(i)}_2)$ such that $\query^{(i)}_1$ and $\query^{(i)}_2$ are the optimal trajectories with respect to $\weights$ and $\weights'$, respectively. By observing feedback to such queries it greedily improves the probabilistic worst case error. In contrast to the mutual information based approach, obtaining queries through max regret optimization requires $\query^{(i)}_1$ and $\query^{(i)}_2$ to be optimal trajectories for some users $(\weights,\sensitivityThreshold)$ and $(\weights',\sensitivityThreshold')$. 
On the other hand, maximum regret does not require a one-step look-ahead and thus no summation over potential feedback values $\queryResponse^{(i)}$, making it computationally lighter.

Optimizations in~\eqref{eq:04_05_entropy_greedy_extended} and \eqref{eq:04_05_maxregret_greedy} now give us two different policies for actively solving the initial reward learning via scale feedback problem posed in Section~\ref{subsec:03_04_formulation}. In the simulations, we compare how the performance of both benefits from scale feedback.

\subsection{Experiments}\label{subsec:04_05_corl21_scale_experiments}
\subsubsection{Simulation Results}
\label{sec:04_05_corl21_scale_simulations}
\begin{figure}[ht]
    \centering
    \includegraphics[width=0.45\textwidth]{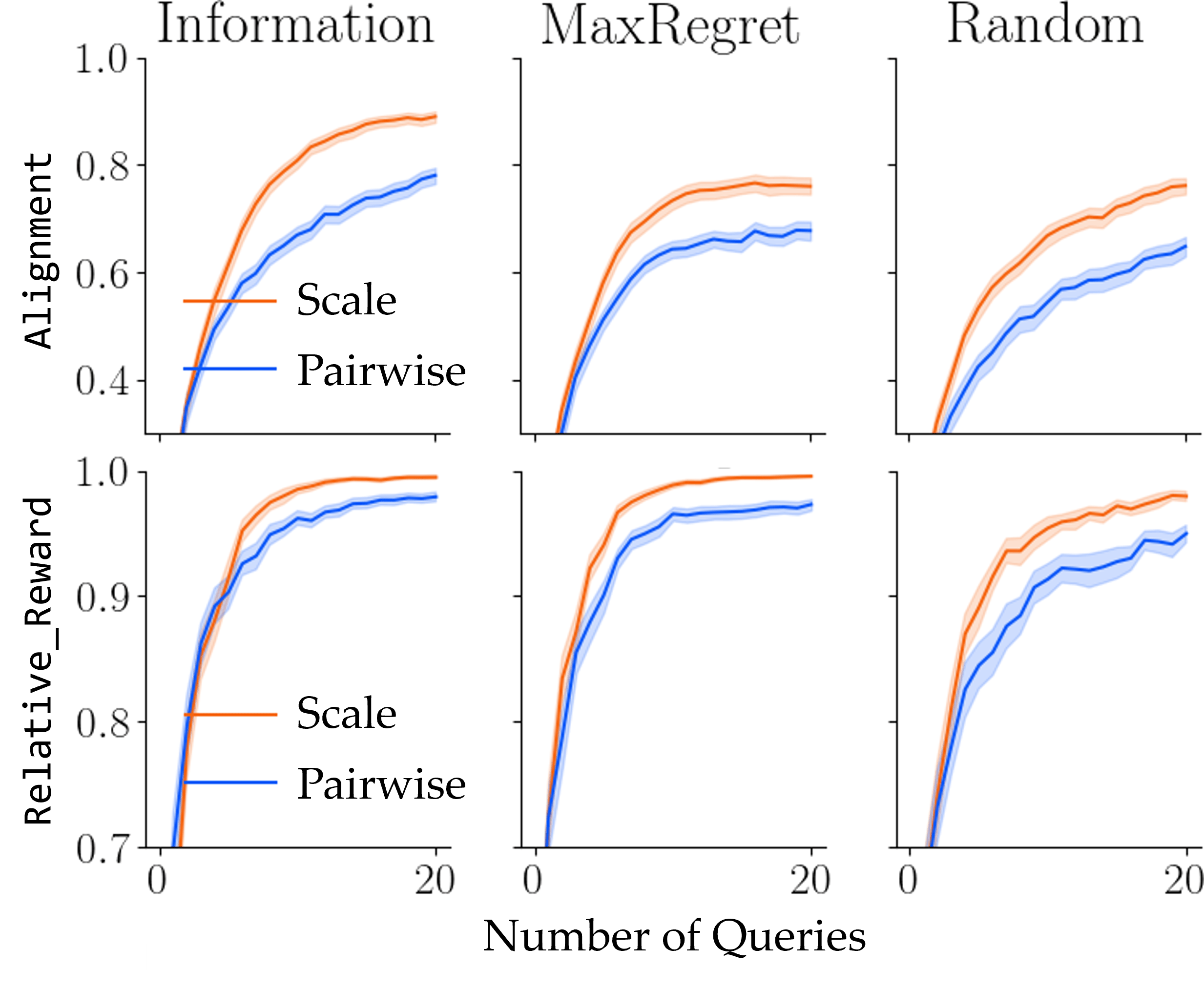}
    \caption{Comparison of scale feedback and weak pairwise comparisons for different active querying methods.}
    \label{fig:04_05_corl21_scale_driver_extended}
\end{figure}

We now present our main simulation results. Additional results can be found in Appendix~\ref{app:99_05_corl21_scale_additional_results}. In all simulations and experiments, we assume a parametric reward function that is linear in trajectory features, similar to Sections~\ref{subsec:04_01_experiments} and \ref{subsec:04_02_experiments}.

\noindent\textbf{Experiment Setup.} We simulate the presented framework using the \emph{Driver} experiment used in \cite{sadigh2017active, wilde2020active,basu2018learning} and Sections~\ref{subsec:04_02_experiments} and \ref{subsec:04_03_gp_experiments}. We modify the setup by adding $6$ new features, obtaining a more challenging $10$-dimensional problem (details on the features and the results for the original \emph{Driver} can be found in Appendices~\ref{app:99_02_corl21_scale_details} and \ref{app:99_05_corl21_scale_additional_results}, respectively). $71$ distinct true reward parameters $\weights^*$ are drawn uniformly at random, and each user is simulated with $\sensitivityThreshold^*\in\{0.25, 0.5, 0.75, 1.00\}$, making it $284$ runs for each method. We set $\scaleNoiseStd = 0.1$ for the noise level. We generate a set of $200$ distinct sample trajectories by drawing random reward parameters $\weights$ and then computing their optimal trajectories. The active query generation methods then optimize over this set. We evaluate learning using the \texttt{Alignment} metric and the \texttt{Relative\_Reward} (see Section~\ref{sec:03_04_scale} for their definitions).

As a baseline we use weak pairwise comparisons (strict pairwise comparisons showed a slightly poorer performance). To ensure a fair comparison, we emulate weak pairwise comparisons by setting the step size to $\sliderStepSize=1$ and use the same noise model for both forms of feedback (as opposed to using an alternative model, such as the one presented in Equation~\eqref{eq:04_01_weak_noisily_optimal}).

\noindent\textbf{Results.}
Figure~\ref{fig:04_05_corl21_scale_driver_extended} shows the \texttt{Alignment} and \texttt{Relative\_Reward} for the \emph{Driver} experiment for mutual information, max regret and random query generation.
We observe that in all cases scale feedback significantly improves the performance over weak pairwise comparisons in both metrics ($p<0.001$ in all cases with two-sample $t$-test). When using the proposed scale feedback, the \texttt{Alignment} after $20$ iterations improves from $0.77$ to $0.86$ for mutual information, from $0.67$ to $0.76$ for max regret, and from $0.64$ to $0.75$ for random queries. The \texttt{Relative\_Reward} improves for mutual information and max regret similarly from $0.97$ to $1.00$, i.e., the learned solution is optimal. Both methods make most progress during the first $10$ iterations. Random queries improve the final relative reward from $0.94$ to $0.97$. Overall, the simulation showcases that scale feedback improves learning, independent of the query selection method. For mutual information and max regret, scale feedback allows for finding optimal solution within a small budget of iterations. Appendix~\ref{app:99_05_corl21_scale_numerical_results} presents the numerical results for all plots in this section. In Appendix~\ref{app:99_05_corl21_scale_additional_results}, we show additional simulation results for higher noise.

\subsubsection{User Study}
\label{subsec:04_05_corl21_scale_study}
Next, we analyze the scale feedback in comparison with pairwise comparisons and under different active querying methods with two user studies.\footnote{A summary video is at \url{https://sites.google.com/view/reward-learning-scale-feedback}, and the code is at \url{https://github.com/Stanford-ILIAD/reward-learning-scale-feedback}.} In both studies, we used $\sliderStepSize=0.1$ for scale queries.

\noindent\textbf{Experiment Setup.} We designed a serving task with a Fetch robot \cite{wise2016fetch} as shown in Figure~\ref{fig:03_04_frontfig}, which we call \emph{FetchDrink}. We generated a dataset of $120$ distinct trajectories. Human subjects were told they should train the robot to bring the drink to the customer in the manner they prefer, paying attention to the following five factors: the drink (out of $3$ options) to be served, the orientation of the pan in front of the robot, moving the drink behind or over the pan, the maximum height of the path, and the speed. The subjects were also informed about the types of queries they will respond to.

\noindent\textbf{Independent Variables.} In the first experiment, we wanted to compare scale feedback and weak pairwise comparisons under random querying, and scale feedback under random and mutual information based querying. Hence, we varied the query type and the querying algorithm among: (i) weak pairwise comparisons with random querying, (ii) scale feedback with random querying, and (iii) scale feedback with mutual information based querying. In the second experiment, we wanted to compare scale and weak pairwise comparisons under mutual information based querying. Hence, we employed: (i) weak pairwise comparisons with mutual information based querying, and (ii) scale feedback with mutual information based querying. For all, we took $\scaleNoiseStd=0.35$ based on pilot trials with different users (see Appendix~\ref{app:99_02_corl21_scale_sigma_selection}).

\noindent\textbf{Procedure.} We recruited $18$ participants ($5$ female, $13$ male, ages 20 -- 55) for the first, and $14$ participants ($5$ female, $9$ male, ages 20 -- 56) for the second experiment. Due to the pandemic conditions, the subjects participated in the study remotely with an online interface as in Figure~\ref{fig:03_04_frontfig}. The study started with an instructions page with a two-question quiz to make sure the participants understood how to use the interface. After reading the instructions, we had the subjects fill a form where they indicated their preferences for each of the five individual factors described above, to encourage them to be consistent in their responses during the data collection.

In the experiments, each participant responded to $10$ queries generated with each of the algorithms. After each of these $10$-query sets, they were shown the optimal trajectory from the dataset with respect to their learned reward function. The participants responded to a $5$-point rating scale survey (1-Strongly Disagree, 5-Strongly Agree) for this trajectory: ``The displayed trajectory fits my preferences on the task." We also collected scale feedback for $10$ more randomly-generated queries (called the test set) to measure performance in each experiment. We randomized the order of these sets (of $10$ queries) to prevent any bias. The interface provided a ``Sync Videos" button to restart both videos for easier comparison.

\noindent\textbf{Dependent Measures.} 
As an objective measure of the learning performance, we calculated the log-likelihood of the test set (of $10$ scale queries\footnote{We present results with a test set that consists of both scale feedback and weak pairwise comparisons in Appendix~\ref{app:99_05_corl21_scale_test_mixture_data}.}) under the posterior $b^{\abs{\dataset}}(\weights,\sensitivityThreshold)$ learned using the $10$ queries generated via each algorithm, i.e., we calculated:
\begin{align}
    \mathtt{Log\!-\!Likelihood} = \log P(\dataset_{\textrm{test}} \mid \dataset) = \log \mathbb{E}_{\weights \mid \dataset}\left[P(\dataset_{\textrm{test}} \mid \weights)\right] 
\end{align}
We also used the responses to the $5$-point rating scale survey questions to measure how well the learned rewards achieve the task. Finally, the users took a post-experiment survey where they rated (from $1$ to $5$) the easiness and expressiveness of weak pairwise comparison and scale feedback questions.

\noindent\textbf{Hypotheses.} We test the following hypotheses.\\
\textbf{H10.} \textit{Scale feedback leads to faster learning than weak pairwise comparisons.}\\
\textbf{H11.} \textit{Querying based on mutual information accelerates learning compared to random querying.}\\
\textbf{H12.} \textit{Users will prefer mutual information based querying over random querying in terms of the optimized trajectories.}\\
\textbf{H13.} \textit{Users will prefer scale feedback over weak pairwise comparisons in terms of the  optimized trajectories.}\\
\textbf{H14.} \textit{Users will rate the scale feedback questions as easy as weak pairwise comparison questions.}\\
\textbf{H15.} \textit{Users will rate the scale feedback questions as expressive as weak pairwise comparison questions.}

\begin{figure}[t]
    \includegraphics[width=1\textwidth]{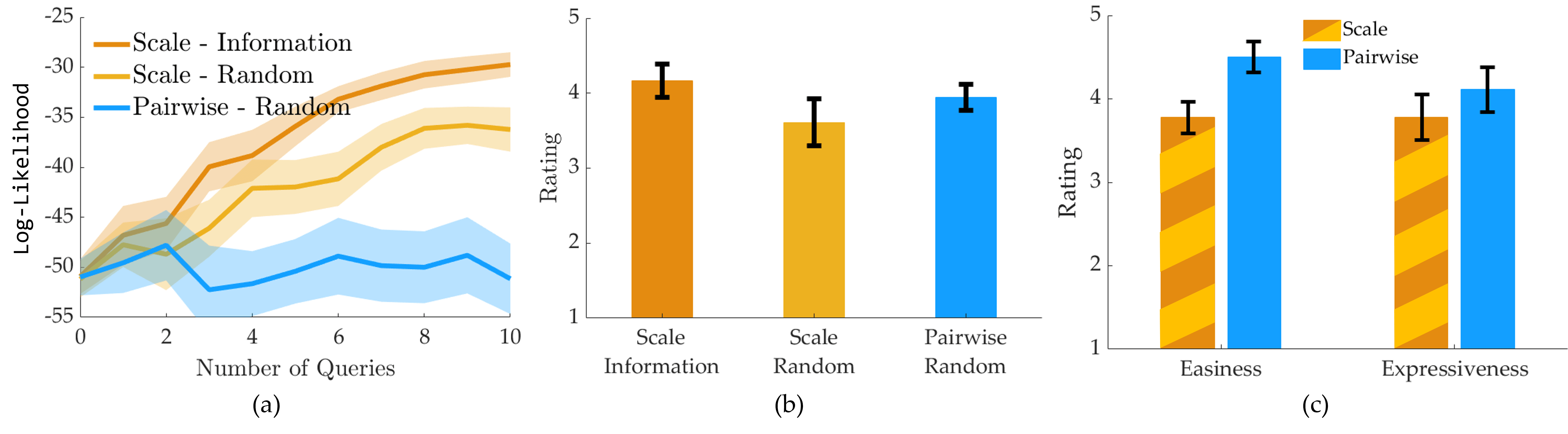}
	\caption{All results are shown for the first experiment (mean$\pm$s.e. over $18$ subjects).}
	\label{fig:04_05_corl21_scale_user_study_results}
\end{figure}

\begin{figure}[t]
    \includegraphics[width=1\textwidth]{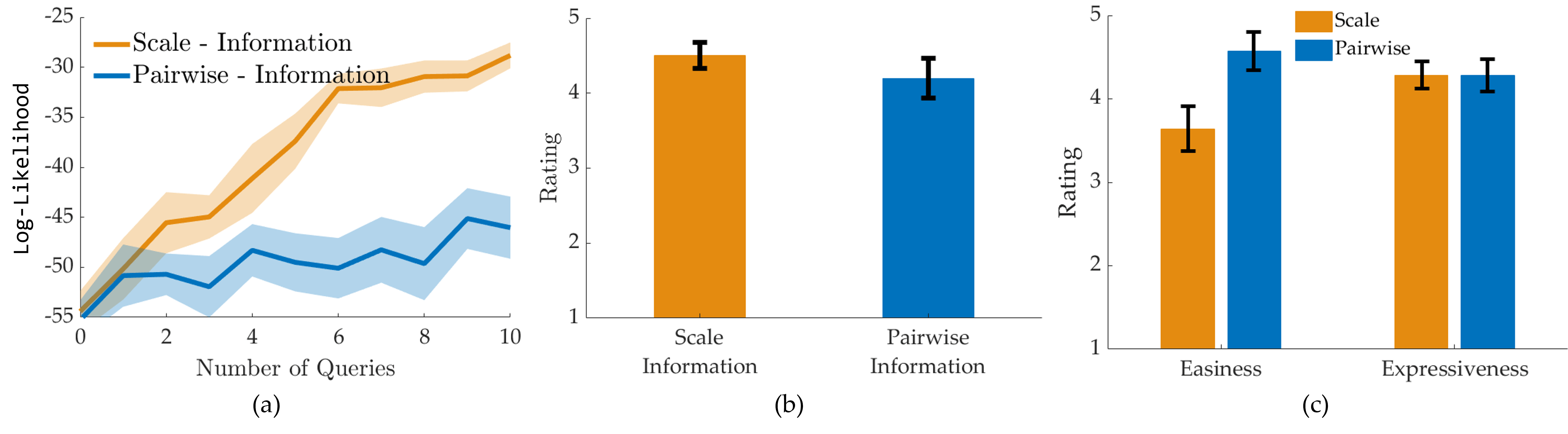}
	\caption{All results are shown for the second experiment (mean$\pm$s.e. over $14$ subjects).}
	\label{fig:04_05_corl21_scale_user_study2_results}
\end{figure} 

\noindent\textbf{Results.}
We present results of the first and the second experiments in Figures~\ref{fig:04_05_corl21_scale_user_study_results} and \ref{fig:04_05_corl21_scale_user_study2_results}, respectively. It can be seen that the log-likelihood of the test set after learning the reward function via scale feedback is higher than learning via weak pairwise comparisons, under both random and mutual information based querying. Besides, mutual information based query generation accelerates the learning and leads to higher log-likelihood values compared to random querying. All of these comparisons are statistically significant with $p<0.001$ (paired-sample $t$-test), so they strongly support \textbf{H10} and \textbf{H11}.

In Figure~\ref{fig:04_05_corl21_scale_user_study_results}b, it can be seen active querying led to learning reward functions that better optimize trajectories compared to random querying -- this comparison was somewhat significant with $p\approx 0.05$, supporting \textbf{H12}. In fact, when we fit a Gaussian distribution to the ratings, we observe that it is $1.95$ times as likely to get a better rating with mutual information based querying than random querying. Surprisingly, learning via weak pairwise comparisons achieved slightly higher reward than learning via scale feedback when queries were randomly selected, and slightly lower reward when queries were generated based on mutual information maximization. However, these comparisons are not statistically significant. This is indeed analogous to the \texttt{Relative\_Reward} comparisons in Figure~\ref{fig:04_05_corl21_scale_driver_extended}: more complex tasks might be needed to better analyze the difference between the two methods. Thus, we neither reject nor accept \textbf{H13}.

Finally, the subjective results in Figure~\ref{fig:04_05_corl21_scale_user_study_results}c and \ref{fig:04_05_corl21_scale_user_study2_results}c suggest that users find the weak pairwise comparisons slightly, but consistently, easier than the scale feedback ($p<0.01$), rejecting \textbf{H14}. This is not surprising, as it is often easier to make a pairwise comparison and the ``About Equal" option in the weak pairwise comparison questions makes them even easier (see Section~\ref{sec:04_02_information_gain}). On the other hand, there was no statistically significant difference in terms of expressiveness of scale and weak pairwise comparison questions, partially supporting \textbf{H15}. In summary, it is interesting that our users perceived the weak pairwise comparison questions as easier and even more expressive at times; even though quantitatively, the scale feedback significantly outperforms the weak pairwise comparisons.

Appendix~\ref{app:99_05_corl21_scale_numerical_results} presents the numerical results for all plots in this section.

%% file: 04_active/06_rankings.tex
\label{sec:04_06_rankings}

The results in the previous sections in this chapter imply the comparative feedback queries made to the experts, $\query^{(i)}$'s, affect how well the posterior will be learned when the reward function is unimodal. In the case of multimodal rewards where we learn via ranking queries as we discussed in Section~\ref{sec:03_05_rankings}, this is still true, and can be seen from Equation~\eqref{eq:03_05_bayesian_learning}.

Specifically, a query $\query^{(i)}$ is desirable if observing the (anonymous) response to that, $\queryResponse^{(i)}$, yields high information about the underlying model parameters, $\weights_\modeIndex$ and $\mixingCoefficient_\modeIndex$ for all $\modeIndex\in[\numberOfModes]$. Therefore, we again propose using a mutual information objective to adaptively select the most informative query at each querying step $i$, generalizing the approach we presented in Section~\ref{sec:04_02_information_gain}.

\subsection{Formulation}
Assume at a fixed round $i$, we have made past ranking query observations $\rankingDataset=\bigl\{\query^{(i')},\queryResponse^{(i')}\bigr\}_{i'=1}^{i-1}$, possibly with other types of feedback to obtain the belief distribution $\belief^{i-1}$. The desired query is then \begin{align}
    \query^{(i)}_* = \argmax_{\query} I(\queryResponse; \weights, \mixingCoefficient \mid \query, \belief^{i-1})
\end{align}
where $I(\cdot;\cdot)$ denotes mutual information and $\queryResponse$ is the response to the query $\query$. Equivalently,
\begin{align}
    \query^{(i)}
    &=\argmin_{\query}\mathbb{E}_{\queryResponse', \weights', \mixingCoefficient' \sim \queryResponse, \weights, \mixingCoefficient \mid \query,\belief^{i-1}}\log\frac{\mathop{\mathbb E}_{\weights'',\mixingCoefficient'' \sim \weights,\mixingCoefficient\mid\belief^{i-1}}P\left(\queryResponse=\queryResponse'\mid \query, \weights=\weights'', \mixingCoefficient=\mixingCoefficient'' \right)}{P\left(\queryResponse=\queryResponse' \mid \query,\weights=\weights', \mixingCoefficient=\mixingCoefficient'\right)}\:.
    \label{eq:04_06_corl21_ranking_objective}
\end{align} 
The details of this derivation are presented in Appendix~\ref{app:99_02_corl21_ranking_derivation}.

\subsection{Overall Algorithm}
\label{subsec:04_06_overall_algorithm}
To efficiently solve the optimization in Equation~\eqref{eq:04_06_corl21_ranking_objective}, we first note that we should use a Monte Carlo approximation since the expectations are taken over continuous variables $\weights, \mixingCoefficient$ and a discrete variable $\queryResponse$ over an intractably large set of $\abs{\query}!$ alternatives. To perform this Monte Carlo integration, we require samples from the posterior $P(\queryResponse, \weights,\mixingCoefficient \mid \query, \belief^{i-1})$. 

Our key insight is that we can obtain joint samples from both posteriors by first sampling $\bar\weights, \bar\mixingCoefficient\sim P(\weights,\mixingCoefficient\mid\rankingDataset)$ and then sampling $\bar\queryResponse \sim P(\queryResponse \mid \query,\weights=\bar\weights, \bar\mixingCoefficient=\mixingCoefficient)$ since $(\weights,\mixingCoefficient)\perp \query \mid \belief^{i-1}$ and $\queryResponse\perp \belief^{i-1} \mid \query, \weights, \mixingCoefficient$. We perform the sampling $\bar\queryResponse\sim P(\queryResponse \mid \query,\weights=\bar\weights, \mixingCoefficient=\bar\mixingCoefficient)$ efficiently using Equation~\eqref{eq:03_05_pmf}. In general, exact sampling from the posterior $P(\weights, \mixingCoefficient \mid \belief^{i-1})$ is intractable. However, we note Equation~\eqref{eq:03_05_bayesian_learning} can be directly evaluated (using Equation~\eqref{eq:03_05_pmf}) and gives $P(\weights, \mixingCoefficient \mid \belief^{i-1})$ up to a proportionality constant factor.

With this unnormalized posterior of Equation~\eqref{eq:03_05_bayesian_learning}, we use the Metropolis-Hastings algorithm as described in Appendix~\ref{app:99_03_corl21_ranking_metropolis_hastings} to generate samples from the posterior $P(\weights, \mixingCoefficient \mid \belief^{i-1})$.

We see our optimization problem simplifies to finding, for $\abs{\weightsSampleSet}$ fixed samples $\bar\weights_j, \bar\mixingCoefficient_j \sim P(\weights, \mixingCoefficient\mid\belief^{i-1})$ and corresponding samples: $\bar\queryResponse_j\sim P(\queryResponse \mid \query,\bar\weights_j, \bar\mixingCoefficient_j)$
\begin{align}
    \query^{(i)} \!=\! \argmin_{\query} \sum_{j=1}^{\abs{\weightsSampleSet}} \bigg[\log\!\sum_{j'=1}^{\abs{\weightsSampleSet}}\! P\left(\queryResponse \!=\! \bar\queryResponse_{j} \mid \query, \weights\!=\!\bar\weights_{j'}, \mixingCoefficient\!=\!\bar\mixingCoefficient_{j'}\right) - \log\! P\left(\queryResponse \!=\!\bar\queryResponse_j \mid \query,\weights\!=\!\bar\weights_j, \mixingCoefficient \!=\! \bar\mixingCoefficient_j\right)\bigg]\:.\label{eq:04_06_corl21_ranking_final}
\end{align}
We solve this optimization using simulated annealing \cite{bertsimas1993simulated} (see Appendix~\ref{app:99_03_corl21_ranking_simulated_annealing}).

\begin{algorithm}[ht]
\caption{Active Querying for Rankings via Mutual Information Maximization}
\label{alg:04_06_corl21_ranking_info}
\begin{algorithmic}[1]
    \Require{Current belief distribution $\belief^{i-1}$}
    \State{$\{\bar\weights_j,\bar\mixingCoefficient_j\}_{j=1}^{\abs{\weightsSampleSet}} \sim P(\weights, \mixingCoefficient \mid \belief^{i-1})$ with Equation~\eqref{eq:03_05_bayesian_learning} via Metropolis-Hastings}
    \State{$\forall j, \bar\queryResponse_j \sim  P(\queryResponse_j \mid \query,\bar\weights_j, \bar\mixingCoefficient_j)$}
    \State{$\query^{(i)}_*\!\gets\!\argmin\limits_{\query} \sum\limits_{j=1}^{\abs{\weightsSampleSet}} \bigg[\log\!\sum\limits_{j'=1}^{\abs{\weightsSampleSet}}\! P\left(\queryResponse \!=\! \bar\queryResponse_{j} \mid \query, \weights\!=\!\bar\weights_{j'}, \mixingCoefficient\!=\!\bar\mixingCoefficient_{j'}\right) - \log\! P\left(\queryResponse \!=\!\bar\queryResponse_j \mid \query,\weights\!=\!\bar\weights_j, \mixingCoefficient \!=\! \bar\mixingCoefficient_j\right)\bigg]$}
    \State{select query $\query^{(i)}_*$}
\end{algorithmic}
\end{algorithm}

Algorithm~\ref{alg:04_06_corl21_ranking_info} goes over the pseudocode of our approach, and we discuss the hyperparameters in our experiments in Appendix~\ref{app:99_03_corl21_ranking_hyperparameters}.

\subsubsection{Analysis} \label{subsec:04_06_corl21_ranking_theory}

We start the analysis by stating the bounds on the required number of trajectories in each ranking query to achieve \emph{generic identifiability}. A Plackett-Luce model over $\trajectorySpace$ is generically identifiable if for any sets of parameters $(\weights, \mixingCoefficient)$ and $(\weights', \mixingCoefficient')$ inducing the same distribution over the responses to all queries of size $\abs{\query}$ on $\trajectorySpace$, the mixing coefficients of $(\weights, \mixingCoefficient)$ and $(\weights', \mixingCoefficient')$ are the same and the induced rewards $\trajectoryRewardFunction_\modeIndex(\trajectory)$ are identical across $\trajectorySpace$ up to a constant additive scaling factor.

\begin{restatable}[\citet{zhao2016learning}]{theorem}{identifiability}
\label{thm:04_06_corl21_ranking_identifiability}
A mixture of $\numberOfModes$ Plackett-Luce models with query size $\abs{\query}$ and $\abs{\trajectorySpace}=\abs{\query}$ is {generically identifiable} if $\numberOfModes\leq\left\lfloor \frac{\abs{\query}-2}{2}\right\rfloor!$.
\end{restatable}

This statement follows directly from \cite{zhao2016learning}, which proves the above bound assuming that each query to the Plackett-Luce mixture is a full ranking over the set of items (i.e. $\abs{\trajectorySpace}=\abs{\query}$). However, the assumption $\abs{\trajectorySpace}=\abs{\query}$ is untenable in the active learning context, as it prevents any active query selection. To apply this result for our active learning algorithms, we relax the condition to $\abs{\trajectorySpace}\geq \abs{\query}$.

\begin{restatable}{corollary}{largeident}
\label{thm:04_06_corl21_ranking_largeident}
A mixture of $\numberOfModes$ Plackett-Luce models with query size $\abs{\query}$ is \emph{generically identifiable} if $\numberOfModes\leq\left\lfloor \frac{\abs{\query}-2}{2}\right\rfloor!$.
\end{restatable}

We prove Corollary~\ref{thm:04_06_corl21_ranking_largeident} in Appendix~\ref{app:99_01_corl21_ranking_identproof}. In our context, generic identifiability implies if the human response is modelled by a Plackett-Luce mixture, our Algorithm~\ref{alg:04_06_corl21_ranking_info} will be able to recover its true parameters (up to a constant additive factor for the rewards) in the limit of infinite queries.

\begin{restatable}{remark}{optimality}
\label{thm:04_06_corl21_ranking_remark}
Greedy selection of queries maximizing mutual information in Equation~\eqref{eq:04_06_corl21_ranking_objective} is not necessarily within a constant factor of optimality.
\end{restatable}

Appendix~\ref{app:99_01_corl21_ranking_remark} justifies Remark~\ref{thm:04_06_corl21_ranking_remark}. In fact, greedy optimization of mutual information for adaptive active learning can be significantly worse than a constant factor of optimality in pathological settings \cite{golovin2010near}. Despite its lack of theoretical guarantees, mutual information is a commonly used effective approach in adaptive active learning \cite{houlsby2011bayesian,zheng2005efficient}. Although other approaches like volume removal satisfy adaptive submodularity \cite{sadigh2017active}, they fail in settings with noisy responses by selecting high-noise low-information queries as we showed in Section~\ref{sec:04_01_volume_removal}, and in practice result in far worse performance than mutual information.

\subsection{Experiments}
\label{sec:04_06_experiments}
\begin{figure}[ht]
    \centering
    \includegraphics[width=0.65\linewidth]{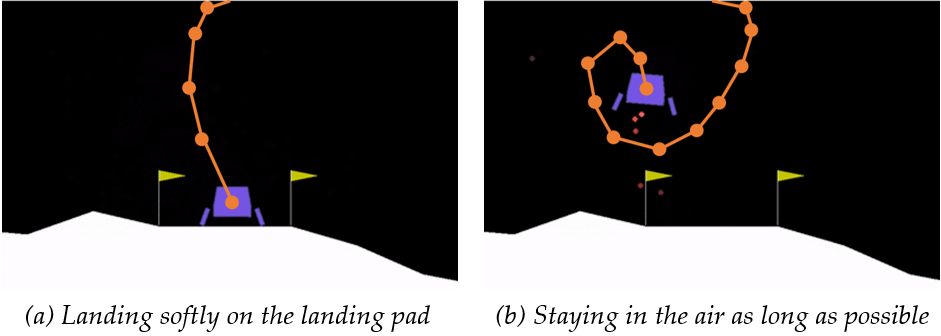}
    \caption{The \emph{LunarLander} environment is visualized with the two tasks. Sample trajectories associated with these tasks are shown.}
    \label{fig:04_06_lunarlander_visual}
\end{figure}
Having presented our learning and active querying algorithms, we now evaluate their performance in comparison with other alternatives. In all simulations and experiments, we assume $\comparisonRationalityCoefficient=1$, and $\trajectoryRewardFunction_{\weights_{\modeIndex}}(\trajectory) = \weights_{\modeIndex}^\top \trajectoryFeaturesFunction(\trajectory)$, i.e., the trajectory reward function of each individual expert is linear in trajectory features. We start with describing the two tasks we experimented with:\\
\noindent\emph{LunarLander.} We used 1000 trajectories in OpenAI Gym's discrete action space LunarLander environment~\cite{brockman2016openai} shown in Figure~\ref{fig:04_06_lunarlander_visual} (see Appendix~\ref{app:99_04_corl21_ranking_lunar_traj} for details on how those trajectories were generated).\\
\noindent\emph{FetchBanana.} We generated $351$ distinct trajectories of the Fetch robot~\cite{wise2016fetch} putting the banana on the shelves as shown in Fig.~\ref{fig:03_05_robot_experiment_visual} (see Appendix~\ref{app:99_04_corl21_ranking_fetch_traj} for details).

\subsubsection{Methods}\label{subsec:04_06_benchmark_methods}
We compare our active querying via mutual information (MI) discussed in Algorithm~\ref{alg:04_06_corl21_ranking_info} with two baselines.
A simple benchmark for active learning is \emph{random} query selection without replacement. We also benchmark against the \emph{volume removal} (VR) method, which we described in Section~\ref{sec:04_01_volume_removal}. See Appendix~\ref{app:99_04_corl21_ranking_baselines} for further details of how we implemented these two baselines.

\subsubsection{Metrics}
We want to evaluate both the active querying and the learning performance. The former requires metrics that assess the quality of the algorithm's selected queries $\rankingDataset = \{(\query^{(i)},\queryResponse^{(i)})\}_i$ in terms of the information they provide on the model parameters $(\weights,\mixingCoefficient)$. We use two such metrics: \texttt{Mean Squared Error (MSE)} and \texttt{Log-Likelihood}. Since both active and non-active methods are expected to reach the same performance with a large number of queries, we look at the area under the curve (AUC) of these two metrics over number of queries. To evaluate the learning performance, we quantify the success of a robot, which learned a multimodal reward, via the \texttt{Learned\_Policy\_Rewards} on the actual task.

\noindent\texttt{MSE.} Suppose we know the human is truly modeled by $(\weights^*, \mixingCoefficient^*)$ adhering to the assumed model class of Section~\ref{subsec:03_05_formulation}. Given a set of queries and responses in $\rankingDataset$, we can compute a maximum likelihood estimate $(\hat\weights, \hat\mixingCoefficient)$ of the model parameters using Equation~\eqref{eq:03_05_pmf}. The \texttt{MSE} is then the squared error between $(\hat\weights, \hat\mixingCoefficient)$ and $(\weights^*, \mixingCoefficient^*)$ (see Appendix~\ref{app:99_04_corl21_ranking_mse}).

While this metric cannot be evaluated with real humans, we can use this metric with synthetic human models (model with known parameters $(\weights^*, \mixingCoefficient^*)$) in simulation. A lower \texttt{MSE} score means the selected queries $\rankingDataset$ allow us to better learn a multimodal Plackett-Luce model close to the true model $(\weights^*, \mixingCoefficient^*)$.

\noindent\texttt{Log-Likelihood.} This metric measures the \texttt{Log-Likelihood} of the response to a random query given the past observations $\rankingDataset$. If the past observations $\rankingDataset$ are informative, the true response to a random query $\query$ will in expectation be more likely, meaning the \texttt{Log-Likelihood} metric will be greater. See Appendix~\ref{app:99_04_corl21_ranking_ll} for details on how we compute this metric.

\noindent\texttt{Learned\_Policy\_Reward.} We take the maximum likelihood estimate of each reward parameters vector and train a DQN policy using them \cite{mnih2013playing}.\footnote{As we are using a real Fetch robot for our experiments and it would be infeasible and unsafe to train DQN on Fetch, this metric is limited to our simulations, i.e., \emph{LunarLander} in our experiments.} We then run these learned policies on the actual environment with the corresponding true reward functions (see Appendix~\ref{app:99_04_corl21_ranking_learned_policy_rewards}) to obtain the \texttt{Learned\_Policy\_Reward} values.

\subsubsection{Results}

\noindent\textbf{Multimodal Learning is Necessary.} We first compare unimodal and multimodal models to show the insufficiency of unimodal rewards when the data come from a mixture. To leave out any possible bias due to active querying, we make this comparison using random querying.

\begin{figure}[ht]
    \centering
    \includegraphics[width=0.6\textwidth]{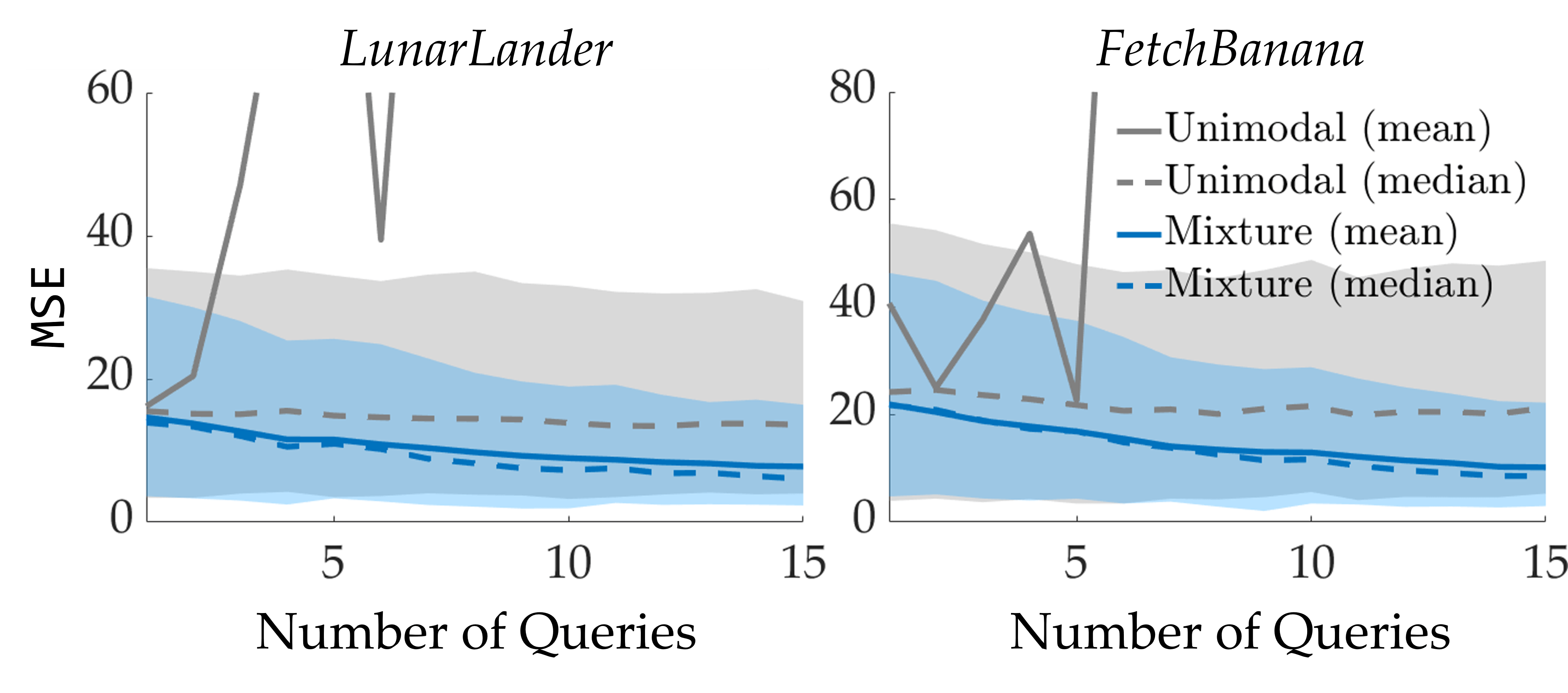}
    \caption{Unimodal and bimodal reward learning models are compared under \texttt{MSE}. Both mean and median values (over $100$ runs) are shown.
    Shaded regions show the first and the third quartiles.
    }
    \label{fig:04_06_unimodal_vs_bimodal}
\end{figure}

We let the true reward function have $\numberOfModes=2$ modes and set a query size of $\abs{\query}=6$ items for identifiability as Section~\ref{subsec:04_06_corl21_ranking_theory} suggests, and for acquiring high information from each query. We simulate $100$ pairs of experts whose reward function parameters $\weights_\modeIndex$ and the mixing coefficients $\mixingCoefficient_\modeIndex$ are sampled from the prior $P(\weights,\mixingCoefficient)$. Having these simulated experts respond to $15$ queries, we report the \texttt{MSE} in Figure~\ref{fig:04_06_unimodal_vs_bimodal}.

The unimodal reward model causes an unstably increasing \texttt{MSE}. This is mostly due to the outliers where the reward function parameters $\weights_1$ and $\weights_2$ are far away from each other and the unimodal reward fails to learn any of them. We therefore also plot the median values and quartiles in Figure~\ref{fig:04_06_unimodal_vs_bimodal}. While the bimodal reward model learned using our proposed approach decreases the \texttt{MSE} over time, the unimodal model has a roughly constant \texttt{MSE}, which suggests it is unable to learn when the data come from a mixture.

We present an additional unimodal learning baseline evaluated on the user study data in Appendix~\ref{app:99_04_corl21_ranking_new_baseline}.

\begin{figure}[ht]
    \centering
    \includegraphics[width=0.6\textwidth]{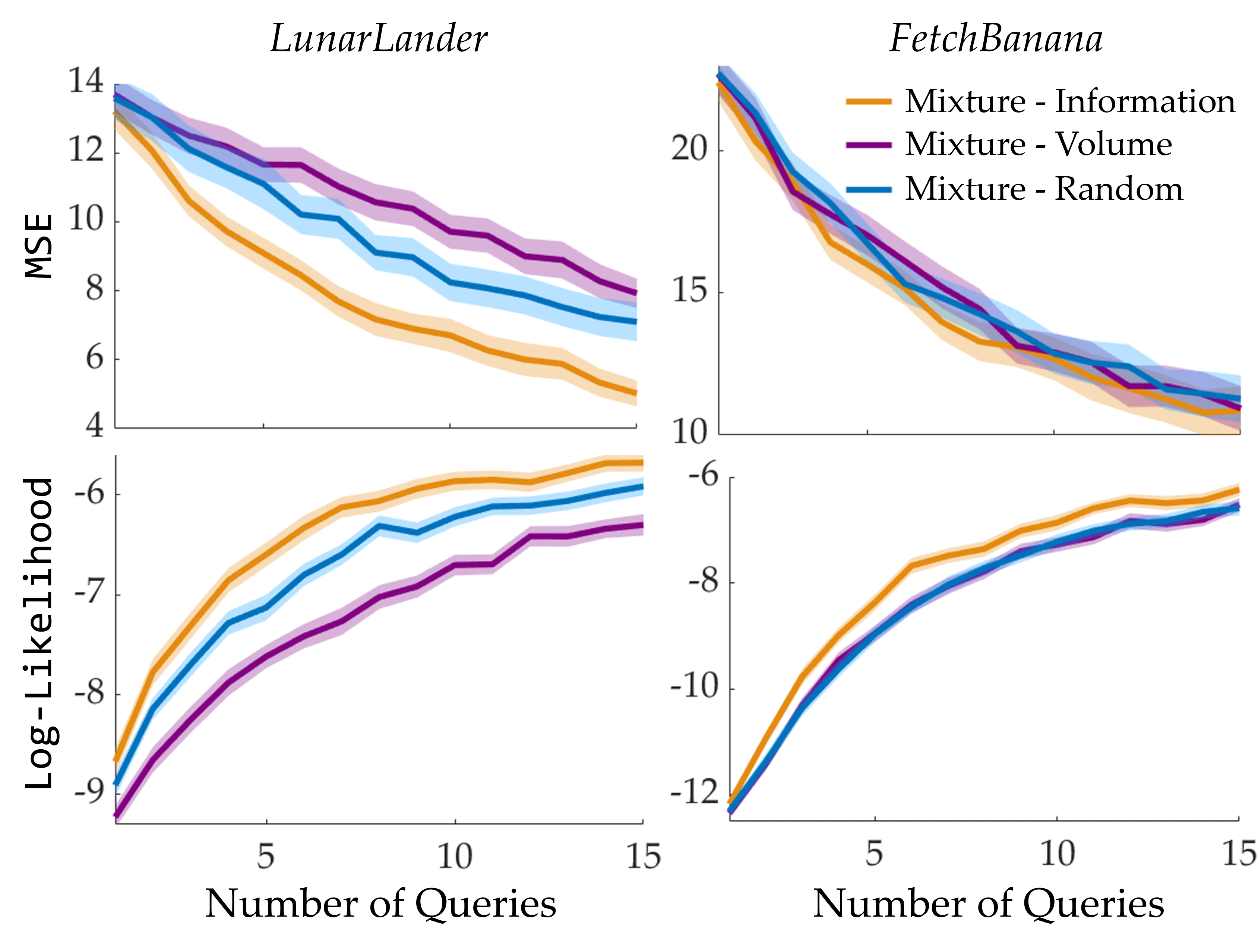}
    \caption{Different querying methods are compared with the (top) \texttt{MSE} and (bottom) \texttt{Log-Likelihood} metrics (mean$\pm$se over $75$ runs).}
    \label{fig:04_06_bimodal}
\end{figure}

\noindent\textbf{Active Querying with Mutual Information is Data-Efficient.} We next compare our mutual information based active querying approach with the other baselines. For this, we use the same experiment setup as above with $\numberOfModes=2$ reward function modes and ranking queries of size $\abs{\query}=6$, and simulate $75$ pairs of human experts. We present the results in terms of \texttt{MSE} in Figure~\ref{fig:04_06_bimodal}. In \emph{LunarLander}, the mutual information objective significantly outperforms both random querying and volume removal in terms of the AUC \texttt{MSE} ($p<0.005$, paired-sample $t$-test). Notably, volume removal performs even worse than the random querying method, which might be due to the known issues of volume removal optimization as briefly discussed in Appendix~\ref{app:99_04_corl21_ranking_volume} and in more detail in Section~\ref{sec:04_01_volume_removal}. On the other hand, the difference is not statistically significant in the \emph{FetchBanana} experiment, which might be due to the small trajectory dataset, or because almost all trajectories in the dataset minimize or maximize some of the trajectory features, accelerating and simplifying learning under the linear reward assumption. See Appendix~\ref{app:99_04_corl21_ranking_fetch_traj} for details about the trajectory features and how we generated the trajectory dataset.

We further analyze the querying methods in this multimodal setting under the \texttt{Log-Likelihood} metric in Figure~\ref{fig:04_06_bimodal}. Mutual information based querying significantly outperforms random querying and volume removal based querying in both experiments with respect to the AUC \texttt{Log-Likelihood} ($p<0.005$).
With respect to the final \texttt{Log-Likelihood}, mutual information reduces the amount of required data in \emph{LunarLander} by about $35\%$ compared to random querying and about $60\%$ to volume removal. Similarly in the \emph{FetchBanana}, the improvement is approximately $25\%$ over both baselines.

Appendix~\ref{app:99_04_corl21_ranking_syn} presents two additional experiments: one which clearly shows the effectiveness of our approach for learning a mixture of more than two reward functions (specifically, $\numberOfModes=5$), and one which studies the robustness against misspecified $\numberOfModes$.

\noindent\textbf{Mutual Information Maximization Leads to Better Learning.} Having seen the superior predictive performance of the reward learned via mutual information maximization, we next assess its performance in the actual environment. As random querying outperforms volume removal in terms of \texttt{Log-Likelihood} and \texttt{MSE} as in Figure~\ref{fig:04_06_bimodal}, we compare the mutual information based method with random querying.

\begin{figure}[ht]
    \centering
    \includegraphics[width=0.6\textwidth]{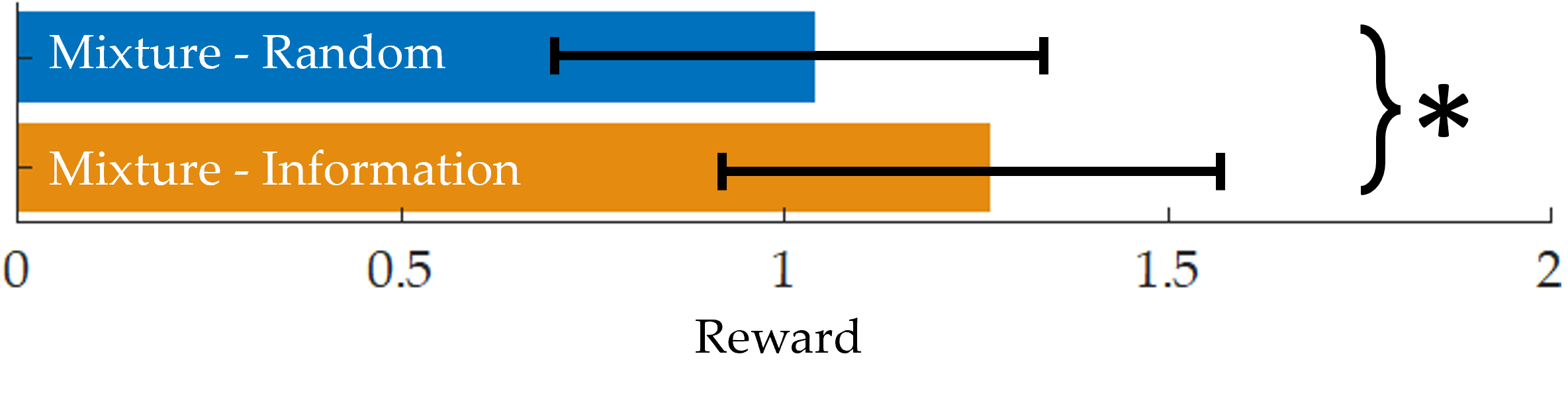}
    \caption{Mutual information based and random querying methods are compared with the \texttt{Learned\_Policy\_Reward} values (mean$\pm$se over $75$ runs which correspond to $150$ randomly generated reward function parameters) in \emph{LunarLander}.}
    \label{fig:04_06_bimodal_dqn}
\end{figure}

For this, we run the multimodal reward learning with $75$ pairs of randomly generated reward function parameters ($\numberOfModes=2$ and $\abs{\query}=6$). For each of the $150$ individual reward functions, we compute the \texttt{Learned\_Policy\_Reward} values. Figure~\ref{fig:04_06_bimodal_dqn} shows the results. While the standard errors in the plots seem high, this is mostly because optimal trajectories for different reward parameters differ substantially in terms of rewards, which causes an irreducible variance. However, since the underlying true rewards are the same between the mutual information based and random querying methods, we ran the paired sample $t$-test between the results and observed statistical significance ($p<0.05$). This means although the \texttt{Learned\_Policy\_Reward} values between different runs differ substantially, the reward function learned via the mutual information method leads to better task performance compared to random querying.

\subsection{User Studies}

We now empirically analyze the performance of our algorithm with two online user studies. We again used the \emph{LunarLander} and \emph{FetchBanana} environments. We provide a summary and a video of the user studies and their results at \url{https://sites.google.com/view/multimodal-reward-learning}.

\noindent\textbf{Experimental Setup.} For \emph{LunarLander}, subjects were presented with either of the following instructions at every ranking query: ``Land softly on the landing pad" or ``Stay in the air as long as possible". We randomized these instructions such that users get one of them with $0.6$ and the other with $0.4$ probability. We kept the presented instructions hidden from the learning algorithms so that they need to learn a multimodal reward without knowing which mode each ranking belongs to.

For the \emph{FetchBanana} environment, we recorded the $351$ trajectories on the real robot as short video clips so that the experiment can be conducted online under the pandemic regulations. Human subjects participated in the experiment as groups of two to test learning from multiple users. Each participant was instructed that the robot needs to put the banana in one of the shelves and different shelves have different conditions (the same as in our running example in Section~\ref{sec:03_05_rankings}, see Figure~\ref{fig:03_05_robot_experiment_visual}, Appendix~\ref{app:99_04_corl21_ranking_desc}).

After emphasizing there is no one correct choice and it only depends on their preferences, we asked each participant to indicate their preferences between the shelves on an online form. Afterwards, each group of two subjects responded to $30$ ranking queries in total where each query consisted of $6$ trajectories. We selected who responds to each query randomly, with probabilities $0.6$ and $0.4$.

Appendix~\ref{app:99_04_corl21_ranking_ui} presents details on the user interface used in our experiments.

\noindent\textbf{Independent Variables.} We varied the querying algorithm: active querying with mutual information and random querying. We excluded the volume removal based method to reduce the experiment completion time for the subjects, as it already performed worse than random querying in our simulations.

\noindent\textbf{Procedure.} We conducted the experiments as a within-subjects study. We recruited $24$ participants (ages 19 -- 56; $9$ female, $15$ male) for \emph{LunarLander} and $26$ participants (ages 19 -- 56; $11$ female, $15$ male) for the \emph{FetchBanana}. Each subject in the \emph{LunarLander}, and each group of two subjects in the \emph{FetchBanana} experiment responded to $40$ ranking queries; $15$ with each algorithm and $10$ random queries for evaluation at the end. The order of the first $30$ queries was randomized to prevent bias.

\noindent\textbf{Dependent Measures.} Learning the multimodal reward functions via the $15$ rankings collected by each algorithm, we measured the \texttt{Log-Likelihood} of the final $10$ rankings collected for evaluation.

\noindent\textbf{Hypotheses.} With \emph{LunarLander} and \emph{FetchBanana}, we test the following hypotheses respectively:\\
    \textbf{H16.} \textit{Querying the participants, who are trying to teach two different tasks, actively with mutual information maximization will lead to faster learning than random querying.}\\
    \textbf{H17.} \textit{While learning from two people with different preferences, active querying with mutual information maximization will lead to faster learning than random querying.}

\begin{figure}[ht]
    \centering
    \includegraphics[width=0.6\textwidth]{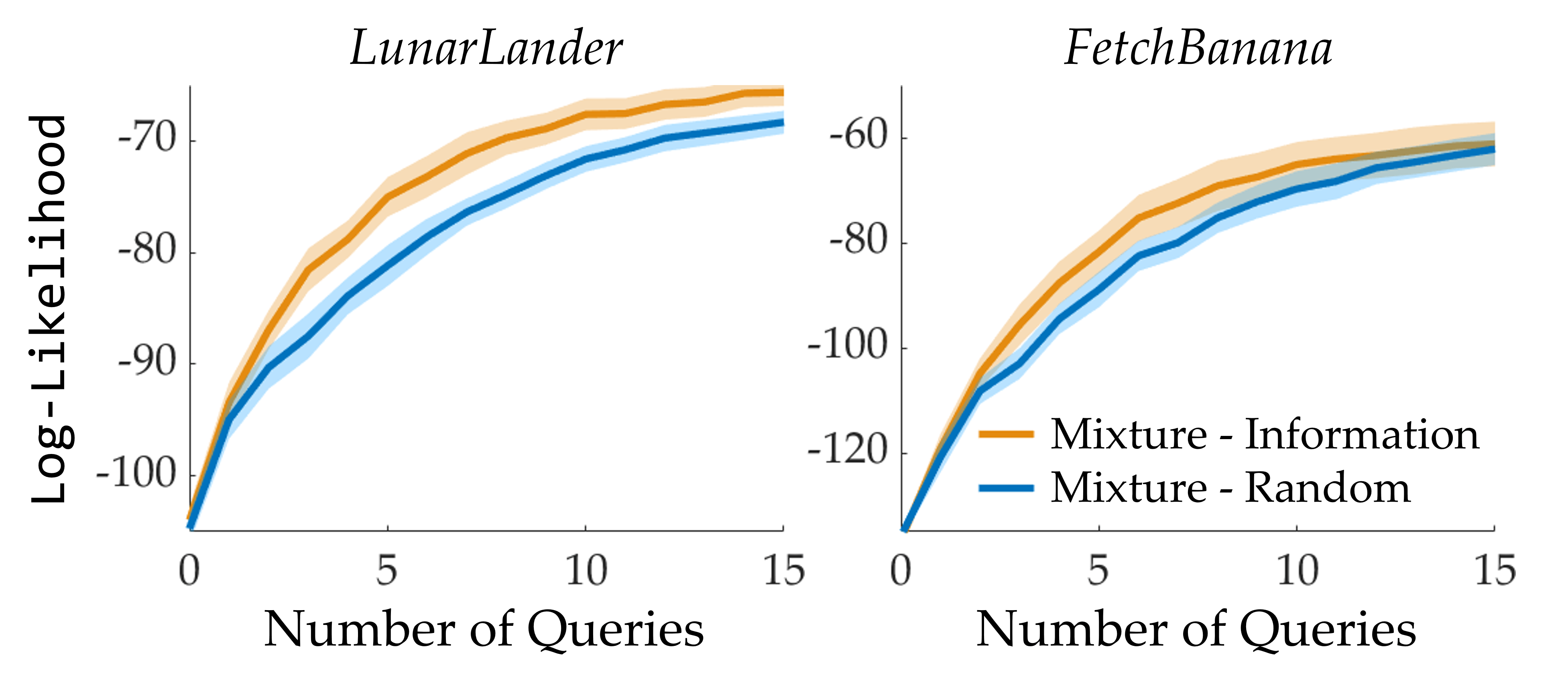}
    \caption{User study results (mean$\pm$se over $24$ users for \emph{LunarLander} and $13$ groups for \emph{FetchBanana}).}
    \label{fig:04_06_user_studies_loglikelihood}
\end{figure}

\noindent\textbf{Results.} Figure~\ref{fig:04_06_user_studies_loglikelihood} visualizes how \texttt{Log-Likelihood} of the evaluation queries changes over the course of learning by both algorithms. Active querying with mutual information maximization leads to significantly faster learning compared to random querying in \emph{LunarLander}. Indeed, the difference in AUC \texttt{Log-Likelihood} is statistically significant ($p<0.05$). Furthermore, the active querying method enabled reaching the final performance of random querying after only $9$ or $10$ queries, for around a $35\%$ reduction in the amount of data needed, supporting \textbf{H16}.

As \emph{FetchBanana} experiments have an easier task with a small number of variables between the trajectories, both querying methods converge to similar performances by the end of $15$ queries. However, active querying accelerates learning in the early stages---the difference in AUC \texttt{Log-Likelihood} is again statistically significant ($p<0.05$). Looking at the final performance with random querying, improvement in data efficiency is about $10\%$, supporting \textbf{H17}.

%% file: 04_active/07_hierarchical.tex
\label{sec:04_07_hierarchical}

A special case of multimodal rewards we studied in Section~\ref{sec:03_06_hierarchical} is a setting where the user's reward function transitions from one mode to another based on a parametric model that depends on the behavior observed in the previous trajectories. We showed that using hierarchical queries, such non-stationary rewards could be learned along with the transition models. In this section, we extend that learning algorithm with active querying based on the maximum volume removal approach we presented in Section~\ref{sec:04_01_volume_removal}. One could also use the mutual information maximization based active querying technique from Section~\ref{sec:04_02_information_gain} to avoid the shortcomings of maximum volume removal, which we discussed in detail in Section~\ref{sec:04_01_volume_removal}.

Similar to Sections~\ref{sec:04_02_information_gain} through \ref{sec:04_06_rankings}, and \cite{basu2018learning}, we will actively select hierarchical choice queries from a query database to learn a probability distribution over \emph{reward dynamics}: a collection of unimodal reward functions representing different moods and a set of parameters for the transitions between the moods. 

In this section, we provide this algorithm which actively selects hierarchical choice queries in order to efficiently learn reward dynamics. We evaluate our active querying algorithm on an autonomous driving example, which we also used as a running example in Section~\ref{sec:03_06_hierarchical}. We show in simulations that we can efficiently learn changes in preferences when preferences indeed vary based on behavior and interactions in the environment.

\subsection{Active Querying based on Maximum Volume Removal}
In this section, $\hierarchicalDataset$ denotes the dataset of all the hierarchical choice queries and their responses up to the current iteration $i$. Each response tuple $(\queryResponse^{(i,1)},\queryResponse^{(i,2)})$ removes some volume from the hypothesis space of $(\weights,\utilityWeights)$, where, volume removed is given as the difference between the unnormalized posterior distribution over $(\weights,\utilityWeights)$, and its prior distribution. We have a belief over what the user responses could be. We leverage this probabilistic model to \emph{actively} select a query at each iteration that will maximize the expected volume removal. We again restrict our implementation to the case where $\abs{\query^{(i)}}=3$, i.e., we first show a trajectory to the user, and then ask $2$ best-of-many choice questions where the trajectories in the first question start from the last state of the first trajectory and the trajectories in the second question start from the last state of the user's choice in the first question. Then, we need to solve the following optimization problem:
\begin{align}
(\query^{(i,0)}_*,\query^{(i,1)}_*,\query^{(i,2)}_*) = \argmax_{\query^{(i,0)},\query^{(i,1)},\query^{(i,2)}} & \mathbb{E}_{\queryResponse^{(i,1)},\queryResponse^{(i,2)}\mid \hierarchicalDataset,\query^{(i,0)},\query^{(i,1)},\query^{(i,2)}}\Big[\nonumber \\ &\mathbb{E}_{\weights,\utilityWeights \mid \hierarchicalDataset}\left[1 - P(\queryResponse^{(i,1)}, \queryResponse^{(i,2)} \mid \weights, \utilityWeights, \query^{(i,0)},\query^{(i,1)},\query^{(i,2)})\right]\Big]
\end{align}
To compute the inner expectation, we first sample $(\weights, \utilityWeights)$ from $P(\weights,\utilityWeights\mid\hierarchicalDataset)$ using Markov Chain Monte Carlo methods. Similar to previous sections, we use Metropolis-Hastings algorithm \cite{chib1995understanding}. Now, we let $\bar{\weights}$ and $\bar{\utilityWeights}$ represent those samples, so:
\begin{align*}
(\query^{(i,0)}_*,\query^{(i,1)}_*,\query^{(i,2)}_*) \asymeq \argmin_{\query^{(i,0)},\query^{(i,1)},\query^{(i,2)}} & \mathbb{E}_{\queryResponse^{(i,1)},\queryResponse^{(i,2)} \mid \hierarchicalDataset, \query^{(i,0)}, \query^{(i,1)}, \query^{(i,2)}}\Big[\\ 
&\sum_{\bar{\weights},\bar{\utilityWeights}} P(\queryResponse^{(i,1)}, \queryResponse^{(i,2)} \mid \bar{\weights},\bar{\utilityWeights},\query^{(i,0)},\query^{(i,1)},\query^{(i,2)})\Big]
\end{align*}
where $\asymeq$ denotes asymptotic equality as the number of samples $(\bar{\weights}, \bar{\utilityWeights})$ goes to infinity. Expanding $\mathbb{E}_{\queryResponse^{(i,1)},\queryResponse^{(i,2)} \mid \hierarchicalDataset, \query^{(i,0)}, \query^{(i,1)}, \query^{(i,2)}}$, the minimization objective is
\begin{align*}
\sum_{\queryResponse^{(i,1)},\queryResponse^{(i,2)}}P(\queryResponse^{(i,1)},\queryResponse^{(i,2)} \mid \query^{(i,0)},\query^{(i,1)},\query^{(i,2)},\hierarchicalDataset)\sum_{\bar{\weights},\bar{\utilityWeights}} P(\queryResponse^{(i,1)},\queryResponse^{(i,2)} \mid \bar{\weights},\bar{\utilityWeights},\query^{(i,0)},\query^{(i,1)},\query^{(i,2)})
\end{align*}
where the first sum is over all possible response sequences, i.e., $\query^{(i,1)} \times \query^{(i,2)}$. By the law of large numbers, we can write the optimization as:
\begin{align}
&\argmin_{\query^{(i,0)},\query^{(i,1)},\query^{(i,2)}} \sum_{(\queryResponse^{(i,1)},\queryResponse^{(i,2)})\in \query^{(i,1)} \times \query^{(i,2)}}\left(\sum_{\bar{\weights},\bar{\utilityWeights}} p(\queryResponse^{(i,1)},\queryResponse^{(i,2)} \mid \bar{\weights},\bar{\utilityWeights},\query^{(i,0)},\query^{(i,1)},\query^{(i,2)})\right)^2\:,
\end{align}
because $P(\queryResponse^{(i,1)}, \queryResponse^{(i,2)} \mid \query^{(i,0)}, \query^{(i,1)}, \query^{(i,2)}, \hierarchicalDataset) = \mathbb{E}_{\bar{\weights}, \bar{\utilityWeights}}\left[P(\queryResponse^{(i,1)}, \queryResponse^{(i,2)} \mid \bar{\weights}, \bar{\utilityWeights}, \query^{(i,0)}, \query^{(i,1)}, \query^{(i,2)})\right]$
where the probability expression in the objective function is already derived in Section~\ref{subsec:03_06_reward_dynamics}.

\subsection{Simulations and Experiments}
\subsubsection{Problem Domain}
We focus on learning non-stationary reward functions for driving, using a version of the \emph{Driver} environment presented earlier in Section~\ref{subsec:04_02_experiments}. Each mode of the learned \emph{reward dynamics} weighs $5$ features for driving that attempt to encode safety and traffic rules: one feature for penalizing closeness to the edges of the road, one for velocity, and three more features of proximity to lane centers, to other cars and alignment with the road, similar to \cite{sadigh2017active}.\footnote{While traffic rules are not explicitly modeled in our simulation, many of the features can be weighed appropriately to encode them. For example, the feature for velocity measures the deviation from the maximum allowed speed which can be easily adapted to the road type.} Similar to what we did in Section~\ref{sec:04_06_rankings} with multimodal reward functions, we assume each mode corresponds to a reward function that is linear in these trajectory features.

Each sub-query consists of the driving environment and a pair of trajectories of the ego car whose preferred behavior we are learning (i.e., $\abs{\query^{(i,j)}}=2$ for all iterations $i$ and $j\in\{1,2\}$), and another car in the scenario. Our query database consists of $10,\!000$ randomly generated hierarchical choice queries.

\subsubsection{Dependent Measures}
In our implementation, we learned $\Delta\utilityWeights := \utilityWeights_1 - \utilityWeights_2$, instead of $\utilityWeights$, as it has fewer parameters.\footnote{Note we cannot do the same trick for $\weights$, because while $\utilityWeights$ in Equation~\eqref{eq:03_06_simplified_response_model} can be simplified to $\Delta\utilityWeights$, there is no such simplification on $\weights$.} The same approach generalizes to any $\numberOfModes$ --- we can simply subtract $\utilityWeights_{\numberOfModes}$ from $\utilityWeights_\modeIndex$ for $\forall \modeIndex\in[\numberOfModes-1]$ to reduce the number of parameters to be learned.

We measure the performance of hierarchical preference learning in terms of expected dot product between learned weights and true weight as in previous sections and \cite{sadigh2017active}, separately for each component of $(\weights,\Delta\utilityWeights)$:
\begin{align}
\texttt{Alignment}(\weights_1) = \mathbb{E}\left[\frac{\hat{\weights_1}.\weights_1^*}{\norm{\hat{\weights_1}}_2\norm{\weights_1^*}_2}\right]
\end{align}
$\hat{\weights_1}$ and $\weights_1^*$ are the estimated and true weights, respectively, and the expectation is taken over the sampled $\hat{\weights_1}$ values. We define \texttt{Alignment} similarly for $\weights_2$ and $\Delta\utilityWeights$. Hence, it is a measure of convergence, as its value being close to $1$ indicates the learned weights are close to the true weights.

\begin{figure}[ht]
    \centering
    \includegraphics[width=0.6\textwidth]{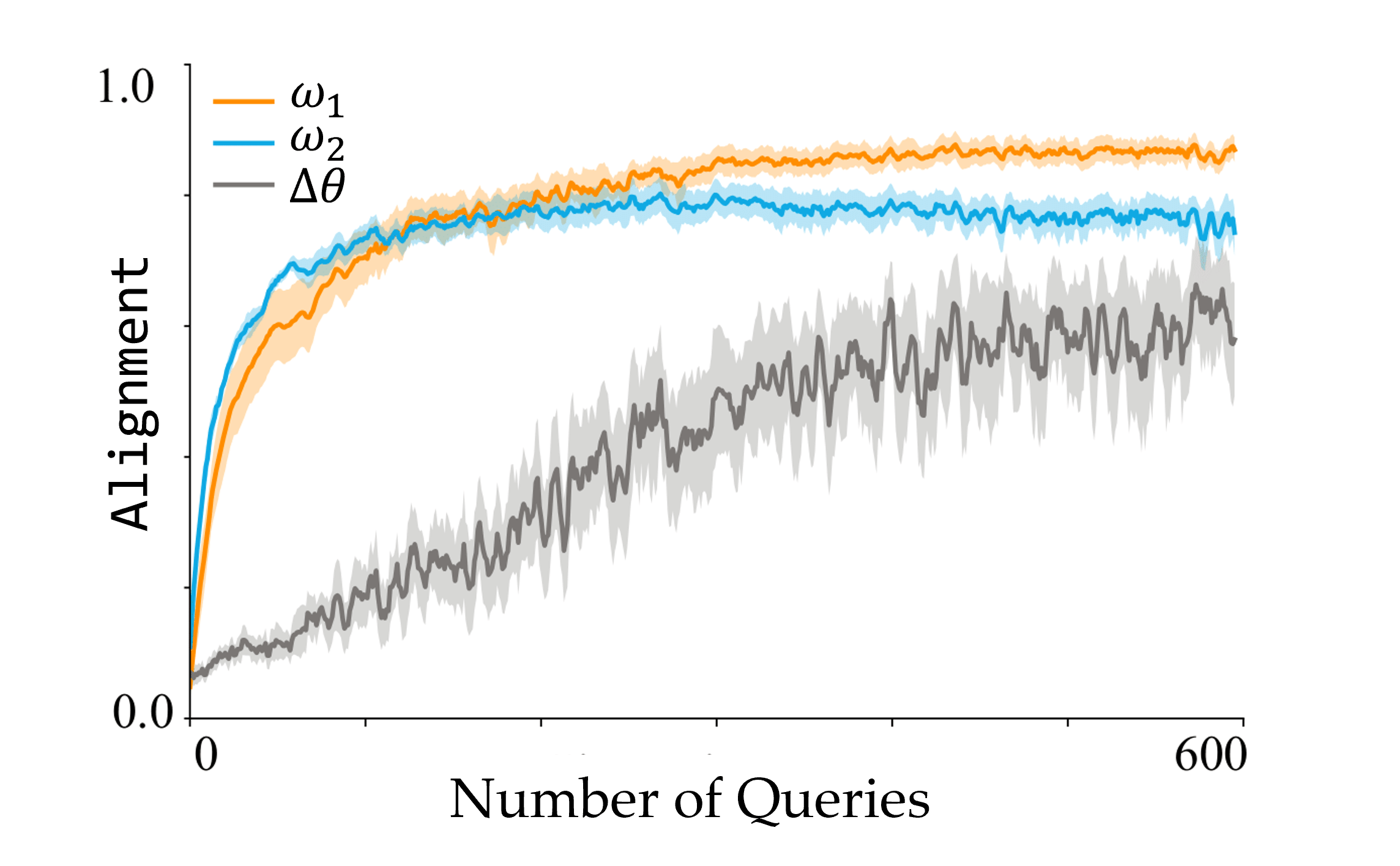}
    \caption{\texttt{Alignment} value shows that our algorithm converges well for non-driving data with non-active query selection when the simulated user is \emph{oracle}. Here we show an average \texttt{Alignment} over 5 different ground truth \emph{reward dynamics}.}
    \label{fig:04_07_m_random}
\end{figure}

\subsubsection{Experiments with Random Data}
We first conduct experiments with completely random and independent sub-queries without the driving environment. We assume we can generate queries in an unconstrained way such that any $\trajectoryFeaturesFunction$-vector is possible, i.e. there is no dynamics constraint in the generation of queries. This is similar to the \emph{LDS} environment we used in Section~\ref{subsec:04_02_experiments}. Here, we simulate \emph{oracle users}: users who are perfectly aware of their true reward dynamics. That is, they always behave (change mode and respond) with respect to the higher probability out of response and mode transition models. In Figure~\ref{fig:04_07_m_random}, the average results of $5$ different simulated oracle users show convergence of $(\weights_1, \weights_2, \Delta\utilityWeights)$ whose true values were independently drawn from standard normal distribution independently for each entry.

\subsubsection{Experiments with Driving Data} 
\begin{figure}[ht]
    \centering
    \includegraphics[width=0.75\textwidth]{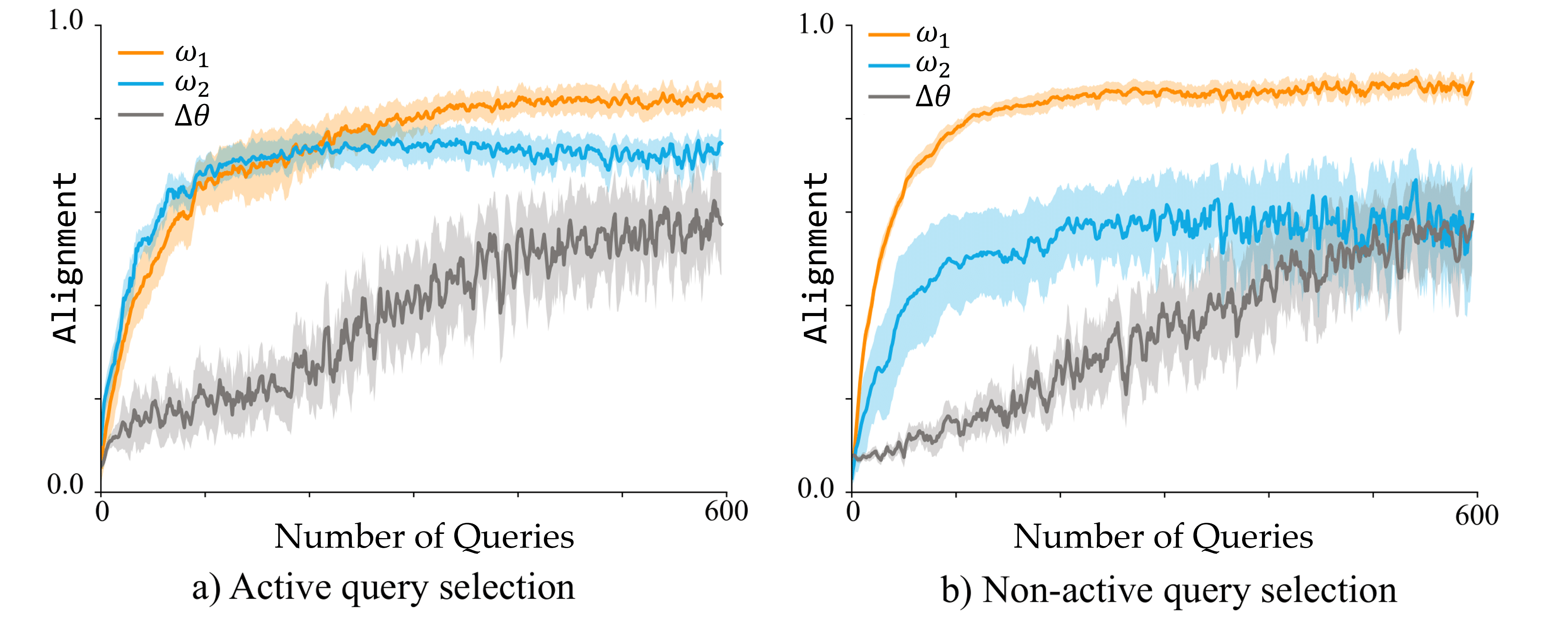}
    \caption{\texttt{Alignment} values show that our algorithm with active query selection (left) can learn reward dynamics faster than non-active query selection (right) when the simulated user is \emph{oracle}. Here we show an average $\texttt{Alignment}$ over $5$ different ground truth reward dynamics.}
    \label{fig:04_07_m_oracle}
\end{figure}

\noindent\textbf{Active versus non-active query selection.}
We compare the performance of our active query selection algorithm with a non-active baseline where we uniformly sample the queries from the discrete database of $10,\!000$ queries. Here our simulated users are always oracle. We test the following hypothesis:\\
\textbf{H18.} \textit{The reward dynamics learned with our active query selection algorithm converges to the true parameters faster compared to the non-active baseline.}
Our results in Figure~\ref{fig:04_07_m_oracle}
support this hypothesis by demonstrating that active query selection accelerated the learning of one of the modes ($\weights_2$) compared to the non-active baseline.

\noindent\textbf{Testing different mode preferences.}
Next we simulate $5$ noisy users, who choose between options with respect to $P(\queryResponse \mid \weights,\utilityWeights,\query)$ as we defined in Equation~\eqref{eq:03_06_simplified_response_model}. Our algorithm actively selects queries from the same discrete dataset of size $10,\!000$ as in the case of oracle users. We first set the following hypothesis:\\
\textbf{H19.} \textit{Our algorithm learns the reward dynamics even when the users are noisy.}

We also test the performance of our algorithm for different mode likelihoods, i.e. probability of transitioning to a given mode. We manipulated the ground truth reward dynamics to reflect different mode likelihoods. For example, one user might be in one mode 80\% of the time while another user has equal chances of being in one of the two modes. Although this might actually affect the priors $P_1$ and $P_2$ as we explained in Section~\ref{subsec:03_06_reward_dynamics}, we still adopted the derivations based on uniform prior to test the robustness of our framework. Therefore, we test the following hypothesis:\\
\textbf{H20.} \textit{Our algorithm learns the reward function parameters $\weights$ that correspond to both modes, and it converges faster for $\weights_\modeIndex$ if the user is more likely to be in mode $\modeIndex$.}

\begin{figure}[ht]
    \centering
    \includegraphics[width=0.75\textwidth]{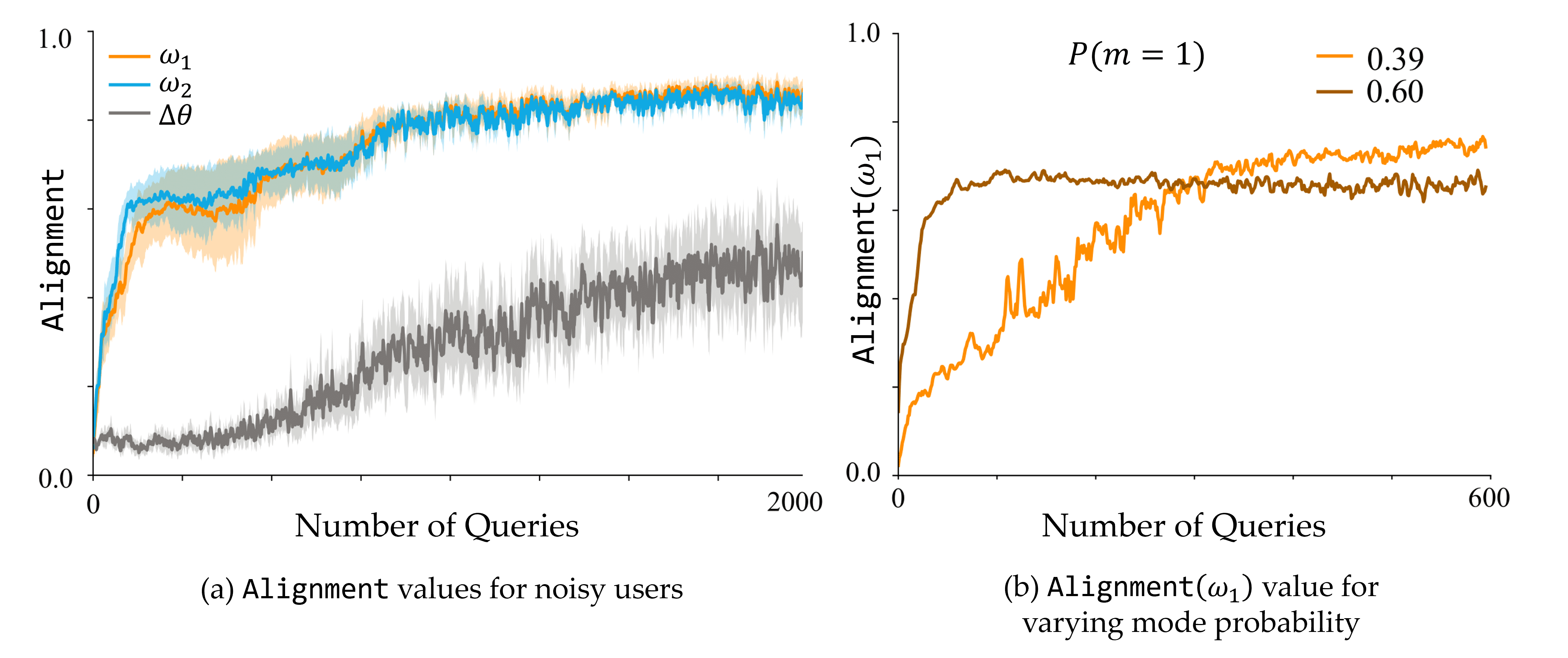}
    \caption{\texttt{Alignment} value shows even when the users are noisy our algorithm can learn the true \emph{reward dynamics} (left) and that as the probability being at mode $1$ increases, $\weights_1$ converges faster (right).}
    \label{fig:04_07_m_noisy_modechange}
\end{figure}

Figure~\ref{fig:04_07_m_noisy_modechange}a shows that our algorithm was able to learn $\weights_1$, $\weights_2$ and $\Delta\utilityWeights$ even when the users are noisy, supporting \textbf{H19}. We note that in general we learn $\Delta\utilityWeights$ slowly and we need more queries for its convergence. The second plot shows that we are able to learn the reward weights $\weights_\modeIndex$ of a mode $\modeIndex$ regardless of its likelihood probability being high or low. The same plot also shows the algorithm converges faster for the modes that are visited more often. This is intuitive, as the algorithm is able to gather more information about those modes, even though it does not perfectly know that the user is in the corresponding mood. Hence, $\textbf{H20}$ has strong empirical support.

\subsubsection{User Study}
\noindent\textbf{Hypotheses.} We test the following hypotheses with the user study:\\
\textbf{H21.} \textit{Our algorithm learns weights that can represent the driving behavior of the users.}\\
\textbf{H22.} \textit{Some people indeed change preferences depending on the driving behaviors of the interacting agents.}

\begin{figure}[ht]
    \centering
    \includegraphics[width=0.75\textwidth]{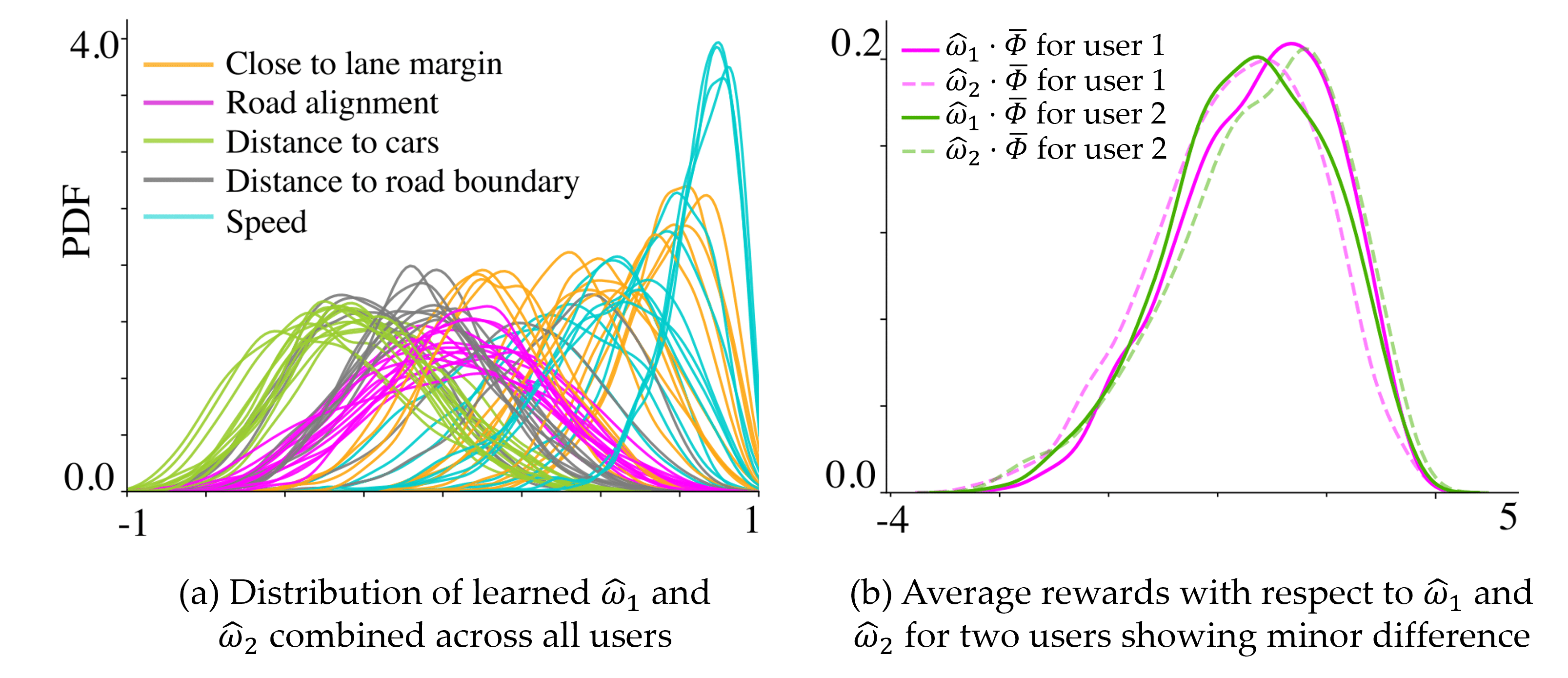}
    \caption{Distribution of $\hat{\weights}_1$ and $\hat{\weights}_2$ across all users for individual features. (a) User preferences vary widely for adherence to lane center and distance to road boundaries, but are very similar for efficiency (speed) and safe driving (collision avoidance). (b) While we did not learn significantly different $\weights_1$ and $\weights_2$ for individual users, the average reward with respect to $\hat{\weights}_1$ and $\hat{\weights}_2$ differs slightly for some of our study participants.}
    \label{fig:04_07_jointdist}
\end{figure}

\noindent\textbf{Study Design.}
To validate our hypotheses, we collected data from 10 real users in a within-subjects study. We first learn a general reward dynamics $(\hat{\weights}, \hat{\utilityWeights})$ using $50$ hierarchical queries. We then use the posterior distributions over these parameters to jump-start the process for each subject with a reasonable prior that better represents legally correct driving. During validation, we ask users to provide ratings for trajectories locally optimized with respect to the learned reward dynamics. We compare the expressiveness of the learned reward parameters ($\hat{\weights}_1,\hat{\weights}_2$) against their perturbed versions ($\weights_1^p, \weights_2^p$). We sampled these perturbed versions from Gaussian distributions centered around ($\hat{\weights}_1,\hat{\weights}_2$) and a standard deviation of $0.5 \times \norm{\hat{\weights}_1}$ and $0.5 \times \norm{\hat{\weights}_2}$. While creating perturbed versions of $\hat{\weights}_1$ and $\hat{\weights}_2$, as an attempt to ensure legally correct driving, we constrain the weights for the features that correspond to staying within the road and avoiding collision with cars. We also compare with $\weights_{\textrm{r}}$ sampled from a Gaussian distribution centered on either $\hat{\weights}_1$ or $\hat{\weights}_2$ with a standard deviation of $2 \times \norm{\hat{\weights}_\modeIndex}$ with the corresponding mode index $\modeIndex$. Each rating question consists of two parts. The first part is similar to $\query^{(i,0)}$ of the learning step, where we show user one trajectory demonstration of the robot as an attempt to set their initial mode. In the next part, we show users $5$ trajectories continued from the first part, optimal with respect to $5$ reward functions parameterized with: $\hat{\weights}_1$, $\hat{\weights}_2$, their perturbed versions $\weights_1^p$ and $\weights_2^p$, and $\weights_{\textrm{r}}$. For each of the $5$ trajectories, we ask users a 7-point rating scale question: \emph{Indicate your level of agreement with the following statement: I would like to ride this car}.

In \textbf{H21} we claim 1) users will give the highest overall rating to the trajectories that are optimal with respect to $\trajectoryRewardFunction_{\hat{\weights}_1}$ and/or $\trajectoryRewardFunction_{\hat{\weights}_2}$ most of the time, and 2) if probability of being at mode $\modeIndex$ is very high, we expect people to give the highest rating to trajectories that suit to $\trajectoryRewardFunction_{\hat{\weights}_\modeIndex}$. To validate the first part, we repeat the same demonstration across several rating queries preserving the trajectory of the other car in the environment alike and changing the trajectory of the ego agent, varying between different local optima with respect to $\weights_1$, $\weights_2$ and the other weights. We randomize demonstration trajectories across the rating questions. In \textbf{H22} we hypothesize that subject to different interactions in the environment, users will sometimes give higher rating to trajectory optimal with respect to $\trajectoryRewardFunction_{\hat{\weights}_1}$ and sometimes to those optimal with respect to $\trajectoryRewardFunction_{\hat{\weights}_2}$.

\noindent\textbf{Results.}
We found that the users have somewhat similar preferences: proximity to cars has high negative weight, and speed has high positive weight showing that people generally prefer safe and efficient driving (see Figure~\ref{fig:04_07_jointdist}). On the other hand, features that encode staying on the road and alignment with the road vary more. While the general direction of the feature weights is similar between $\hat{\weights}_1$ and $\hat{\weights}_2$ for each user, there is some difference in the magnitudes. We computed the percentage difference between average reward with respect to $\trajectoryRewardFunction_{\hat{\weights}_1}$ and $\trajectoryRewardFunction_{\hat{\weights}_2}$ as $\frac{\hat{\weights}_1\cdot\bar{\trajectoryFeaturesFunction}-\hat{\weights}_2\cdot\bar{\trajectoryFeaturesFunction}}{\hat{\weights}_1\cdot\bar{\trajectoryFeaturesFunction}}$, where $\bar{\trajectoryFeaturesFunction}$ is the average feature values of the trajectories in our query database. This gave us the percentage difference in the average reward. We found that of all the users the maximum difference is 12\% and the minimum difference is 6\%. While we also learned $\Delta\utilityWeights$, it becomes relatively unimportant here, as $\weights_1$ and $\weights_2$ are very close.

\begin{figure}[t]
    \centering
    \includegraphics[width=0.8\textwidth]{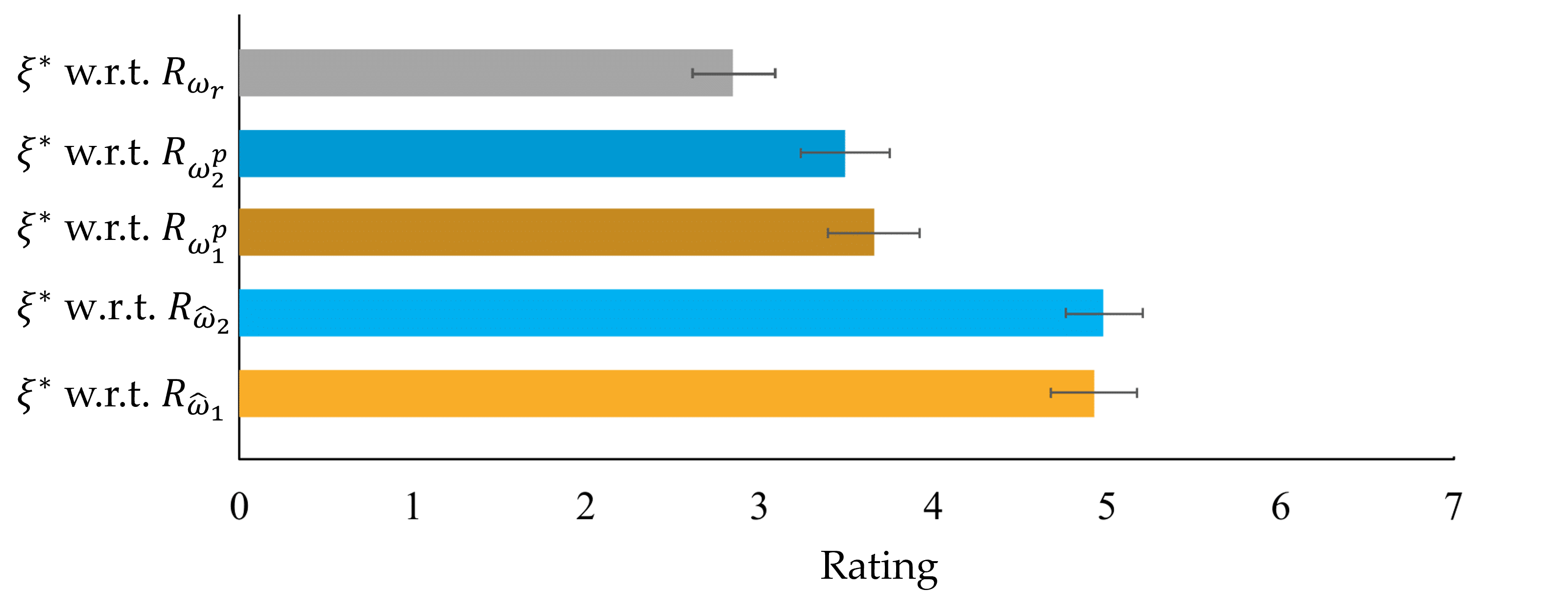}
    \caption{Most users gave high ratings to the trajectories optimal with respect to $\trajectoryRewardFunction_{\hat{\weights}_1}$ and $\trajectoryRewardFunction_{\hat{\weights}_2}$ and low ratings to trajectories optimal with respect to their perturbed versions $\trajectoryRewardFunction_{\weights_1^p}$ and $\trajectoryRewardFunction_{\weights_2^p}$ and the lowest rating to the trajectories that were optimal with respect to a reward function that is parameterized randomly $\trajectoryRewardFunction_{\weights_r}$.}
    \label{fig:04_07_ratings}
\end{figure}

As it can be seen in Figure~\ref{fig:04_07_ratings}, the users gave the highest scores to the trajectories that are optimal with respec to $\trajectoryRewardFunction_{\hat{\weights}_1}$ and $\trajectoryRewardFunction_{\hat{\weights}_2}$ with statistical significance. This suggests an empirical evidence for \textbf{H21}. While we also observed that users sometimes gave high ratings to the trajectories of $\trajectoryRewardFunction_{\hat{\weights}_1}$ and sometimes to those of $\trajectoryRewardFunction_{\hat{\weights}_2}$, we have not observed a significant dependence on the modes. This is due to the fact that the learned weights were very close to each other as they represent the legal driving behavior, which is a very small subset of all the reward space. Further, our simulation environment may not be realistic enough to elicit emotions like anger, frustration etc. that cause behavioral changes in different traffic situations \cite{shinar1998aggressive,zhang2015dimensions,lotz2018recognizing}. Therefore, our results are inconclusive about \textbf{H22}.

With this section, we completed extending the learning methods we presented in Chapter~\ref{chap:learning} with active querying techniques. In the next section, we describe how these techniques can be executed in batch settings where multiple questions are actively generated at the same time.

%% file: 04_active/08_batch.tex
\label{sec:04_08_batch}
Two important drawbacks of active query generation are the following: (i) the robot needs to optimize for each and every query, (ii) the querying process cannot be parallelized, i.e., even if multiple users are available to give comparative feedback, the robot needs to query them sequentially because each query is actively generated based on the responses to all the previous questions. Therefore, even though active querying leads to significant gains in terms of data-efficiency, it might hurt time-efficiency, especially when optimizing queries is difficult. In some cases, it is desirable to be able to quickly generate queries and ask them to multiple users in parallel.

We thus propose using methods that generate a \emph{batch} of comparison queries optimized at the same time as opposed to generating queries one after the other. These batch methods not only improve time-efficiency, but also have other computational benefits. For example, they can help when fitting the learning model is expensive, e.g., as in Gaussian processes (as in Sections~\ref{sec:04_03_gp} and \ref{sec:04_04_roial}), as the model should be retrained only after all queries in the batch are responded, rather than after every single query. In addition, these methods are parallelizable, which is a desirable feature when the robot is learning from multiple humans.

While larger batches amplify these advantages, they can hurt data-efficiency, because new queries become less optimized with respect to the queries made earlier (and so the learned model so far). Hence, there is a direct tradeoff between \emph{the required number of queries} and the \emph{time it takes to generate each query}. Besides, it is challenging to decide how an informative batch must be generated.  While a batch of random queries hurts data-efficiency, finding the optimal batch is computationally intractable because it requires an exhaustive search over all possible human responses to the queries in the batch.

Ideally, we would like to generate queries actively for the highest data-efficiency while generating each query time-efficiently. In this section, we propose a new set of algorithms --- \emph{batch active comparison-based learning} methods --- that balance this tradeoff between the number of queries it requires to learn human preferences and the time it spends on generation of each query.

To this end, we actively generate each batch based on the data collected so far. We focus on pairwise comparison queries due to their simplicity, but the same techniques can be easily extended to any query type we discussed in this thesis. In our framework with pairwise comparisons, we select and query $\batchSize$ pairs of trajectories, to be compared by the user or users, at once. Since $\batchSize$ queries are generated at once, our framework is parallelizable for data collection as opposed to standard active querying methods that require data to come sequentially.

\begin{figure*}[t]
	\centering
	\includegraphics[width=\textwidth]{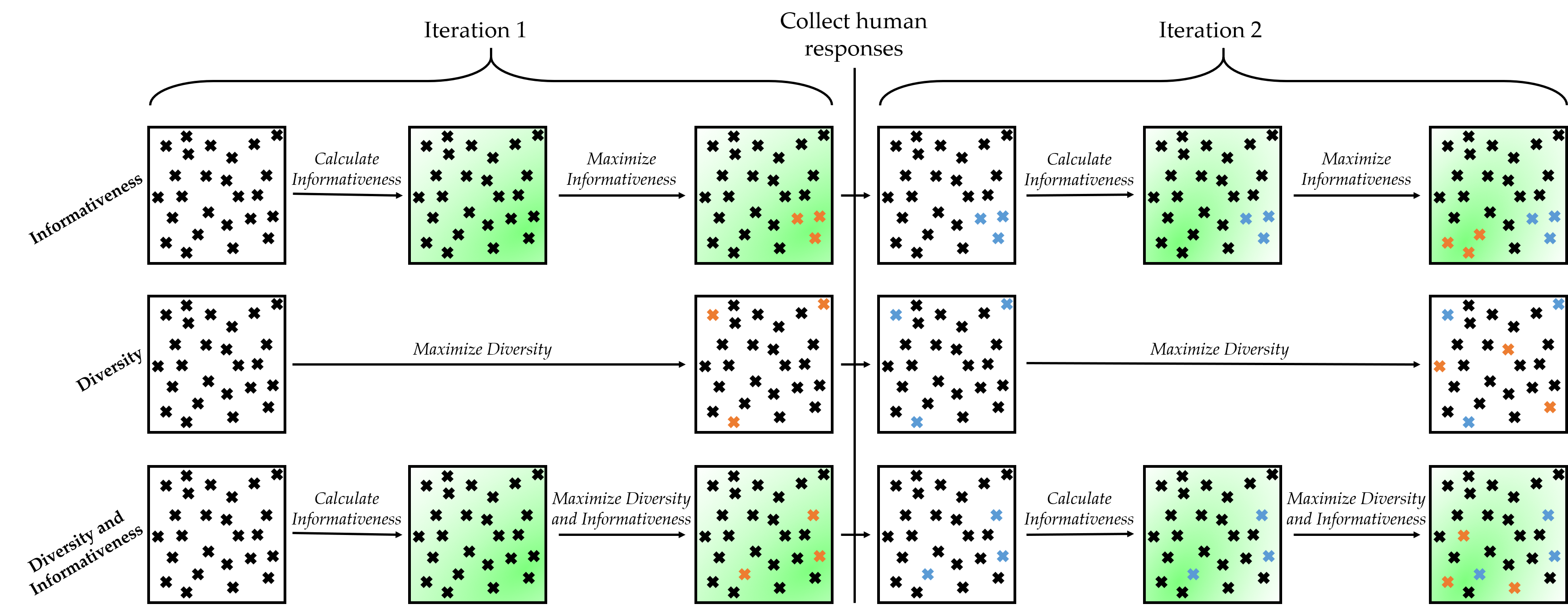}
	\caption{Batches should be both diverse and informative in batch active learning. Here, a hypothetical batch selection problem is visualized. Each cross represents a query. Similar queries are close to each other. Orange shows the queries selected in that iteration, and blue shows the queries for which the human responses have already been collected in the previous iterations. Green color represents informativeness: darker regions correspond to the queries with high informativeness based on the information collected until that iteration. \textbf{(Top)} Maximizing only informativeness generates batches that include very similar queries which, when queried together, carry redundant information. \textbf{(Middle)} Maximizing only diversity does not take informativeness into account at all, and so is wasteful as it selects some queries that are not informative. \textbf{(Bottom)} A good batch active learning algorithm should both select informative queries and avoid redundancy.}
	\label{fig:04_08_front_fig}
\end{figure*}

What makes batch active learning more difficult than standard active learning problems is that we cannot select the queries by simply maximizing their informativeness. Since a batch of queries is selected all at once, they must be selected without any information about the user responses to the queries within that batch. The batch active learning methods should then try to maximize the diversity between the queries in order to avoid selecting very similar queries in a single batch \cite{yang2015multi,cardoso2017ranked}. Therefore, a good batch active learning method must produce batches that consist of both dissimilar and informative queries. This is visualized in Figure~\ref{fig:04_08_front_fig}.

The problem of actively generating a batch of data is well-studied in other machine learning problems such as classification \cite{wei2015submodularity,elhamifar2016dissimilarity, yang2019single}, where decision boundaries may inform the active learning algorithms. While this may simplify the problem, it is not applicable in our setting, where we attempt to actively learn a reward function for dynamical systems using pairwise comparison queries as opposed to data point - label pairs where the labels are directly associated with the corresponding data points.

While existing batch active learning methods are not readily applicable in our problem, we have the same challenge of generating both informative and diverse batches. For this, determinantal point processes (DPP) are a natural fit. DPPs are a mathematical tool that is often used for generating diverse batches from a set of items \cite{kulesza2012determinantal} and are used to generate batches in other machine learning applications, such as for improving the convergence of stochastic gradient descent \cite{zhang2017determinantal,zhang2019active}. Here, we propose using DPPs to generate not only diverse but also informative batches in active preference-based reward learning.

We summarize our contributions in this section as:
\begin{enumerate}[nosep]
    \item Developing a batch active learning algorithm based on determinantal point processes (DPP) that leads to the highest performance by balancing the tradeoff between the informativeness and diversity of queries.
    \item Designing a set of approximation algorithms for efficient batch active learning to learn about human preferences from pairwise comparison queries.
    \item Experimenting and comparing approximation methods for batch active learning in complex comparison based learning tasks.
    \item Showcasing our framework in predicting human users' preferences in simulated autonomous driving and robotics tasks.
\end{enumerate}

For the rest of the section, we will start with formalizing the problem. We will then present how standard active methods select queries. After introducing the general batch-mode active learning idea, we propose our methods for batch selection. First, we propose a method based on determinantal point processes. Next, we propose more time-efficient alternatives which might be preferable when batch sizes are large or to avoid hyperparameter tuning. After proposing these different approaches, we give theoretical guarantees for three of them. Finally, we present our experiments with both simulated and real users.

\subsection{Formulation}
We consider the setup we presented in Section~\ref{sec:03_01_pairwise_comparisons}, for the special case of pairwise comparisons. which we later extended with maximum volume removal based active querying in Section~\ref{sec:04_01_volume_removal}. In this section, we will again use the maximum volume removal based method, but the batch generation algorithms we propose are agnostic to the choice of the acquisition function; so for example, mutual information (see Section~\ref{sec:04_02_information_gain}) or max regret (see Section~\ref{sec:04_05_scale}) can also be used in practice. Similarly, although we focus on parametric reward functions in this section, the algorithms we propose can also be used for non-parametric reward functions. THe batch active querying algorithms require only two things: (i) a score for each query that represents queries' informativeness, and (ii) a similarity metric between the queries.

In the setup for this section, we start with a prior belief $\belief^0$ over the reward function parameters $\weights$. This prior might be initialized with some domain knowledge and/or some expert demonstrations. Most active querying techniques, as we discussed in previous sections, relies on samples from this belief; and then in later querying iterations, samples from the updated beliefs $\belief^i$. However, we do not know the shape of this distribution. As a result, we used Metropolis-Hastings \cite{chib1995understanding} in the previous sections to get the samples. In this section, since our goal is to increase the time-efficiency of active querying, our first change in the algorithm is to use a more efficient sampling technique. For this, we approximate $P(\queryResponse \mid \query, \weights)$, which we modeled in Equation~\ref{eq:03_01_noisily_optimal} with a log-concave function whose mode always evaluates to $1$:
\begin{align}
	P(\queryResponse \mid \query=(\trajectory_1,\trajectory_2), \weights) &= \min(1,\exp((-1)^{\mathbb{I}(\queryResponse=\trajectory_2)} (\trajectoryRewardFunction(\trajectory_1)-\trajectoryRewardFunction(\trajectory_2))))\:,
	\label{eq:04_08_human_noise_approx} 
\end{align}
where $\mathbb{I}$ denotes the indicator function. This allows us to efficiently use an adaptive Metropolis algorithm \cite{haario2001adaptive} for sampling.

Our goal is to learn the human's reward function, or equivalently $\weights$, in a both data-efficient and time-efficient way. To this end, we develop batch-active comparison-based reward learning methods, which actively generate a batch of pairwise comparison queries based on the previous queries and the human's responses to them.

Our insight is that we can in fact balance between the number of queries required for convergence to $\trajectoryRewardFunction_\weights$ and the time required to generate each query. We construct this balance by introducing a \emph{batch active learning} approach, where $\batchSize$ queries are simultaneously generated at a time based on the current estimate of $\weights$. The batch approach can significantly reduce the total time required for the satisfactory estimation of $\weights$ at the expense of increasing the number of queries needed for convergence to true $\trajectoryRewardFunction_\weights$.

To obtain a batch of queries that are informative, we need to find queries that optimize an acquisition function, e.g., volume removal as computed by the objective of Equation~\eqref{eq:04_01_VR2}. We again fall back to a discretization method for generating batches of queries: we discretize the space of all possible pairwise comparison queries by randomly sampling $\numberOfQueries$ pairs of feasible trajectories from $\trajectorySpace$. While increasing $\numberOfQueries$ may lead to more accurate optimization results, the computation time also increases linearly with $\numberOfQueries$.

The batch active learning problem we are trying to solve is then an optimization that attempts to find the $\batchSize$ pairwise comparison queries out of $\numberOfQueries$ that will maximize the volume removal in the worst case in terms of the human's responses (or the expected volume removal --- see the equivalence in Appendix~\ref{app:99_01_ijrr_dpp_worst}). However, such a problem is often computationally hard (see \cite{cuong2013active} and \cite{chen2013near} for the proofs with similar objectives), requiring an exhaustive search which is intractable in practice as the search space is exponentially large \cite{guo2008discriminative}.
	
\begin{algorithm}[ht]
	\caption{Batch Active Comparison-based Learning}
	\label{alg:04_08_batch_active_learning}
	\begin{algorithmic}[1]
		\State Generate query dataset $\datasetOfQueries = ({\trajectory_i}_1,{\trajectory_i}_2)_{i=1}^{\numberOfQueries}$ where each trajectory comes from $\trajectorySpace$
		\State $\comparisonDataset \gets \emptyset$
		\For{$\textrm{iteration } i = 1, 2, \dots$}
		\State Get samples $\bar\weights\sim \belief^{(i-1)k}(\weights)$
		\State Generate a query batch of size $\batchSize$ using $\bar\weights$ from the query set
		\State Get the human response for each query in the batch
		\State Update the belief $\belief^{(i-1)k}(\weights)$ with the new data to get $\belief^{ik}(\weights)$
		\EndFor
		\State\Return $\mathbb{E}\left[\weights \mid \comparisonDataset\right]$
	\end{algorithmic}
\end{algorithm}

In the subsequent sections, we present our batch generation algorithms that attempt to find approximately optimal batches using various techniques and time-efficient heuristics. Algorithm~\ref{alg:04_08_batch_active_learning} gives an overview of the overall batch active comparison-based learning approach: line 1 discretizes the space of queries, line 4 samples a set of $\weights$ from the belief distribution $\belief^{(i-1)k}(\weights)=P(\weights \mid \comparisonDataset)$. Line 5 produces a batch of queries, for which we present several methods in the subsequent sections. After the human responses are collected for the queries in the batch in line 6, the posterior belief is updated in line 7 with respect to Equation~\eqref{eq:03_01_DP7}.

\subsection{DPP-based Batch Active Learning}\label{subsec:04_08_dpp_bal}

Determinantal point processes (DPP) are a class of distributions that promote diversity. They are a natural fit for our problem as they can be tuned to balance the tradeoff between diversity and how desirable each item is. In our approach, we regard the set of queries as the item set of DPPs. We first start with presenting the necessary background on DPPs.

\subsubsection{Background}
A point process is a probability measure on a ground set $\datasetOfQueries$ over finite subsets of $\datasetOfQueries$. In our batch active comparison-based learning framework, $\datasetOfQueries$ is a set of queries. We let $\abs{\datasetOfQueries}=\numberOfQueries$.

An $\dppKernel$-ensemble defines a DPP through a real, symmetric and positive semidefinite (PSD) $\numberOfQueries$-by-$\numberOfQueries$ kernel matrix $\dppKernel$ \cite{borodin2005eynard}. Then, sampling a subset $\batch=\batchExample\subseteq \datasetOfQueries$ has the probability
\begin{align}
P(\batch=\batchExample) \propto \det{\dppKernel_\batchExample}
\label{eq:04_08_dpp}
\end{align}
where $\dppKernel_\batchExample$ is an $\abs{\batchExample}$-by-$\abs{\batchExample}$ matrix that consists of the rows and columns of $\dppKernel$ that correspond to the queries in $\batchExample$. For instance, if $\batchExample=\{i,j\}$, i.e., $\batchExample$ is a set consisting of $i^{\textrm{th}}$ and $j^{\textrm{th}}$ queries in $\datasetOfQueries$, then
\begin{align*}
P(\batch=\batchExample)\propto \dppKernel_{ii}\dppKernel_{jj} - \dppKernel_{ij}\dppKernel_{ji}.
\end{align*}
We can consider $\dppKernel_{ij}=\dppKernel_{ji}$ as a similarity measure between the queries $i$ and $j$ in the set. The nonnegativeness of the second term in the above expression shows an example of \emph{repulsiveness} property of DPPs. This property makes DPPs the ubiquitous tractable point process to model negative correlations, and useful for generating diverse batches.

As $\det{\dppKernel_\batchExample}$ can be positive for various $\batchExample$ with different cardinalities, we do not know $\abs{\batchExample}$ in advance. There is an extension of DPPs referred to as $\batchSize$-DPP where it is guaranteed that $\abs{\batchExample}=\batchSize$, and Equation~\eqref{eq:04_08_dpp} remains valid \cite{kulesza2011k}. In this section, we employ $\batchSize$-DPPs and refer to them as DPPs for brevity.

\begin{figure}[ht]
	\centering
	\includegraphics[width=0.5\textwidth]{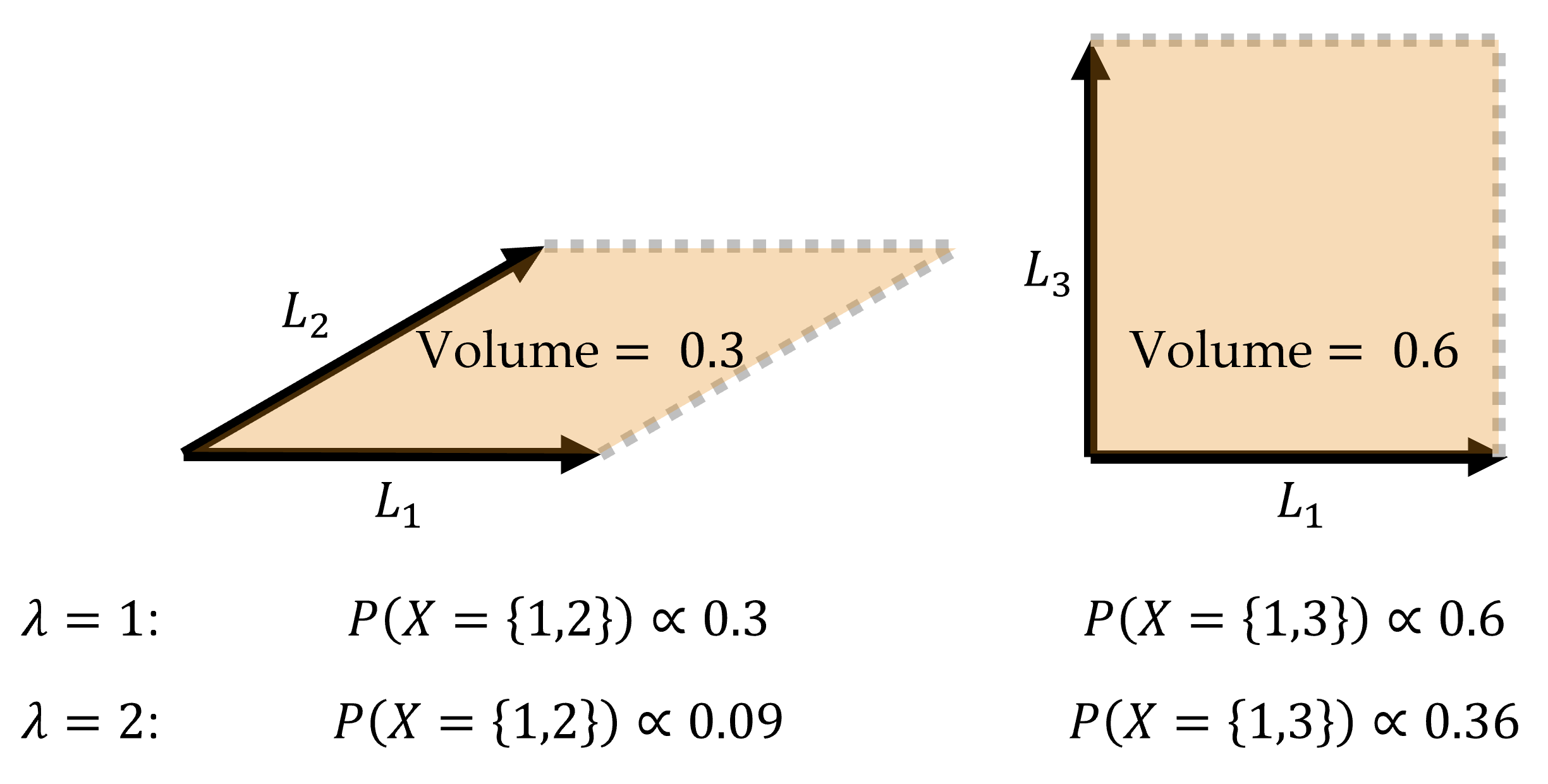}
	\caption{The effect of $\diversityParameter$ is visualized. The columns of the matrix $\dppKernel$ have the same magnitude here; however $\{1,3\}$ is a more diverse set than $\{1,2\}$. When $\diversityParameter=1$, $\{1,3\}$ is two times more likely to be sampled from the DPP distribution than $\{1,2\}$. When we increase $\diversityParameter$ to $2$, this ratio increases to $4$, since more diverse sets are boosted against the less diverse sets.}
	\label{fig:04_08_dpp_alpha}
\end{figure}

Now, we explain what parameters we can have in an $\dppKernel$-ensemble DPP. We note that
\begin{align}
P(\batch=\batchExample)\propto \det \dppKernel_\batchExample = \textrm{Vol}(\{\dppKernel_i\}_{i\in \batchExample})\:,
\end{align}
so the probability is proportional to the square of the associated volume.\footnote{Volume here refers to the volume of the parallelepiped spanned by the columns of $\dppKernel$, whereas the volume removal in the previous sections referred to the change in the belief distribution between the prior and the unnormalized posterior.} In fact, by using a generalized version of DPP, we can approximately achieve \cite{anari2019log,mariet2018exponentiated}:
\begin{align}
P(\batch=\batchExample)\propto \textrm{Vol}^{\diversityParameter}(\{\dppKernel_i\}_{i\in \batchExample})\:,
\end{align}
for $\diversityParameter\geq0$. One can note that higher $\diversityParameter$ enforces more diversity, because the probability of more diverse sets (larger volumes) will be boosted against the less diverse sets. We visualize this in Figure~\ref{fig:04_08_dpp_alpha}.

What remains is to construct the kernel matrix $\dppKernel$. For this, we first define a matrix $\similarityMatrix\in\mathbb{R}^{\numberOfQueries\times \numberOfQueries}$ whose entries measure the similarity between the queries. In our problem, every query $i$ has a feature difference vector $\featureDifference_i = \trajectoryFeaturesFunction({\trajectory_i}_1) - \trajectoryFeaturesFunction({\trajectory_i}_2)$, and close $\featureDifference$'s (in terms of Euclidean distance) correspond to similar queries in terms of the information they provide. Therefore, we let
\begin{align}
    \similarityMatrix_{ij} = \exp\left(-\frac{\norm{\featureDifference_i-\featureDifference_j}_2^2}{2\similarityMatrixHyperparameter^2}\right)\:,
\end{align}
where $\similarityMatrixHyperparameter$ is a hyperparameter. However, we are not restricted to this choice --- we could use distance metrics other than Euclidean distance. We then define the matrix $\dppKernel$ as
\begin{align}
\dppKernel_{ij} = \score_i^{\scoreParameter/\diversityParameter}\similarityMatrix_{ij}\score_j^{\scoreParameter/\diversityParameter}\:,
\end{align}
which is guaranteed to be PSD by the construction of $\similarityMatrix$. Here, $\scoreParameter$ is another hyperparameter and $\score_i\in\mathbb{R}_{\geq0}$ is the \emph{score} of $i^\textrm{th}$ query that represents how much we want that query in our batch. We use these scores to weight the queries based on how much volume they will remove from the belief distribution, as computed by the objective of Equation~\eqref{eq:03_01_DP7}. By increasing $\scoreParameter$ for fixed $\diversityParameter$, we give more importance to the scores than diversity. This enables us set the tradeoff between informativeness and diversity.

\noindent\textbf{Relating the Mode of a DPP with High Diversity and Informativeness.} With proper tuning of $\diversityParameter$ and $\scoreParameter$, the batches that are both diverse and informative will have higher probabilities of being sampled. This motivates us to find the mode of the distribution, i.e., $\argmax_{\batchExample} P(\batch = \batchExample)$, which will guarantee informativeness and diversity. Another advantage of using the mode, instead of a random sample from the distribution, is the fact that it is significantly faster to approximate, even compared to the approximate sampling methods \cite{anari2016monte,li2016fast,mariet2018exponentiated,anari2019log}.

\subsubsection{Approximating the Mode of a DPP}
Finding the mode of a DPP exactly is NP-hard \cite{ko1995exact}. It is hard to even approximate it better than a factor of $2^{x\batchSize}$ for some $x>0$, under a cardinality constraint of size $\batchSize$ \cite{civril2013exponential}. Here, we discuss a greedy optimization algorithm to approximate the mode of a DPP.

In this approach, queries are greedily added to the batch. More formally, to approximate
\begin{align*}
\argmax_\batchExample P(\batch=\batchExample) = \argmax_\batchExample \textrm{Vol}^{\diversityParameter}(\{\dppKernel_i\}_{i\in \batchExample}),
\end{align*}
we greedily add queries to $\batchExample$. Let $\batchExample^{(j)}$ denote the set of selected queries at iteration $j$ of batch generation. We have
\begin{align*}
\batchExample^{(j+1)} = \batchExample^{(j)} \cup \{\argmax_{i'}\textrm{Vol}^{\diversityParameter}(\{\dppKernel_i\}_{i\in A^{(j)}\cup\{i'\}})\}\:,
\end{align*}
which we repeat until we obtain $\batchSize$ queries in $\batchExample$. \citet{ccivril2009selecting} showed that the greedy algorithm always finds a $\batchSize^{O(\batchSize)}$-approximation to the mode.

An important advantage of greedily approximating the mode is that the hyperparameter $\diversityParameter$ becomes irrelevant, as it is just an exponent in the objective in every iteration of batch generation, unless trivially $\diversityParameter=0$. This reduces the burden of hyperparameter tuning. 


\subsubsection{Overall Algorithm}

Having presented the background in DPPs and the method to approximately find the DPP-mode, which corresponds to our diverse and informative batch, we are now ready to present our overall DPP-based batch active comparison-based learning algorithm.

As noted earlier, we work with a discretized set of queries. While this set has $\numberOfQueries$ queries, it might be computationally prohibitive to approximate the DPP mode (even greedily) if $\numberOfQueries$ is large. In such cases, we first reduce the query set into a smaller set $\reducedQuerySet$ by picking the queries which will individually remove the highest volume. Algorithm~\ref{alg:04_08_reduction} presents the pseudocode for this procedure.

\begin{algorithm}[ht]
	\caption{$\textsc{ReduceDataset}(\bar\weights,\datasetOfQueries, \abs{\reducedQuerySet})$}
	\label{alg:04_08_reduction}
	\begin{algorithmic}[1]
		\Statex \textbf{Input: } $\bar\weights_1,\bar\weights_2,\dots$\Comment{Sampled $\weights$ estimates}
		\Statex \textbf{Input: } $\datasetOfQueries:=\left((\trajectory_{1,1},\trajectory_{1,2}),\dots,(\trajectory_{\numberOfQueries,1},\trajectory_{\numberOfQueries,2})\right)$\Comment{Dataset of queries}
		\Statex \textbf{Input:} $\abs{\reducedQuerySet}$\Comment{Desired size of the reduced query set}
		\For{$j=1,\dots,\numberOfQueries$}
		\State $\featureDifference_j \gets \trajectoryFeaturesFunction(\trajectory_{j,1}) - \trajectoryFeaturesFunction(\trajectory_{j,2})$
		\State $\score_j \gets \min_{\queryResponse\in\{\trajectory_{j,1},\trajectory_{j,2}\}}\mathbb{E}_{\bar\weights}\left[1-P(\queryResponse \mid \bar\weights)\right]$\Comment{Volume removal of query $j$ (see Equation~\eqref{eq:04_01_VR1_worst_case})}
		\EndFor
		\State $\reducedQuerySet\gets$ $\featureDifference_j$'s with $\abs{\reducedQuerySet}$ highest $\score_j$ values \Comment{Reduction}
		\State $\mathbf{\score} \gets \score_j$ values corresponding to $\reducedQuerySet$
		\State\Return $\reducedQuerySet, \mathbf{\score}$
	\end{algorithmic}
\end{algorithm}

Afterwards, we approximately compute the mode of the DPP distribution over this reduced set $\reducedQuerySet$ as our batch. Algorithm~\ref{alg:04_08_dpp} presents the DPP-based method. The first for-loop (lines 2 through 7) constructs the DPP kernel, and the second part (lines 8 through 11) generates the batch by greedily approximating the mode of the constructed DPP.

\begin{algorithm}[ht]
	\caption{DPP-based Batch Generation}
	\label{alg:04_08_dpp}
	\begin{algorithmic}[1]
		\Require DPP hyperparameters $\similarityMatrixHyperparameter$ and $\scoreParameter$, sampled $\weights$ estimates $\bar\weights_1,\bar\weights_2,\ldots$
		\State $\reducedQuerySet, \mathbf{\score} \gets \textsc{ReduceDataset}(\bar\weights,\datasetOfQueries, \abs{\reducedQuerySet})$
		\For{$i=1,\dots,\abs{\reducedQuerySet}$}
		\For{$j=1,\dots,\abs{\reducedQuerySet}$}
		\State $\similarityMatrix_{ij} \gets \exp\left(-\frac{\norm{\featureDifference_i-\featureDifference_j}_2^2}{2\similarityMatrixHyperparameter^2}\right)$
		\State $\dppKernel_{ij} \gets \score_i^{\scoreParameter}\similarityMatrix_{ij}\score_j^{\scoreParameter}$
		\EndFor
		\EndFor
		\State $\batchExample \gets \emptyset$ \Comment{Initialize the batch}
		\For{$j=1,\dots,\batchSize$}
		\State $\batchExample \gets \batchExample \cup \{\argmax_{j'}\det{\dppKernel_{\batchExample\cup\{j'\}}}\}$
		\EndFor
		\State\Return $\batchExample$
	\end{algorithmic}
\end{algorithm}

\subsection{Time-Efficient Batch Active Learning Methods}

DPP-based batch active learning method enables us to systematically tune the tradeoff between diversity and informativeness. This approach leads to the best learning performance as we will present in our experiments. However, the DPP method has two important drawbacks: (i) it requires tuning of the hyperparameters $\similarityMatrixHyperparameter$ and $\scoreParameter$, and (ii) even approximating the mode of the DPP might take too much time depending on the batch size $\batchSize$ and the reduced set size $\abs{\reducedQuerySet}$, as we need to compute the matrix $\dppKernel$ and determinants of some of its submatrices at every greedy iteration. Although the former problem may be tolerated as it can be performed offline, the latter may cause problems in practice. DPP-based method is useful for the cases when parallelization in data collection is desired, but the time-efficiency is not crucial.

\begin{figure*}[ht]
	\centering
	\includegraphics[width=\textwidth]{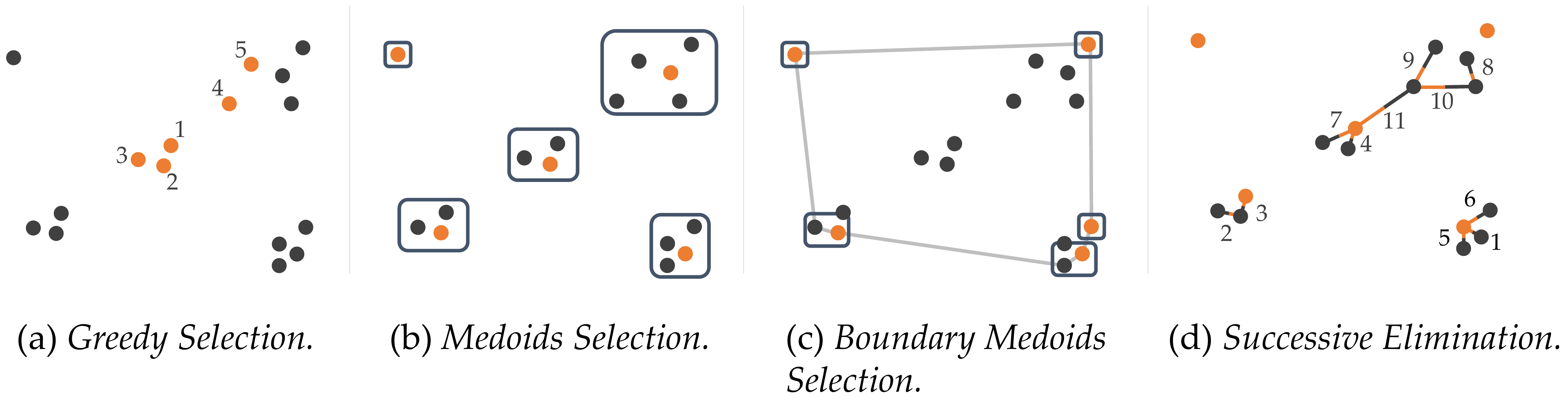}
	\caption{Visualizations of the batch generation process of the proposed time-efficient batch active learning algorithms. In each visual, a simple 2D space with $16$ different $\featureDifference$ values that correspond to the reduced set $\reducedQuerySet$ is shown. The goal is to select a batch of $\batchSize=5$ that will near-optimally maximize the joint volume removal. The selected queries are shown in orange. (a) Greedy Selection. (b) Medoids Selection. The points are selected based on the $\batchSize$-medoids clustering algorithm. (c) Boundary Medoids Selection. The clusters are chosen over the boundary of the convex hull of all samples. (d) Successive Elimination. One point is selected and another is eliminated based on pairwise comparisons of volume removal.
	}
	\label{fig:04_08_method_visuals}
\end{figure*}

However in many cases, we want our approach to be not only parallelizable but also time-efficient. For this purpose, we now describe a set of methods that do not rely on DPPs in increasing order of complexity to provide alternative approximations to the batch active learning problem. Figure~\ref{fig:04_08_method_visuals} visualizes each approach for a small set of queries.

\subsubsection{Greedy Selection}
The simplest method to approximate the optimal batch generation is using a greedy strategy. In the greedy selection approach, we conveniently assume the $\batchSize$ different queries in a batch are independent from each other. Of course this is not a valid assumption, but the independence assumption allows us to choose the $k$-many maximizers of the objective of Equation~\eqref{eq:04_01_VR1_worst_case} among the $\numberOfQueries$ discrete queries.

This method is a specific case of the DPP-based approach with $\diversityParameter=0$ or with $\abs{\reducedQuerySet}=\batchSize$. While this method can easily be employed; it is suboptimal as similar or redundant queries can be selected together in the same batch because these similar queries are likely to lead to high volume removal values. For instance, as shown in Figure~\ref{fig:04_08_method_visuals}a, the $5$ orange queries chosen are all going to be very close to the center where volume removal values are high.

\subsubsection{Medoid Selection}
To avoid the redundancy in the batch created by the greedy selection, we need to increase the dissimilarity between the selected queries. We introduce an approach, \emph{Medoid Selection}, that leverages clustering as a similarity measure between the samples. In this approach, with the goal of picking the most dissimilar queries, we cluster $\featureDifference$-vectors associated with the elements of the reduced set $\reducedQuerySet$ into $\batchSize$ clusters, using standard Euclidean distance. We then restrict ourselves to only selecting one element from each cluster, which prevents us from selecting very similar trajectories.

One can think of using the well-known $\batchSize$-means algorithm \cite{lloyd1982least} for clustering and then selecting the centroid of each cluster. However, these centroids are not necessarily from the reduced set, so they can have lower volume removal values. More importantly, they might be infeasible, i.e., there might not be a pair of trajectories that produce the $\featureDifference$ vectors corresponding to the centroids.

Instead, we use the $\batchSize$-medoids algorithm \cite{kaufman1987clustering, bauckhage2015numpy} which again clusters the queries into $\batchSize$ sets. The main difference between $\batchSize$-means and $\batchSize$-medoids is that $\batchSize$-medoids enables us to select medoids as opposed to the centroids, which are queries \textit{in the set} $\reducedQuerySet$ that minimize the average distance to the other queries in the same cluster.
While $\batchSize$-medoids is known to be a slower algorithm than $\batchSize$-means \cite{velmurugan2010computational}, efficient approximate algorithms exist \cite{bagaria2018medoids}. Figure~\ref{fig:04_08_method_visuals}b shows the medoids selection approach, where $5$ orange queries are selected from the $5$ clusters.

\subsubsection{Boundary Medoid Selection}
We note that picking the medoid of each cluster is not the best option for increasing dissimilarity \textemdash instead, we can further exploit clustering to select queries more effectively. In the \emph{Boundary Medoid Selection} method, we propose restricting the selection to be only from the boundary of the convex hull of the reduced set $\reducedQuerySet$. If feasible, this selection criteria can separate out the selected queries from each other on average. We note that when $\numberOfFeatures$, the dimensionality of $\featureDifference$, is large enough compared to $\batchSize$, most of the clusters will have queries on the boundary. We thus propose the following modifications to the medoid selection algorithm. The first step is to only select the queries that are on the boundary of the convex hull of the reduced set $\reducedQuerySet$. We then apply $\batchSize$-medoids with $\batchSize$ clusters over the queries on the boundary and finally only accept the cluster medoids as the selected batch.
As shown in Figure~\ref{fig:04_08_method_visuals}c, we first find $\batchSize=5$ clusters over the points on the boundary of the convex hull of $\reducedQuerySet$. We note that the number of queries on the boundary of convex hull of $\reducedQuerySet$ can be larger than the number of queries needed in a batch, e.g., there are $7$ points on the boundary; however, we only select the medoids of the $5$ clusters created over these boundary queries shown in orange.

\subsubsection{Successive Elimination}
One of the main objectives of batch generation for active learning as described in the previous methods is to select $\batchSize$ queries that will maximize the average distance among them out of the queries in the reduced set $\reducedQuerySet$. This problem is also referred to as \emph{max-sum diversification} in literature, which is known to be NP-hard \cite{gollapudi2009axiomatic,borodin2012max}. However, there exists a set of algorithms that provide approximate solutions~\cite{cevallos2017local}.

What makes our batch generation problem special and different from standard max-sum diversification is that we can compute the volume removal for each query. As in the DPP-based method, volume removal is a metric that models how much we want a query to be in the final batch. Thus, we propose a novel method that leverages the volume removal values to successively eliminate queries for the goal of obtaining a satisfactory diversified set. We refer to this algorithm as \emph{Successive Elimination}. At every iteration of the algorithm, we select two closest queries (in terms of Euclidean distance of their $\featureDifference$ vectors, but again, other distance metrics between queries could also be used) in the reduced set $\reducedQuerySet$, and remove the one with lower volume removal value. We repeat this procedure until $\batchSize$ points are left in the set, resulting in the $\batchSize$ queries in our final batch, which efficiently increases the diversity among queries.

A pseudo-code of this method is given in Algorithm~\ref{alg:04_08_successive_elimination}. Figure~\ref{fig:04_08_method_visuals}d shows the successive pairwise comparisons between two queries based on their corresponding volume removal. In every pairwise comparison, we eliminate one of the queries, shown with black edge, keeping the query connected with the orange edge. The numbers show the order of comparisons made before finding $\batchSize=5$ queries shown in orange.

\begin{algorithm}[ht]
	\caption{Successive Elimination}
	\label{alg:04_08_successive_elimination}
	\begin{algorithmic}[1]
		\State $\reducedQuerySet, \mathbf{\score} \gets \textsc{ReduceDataset}(\bar\weights,\datasetOfQueries, \abs{\reducedQuerySet})$
		\State $\batchExample \gets \reducedQuerySet$ \Comment{Initialize the batch}
		\While{$\abs{\batchExample}>\batchSize$}
		\State $(\featureDifference_i,\featureDifference_j) \gets \argmin_{\featureDifference_i,\featureDifference_j\in \batchExample} \norm{\featureDifference_i- \featureDifference_j}_2$
		\If{$\score_i < \score_j$}
		\State Remove $\featureDifference_i$ from $\batchExample$
		\Else
		\State Remove $\featureDifference_j$ from $\batchExample$
		\EndIf
		\EndWhile
		\State\Return $\batchExample$
	\end{algorithmic}
\end{algorithm}

We make the code for our batch active learning methods available at \url{https://bit.ly/381brBK}.

\subsection{Theoretical Guarantees}
\begin{theorem}
	Under the following assumptions:
	\begin{enumerate}[nosep]
		\item The error introduced by the approximation given by Equation~\eqref{eq:04_08_human_noise_approx} is ignored,
		\item The error introduced by the sampling of $\bar\weights$'s via adaptive metropolis algorithm is ignored,
	\end{enumerate}
	DPP-based method, greedy selection and successive elimination algorithms remove at least $1-\epsilon$ times as much volume as removed by the best adaptive non-batch strategy after $\batchSize\ln(\frac1{\epsilon})$ times as many queries.
\begin{proof}
	In the DPP-based method, greedy selection and successive elimination, the volume removal maximizer query $\query_*$ out of $\numberOfQueries$ possible queries will always remain in the resulting batch of size $\batchSize$, because: (i) the greedy DPP mode approximation will first add this query to the batch, (ii) greedy selection algorithm will first add this query to the batch, and (iii) the queries will be removed in successive elimination only if they have lower volume removal than some other queries in the set. \citet{sadigh2017active} proved, by using the ideas from adaptive submodular function maximization literature \cite{krause2014submodular}, that if we make the single query $\query_*$ at each iteration, then at least $1-\epsilon$ times as much volume as removed by the best adaptive non-batch strategy will be removed after $\ln(\frac1{\epsilon})$ times as many iterations. The proof is then complete with the pessimistic approach that accepts other $\batchSize-1$ queries will not remove any volume at all.
\end{proof}
\end{theorem}

\subsection{Simulations and Experiments} \label{subsec:04_08_experiments}

\begin{figure*}
	\centering
	\includegraphics[width=\textwidth]{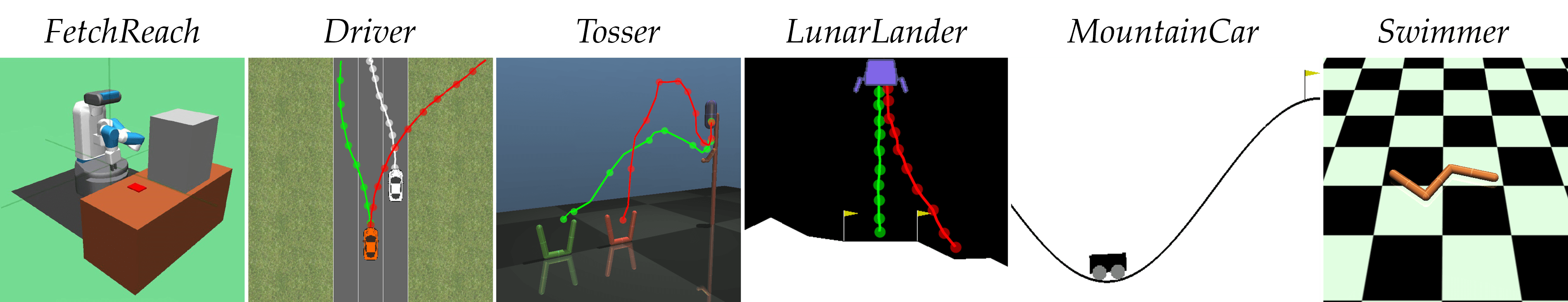}
	\caption{Simulation view of each environment. (a) \emph{FetchReach}, (b) \emph{Driver}, (c) \emph{Tosser}, (d) \emph{LunarLander}, (e) \emph{MountainCar}, (f) \emph{Swimmer}.}
	\label{fig:04_08_experiment_visuals}
\end{figure*}

\noindent\textbf{Experimental Setup.} We performed several simulations and experiments to compare the methods we propose and to demonstrate their performance. In all experiments, we set batch size $\batchSize=10$, reduced query set size $\abs{\reducedQuerySet}=200$, number of $\weights$ samples $\abs{\weightsSampleSet}=1000$, and assumed a linear reward function: $\trajectoryRewardFunction_{\weights}(\trajectory) = \weights^\top \trajectoryFeaturesFunction(\trajectory)$.

\noindent\textbf{Alignment Metric.} For our simulations, we generate synthetic random $\weights^*$ vectors as our true reward function parameters. We again used the \texttt{Alignment} metric in order to compare non-batch active, batch active and random query selection methods, where all queries are selected randomly over all feasible trajectories. As a reminder,
\begin{align}
\texttt{Alignment} = \frac{\weights^*\cdot \hat{\weights}}{\norm{\weights^*}_2\norm{\hat{\weights}}_2}\:,
\end{align}
where $\hat{\weights}$ is $\mathbb{E}[\weights\mid\comparisonDataset]$, the expectation of the learned belief distribution over $\weights$. We remind that this \texttt{Alignment} metric can be used to test convergence, because its value being close to $1$ means the estimate of $\weights$ is very close to (aligned with) the true reward parameters vector. In our experiments, we compare the methods using \texttt{Alignment} and the number of queries made.

\subsubsection{Tasks}
\begin{table}
	\centering
	\begin{threeparttable}
		\caption{Environment Properties}
		\label{tab:04_08_experiment_properties}
		\begin{tabular}{ c  c  c  c  c } 
			\hline
			\bfseries Task Name & $\dim(\action_\timestep)$ & $\horizon$ & $\numberOfFeatures$\\ 
			\hline
			\emph{LDS} & 5 & 1 & 5 \\
			\emph{FetchReach} & 7 & 19 & 4 \\ 
			\emph{Driver} & 2 & 5 & 4 \\
			\emph{Tosser} & 2 & 2 & 4 \\
			\emph{LunarLander}$^*$ & 2 & 5 & 6 \\
			\emph{MountainCar}$^*$ & 1 & 12 & 3 \\
			\emph{Swimmer} & 2 & 12 & 3 \\
			\hline
		\end{tabular}
		\begin{tablenotes}[para,flushleft]
			$^*$ Continuous versions
		\end{tablenotes}
	\end{threeparttable}
\end{table}
We perform experiments in different simulation environments that are summarized in Table~\ref{tab:04_08_experiment_properties} with a list of the variables associated with every environment, where $\horizon$ is the number of time steps in each trajectory, also known as the horizon of the task. The optimization of queries in the non-batch active method (of Section~\ref{sec:04_01_volume_removal}), is hence over $2\times(\horizon\dim(\action))$ with a fixed initial state $\state_0$, where the factor $2$ is because we generate $2$ trajectories for each query. Figure~\ref{fig:04_08_experiment_visuals} visualizes each of the experiment environments with some sample trajectories. Most of these environment were also used in the previous sections, but we now go over them for the the convenience of the reader.

\emph{Linear Dynamical System (LDS).}
We assess the performance of our methods on a simple simulated linear dynamical system:	
\begin{align}
\tilde\state_{\timestep+1} = A \tilde\state_\timestep + B \action_\timestep,\quad \state_\timestep = C\tilde\state_\timestep + D\action_\timestep
\end{align}
For a fair comparison between the proposed methods independent of the dynamical system, we want $\trajectoryFeaturesFunction(\trajectory)$ to uniformly cover its range when the control inputs are uniformly distributed over their possible values. We thus set $A$, $B$ and $C$ to be zero matrices and $D$ to be identity matrix in this section. Then a single step simulation of the system results in the observation $\state_0$, which can be treated as $\trajectoryFeaturesFunction(\trajectory)$. Therefore, the control inputs are equal to the features over trajectories, and optimizing over control inputs or features is equivalent. Despite its name, this environment is not meant to be a real dynamical system -- instead it measures how our batch active learning algorithms would perform in the simplest linear regression via pairwise comparisons problem.

\emph{FetchReach.} We use the simulator for Fetch mobile manipulator robot \cite{wise2016fetch}, visualized in Figure~\ref{fig:04_08_experiment_visuals}a. We use features that correspond to average and final distances to the target object (red block), average distance to the table (brown block), and average distance to the obstacle (gray block).

\emph{Driver.} We use the 2D driving simulator \cite{sadigh2016planning}, shown in Figure~\ref{fig:04_08_experiment_visuals}b. We use features corresponding to distance to the closest lane, speed, heading angle, and distance to the other vehicle in the scenario. Two sample trajectories are shown in red and green in Figure~\ref{fig:04_08_experiment_visuals}b. In addition, the white line shows the fixed trajectory of the other vehicle on the road.

\emph{Tosser.} We use MuJoCo's ``Tosser" \cite{todorov2012mujoco} where a robot tosses a capsule-shaped object. The features we use are maximum horizontal range, maximum altitude, the sum of angular displacements at each time step and final distance to closest basket of the object. The two red and green trajectories in Figure~\ref{fig:04_08_experiment_visuals}c correspond to synthesized queries showing different preferences for what basket to toss the object to.

\emph{LunarLander.} We use OpenAI Gym's continuous version of the ``LunarLander" environment \cite{brockman2016openai} where a spacecraft is controlled. We use features corresponding to final heading angle, final distance to landing pad, total rotation, path length, final vertical speed, and flight duration. Two sample trajectories are shown in red and green in Figure~\ref{fig:04_08_experiment_visuals}d.

\emph{MountainCar.} We use OpenAI Gym's ``MountainCar" \cite{brockman2016openai} where a simple 1D car model is controlled on a hill. The features are maximum range in the positive direction, maximum range in the negative direction, and time to reach the flag (or $\horizon$ if not reached). The environment is shown in Figure~\ref{fig:04_08_experiment_visuals}e.

\emph{Swimmer.} We use OpenAI Gym's ``Swimmer" \cite{brockman2016openai}. We use features corresponding to horizontal displacement, vertical displacement, and total distance traveled. The environment is shown in Figure~\ref{fig:04_08_experiment_visuals}f.

\subsubsection{Comparison of Batch-Active Learning Methods}
We first quantitatively compare the batch-active methods we proposed with each other. We use the \emph{LDS}, \emph{FetchReach}, \emph{Driver} and \emph{Tosser} to demonstrate this comparison. For each of these environments, we create a dataset of $\numberOfQueries=100,\!000$ queries.

Independently for each environment, we randomly generated $200$ different reward functions ($\weights^*$ vectors), $100$ of which are for tuning $\scoreParameter$ in the DPP-based method and the remaining $100$ are for tests of all methods. The same approach can be employed in practice: One can simulate random reward functions for tuning and then deploy the system to learn the reward functions from real users. For both tuning and tests, we simulated noiseless users, who always reveal their true preferences in order to eliminate the effect of noise in the results. We present the further details and results of hyperparameter tuning in the Appendix~\ref{app:99_03_ijrr_dpp_hyperparameters}.

\begin{figure}[ht]
	\centering
	\includegraphics[width=0.6\textwidth]{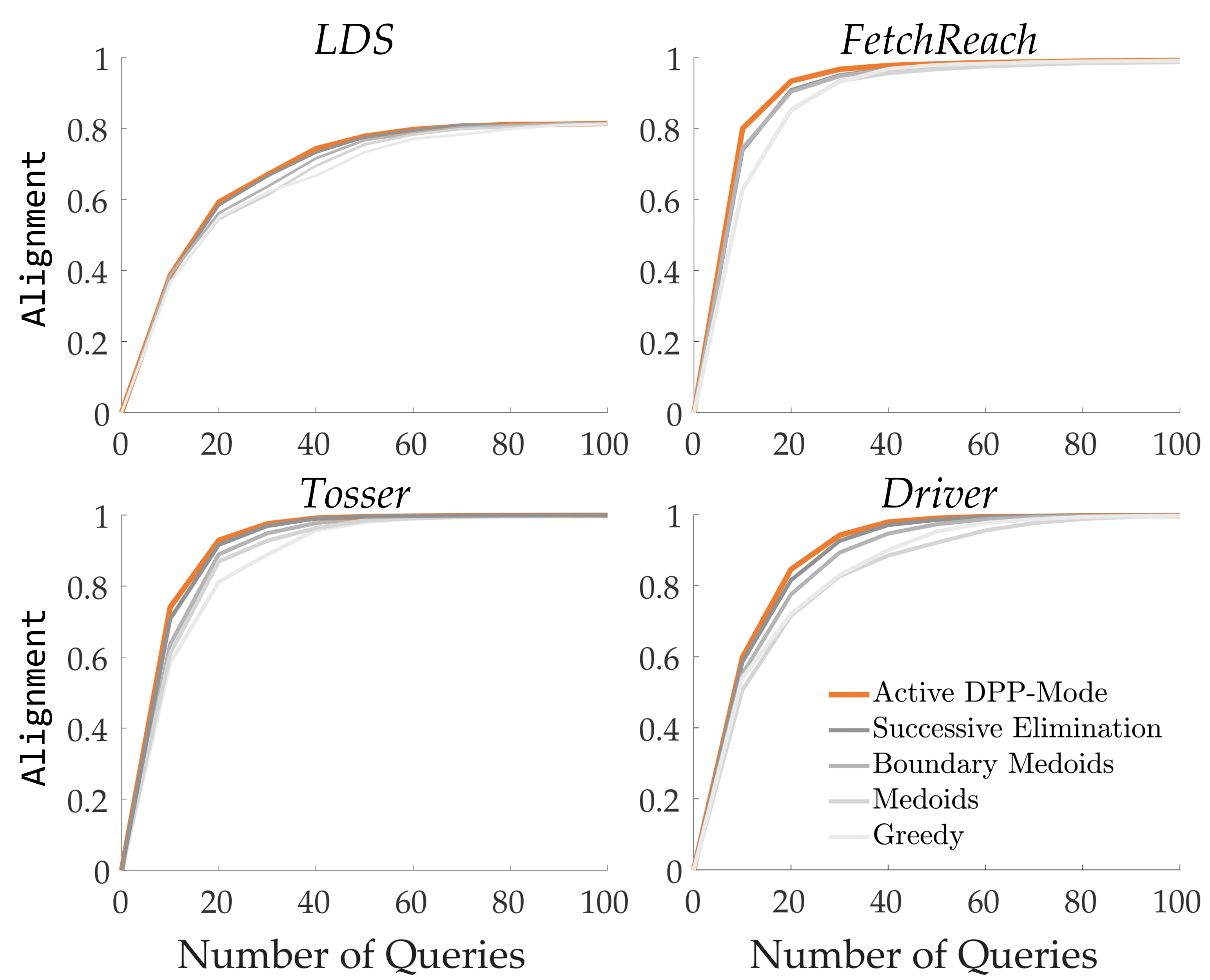}
	\caption{Batch-active learning methods are compared.}
	\label{fig:04_08_dpp_results}
\end{figure}
For each simulated reward function during our tests, we ran $10$ batch generations with each method, summing up to $100$ pairwise comparison queries. We demonstrate the results in Figure~\ref{fig:04_08_dpp_results}. Our results suggest that the DPP-based method significantly outperforms all other methods in all environments ($p<0.05$, Wilcoxon signed-rank test \cite{wilcoxon1945individual}) except for successive elimination in LDS where both algorithms perform comparably.

Among the time-efficient batch active learning methods we proposed, successive elimination method significantly outperforms the others ($p<0.05$) in all environments except \emph{FetchReach} where it performs comparably to boundary medoids, significantly outperforms medoids selection ($p<0.05$), and marginally significantly outperforms greedy selection ($p\approx0.06$). Similarly, boundary medoids approach significantly outperformed medoids selection and greedy selection in all environments ($p<0.05$). Finally, medoids selection and greedy selection performed comparably in all environments, except \emph{Tosser} where medoids selection significantly outperformed the greedy approach ($p<0.05$).

Overall, these results show us the ranking of batch active learning methods from the best to the worst are as follows:
\begin{enumerate}[nosep]
	\item Active DPP-Mode
	\item Successive Elimination
	\item Boundary Medoids
	\item Medoids
	\item Greedy
\end{enumerate}

\subsubsection{Comparison to Non-Batch Active Learning}
We next investigated the average time it required to generate one query. For this, we took a dataset of $\numberOfQueries=500,\!000$ queries. We recorded the batch generation times, and divided it by $\batchSize=10$. To show the advantage of batch-active learning methods, we also ran the same analysis on the non-batch active learning approach that synthesizes trajectories by optimizing over their action spaces along the time horizon. We ran this study for \emph{Driver}, \emph{Tosser}, \emph{LunarLander}, \emph{MountainCar} and \emph{Swimmer} environments. The results are shown in Table~\ref{tab:04_08_average_query_times}. It can be seen that batch active learning methods lead to a great decrease in query generation times compared to the non-batch method, and the DPP-based method is slightly slower than the other batch algorithms. This slowness could be even more significant for larger batches.

\begin{table*}
	\centering
	\begin{threeparttable}
		\caption{Average Query Generation Times (seconds)}
		\label{tab:04_08_average_query_times}
		\begin{tabular}{ c | c | c  c  c  c  c } 
			\hline
			\multirow{2}{*}{\bfseries Environment} & \multirow{2}{*}{\bfseries Non-Batch} & \multicolumn{5}{c}{\bfseries Batch Active Learning}\\
			& & \begin{tabular}{@{}c@{}}\emph{Active} \\ \emph{DPP-Mode}\end{tabular} & \emph{Greedy} & \emph{Medoids} & \begin{tabular}{@{}c@{}}\emph{Boundary} \\ \emph{Medoids}\end{tabular} & \begin{tabular}{@{}c@{}}\emph{Successive} \\ \emph{Elimination}\end{tabular}\\
			\hline
			\emph{Driver} & 79.2 & 5.5 & 5.4 & 5.7 & 5.3 & 5.5 \\ 
			\emph{Tosser} & 149.3 & 5.5 & 4.1 & 4.3 & 3.8 & 3.9 \\
			\emph{LunarLander} & 177.4 & 5.6 & 4.1 & 4.1 & 4.2 & 4.1 \\
			\emph{MountainCar} & 96.4 & 7.1 & 3.8 & 4.0 & 4.0 & 3.8 \\
			\emph{Swimmer} & 188.9 & 10.8 & 3.8 & 3.9 & 4.0 & 4.1 \\
		\end{tabular}
	\end{threeparttable}
\end{table*}

As we observed that successive elimination generates highly informative queries in a time-efficient way without any hyperparameter tuning, we now compare its performance to the non-batch active learning approaches. Specifically, we assessed its performance against non-batch active learning and random query selection where queries are selected uniformly at random.

We again conducted our simulation experiments on all environment but \emph{FetchReach} due to its large action space, which makes the non-batch active learning impractical as it optimizes over the action space along the horizon. For the simulations with \emph{LDS}, we assume that human's preferences are noisy as discussed in Equation~\eqref{eq:04_08_human_noise_approx}. For all other environments, we again assume an oracle user who responds to queries with no error to avoid perturbations due to noise in responses.

\begin{figure}[ht]
	\centering
	\includegraphics[width=0.7\textwidth]{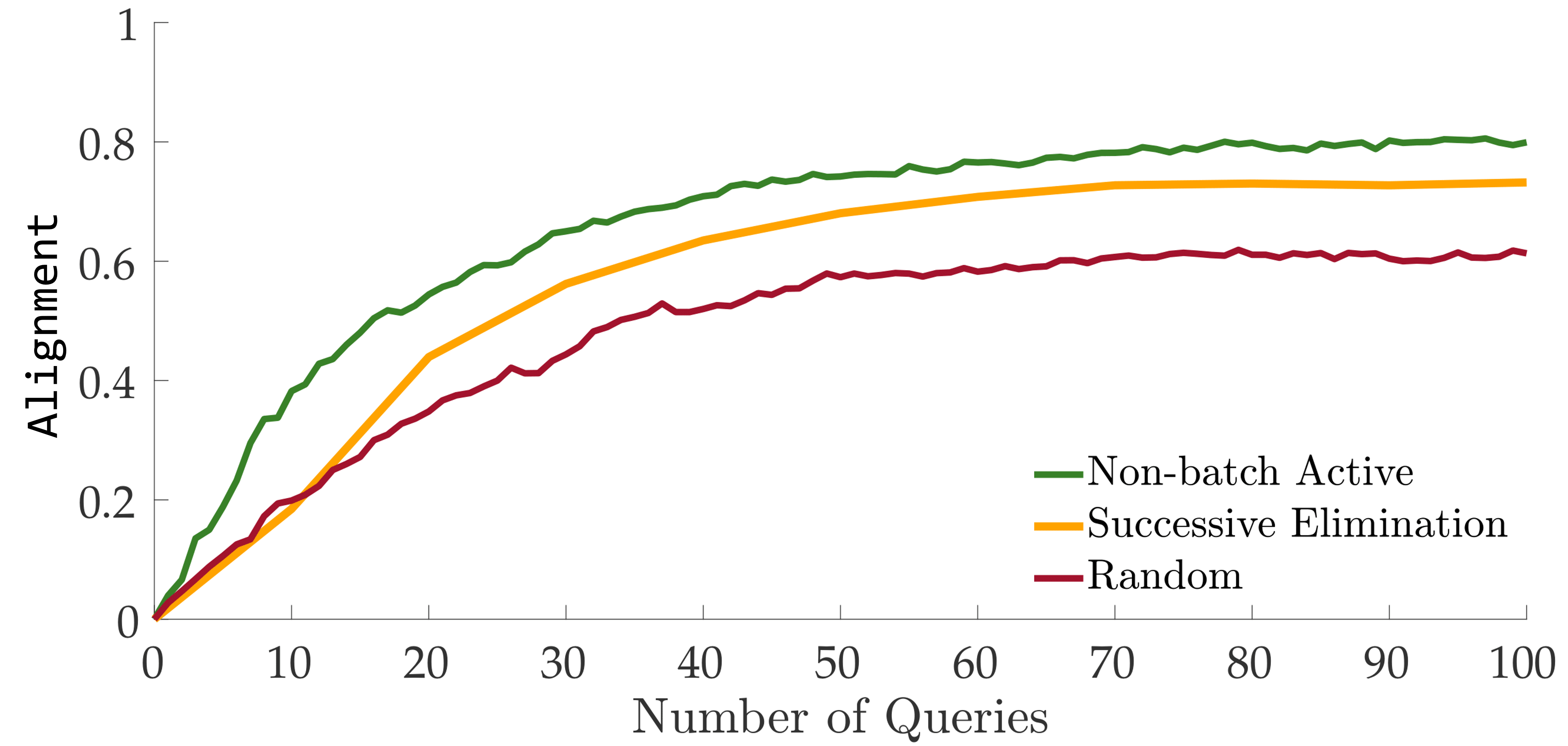}
	\caption{The performance of each algorithm is averaged over $10$ different runs on \emph{LDS} where $\weights^*$ is uniformly randomly generated. Successive elimination performs better than the random querying and worse than the non-batch active method.
	}
	\label{fig:04_08_simulation}
\end{figure}

Figure~\ref{fig:04_08_simulation} shows the number of queries that result in a corresponding \texttt{Alignment} value for each method in the \emph{LDS} environment, averaged over $10$ runs. The non-batch active version significantly outperforms successive elimination ($p<0.05$), as it performs the optimization for each and every query. As expected, both active methods significantly outperform random querying ($p<0.05$).

\begin{figure*}
	\centering
	\includegraphics[width=\textwidth]{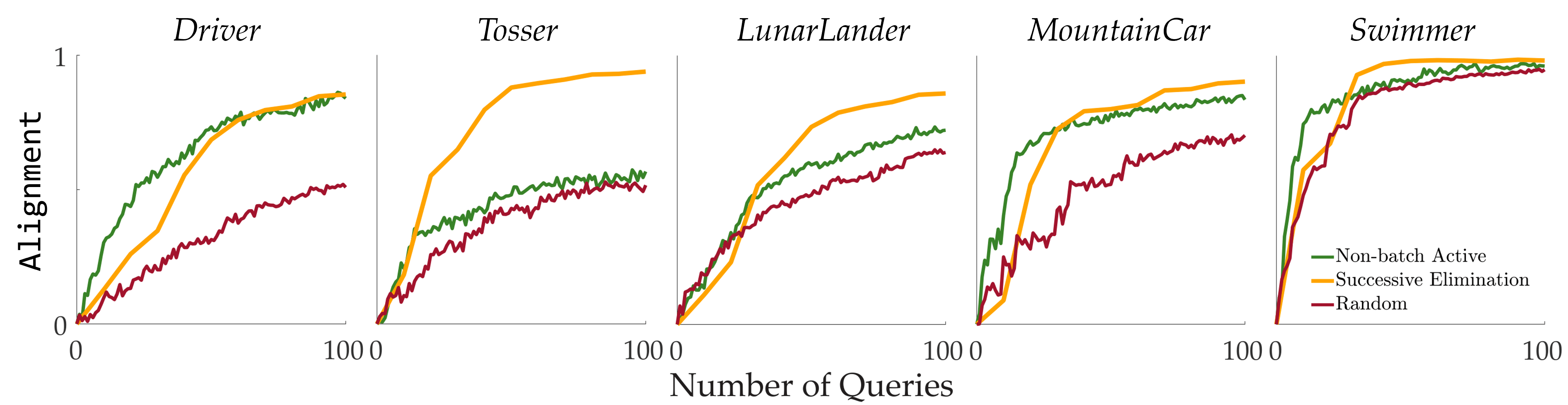}
	\caption{The performance the algorithms is shown. The non-batch active method performs poorly on \textit{LunarLander} and \textit{Tosser}.}
	\label{fig:04_08_query_count}
\end{figure*}

\begin{figure*}
	\centering
	\includegraphics[width=\textwidth]{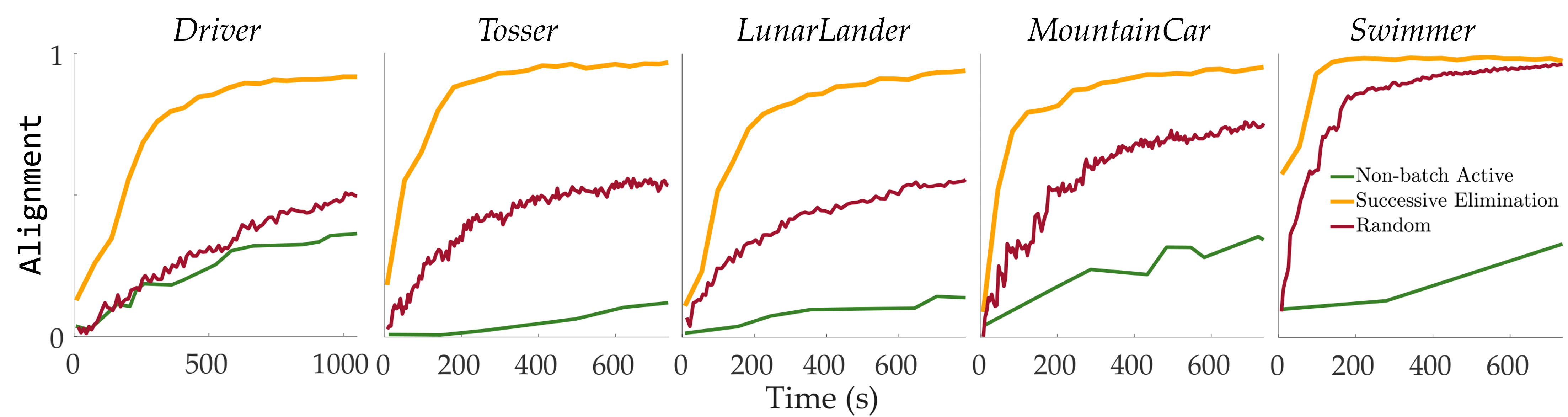}
	\caption{Convergence to $\weights^*$ as a function of time is plotted for each environment. Non-batch active learning method is slow due to the optimization and adaptive metropolis algorithm involved in each iteration, whereas random querying performs poorly due to redundant queries. Successive elimination clearly outperforms both of them.}
	\label{fig:04_08_timing}
\end{figure*}

We show the results of our experiments on the other five environments in Figures~\ref{fig:04_08_query_count} and \ref{fig:04_08_timing}. Figure~\ref{fig:04_08_query_count} shows the convergence to the true reward function parameters $\weights^*$ as the number of queries increases (similar to Figure~\ref{fig:04_08_simulation}). It is interesting to note that non-batch active learning performs suboptimally in \textit{LunarLander} and \textit{Tosser}. We believe this can be due to the non-convex optimization being solved in non-batch methods leading to suboptimal behavior, or because of the known failure cases of the volume removal objective as we discussed in Section~\ref{sec:04_01_volume_removal}. The proposed batch active learning methods overcome this issue thanks to query space discretization and the fact that it does not rely merely on the volume removal values and tries to incorporate diversity among the queries.

Figure~\ref{fig:04_08_timing} evaluates the computation time required for querying. It is clearly visible that batch active learning makes the process much faster than the non-batch active method and random querying. 
\begin{figure}[ht]
	\centering
	\includegraphics[width=0.5\textwidth]{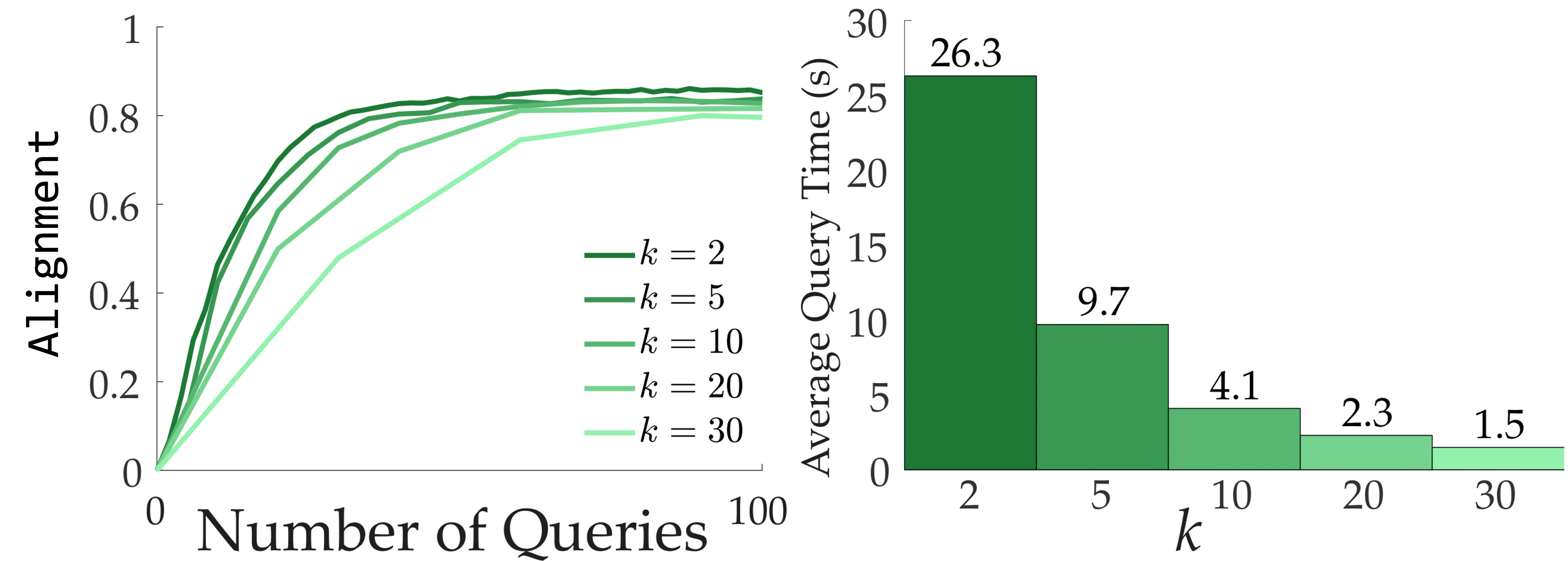}
	\caption{The performance of successive elimination algorithm with varying $\batchSize$ values was averaged over $10$ different runs with \emph{LDS} where $\weights^*$ is uniformly randomly generated and $\abs{\reducedQuerySet}=20\batchSize$. (a) The \texttt{Alignment} values, and (b) average query times. 
	}
	\label{fig:04_08_batch_size}
\end{figure}

Therefore, batch active learning is preferable over other methods as it balances the tradeoff between the number of queries required and the time it takes to compute the queries. This tradeoff can be seen in Figure~\ref{fig:04_08_batch_size} where we simulated \emph{LDS} with varying $\batchSize$ values. For this simulation, we set $\abs{\reducedQuerySet}=20\batchSize$ in accordance with other experiments.

\subsubsection{User Preferences}
In addition to our simulation results using synthetic $\weights^*$ vectors, we perform a user study to learn humans' preferences for the  \emph{Driver} and \emph{Tosser} environments. This experiment is mainly designed to show the ability of our framework to learn humans' preferences. 

\noindent\textbf{Setup.} We recruited $10$ users who responded to $150$ queries generated by successive elimination algorithm for each environment (\emph{Driver} or \emph{Tosser}).

\begin{figure}[ht]
	\centering
	\includegraphics[width=0.5\textwidth]{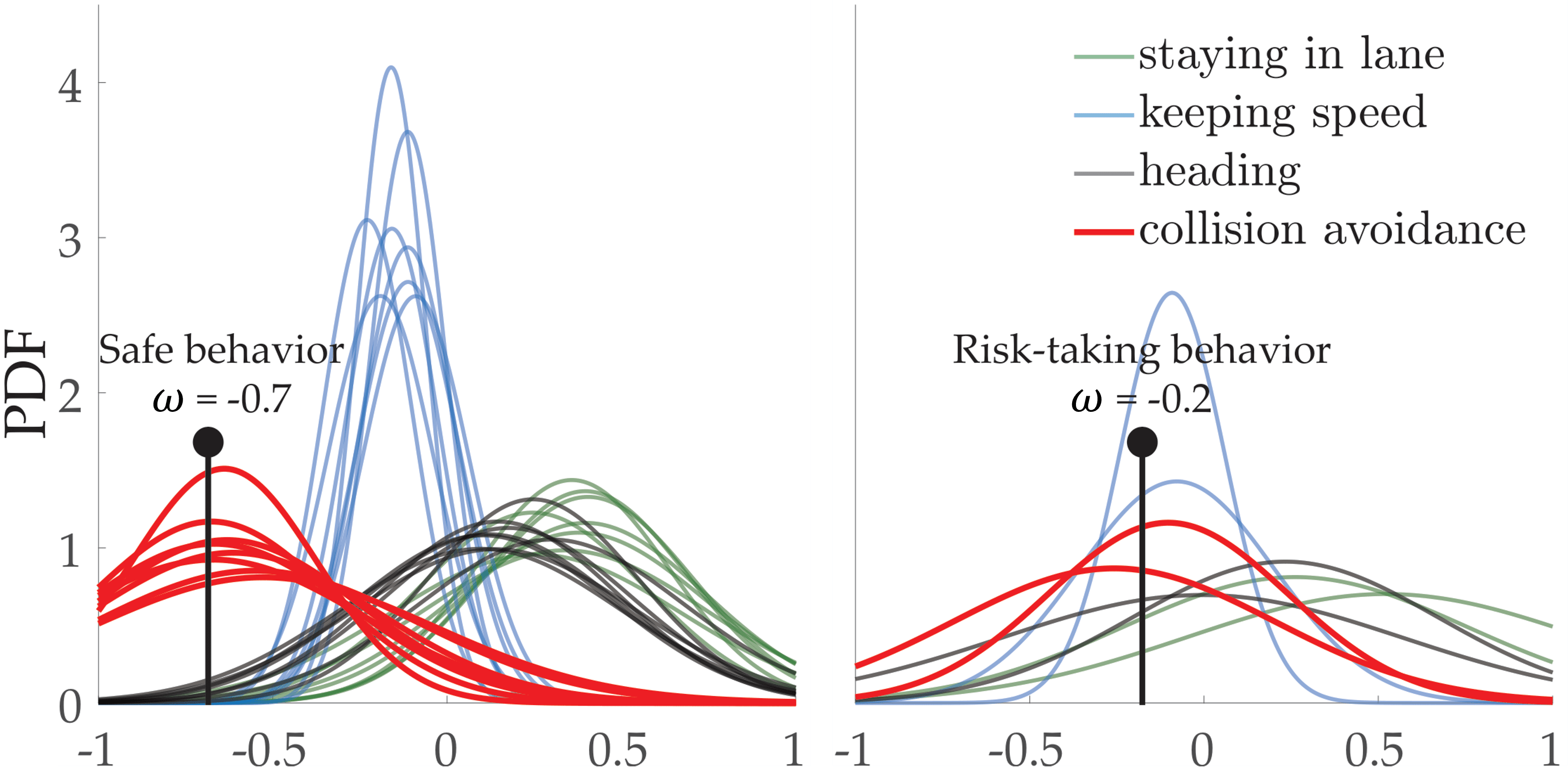}
	\caption{User preferences on \emph{Driver} task are grouped into two sets. While the first set shows the preferences conforming with the natural driving behavior, the second set is comprised of data from two users one of whom preferred collisions with the other car over leaving the road and the other regarded some collisions as near-misses and thought they can be acceptable in order to keep speed. It can be seen that the uncertainty in their learned preferences is higher.}
	\label{fig:04_08_user_study_driver}
\end{figure}

\noindent\textbf{Driver Preferences.}
Using successive elimination, we are able to learn humans' driving preferences. Our results show that the reward functions of users are very close to each other as this task mainly models natural driving behavior. This is consistent with results shown by \citet{sadigh2017active}, where non-batch techniques are used.  We noticed a few differences between the driving behavior as shown in Figure~\ref{fig:04_08_user_study_driver}. This figure shows the distribution of the weights for the four features after $150$ queries. Two of the users (plot on the right) seem to have slightly different preferences about collision avoidance, which may correspond to more aggressive driving behavior. We observed that $70$ queries were enough for converging to safe and sensible driving in the defined scenario. The optimized driving with different number of queries can be watched on \url{https://youtu.be/MaswyWRep5g}.

\begin{figure*}[t]
	\centering
	\includegraphics[width=\textwidth]{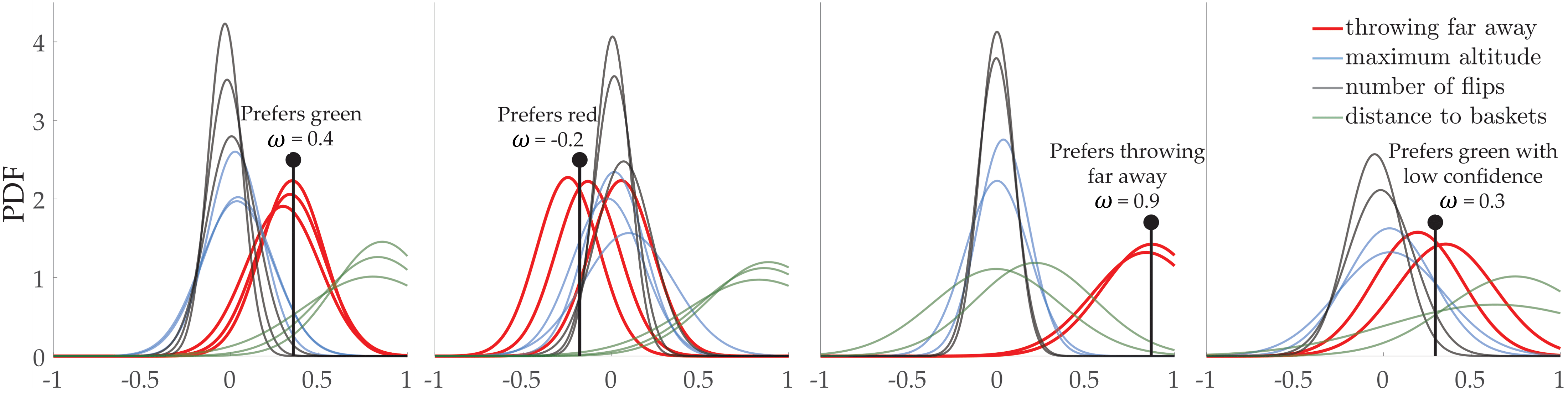}
	\caption{User preferences on \emph{Tosser} task are grouped into four sets. The first set shows the preferences of people who aimed at throwing the ball into the green basket (the distant one) but accepted throwing into the other basket is better than not throwing into any baskets. The second set is comprised of data from three users who preferred the red basket (the closer one). In the third group, the users preferred the green basket over the red one, but also accepted throwing far away is better than throwing into the red basket, because it is an attempt for the green basket. Lastly, the fourth group is similar to the first group; however the confidence over preferences is much less, because the users were not sure about how to compare the cases where the ball was dropped between the baskets in one of the trajectories.}
	\label{fig:04_08_user_study_tosser}
\end{figure*}

\paragraph{Tosser Preferences.} Similarly, we use successive elimination to learn humans' preferences on the \emph{Tosser} task. Figure~\ref{fig:04_08_user_study_tosser} shows we learn interesting tossing preferences. For demonstration purposes, we optimize the control inputs with respect to the preferences of two of the users, one of whom prefers the green basket while the other prefers the red one (see Figure~\ref{fig:04_08_experiment_visuals}c). We note that $100$ queries were enough to see reasonable convergence in this task. The evolution of the learning can be watched on \url{https://youtu.be/cQ7vvUg9rU4}.

%% file: 04_active/09_summary.tex
\label{sec:04_09_summary}

In this chapter, we built on the techniques we presented in Chapter~\ref{chap:learning}: we showed how different comparative feedback types, which robots can leverage to learn reward functions, can be actively collected from humans for better data- and time-efficiency. We first started with reviewing an existing acquisition function from the literature, namely volume removal \cite{sadigh2017active}, and showed its drawbacks and failure cases in Section~\ref{sec:04_01_volume_removal}. We then proposed an alternative objective in Section~\ref{sec:04_02_information_gain} to optimize for data-efficiency: mutual information, a widely used notion from information theory \cite{cover2012elements}. We showed it does not suffer from the same problems as volume removal even though it has the same computational complexity. We then adopted the same structure as in Chapter~\ref{chap:learning} to extend each comparative query and reward function type (parameteric and non-parametric) with active querying techniques, one by one from Section~\ref{sec:04_03_gp} to \ref{sec:04_07_hierarchical}. Finally in Section~\ref{sec:04_08_batch}, we showed how one can actively generate multiple queries at the same time, i.e., in batches, for parallelizability and better time-efficiency as it avoids solving an optimization problem for each and every query.

%% file: 05_conclusion/00_intro.tex
We envision a world where robots and agents powered with artificial intelligence seamlessly interact with humans and each other, which may include collaborating, competing, teaching and influencing. We are convinced that they need to be able to learn the objectives or the preferences of other agents to achieve such interactions. Because the information about an agent's objective enables a robot to better predict their behavior, which in turn, enables the robot to condition their interaction on these predictions. Our preliminary works empirically proved this concept in various settings such as human-robot collaboration \cite{biyik2022partner}, autonomous driving \cite{cao2021leveraging}, traffic network optimization \cite{lazar2021learning,biyik2018altruistic,biyik2021incentivizing}, or multi-agent learning \cite{wang2021emergent,beliaev2020emergent,zhu2020multi}. Perhaps more importantly, learning humans' objectives in a task means learning how to perform the task itself. This has also been the theme of the simulations and experiments we conducted in this thesis.

Overall, this thesis is an important step towards the goal of reliably learning humans' objectives in various tasks. For more and easier accessibility, we released a software library that implements many of the methods we proposed in this thesis \cite{biyik2022aprel}. However, both in that library and in this thesis, we only focused on learning from comparative feedback (possibly in addition to demonstrations) and how to elicit human preferences using such feedback. In real world, when people interact with each other, they often use many more sources of information, such as gaze, gestures, language, facial expressions, etc. Robots are still not fully capable of using these other forms of feedback. While such high-level challenges exist and are yet to be solved, we would like to conclude this thesis by focusing on a discussion of the limitations and future directions of the methods we developed.

%% file: 05_conclusion/01_challenges.tex
Although we proposed several techniques for actively learning from comparative feedback and empirically showed they can be used to learn humans' preferences in different tasks, those techniques are limited in various ways. In this section, we would like to focus on these limitations for both learning and active querying.

\subsection{Limitations and Future Work in Learning}
In this thesis, we worked with two different assumptions about the reward function that we are trying to learn: it is either a parametric or a non-parametric reward function. In the former, we were constrained to work with functions that have a small number of parameters in practice, because our learning scheme is fully Bayesian. However, many applications in robotics may require more complex reward functions, such as those modeled with deep neural networks. We are not able to efficiently (in terms of both data and time) learn such complex functions using the techniques in this thesis. Even though alternative learning methods that rely on gradient based learning exist, e.g. \cite{brown2019deep}, they often cannot give reliable uncertainties about the learned reward which limits their usability and makes active querying difficult.

In the latter, we used Gaussian processes to model non-parametric reward functions. Although this relaxes some of the assumptions about the functional form of the reward, it brings its own challenges in practice: how large can the input to the Gaussian process be? Even though this approach enabled us to learn complex rewards, we were now limited in terms of the dimensionality of the input (the number of trajectory features), because Gaussian processes are difficult to fit with large amounts of data and increasing the input dimensionality makes it more data-hungry.

Both of these limitations point out another challenge: where do the trajectory features come from? Often, we rely on experts to design such features, but these reward models are prune to errors in feature functions. Although our preliminary works show new features could be discovered using comparative feedback \cite{katz2021preference}, we still rely on the initial set of hand-designed features. Future work should investigate supervised and unsupervised techniques for learning features for reward functions.

Moreover, we adopted and used different models about how humans respond to comparative feedback queries. All of these models implicitly assume humans are rational decision makers: in all models, the most likely human response is the true response based on their underlying reward function. However, humans are bounded rational in various settings and conditions \cite{simon1990bounded,gigerenzer2002bounded,halpern2014decision}. Our work based on cumulative prospect theory \cite{kahneman2013prospect,tversky1992advances} showed humans take consistently suboptimal actions when the system involves some risk \cite{kwon2020when}, even when there are only two action choices, just like the pairwise comparisons setting we have in this thesis. Future work should incorporate this suboptimality or irrationality of humans into the reward learning from comparative feedback framework.\footnote{See the recent work by \citet{chan2021human} that attempts to model various reasons of irrationality and incorporate them for reward inference.}

An important limitation of our multimodal reward learning techniques presented in Sections~\ref{sec:03_05_rankings} and \ref{sec:03_06_hierarchical} is the fact that we assumed we know the number of modes in the reward function. Although this might be the case when learning rewards from multiple humans with different objectives, it is unrealistic if we are trying to learn a multimodal reward from one person who has non-stationary preferences. Our techniques could be easily modified to handle such scenarios: similar to clustering algorithms, one can experiment with varying number of modes and then take the simplest model that gives reasonable performance. However, this approach will prevent the robots from using active querying techniques as they rely on a fixed number of modes.

Finally, an interesting research direction is about the interfaces that can be used for collecting comparative feedback. In Section~\ref{sec:03_04_scale}, we showed we can use a slider bar to collect scale feedback, which gives more information than pairwise comparisons. However, extending this to higher number of trajectories within a query is challenging. For three trajectories, one could think of a 2D plane on which the user selects a point whose distance to the corners indicate how much the user prefers each trajectory. Going beyond three trajectories requires more and more complex interfaces, and perhaps hardware. Even more interestingly, one could leverage the fact that the systems we are trying to teach via comparative feedback are robots that are embodied. This may open new possibilities such as giving feedback to the robot about different parts of the space it is operating in or about different segments of its trajectory (see \cite{cui2018active}).

\subsection{Limitations and Future Work in Active Querying}
All of the techniques we proposed for active querying had an implicit assumption: the robot is in an offline training phase. Thanks to this assumption, we are able to optimize our queries to humans specifically for data-efficiency. In other words, we do not have to worry about whether the trajectories we are demonstrating to humans are good or bad: we can show bad trajectories just for the sake of learning. However, in many cases, it is desirable to ask questions while, at the same time, trying to perform the task. Such an online setting would require optimizing when to ask a question during the task, and other acquisition functions for deciding what question to ask.

It is not only that we did not have to worry about the quality of trajectories, but we also did not formulate any hard safety constraints (even though we formulated some soft constraints in Section~\ref{sec:03_03_roial} by constructing a region of avoidance in the trajectory space). This prevents one from using learning from comparative feedback techniques in safety critical systems. One possibility is to first utilize safe exploration techniques, e.g., \cite{berkenkamp2016safe,sui2015safe,biyik2019efficient}, to construct the space of trajectories, but this requires a well-defined safety function. Future work should investigate how safety could also be learned using comparative feedback, similar to the related works on constraint learning from demonstrations \cite{chou2018learning}.

%% file: 05_conclusion/02_closing_thoughts.tex
This thesis is only a step that brings ideas from robotics, machine learning, information theory and control theory to address the reward learning problem in artificial intelligence which we believe to be an important challenge to achieve seamless human-robot (or more generally human-AI) interaction. As we discussed in this chapter, there are still many limitations and challenges that need to be addressed for achieving such interactions. We believe these challenges will require collaborations between researchers from a wide range of fields, such as machine learning, robotics, computational psychology, human-robot interaction, behavioral economics, formal methods and control theory.

%% file: 99_appendix/01_proofs.tex
\section{Proof of Proposition~\ref{prop:03_04_upper_bound}}\label{app:03_04_proposition_proof}

\begin{proof}
To prove the statement, we show the feasible set obtained from scale feedback is a subset of the feasible set from choice feedback. We note $\maxRewardGap^*>0$ for any non-trivial problem instance, as otherwise every trajectory would be equally optimal for any $\weights^*$.
For one of the queries that form $\scaleDataset$ and $\comparisonDataset$, say query $i$, we assume the user prefers $\trajectory_1^{(i)}$ over $\trajectory_2^{(i)}$ without loss of generality, implying $\bar{\queryResponse}^{(i)}\geq0$.
For this query, pairwise comparison feedback defines a feasible set $\halfspace^{(i)}_{\mathtt{Comparison}}=\{\weights \mid \trajectoryRewardFunction_{\weights}(\trajectory_1^{(i)})-\trajectoryRewardFunction_{\weights}(\trajectory_2^{(i)})\geq0\}$.
First, we consider $\bar{\queryResponse}^{(i)}=1$. This yields $\halfspace^{(i)}_{\mathtt{Scale}}=\{\weights \mid \trajectoryRewardFunction_{\weights}(\trajectory_1^{(i)})-\trajectoryRewardFunction_{\weights}(\trajectory_2^{(i)})\geq\sensitivityThreshold\maxRewardGap(\weights)\}$. Since both $\sensitivityThreshold>0$ and $\maxRewardGap(\weights)\geq0$, we obtain $\halfspace^{(i)}_{\mathtt{Scale}}\subseteq \halfspace^{(i)}_{\mathtt{Comparison}}$.
For the case $\bar{\queryResponse}^{(i)}\in[0,1)$, we have $\halfspace^{(i)}_{\mathtt{Scale}}=\{\weights \mid \trajectoryRewardFunction_{\weights}(\trajectory_1^{(i)})-\trajectoryRewardFunction_{\weights}(\trajectory_2^{(i)})=\bar{\queryResponse}^{(i)}\sensitivityThreshold\maxRewardGap(\weights)\}$; the right hand side is non-negative and thus any $\weights$ satisfying the equality must satisfy $\trajectoryRewardFunction_{\weights}(\trajectory_1^{(i)})-\trajectoryRewardFunction_{\weights}(\trajectory^{(i)}_2)\geq0$. This also implies $\halfspace^{(i)}_{\mathtt{Scale}}\subseteq \halfspace^{(i)}_{\mathtt{Comparison}}$.
As $\mathtt{Err}^{\max}(\weights^*, \scaleDataset)$ maximizes over $\halfspace_{\mathtt{Scale}}$, which is the intersection of $\halfspace^{(i)}_{\mathtt{Scale}}$'s over queries, while $\mathtt{Err}^{\max}(\weights^*, \comparisonDataset)$ maximizes over $\halfspace_{\mathtt{Comparison}}$, $\mathtt{Err}^{\max}(\weights^*, \scaleDataset)$ cannot attain a larger value than $\mathtt{Err}^{\max}(\weights^*, \comparisonDataset)$.
\end{proof}

\section{Volume Removal Equivalence when $\abs{\query^{(i)}}=2$}\label{app:99_01_ijrr_dpp_worst}
While presenting the volume removal optimization for active learning in Equations~\eqref{eq:04_01_VR1} and \eqref{eq:04_01_VR1_worst_case}, we stated we could maximize the worst-case volume removal when $\abs{\query^{(i)}}=2$ so that a large amount of volume will be removed regardless of the human's response. This objective is in fact equal to the expected volume removal objective while learning from pairwise comparisons.

\begin{theorem}
While learning from pairwise comparisons, the worst-case volume removal maximization in Equation~\eqref{eq:04_01_VR1_worst_case} is equivalent to expected volume removal maximization presented in Equation~\eqref{eq:04_01_VR1}:
\begin{align*}
    \max_{\query^{(i)}=\{\trajectory_1, \trajectory_2\}}\:\mathbb{E}_\queryResponse^{(i)}\left[\int_{\weights}\left(\belief^{i-1}(\weights) - P(\queryResponse^{(i)}\mid \query^{(i)}\weights)\belief^{i-1}(\weights)\right)d\weights\right]
\end{align*}

\begin{proof}
We first work on the worst-case optimization objective:
\begin{align*}
    \min_{\queryResponse^{(i)}\in\query^{(i)}} & \int_{\weights}\left(\belief^{i-1}(\weights) - P(\queryResponse^{(i)}\mid\query^{(i)},\weights)\belief^{i-1}(\weights)\right)d\weights\\
    &= \min_{\queryResponse^{(i)}\in\query^{(i)}} \left(1-\int_{\weights}P(\queryResponse^{(i)}\mid\query^{(i)},\weights)\belief^{i-1}(\weights)d\weights\right)\\
    &= \min_{\queryResponse^{(i)}\in\query^{(i)}} \left(1-\mathbb{E}_{\weights\sim\belief^{i-1}(\weights)}\left[P(\queryResponse^{(i)}\mid\query^{(i)},\weights)\right]\right)\\
    &= \min_{\queryResponse^{(i)}\in\query^{(i)}} \left(1-P(\queryResponse^{(i)}\mid \query^{(i)}, \belief^{i-1})\right)\\
    &= \min \left\{P(\queryResponse^{(i)}=\query^{(i)}_1 \mid \query^{(i)}, \belief^{i-1}), P(\queryResponse^{(i)}=\query^{(i)}_2 \mid \query^{(i)}, \belief^{i-1})\right\}\:,
\end{align*}
where the last equation is because the queries are pairwise and $P(\queryResponse^{(i)}=\query^{(i)}_1 \mid \query^{(i)}, \belief^{i-1})+P(\queryResponse^{(i)}=\query^{(i)}_2 \mid \query^{(i)}, \belief^{i-1})=1$. Hence, we are trying to find queries for which the minimum of those two terms is maximized. An optimal query is therefore one for which our model predicts $P(\queryResponse^{(i)}=\query^{(i)}_1 \mid \query^{(i)}, \belief^{i-1})=P(\queryResponse^{(i)}=\query^{(i)}_2 \mid \query^{(i)}, \belief^{i-1})=0.5$. Intuitively, we are looking for queries where our model is highly unsure about the human's response. We know that $0 \leq P(\queryResponse^{(i)}=\query^{(i)}_1 \mid \query^{(i)}, \belief^{i-1}), P(\queryResponse^{(i)}=\query^{(i)}_2 \mid \query^{(i)}, \belief^{i-1}) \leq 1$. Hence, we note that this optimization is equivalent to optimizing $P(\queryResponse^{(i)}=\query^{(i)}_1 \mid \query^{(i)}, \belief^{i-1})P(\queryResponse^{(i)}=\query^{(i)}_2 \mid \query^{(i)}, \belief^{i-1})$.

Next, we work on the expected volume removal objective:
\begin{align*}
    \mathbb{E}_{\queryResponse^{(i)}\mid\query^{(i)},\belief^{i-1}} &\int_{\weights}\left(\belief^{i-1}(\weights) - P(\queryResponse^{(i)}\mid\query^{(i)},\weights)\belief^{i-1}(\weights)\right)d\weights\\
    &= \mathbb{E}_{\queryResponse^{(i)}\mid\query^{(i)},\belief^{i-1}} \left[1-\int_{\weights}P(\queryResponse^{(i)}\mid\query^{(i)},\weights)\belief^{i-1}(\weights)d\weights\right]\\
    &= \mathbb{E}_{\queryResponse^{(i)}\mid\query^{(i)},\belief^{i-1}} \left[1-\mathbb{E}_{\weights\sim\belief^{i-1}(\weights)}\left[P(\queryResponse^{(i)}\mid\query^{(i)},\weights)\right]\right]\\
    &= \mathbb{E}_{\queryResponse^{(i)}\mid\query^{(i)},\belief^{i-1}} \left[1-P(\queryResponse^{(i)} \mid \query^{(i)}, \belief^{i-1})\right]\\
    &= 2P(\queryResponse^{(i)}=\query^{(i)}_1\mid \query^{(i)}, \belief^{i-1})P(\queryResponse^{(i)}=\query^{(i)}_2\mid \query^{(i)}, \belief^{i-1})\:,
\end{align*}
and so optimizing this objective is equivalent to optimizing $P(\queryResponse^{(i)}=\query^{(i)}_1\mid \query^{(i)}, \belief^{i-1})P(\queryResponse^{(i)}=\query^{(i)}_2\mid \query^{(i)}, \belief^{i-1})$ again.
\end{proof}
\end{theorem}

\section{Proof of Theorem~\ref{thm:04_02_optimal_stopping}}\label{app:99_01_ijrr_active_optimal_stopping}
\begin{proof}
    We need to show if the global optimum is negative, then any longer-horizon optimization will also give negative reward (difference between information gain and the cost) in expectation. Let $\query^{(i)}_*$ denote the global optimizer. For any $i'\geq0$,
	\begin{align}
	I&(\queryResponse^{(i)},\dots,\queryResponse^{(i+i')} ; \weights \mid  \query^{(i)},\dots,\query^{(i+i')}) - \sum_{j=0}^{i'}\costFunction(\query^{(i+j)}) \nonumber\\
	&= I(\queryResponse^{(i)} ; \weights \mid  \query^{(i)}) + \ldots + \nonumber\\&\quad I(\queryResponse^{(i+i')} ; \weights \mid \queryResponse^{(i)}, \ldots, \queryResponse^{(i+i'-1)}, \query^{(i)},\dots,\query^{(i+i')}) - \sum_{j=0}^{i'}\costFunction(\query^{(i+j)}) \nonumber\\
	&\leq I(\queryResponse^{(i)} ; \weights \mid  \query^{(i)}) + \ldots + I(\queryResponse^{(i+i')} ; \weights \mid  \query^{(i+i')}) - \sum_{j=0}^{i'}\costFunction(\query^{(i+j)}) \nonumber\\
	&\leq (i'+1)\left[I(\queryResponse^{(i)} ; \weights \mid  \query^{(i)}_*) - \costFunction(\query^{(i)}_*)\right] < 0
	\end{align}
    where the first inequality is due to the submodularity of the mutual information, and the second inequality is because $\query^{(i)}_*$ is the global maximizer of the greedy objective. The other direction of the proof is very clear: If the global optimizer is nonnegative, then querying $\query^{(i)}_*$ will not decrease the cumulative active learning reward in expectation, so stopping is not optimal.
\end{proof}

\section{Proof of Corollary~\ref{thm:04_06_corl21_ranking_largeident}}
\label{app:99_01_corl21_ranking_identproof}

\begin{proof}
Suppose we have such a mixture of $\numberOfModes$ Plackett-Luce models that is not identifiable. Then, there must exist two distinct sets of parameters $(\weights,\mixingCoefficient)$ and $(\weights',\mixingCoefficient')$ such that for every query $\query$, the induced ranking distributions $\queryResponse_1$ and $\queryResponse_2$ respectively are identical. But since $(\weights,\mixingCoefficient)$ and $(\weights',\mixingCoefficient')$ are distinct, there is either (1) two mixing coefficients in $(\weights,\mixingCoefficient)$ and $(\weights',\mixingCoefficient')$ that disagree or (2) two trajectories $\trajectory_1$ and $\trajectory_2$ that have a different difference in rewards across $(\weights,\mixingCoefficient)$ and $(\weights',\mixingCoefficient')$ under one of the reward functions. Let $\bar\query$ with corresponding ranking distribution $\bar\queryResponse$ be an arbitrary query in case (1) and an arbitrary query containing $\trajectory_1$ and $\trajectory_2$ in case (2). Note that $\bar\queryResponse$ is the marginal distribution of the overall Plackett-Luce distribution, which by construction is a mixture of $\numberOfModes$ Plackett-Luce models with parameters $(\weights,\mixingCoefficient)$ and $(\weights',\mixingCoefficient')$, restricted to the trajectories in $\bar\query$. But now there are two distinct sets of parameters representing the distribution over the full ranking of $\query$ since we know $(\weights,\mixingCoefficient)$ and $(\weights',\mixingCoefficient')$ differ on the restricted set of trajectories $\trajectorySpace'=\query$ (either because they have differing mixing coefficients or because their induced rewards on $\trajectorySpace'$ are not a within a constant additive factor of each other since $\trajectory_1$ and $\trajectory_2$ are in $\trajectorySpace'$). But we know $\abs{\trajectorySpace'}=\abs{\query}$, so this finding contradicts the fact that $\bar\query$ must be identifiable by Theorem~\ref{thm:04_06_corl21_ranking_identifiability}. We conclude every mixture of $\numberOfModes$ Plackett-Luce models is identifiable subject to the query size bounds in the statement of this corollary.
\end{proof}

\section{Justification for Remark~\ref{thm:04_06_corl21_ranking_remark}}
\label{app:99_01_corl21_ranking_remark}
Here, we define the optimal adaptive set of queries $\rankingDataset^*$ to be the one which, in expectation, minimizes the uncertainty over model parameters $H(\weights,\mixingCoefficient \mid \rankingDataset^*)$. It is a well-known result that for \emph{adaptive submodular} functions, greedy optimization yields results that are within a constant factor $(1-\frac{1}{e})$ of optimality \cite{golovin2011adaptive}. While our mutual information objective in Eqn.~\eqref{eq:04_06_corl21_ranking_objective} is adaptive submodular in the non-adaptive setting (where all queries $\query$ are selected before observing their results), in our adaptive setting these guarantees no longer hold (conditional entropy is only submodular with respect to conditioned variables if those variables are unobserved).

%% file: 99_appendix/02_derivations.tex
\section{Mutual Information Derivation for Section~\ref{sec:04_02_information_gain}}
\label{app:99_02_mutual_information_derivation}

We first present the full derivation of Equation~\eqref{eq:04_02_IG2},
\begin{align*}
\query_*^{(i)} = \argmax_{\query^{(i)}=\{\trajectory_1,\dots,\trajectory_{\abs{\query^{(i)}}}\}} I(\queryResponse^{(i)} ; \weights \mid \query^{(i)}, \belief^{i-1})\:.
\end{align*}
We first write the mutual information as the difference between two entropy terms:
\begin{align}
I&(\queryResponse^{(i)} ; \weights \mid \query^{(i)}, \belief^{i-1}) = H(\weights \mid \query^{(i)}, \belief^{i-1}) - \mathbb{E}_{\queryResponse^{(i)}\mid \query^{(i)},\belief^{i-1}}\left[H(\weights \mid \queryResponse^{(i)},\query^{(i)},\belief^{i-1})\right]\:.
\end{align}
Next, we expand the entropy expressions and use $P(\queryResponse^{(i)}\mid \query^{(i)},\belief^{i-1})P(\weights\mid \queryResponse^{(i)},\query^{(i)},\belief^{i-1})=P(\weights,\queryResponse^{(i)}\mid \query^{(i)},\belief^{i-1})$ to combine the expectations for the second term to get:
\begin{align}
H&(\weights \mid \query^{(i)}, \belief^{i-1}) - \mathbb{E}_{\queryResponse^{(i)}\mid \query^{(i)},\belief^{i-1}}\left[H(\weights \mid \queryResponse^{(i)},\query^{(i)},\belief^{i-1})\right] \nonumber\\
&= -\mathbb{E}_{\weights\mid \query^{(i)},\belief^{i-1}}\left[\log_2 P(\weights\mid \query^{(i)},\belief^{i-1})\right] + \mathbb{E}_{\weights,\queryResponse^{(i)}\mid \query^{(i)},\belief^{i-1}}\left[\log_2\left(P(\weights \mid \queryResponse^{(i)},\query^{(i)},\belief^{i-1})\right)\right]\:.
\end{align}
Since the first term is independent from $\queryResponse^{(i)}$, we can write this expression as
\begin{align}
&\mathbb{E}_{\weights,\queryResponse^{(i)}\mid \query^{(i)},\belief^{i-1}}\left[\log_2 P(\weights \mid \queryResponse^{(i)},\query^{(i)},\belief^{i-1}) - \log_2 P(\weights\mid \query^{(i)},\belief^{i-1})\right] \nonumber\\
&= \mathbb{E}_{\weights, \queryResponse^{(i)}\mid \query^{(i)},\belief^{i-1}}\left[\log_2 P(\queryResponse^{(i)} \mid \query^{(i)},\belief^{i-1},\weights) - \log_2 P(\queryResponse^{(i)}\mid \query^{(i)},\belief^{i-1})\right] \nonumber\\
&= \mathbb{E}_{\weights, \queryResponse^{(i)}\mid \query^{(i)},\belief^{i-1}}\Bigg[\log_2 P(\queryResponse^{(i)} \mid \query^{(i)},\weights) - \log_2\left(\int P(\queryResponse^{(i)}\mid \query^{(i)},\weights')P(\weights' \mid \query^{(i)},\belief^{i-1})d\weights'\right)\Bigg] \;,
\end{align}
where the integral is taken over all possible values of $\weights$.

Having $\weightsSampleSet$ as a set of samples drawn from the prior $\belief^{i-1}$,
\begin{align}
I(\queryResponse^{(i)} ; \weights \mid \query^{(i)}) &\asymeq \mathbb{E}_{\weights, \queryResponse^{(i)}\mid \query^{(i)}, \belief^{i-1}}\Bigg[\log_2 P(\queryResponse^{(i)} \mid \query^{(i)},\weights) - \log_2\left(\frac{1}{\abs{\weightsSampleSet}} \sum_{\weights'\in\weightsSampleSet} P(\queryResponse^{(i)}\mid \query^{(i)},\weights')\right)\Bigg] \nonumber\\
&= \mathbb{E}_{\weights, \queryResponse^{(i)}\mid \query^{(i)}, \belief^{i-1}}\left[\log_2\frac{\abs{\weightsSampleSet}\cdot P(\queryResponse^{(i)} \mid \query^{(i)},\weights)}{ \sum_{\weights'\in\weightsSampleSet} P(\queryResponse^{(i)}\mid \query^{(i)},\weights')}\right] \nonumber\\
&= \mathbb{E}_{\weights\mid \query^{(i)}, \belief^{i-1}} \left[\mathbb{E}_{\queryResponse^{(i)}\mid \query^{(i)},\weights}\left[\log_2\frac{\abs{\weightsSampleSet}\cdot P(\queryResponse^{(i)} \mid \query^{(i)},\weights)}{ \sum_{\weights'\in\weightsSampleSet} P(\queryResponse^{(i)} \mid \query^{(i)},\weights')}\right]\right] \nonumber\\
&= \mathbb{E}_{\weights\mid \query^{(i)}, \belief^{i-1}}\Bigg[ \sum_{\queryResponse^{(i)} \in \query^{(i)}} P(\queryResponse^{(i)} \mid \query^{(i)}, \weights) \log_2\frac{\abs{\weightsSampleSet}\cdot P(\queryResponse^{(i)} \mid \query^{(i)},\weights)}{ \sum_{\weights' \in\weightsSampleSet} P(\queryResponse^{(i)} \mid \query^{(i)}, \weights')} \Bigg] \nonumber\\
&\asymeq \frac{1}{\abs{\weightsSampleSet}} \sum_{\queryResponse^{(i)} \in \query^{(i)}} \sum_{\bar{\weights} \in\weightsSampleSet} P(\queryResponse^{(i)} \mid \query^{(i)}, \bar{\weights}) \log_2\frac{\abs{\weightsSampleSet}\cdot P(\queryResponse^{(i)} \mid \query^{(i)},\bar{\weights})}{ \sum_{\weights' \in\weightsSampleSet} P(\queryResponse^{(i)} \mid \query^{(i)}, \weights')} \;, \tag{see~\ref{eq:04_02_IG2}}
\end{align}
where, in the last step, we use the sampled $\weights$'s to compute the expectation over $\weights\mid \query^{(i)}, \belief^{i-1}$, which is equivalent to $\weights \mid \belief^{i-1}$. This completes the derivation.

\subsection{Extension to User-Specific and Unknown $\minimumPerceivableDifference$}\label{app:99_02_ijrr_active_unknown_parameter}
We now derive the mutual information optimization when the minimum perceivable difference parameter $\minimumPerceivableDifference$ of the extended human model (for weak pairwise comparison queries) we introduced in Section~\ref{subsec:04_02_experiments} is unknown. One can also attempt to learn the rationality coefficient $\comparisonRationalityCoefficient$. Therefore, for generality, we denote all human model parameters that will be learned as a vector $\humanParametersAppendix$. Furthermore, we denote the belief over $(\weights,\humanParametersAppendix)$ before iteration $i$ as $\belief_+^{i-1}$. Since our true goal is to learn $\weights$, the optimization now becomes:
\begin{align}
\query^{(i)}_* &= \argmax_{\query^{(i)}=\{\trajectory_1,\dots,\trajectory_{\abs{\query^{(i)}}}\}} \mathbb{E}_{\humanParametersAppendix\mid \query^{(i)},\belief_+^{i-1}}\left[I(\queryResponse^{(i)};\weights \mid \query^{(i)},\belief_+^{i-1})\right]
\end{align}
We now work on this objective as follows:
\begin{align}
\mathbb{E}&_{\humanParametersAppendix \mid \query^{(i)},\belief_+^{i-1}}\left[I(\queryResponse^{(i)};\weights \mid \query^{(i)},\belief_+^{i-1})\right] \nonumber\\
&= \mathbb{E}_{\humanParametersAppendix \mid \query^{(i)},\belief_+^{i-1}}\Big[H(\weights \mid \humanParametersAppendix, \query^{(i)},\belief_+^{i-1}) - \mathbb{E}_{\queryResponse^{(i)}\mid \humanParametersAppendix, \query^{(i)},\belief_+^{i-1}}\left[H(\weights \mid \queryResponse^{(i)}, \humanParametersAppendix, \query^{(i)},\belief_+^{i-1})\right]\Big] \nonumber\\
&= \mathbb{E}_{\humanParametersAppendix\mid \query^{(i)},\belief_+^{i-1}}\left[H(\weights \mid \humanParametersAppendix, \query^{(i)},\belief_+^{i-1})\right] - \mathbb{E}_{\humanParametersAppendix, \queryResponse^{(i)} \mid \query^{(i)},\belief_+^{i-1}}\left[H(\weights \mid \queryResponse^{(i)}, \humanParametersAppendix, \query^{(i)},\belief_+^{(i)})\right] \nonumber\\
&= -\mathbb{E}_{\humanParametersAppendix,\weights \mid \query^{(i)},\belief_+^{i-1}}\left[\log_2 P(\weights \mid \humanParametersAppendix, \query^{(i)},\belief_+^{i-1})\right] + \mathbb{E}_{\humanParametersAppendix, \queryResponse^{(i)},\weights\mid \query^{(i)},\belief_+^{i-1}}\left[\log_2 P(\weights \mid \queryResponse^{(i)}, \humanParametersAppendix, \query^{(i)},\belief_+^{i-1})\right] \nonumber\\
&= \mathbb{E}_{\humanParametersAppendix, \queryResponse^{(i)}, \weights\mid \query^{(i)},\belief_+^{i-1}}\big[\log_2 P(\weights \mid \queryResponse^{(i)}, \humanParametersAppendix, \query^{(i)},\belief_+^{i-1}) - \log_2 P(\weights \mid \humanParametersAppendix, \query^{(i)},\belief_+^{i-1})\big] \nonumber\\
&= \mathbb{E}_{\humanParametersAppendix,\queryResponse^{(i)},\weights\mid \query^{(i)},\belief_+^{i-1}}\big[\log_2 P(\queryResponse_i \mid \weights, \humanParametersAppendix, \query^{(i)}, \belief_+^{i-1}) - \log_2 P(\queryResponse^{(i)} \mid \humanParametersAppendix, \query^{(i)},\belief_+^{i-1})\big] \nonumber\\
&= \mathbb{E}_{\humanParametersAppendix,\queryResponse^{(i)},\weights\mid \query^{(i)},\belief_+^{i-1}} \!\big[\log_2 P(\queryResponse^{(i)} \mid \weights, \humanParametersAppendix, \query^{(i)},\belief_+^{i-1}) \!-\! \log_2 P(\humanParametersAppendix,\queryResponse{(i)} \mid \query^{(i)},\belief_+^{i-1}) \!+\! \log_2 P(\humanParametersAppendix \mid \query^{(i)},\belief_+^{i-1})\big]
\end{align}
Noting that $P(\humanParametersAppendix \mid \query^{(i)},\belief_+^{i-1}) = P(\humanParametersAppendix \mid \belief_+^{i-1})$, we drop the last term because it does not involve the optimization variable $\query^{(i)}$. Also noting $P(\queryResponse^{(i)} \mid \weights, \humanParametersAppendix, \query^{(i)},\belief_+^{i-1}) = P(\queryResponse^{(i)} \mid \weights, \humanParametersAppendix, \query^{(i)})$, the new objective is:
\begin{align}
& \mathbb{E}_{\humanParametersAppendix,\queryResponse^{(i)},\weights\mid \query^{(i)},\belief_+^{i-1}}\left[\log_2 P(\queryResponse^{(i)} \mid \weights, \humanParametersAppendix, \query^{(i)})- \log_2 P(\humanParametersAppendix,\queryResponse^{(i)} \mid \query^{(i)},\belief_+^{i-1})\right] \nonumber\\
&\asymeq \frac{1}{\abs{\weightsSampleSet^+}} \sum_{(\bar{\weights},\bar{\humanParametersAppendix}) \in \weightsSampleSet^+}\sum_{\queryResponse^{(i)}\in \query^{(i)}}P(\queryResponse^{(i)}\mid \bar{\weights},\bar{\humanParametersAppendix},\query^{(i)})\left[\log_2 P(\queryResponse^{(i)} \mid \bar{\weights}, \bar{\humanParametersAppendix}, \query^{(i)}) - \log_2 P(\bar{\humanParametersAppendix},\queryResponse^{(i)} \mid \query^{(i)},\belief_+^{i-1})\right]
\end{align}
where $\weightsSampleSet^+$ is a set containing samples from $\belief_+^{i-1}$. Since $P(\bar{\humanParametersAppendix},\queryResponse^{(i)} \mid \query^{(i)},\belief_+^{i-1}) = \int P(\queryResponse^{(i)} \mid \bar{\humanParametersAppendix}, \weights', \query^{(i)})P(\bar{\humanParametersAppendix}, \weights' \mid \query^{(i)},\belief_+^{i-1})d\weights'$ where the integration is over all possible values of $\weights$, we can write the second logarithm term as:
\begin{align}
\log_2\left(\frac{1}{\abs{\weightsSampleSet^+}} \sum_{\weights'\in \weightsSampleSet(\bar{\humanParametersAppendix})}P(\queryResponse^{(i)} \mid \bar{\humanParametersAppendix},\weights', \query^{(i)})\right)
\end{align}
with asymptotic equality, where $\weightsSampleSet(\bar{\humanParametersAppendix})$ is the set that contains samples from $\belief_+^{i-1}$ with fixed $\bar{\humanParametersAppendix}$. Note that while we can actually compute this objective, it is computationally much heavier than the case without $\humanParametersAppendix$, because we need to sample $\weights$ for each $\bar{\humanParametersAppendix}$ sample.

One property of this objective that will ease the computation is the fact that it is parallelizable. An alternative approach is to actively learn $(\weights,\humanParametersAppendix)$ instead of just $\weights$. This will of course cause some performance loss, because we are only interested in $\weights$. However, if we learn them together, the derivation follows the derivation of Equation~\eqref{eq:04_02_IG2}, which we already presented, by simply replacing $\weights$ with $(\weights,\humanParametersAppendix)$, and the final optimization becomes:
\begin{align*}
\argmax_{\query^{(i)}=\{\trajectory_1,\dots,\trajectory_{\abs{\query^{(i)}}}\}} \frac{1}{\abs{\weightsSampleSet^+}} \sum_{\queryResponse^{(i)} \in \query^{(i)}}\sum_{(\bar{\weights},\bar{\humanParametersAppendix})\in\weightsSampleSet^+} P(\queryResponse^{(i)}\mid \query^{(i)},\bar{\weights},\bar{\humanParametersAppendix})\log_2\frac{\abs{\weightsSampleSet^+}\cdot P(\queryResponse^{(i)} \mid \query^{(i)}, \bar{\weights},\bar{\humanParametersAppendix})}{ \sum_{(\weights',\humanParametersAppendix')\in\weightsSampleSet^+} P(\queryResponse_i\mid \query_i,\weights',\humanParametersAppendix')}
\end{align*}

\section{Mutual Information Derivation for Section~\ref{sec:04_03_gp}}\label{app:99_02_rss20_gp_derivation}
	Let $\posteriorCovarianceAppendix$ be the posterior covariance matrix between $f(\trajectoryFeaturesFunction^{(1)})$ and $f(\trajectoryFeaturesFunction^{(2)})$. And let
	\begin{align*}
	\posteriorCovarianceAppendix^{-1} = \begin{bmatrix} c&d \\ d&c'  \end{bmatrix}.
	\end{align*}
	Note that the $c$ here is not related to the cost function $\costFunction$ we used in Section~\ref{sec:04_02_information_gain}. Throughout the derivation, all integrals are calculated over $\mathbb{R}$, but we drop it to simplify the notation. $h$ denotes the binary entropy function, and $\standardNormalCdf$ is the cdf of the standard normal distribution. We use $\gpRewardFunction_1$ and $\gpRewardFunction_2$ to denote $\gpRewardFunction(\trajectoryFeaturesFunction^{(1)})$ and $\gpRewardFunction(\trajectoryFeaturesFunction^{(2)})$, respectively. We write the first entropy term in the optimization \eqref{eq:04_03_problem} as:
	\begin{align}
	H&(\queryResponse \mid \trajectoryFeaturesFunction^{(1)}, \trajectoryFeaturesFunction^{(2)}, \mathbf{\query}, \mathbf{\queryResponse}) \nonumber\\
	&= h\left(\int \int\standardNormalCdf\left(\frac{\gpRewardFunction_1-\gpRewardFunction_2}{\sqrt{2}{\preferenceNoiseStd}}\right)\mathcal{N}([\gpRewardFunction_1,\gpRewardFunction_2]\mid[\gpPosteriorMean^{(1)},\gpPosteriorMean^{(2)}], \posteriorCovarianceAppendix)d\gpRewardFunction_2d\gpRewardFunction_1\right) \nonumber\\
	&= h\left(\frac{\sqrt{cc'-d^2}}{2\pi }\int \int \standardNormalCdf\left(\frac{\gpRewardFunction_1-\gpRewardFunction_2}{\sqrt{2}{\preferenceNoiseStd}}\right)e^{-\frac{1}{2}(c(\gpRewardFunction_1-\gpPosteriorMean^{(1)})^2 + c'(\gpRewardFunction_2-\gpPosteriorMean^{(2)})^2+2d(\gpRewardFunction_1-\gpPosteriorMean^{(1)})(\gpRewardFunction_2-\gpPosteriorMean^{(2)}))}d\gpRewardFunction_2d\gpRewardFunction_1\right) \nonumber\\
	&= h\left(\frac{\sqrt{cc'-d^2}}{2\pi}\int \int \standardNormalCdf\left(\frac{\gpRewardFunction_1-\gpRewardFunction_2}{\sqrt{2}{\preferenceNoiseStd}}\right)e^{-\frac{1}{2}(c((\gpRewardFunction_1-\gpPosteriorMean^{(1)})^2+\frac{2d}{c}(\gpRewardFunction_1-\gpPosteriorMean^{(1)})(\gpRewardFunction_2-\gpPosteriorMean^{(2)}))+c'(\gpRewardFunction_2-\gpPosteriorMean^{(2)})^2)}d\gpRewardFunction_1d\gpRewardFunction_2\right) \nonumber\\
	&= h\left(\frac{\sqrt{cc'-d^2}}{2\pi}\int \int \standardNormalCdf\left(\frac{\gpRewardFunction_1-\gpRewardFunction_2}{\sqrt{2}{\preferenceNoiseStd}}\right)e^{-\frac{1}{2}(c(\gpRewardFunction_1-\gpPosteriorMean^{(1)}+\frac{d}{c}(\gpRewardFunction_2-\gpPosteriorMean^{(2)}))^2-\frac{d^2}{c}(\gpRewardFunction_2-\gpPosteriorMean^{(2)})^2+c'(\gpRewardFunction_2-\gpPosteriorMean^{(2)})^2)}d\gpRewardFunction_1d\gpRewardFunction_2\right) \nonumber\\
	&= h\left(\frac{\sqrt{cc'-d^2}}{2\pi}\int e^{-\frac{1}{2}c'(\gpRewardFunction_2-\gpPosteriorMean^{(2)})^2}e^{\frac{1}{2}\frac{d^2}{c}(\gpRewardFunction_2-\gpPosteriorMean^{(2)})^2} \int \standardNormalCdf\left(\frac{\gpRewardFunction_1-\gpRewardFunction_2}{\sqrt{2}{\preferenceNoiseStd}}\right)e^{-\frac{1}{2}(c(\gpRewardFunction_1-\gpPosteriorMean^{(1)}+\frac{d}{c}(\gpRewardFunction_2-\gpPosteriorMean^{(2)}))^2)}d\gpRewardFunction_1d\gpRewardFunction_2\right) \nonumber\\
	&= h\left(\frac{\sqrt{cc'-d^2}}{2\pi}\int e^{-\frac{1}{2}\frac{c'c-d^2}{c}(\gpRewardFunction_2-\gpPosteriorMean^{(2)})^2} \int \frac{\standardNormalCdf\left(\frac{\gpRewardFunction_1-\gpRewardFunction_2}{\sqrt{2}{\preferenceNoiseStd}}\right)e^{-\frac{1}{2}(c(\gpRewardFunction_1-\gpPosteriorMean^{(1)}+\frac{d}{c}(\gpRewardFunction_2-\gpPosteriorMean^{(2)}))^2)}}{\frac{\sqrt{2\pi}}{\sqrt c}} \frac{\sqrt{2\pi}}{\sqrt c}d\gpRewardFunction_1d\gpRewardFunction_2\right) \nonumber\\
	&= h\left(\frac{\sqrt{cc'-d^2}}{\sqrt{2\pi}\sqrt{c}}\int e^{-\frac{1}{2}\frac{c'c-d^2}{c}(\gpRewardFunction_2-\gpPosteriorMean^{(2)})^2} \int \frac{\standardNormalCdf\left(\frac{\gpRewardFunction_1-\gpRewardFunction_2}{\sqrt{2}{\preferenceNoiseStd}}\right)e^{-\frac{1}{2}(c(\gpRewardFunction_1-\gpPosteriorMean^{(1)}+\frac{d}{c}(\gpRewardFunction_2-\gpPosteriorMean^{(2)}))^2)}}{\frac{\sqrt{ 2\pi}}{\sqrt c}}d\gpRewardFunction_1d\gpRewardFunction_2\right)
	\end{align}
	Using the mathematical identity $\int_{x} \phi(x) N(x|\gpPosteriorMean, {\preferenceNoiseStd}^2) dx = \phi(\frac{\gpPosteriorMean}{\sqrt{1+{\preferenceNoiseStd}^2}})$, we obtain
	\begin{align}
	H&(q \mid \trajectoryFeaturesFunction^{(1)}, \trajectoryFeaturesFunction^{(2)}, \mathbf{\query}, \mathbf{\queryResponse}) \nonumber\\
	&= h\left(\frac{\sqrt{cc'-d^2}}{\sqrt{2\pi}\sqrt{c}}\int e^{-\frac{1}{2}\frac{c'c-d^2}{c}(\gpRewardFunction_2-\gpPosteriorMean^{(2)})^2} \standardNormalCdf\left(\frac{\gpPosteriorMean^{(1)}-\frac{d}{c}\gpRewardFunction_2+\frac{d}{c}\gpPosteriorMean^{(2)}-\gpRewardFunction_2}{\sqrt{2}{\preferenceNoiseStd}\sqrt{1+\frac{1}{2c{\preferenceNoiseStd}^2}}}\right)d\gpRewardFunction_2\right) \nonumber\\
	&= h\left(\frac{\sqrt{cc'-d^2}}{\sqrt{2\pi}\sqrt{c}}\int e^{-\frac{1}{2}\frac{c'c-d^2}{c}(\gpRewardFunction_2-\gpPosteriorMean^{(2)})^2} \standardNormalCdf\left(\frac{\gpPosteriorMean^{(1)}+\frac{d}{c}\gpPosteriorMean^{(2)}-(\frac{d}{c}+1)\gpRewardFunction_2}{\sqrt{2}{\preferenceNoiseStd}\sqrt{1+\frac{1}{2c{\preferenceNoiseStd}^2}}}\right)d\gpRewardFunction_2\right) \nonumber\\
	&= h\left(\frac{\sqrt{cc'-d^2}}{\sqrt{2\pi}\sqrt{c}}\int e^{-\frac{1}{2}\frac{c'c-d^2}{c}(\gpRewardFunction_2-\gpPosteriorMean^{(2)})^2} \frac{\standardNormalCdf\left(\frac{-(\frac{d}{c}+1)(\gpRewardFunction_2-\frac{\gpPosteriorMean^{(1)}+\frac{d}{c}\gpPosteriorMean^{(2)}}{\frac{d}{c}+1})}{\sqrt{2}{\preferenceNoiseStd}\sqrt{1+\frac{1}{2c{\preferenceNoiseStd}^2}}}\right)}{\frac{\sqrt{2\pi c}}{\sqrt{c'c-d^2}}} \frac{\sqrt{2\pi c}}{\sqrt{c'c-d^2}}d\gpRewardFunction_2\right)
	\end{align}
	Using the same identity again,
	\begin{align}
	H(q \mid \trajectoryFeaturesFunction^{(1)}, \trajectoryFeaturesFunction^{(2)}, \mathbf{\query}, \mathbf{\queryResponse}) &= h\left(\standardNormalCdf\left(\frac{\gpPosteriorMean^{(1)}-\gpPosteriorMean^{(2)}}{\sqrt{2}{\preferenceNoiseStd} \sqrt{1+\frac{1}{2c{\preferenceNoiseStd}^2}}\sqrt{1+\frac{c}{2{\preferenceNoiseStd}^2+\frac{1}{c}}\frac{(1+\frac{d}{c})^2}{c'c-d^2}}}\right)\right)
	\end{align}
	One can then expand the expression in the denominator and use the facts that $\mathrm{Var}(f(\trajectoryFeaturesFunction^{(1)}))=\frac{c'}{cc'-d^2}$, $\mathrm{Var}(f(\trajectoryFeaturesFunction^{(2)}))=\frac{c}{cc'-d^2}$ and $\mathrm{Cov}(f(\trajectoryFeaturesFunction^{(1)}),f(\trajectoryFeaturesFunction^{(2)}))=\frac{-d}{cc'-d^2}$ to obtain
	\begin{align}
	H(q \mid \trajectoryFeaturesFunction^{(1)}, \trajectoryFeaturesFunction^{(2)}, \mathbf{\query}, \mathbf{\queryResponse})=h\left(\standardNormalCdf\left(\frac{\gpPosteriorMean^{(1)}-\gpPosteriorMean^{(2)}}{\sqrt{2{\preferenceNoiseStd}^2 + \gpGFunction(\trajectoryFeaturesFunction^{(1)},\trajectoryFeaturesFunction^{(2)})}}\right)\right).
	\end{align}
	where $\gpGFunction(\trajectoryFeaturesFunction^{(1)},\trajectoryFeaturesFunction^{(2)}) = \mathrm{Var}(f(\trajectoryFeaturesFunction^{(1)})) + \mathrm{Var}(f(\trajectoryFeaturesFunction^{(2)})) - 2\mathrm{Cov}(f(\trajectoryFeaturesFunction^{(1)}),f(\trajectoryFeaturesFunction^{(2)}))$
	
We next make the derivation for the second entropy term. To simplify the notation, we let ${\preferenceNoiseStd}'^2=\frac{\pi ln(2)}{2}, {\preferenceNoiseStd}''^2 = {\preferenceNoiseStd}'^2 +\frac{1}{c}$, and ${\preferenceNoiseStd}_b^2 = \frac{c(1+\frac{d}{c})^2}{c'c-d^2}$. By performing a linearization over the logarithm of the second entropy term as in \cite{houlsby2011bayesian},
\begin{align}
	&\mathbb{E}_{\gpRewardFunction\sim P(\gpRewardFunction \mid \mathbf{\query}, \mathbf{\queryResponse})}\left[H(q \mid \trajectoryFeaturesFunction^{(1)}, \trajectoryFeaturesFunction^{(2)}, \gpRewardFunction)\right] \nonumber\\
	&\approx \frac{\sqrt{c'c-d^2}}{2\pi}\int \int e^{-\frac{(\gpRewardFunction_1-\gpRewardFunction_2)^2}{\pi ln(2)}}e^{-\frac{1}{2}(c(\gpRewardFunction_1-\gpPosteriorMean^{(1)})^2 + c'(\gpRewardFunction_2-\gpPosteriorMean^{(2)})^2+2d(\gpRewardFunction_1-\gpPosteriorMean^{(1)})(\gpRewardFunction_2-\gpPosteriorMean^{(2)}))}d\gpRewardFunction_1d\gpRewardFunction_2 \nonumber\\
	&= \frac{\sqrt{c'c-d^2}}{2\pi}\int e^{-\frac{1}{2}c'(\gpRewardFunction_2-\gpPosteriorMean^{(2)})^2} \int e^{-\frac{(\gpRewardFunction_1-\gpRewardFunction_2)^2}{2{\preferenceNoiseStd}'^2 }}e^{-\frac{1}{2}(c(\gpRewardFunction_1-\gpPosteriorMean^{(1)})^2 +2d(\gpRewardFunction_1-\gpPosteriorMean^{(1)})(\gpRewardFunction_2-\gpPosteriorMean^{(2)}))}d\gpRewardFunction_1d\gpRewardFunction_2 \nonumber\\
	&=\frac{\sqrt{c'c-d^2}}{2\pi}\int e^{-\frac{1}{2}c'{\gpRewardFunction_2}^2} \int e^{-\frac{(\gpRewardFunction_1+\gpPosteriorMean^{(1)}-\gpRewardFunction_2-\gpPosteriorMean^{(2)})^2}{2{\preferenceNoiseStd}'^2}}e^{-\frac{1}{2}c{\gpRewardFunction_1}^2 -d \gpRewardFunction_1 \gpRewardFunction_2}d\gpRewardFunction_1d\gpRewardFunction_2 \nonumber\\
	&=\frac{\sqrt{c'c-d^2}}{2\pi}\int e^{-\frac{1}{2}c'{\gpRewardFunction_2}^2} \int e^{-\frac{(\gpRewardFunction_1+\gpPosteriorMean^{(1)}-\gpRewardFunction_2-\gpPosteriorMean^{(2)})^2}{2{\preferenceNoiseStd}'^2}}e^{-\frac{1}{2}c(\gpRewardFunction_1+\frac{d}{c}\gpRewardFunction_2)^2+\frac{1}{2}\frac{d^2}{c}{\gpRewardFunction_2}^2}d\gpRewardFunction_1d\gpRewardFunction_2 \nonumber\\
	&=\frac{\sqrt{c'c-d^2}}{2\pi}\int e^{-\frac{1}{2}\frac{c'c-d^2}{c}{\gpRewardFunction_2}^2} \int e^{-\frac{(\gpRewardFunction_1-\gpRewardFunction_2+\gpPosteriorMean^{(1)}-\gpPosteriorMean^{(2)})^2}{2{\preferenceNoiseStd}'^2}}e^{-\frac{1}{2}c(\gpRewardFunction_1+\frac{d}{c}\gpRewardFunction_2)^2}d\gpRewardFunction_1d\gpRewardFunction_2
\end{align}
By the change of variables for the inner integral with $u=\gpRewardFunction_1 + \frac{d}{c}\gpRewardFunction_2$,
\begin{align}
	\mathbb{E}_{\gpRewardFunction\sim P(\gpRewardFunction \mid \mathbf{\query}, \mathbf{\queryResponse})}\left[H(q \mid \trajectoryFeaturesFunction^{(1)}, \trajectoryFeaturesFunction^{(2)}, \gpRewardFunction)\right] &=\frac{\sqrt{c'c-d^2}}{2\pi}\int e^{-\frac{c'c-d^2}{2c}{\gpRewardFunction_2}^2} \int_u e^{-\frac{(u-\frac{d}{c}\gpRewardFunction_2+\gpPosteriorMean^{(1)}-\gpRewardFunction_2-\gpPosteriorMean^{(2)})^2}{2{\preferenceNoiseStd}'^2}}e^{-\frac{1}{2}cu^2}dud\gpRewardFunction_2 \nonumber\\
	&=\frac{\sqrt{c'c-d^2}}{2\pi}\int e^{-\frac{c'c-d^2}{2c}{\gpRewardFunction_2}^2} \int_u e^{-\frac{((1+\frac{d}{c})\gpRewardFunction_2-u-\gpPosteriorMean^{(1)}+\gpPosteriorMean^{(2)})^2}{2{\preferenceNoiseStd}'^2}}e^{-\frac{1}{2}cu^2}dud\gpRewardFunction_2
\end{align}
By another change of variables for the outer integral with $v=\frac{\gpRewardFunction_2}{1 + d/c}$,
\begin{align}
	\mathbb{E}_{\gpRewardFunction\sim P(\gpRewardFunction \mid \mathbf{\query}, \mathbf{\queryResponse})}\!\left[H(q \mid \trajectoryFeaturesFunction^{(1)}, \trajectoryFeaturesFunction^{(2)}, \gpRewardFunction)\right]\!=\!\frac{1}{1+\frac{d}{c}}\frac{\sqrt{c'c-d^2}}{2\pi}\!\int_v\! e^{-\frac{c'c-d^2}{2c}\frac{v^2}{(1+\frac{d}{c})^2}} \!\int_u\! e^{-\frac{(v-u+\gpPosteriorMean^{(2)}-\gpPosteriorMean^{(1)})^2}{2{\preferenceNoiseStd}'^2}}e^{-\frac{1}{2}cu^2}dudv\:.
\end{align}
By identifying the inner integral as a convolution of two Gaussians, we get
\begin{align}
	&\mathbb{E}_{\gpRewardFunction\sim P(\gpRewardFunction \mid \mathbf{\query}, \mathbf{\queryResponse})}\left[H(q \mid \trajectoryFeaturesFunction^{(1)}, \trajectoryFeaturesFunction^{(2)}, \gpRewardFunction)\right] \nonumber\\ &=\frac{1}{1+\frac{d}{c}}\frac{\sqrt{c'c-d^2}}{2\pi}2\pi{\preferenceNoiseStd}'\frac{1}{\sqrt{c}}\int_v e^{-\frac{1}{2}\frac{c'c-d^2}{c}\frac{v^2}{(1+\frac{d}{c})^2}} \frac{1}{\sqrt{2\pi}\sqrt{{\preferenceNoiseStd}'^2+\frac{1}{c}}} e^{-\frac{1}{2}\frac{(v-(\gpPosteriorMean^{(1)}-\gpPosteriorMean^{(2)}))^2}{{\preferenceNoiseStd}'^2+\frac{1}{c}}}dv \nonumber\\
	&=\frac{1}{1+\frac{d}{c}}\sqrt{c'c-d^2}{\preferenceNoiseStd}'\frac{1}{\sqrt{c}}\frac{1}{\sqrt{2\pi}\sqrt{{\preferenceNoiseStd}'^2+\frac{1}{c}}}\int_v e^{-\frac{1}{2}\frac{v^2}{{\preferenceNoiseStd}_b^2}}  e^{-\frac{1}{2}\frac{(v-(\gpPosteriorMean^{(1)}-\gpPosteriorMean^{(2)}))^2}{{\preferenceNoiseStd}''^2}}dv\:.
\end{align}
By repeating the same convolution trick for the second integral,
\begin{align}
	&\mathbb{E}_{\gpRewardFunction\sim P(\gpRewardFunction \mid \mathbf{\query}, \mathbf{\queryResponse})}\left[H(q \mid \trajectoryFeaturesFunction^{(1)}, \trajectoryFeaturesFunction^{(2)}, \gpRewardFunction)\right] \nonumber\\
	&=\frac{1}{1+\frac{d}{c}}\sqrt{c'c-d^2}{\preferenceNoiseStd}'\frac{1}{\sqrt{c}}\frac{1}{\sqrt{2\pi}\sqrt{{\preferenceNoiseStd}'^2+\frac{1}{c}}} 2\pi {\preferenceNoiseStd}_b {\preferenceNoiseStd}'' \frac{1}{\sqrt{2\pi}\sqrt{{\preferenceNoiseStd}_b^2+{\preferenceNoiseStd}''^2}}e^{-\frac{1}{2}\frac{(\gpPosteriorMean^{(1)}-\gpPosteriorMean^{(2)})^2}{{\preferenceNoiseStd}_b^2+{\preferenceNoiseStd}''^2}} \nonumber\\
	&=\frac{1}{1+\frac{d}{c}}\sqrt{c'c-d^2}{\preferenceNoiseStd}'\frac{1}{\sqrt{c}}\frac{1}{\sqrt{{\preferenceNoiseStd}'^2+\frac{1}{c}}}  {\preferenceNoiseStd}_b {\preferenceNoiseStd}'' \frac{1}{\sqrt{{\preferenceNoiseStd}_b^2+{\preferenceNoiseStd}''^2}}e^{-\frac{1}{2}\frac{(\gpPosteriorMean^{(1)}-\gpPosteriorMean^{(2)})^2}{{\preferenceNoiseStd}_b^2+{\preferenceNoiseStd}''^2}}\:.
\end{align}
Again, we express this in terms of covariance and variance expressions:
\begin{align}
	\mathbb{E}_{\gpRewardFunction\sim P(\gpRewardFunction \mid \mathbf{\query}, \mathbf{\queryResponse})}\left[H(q \mid \trajectoryFeaturesFunction^{(1)}, \trajectoryFeaturesFunction^{(2)}, \gpRewardFunction)\right] = \frac{\sqrt{\pi\ln(2){\preferenceNoiseStd}^2}\exp\left(-\frac{(\gpPosteriorMean^{(1)} - \gpPosteriorMean^{(2)})^2}{\pi\ln(2){\preferenceNoiseStd}^2 + 2\gpGFunction(\trajectoryFeaturesFunction^{(1)},\trajectoryFeaturesFunction^{(2)})}\right)}{\sqrt{\pi\ln(2){\preferenceNoiseStd}^2 + 2\gpGFunction(\trajectoryFeaturesFunction^{(1)},\trajectoryFeaturesFunction^{(2)})}}\:.
\end{align}
This completes the derivation.

\section{Mutual Information Derivation for Section~\ref{sec:04_06_rankings}}
\label{app:99_02_corl21_ranking_derivation}
We present the derivation of the formula for computing the maximum mutual information query $\query^*$.
Assume at a fixed round $i$ we have made past ranking query observations $\rankingDataset=\{\query^{(i')},\queryResponse^{(i')}\}_{i'=1}^{i-1}$, and possibly other types of feedback to have the belief distribution $\belief^{i-1}$. The desired query is then 
\begin{align}
\label{eq:99_02_corl21_ranking_initial_objective}
    \query^*=\argmax_\query I(\queryResponse; \weights, \mixingCoefficient \mid \query, \belief^{i-1}),
\end{align} 
where $I(\cdot;\cdot)$ denotes mutual information and $\queryResponse$ is the response to the query $\query$. Equivalently, denoting conditional entropy with $H(\cdot\mid\cdot)$, we note
\begin{align*}
    I(\queryResponse; \weights, \mixingCoefficient \mid \query, \belief^{i-1}) &= H(\weights, \mixingCoefficient \mid \belief^{i-1}) - \mathbb{E}_{\queryResponse' \sim \queryResponse\mid \query, \belief^{i-1}} \bigg[ H(\weights, \mixingCoefficient \mid \query, \queryResponse=\queryResponse',\belief^{i-1})\bigg]\:,
\end{align*}
which allows us to write the optimization in Equation~\eqref{eq:99_02_corl21_ranking_initial_objective} equivalently as
\begin{align*}
    \query^* &= \argmin_{\query} \mathbb{E}_{\queryResponse'\sim\queryResponse\mid \query, \belief^{i-1}}\bigg[H(\weights,\mixingCoefficient\mid \query,\queryResponse=\queryResponse',\belief^{i-1})\bigg]\:.
\end{align*}

We further simplify this minimization objective by denoting the joint distribution over $\queryResponse$ and $(\weights,\mixingCoefficient)$ conditioned on $\query$ and $\belief^{i-1}$ as $P(\queryResponse,\weights,\mixingCoefficient\mid \query,\belief^{i-1})$ and expanding the entropy term:

\begin{align*}
    \query^*&=\argmin_{\query} \mathbb{E}_{\queryResponse',\weights',\mixingCoefficient'\sim\queryResponse,\weights,\mixingCoefficient\mid \query,\belief^{i-1}}\log\frac{P(\queryResponse=\queryResponse'\mid \query, \belief^{i-1})}{P(\queryResponse=\queryResponse'\mid \query,\weights=\weights',\mixingCoefficient=\mixingCoefficient']}\\
    &=\argmin_{\query}\mathbb{E}_{\queryResponse',\weights',\mixingCoefficient'\sim\queryResponse,\weights,\mixingCoefficient\mid \query,\belief^{i-1}}\log\frac{\mathbb{E}_{\weights'',\mixingCoefficient''\sim\weights,\mixingCoefficient\mid\belief^{i-1}}P(\queryResponse=\queryResponse'\mid \query, \weights = \weights'', \mixingCoefficient=\mixingCoefficient'']}{P(\queryResponse=\queryResponse'\mid \query,\weights=\weights',\mixingCoefficient=\mixingCoefficient']}\:. \tag{see~\ref{eq:04_06_corl21_ranking_objective}}
\end{align*}

%% file: 99_appendix/03_implementation_details.tex
\section{Metropolis-Hastings for Section~\ref{sec:04_06_rankings}} \label{app:99_03_corl21_ranking_metropolis_hastings}

\begin{figure}[ht]
    \centering
    \includegraphics[width=0.6\linewidth]{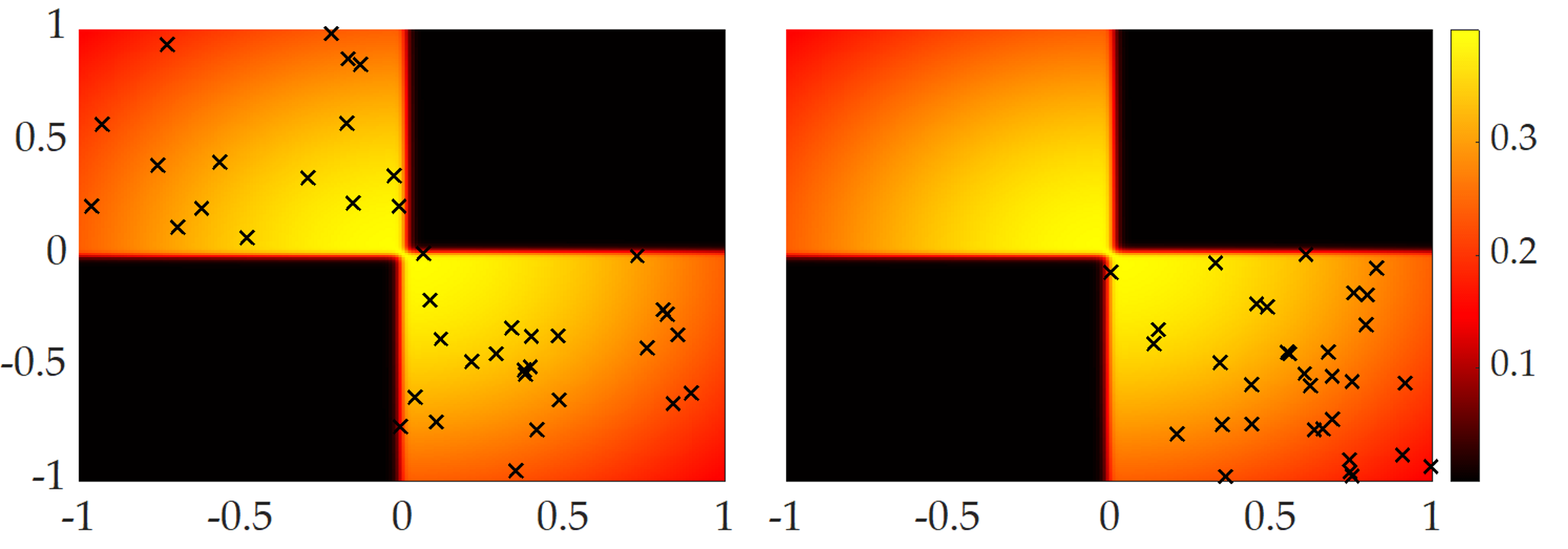}
    \caption{Multi-chain Metropolis-Hastings sampling \textbf{(left)} gives more representative samples from the distribution compared to the single-chain variant \textbf{(right)}.}
    \label{fig:99_03_corl21_ranking_MH_sampling}
\end{figure}

To sample from $P(\weights,\mixingCoefficient\mid\belief^{i-1})$ using Equation~\eqref{eq:03_05_bayesian_learning}, we use the Metropolis-Hastings algorithm \cite{chib1995understanding}, running $N_{\textrm{MH}}$ chains simultaneously for $H_{\textrm{MH}}$ iterations.
To avoid autocorrelation between samples, unlike in conventional Metropolis-Hastings we only use the last state in each chain as a sample.
In contrast, for conventional Metropolis-Hastings, multiple samples would be drawn from a single chain at set intervals after a short \emph{burn-in} period. As we see in Fig.~\ref{fig:99_03_corl21_ranking_MH_sampling}, for our multimodal Plackett-Luce posteriors, performing multi-chain Metropolis-Hastings yields posterior samples that are far more evenly distributed across different posterior modes. Thus, to achieve well-distributed posterior samples, we set our effective burn-in period to be $H_{\textrm{MH}}-1$, taking only the last sample from each chain.

For two states in the chain $\weights,\mixingCoefficient$ and $\weights',\mixingCoefficient'$, our proposal distribution is then $$P_{\textrm{MH}}(\weights',\mixingCoefficient'\mid\weights,\mixingCoefficient) = \prod_{\modeIndex=1}^\numberOfModes \standardNormalPdf_{\textrm{MH}}(\weights_\modeIndex - \weights'_\modeIndex),$$ where $\standardNormalPdf_{\textrm{MH}}$ is the pdf of the zero-mean Gaussian with the covariance matrix $\sigma_{\textrm{MH}}^2 I$.

The posterior distribution in Figure~\ref{fig:99_03_corl21_ranking_MH_sampling} is that of a 2-mode Plackett-Luce mixture with fixed uniform mixing coefficients and 1-D weights conditioned on the observations $50\succ -50$ and $-50\succ 50$. The single-chain algorithm ran for 2000 steps with a burn-in period of 200 steps after which every $18^\textrm{th}$ sample was selected, while the multi-chain algorithm used 100 chains for 20 iterations each, taking only the last sample from each chain.

\section{Simulated Annealing for Section~\ref{sec:04_06_rankings}}\label{app:99_03_corl21_ranking_simulated_annealing}
For our simulated annealing, we run $N_{\textrm{SA}}$ chains in parallel for $H_{\textrm{SA}}$ iterations each, returning the best query $\query$ found across each run. We define the transition proposal distribution $P_{\textrm{SA}}(\query'\mid \query)$ to be a positive constant if $\query'$ and $\query$ differ by one trajectory and $0$ otherwise. We run with a starting temperature of $T^0_{\textrm{SA}}$, cooling by a factor of $\gamma_{\textrm{SA}}$ with each subsequent iteration past the first.

\section{Hyperparameters for Section~\ref{sec:04_06_rankings}}
\label{app:99_03_corl21_ranking_hyperparameters}

We use the hyperparameters in Table~\ref{table:99_03_corl21_ranking_hyperparameters} for the simulated annealing and Metropolis-Hastings algorithms, whose details are provided in Appendix~\ref{app:99_03_corl21_ranking_metropolis_hastings} and Appendix~\ref{app:99_03_corl21_ranking_simulated_annealing}, respectively.
\begin{table}[H]
\caption{Hyperparameters}
\smallskip
\begin{displaymath}
\begin{array}{cc}
    \toprule
    \makebox[5em]{Constant} & \makebox[5em]{Value} \\ \midrule
    \rule{0pt}{2ex} 
     N_{\textrm{MH}} & 100 \\
     H_{\textrm{MH}} &  200 \\
     \sigma_{\textrm{MH}} & 0.15 \\
     N_{\textrm{SA}} & 10 \\
     H_{\textrm{SA}} & 30 \\
     T^0_{\textrm{SA}} & 10 \\
     \gamma_{\textrm{SA}} & 0.9\\\bottomrule
\end{array}
\end{displaymath}
\label{table:99_03_corl21_ranking_hyperparameters}
\end{table}

\section{Hyperparameter Tuning for DPPs in Section~\ref{subsec:04_08_experiments}}\label{app:99_03_ijrr_dpp_hyperparameters}
We introduced the hyperparameters $\diversityParameter$, $\similarityMatrixHyperparameter$ and $\scoreParameter$ for the DPP-based method. However, using the mode of the DPP distribution as the batch eliminates $\diversityParameter$, as it does not affect the results unless trivially $\diversityParameter=0$. Hence, we need to tune $\similarityMatrixHyperparameter$ and $\scoreParameter$ only.

As $\scoreParameter$ is enough in our proposed algorithm to adjust the trade-off between diversity and high volume removal, we use the following heuristic for setting $\similarityMatrixHyperparameter$ to avoid extra computational burden. We simply set $\similarityMatrixHyperparameter$ to be the expected distance between two nearest points (in terms of Euclidean distance) when $\batchSize$ points are selected uniformly at random in the space $[0,1]^\numberOfFeatures$ where $\numberOfFeatures$ is the number of features, i.e., $\numberOfFeatures=\dim(\trajectoryFeaturesFunction(\trajectory))$.

For a human user and a given dynamical system, we cannot try different hyperparameter values, because we need to query the human many times to get the responses under different hyperparameters. Therefore, to perform tuning for and $\scoreParameter$, we simulate $100$ \emph{synthetic true reward weights} separately for each environment.

\begin{figure}[ht]
	\centering
	\includegraphics[width=0.5\textwidth]{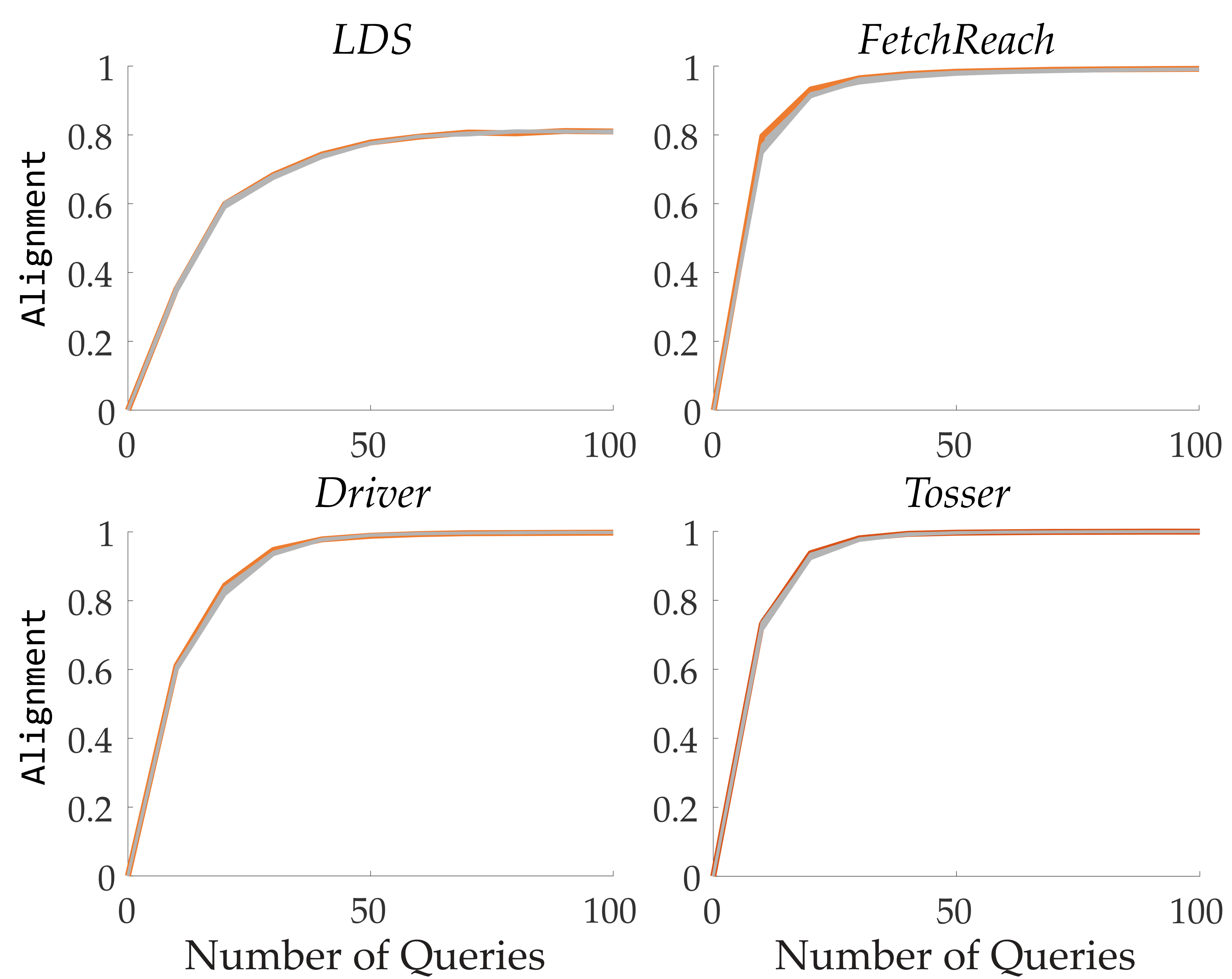}
	\caption{Tuning results for the DPP-based method for various $\scoreParameter$ under each environment.}
	\label{fig:99_03_ijrr_dpp_tuning}
\end{figure}

We tuned $\scoreParameter$ separately for our experiment environments \emph{LDS}, \emph{FetchReach}, \emph{Tosser} and \emph{Driver} where each $\scoreParameter$ has been experimented with $100$ different synthetic true reward functions.

Figure~\ref{fig:99_03_ijrr_dpp_tuning} shows how the \texttt{Alignment} value changes with different number of queries for varying $\scoreParameter$. We highlighted the selected $\scoreParameter$ parameters in the plots.

As it can be seen from the results, the effect of $\scoreParameter$ on performance was slight for the environments we experimented on. It is therefore difficult to select the ``best" $\scoreParameter$. We qualitatively selected $\scoreParameter=1$ for \emph{LDS} and \emph{Tosser}, $\scoreParameter=4$ for \emph{FetchReach}, and $\scoreParameter=0$ for \emph{Driver} based on their slight advantage in learning rate with respect to the number of queries. While it is possible that other environments have heavier dependencies on $\scoreParameter$, our results empirically suggest that the method is robust to the choice of $\scoreParameter$, which can ease the use of our DPP-based batch generation algorithm.

%% file: 99_appendix/04_experiment_details.tex
\section{Environment Features for Sections~\ref{subsec:04_01_experiments} and \ref{subsec:04_02_experiments}}\label{app:99_04_ijrr_features}
\subsection{\emph{FetchReach}}
We present the full set of features below. In Section~\ref{subsec:04_01_experiments}, we used the first three of these features whereas Section~\ref{subsec:04_02_experiments} uses all features to make it a more difficult problem.
    \begin{itemize}[nosep]
    	\item The average of $e^{-c_1d_1}$ over the trajectory where $d_1$ is the distance between the end effector and the goal object, and $c_1=1$.
		\item The average of $e^{-c_2d_2}$ over the trajectory where $d_2$ is the vertical distance between the end effector and the table, and $c_2=1$.
		\item The average of $e^{-c_3d_3}$ over the trajectory where $d_3$ is the distance between the end effector and the obstacle, and $c_3=1$.
		\item The average of the end effector speed over the trajectory.
    \end{itemize}

\subsection{\emph{Driver}}\label{app:99_04_ijrr_features_driver}
    \begin{itemize}[nosep]
    	\item The average of $e^{-c_4d_4^2}$ over the trajectory, where $d_4$ is the shortest distance between the ego car and a lane center, and $c_4=30$.
    	\item The average of $(v_1-1)^2$ over the trajectory, where $v_1$ is the speed of the ego car.
    	\item The average of $\cos(\theta_1)$ over the trajectory, where $\theta_1$ is the angle between the directions of ego car and the road.
    	\item The average of $e^{-c_5d_5^2-c_6d_6^2}$ over the trajectory, where $d_5$ and $d_6$ are the horizontal and vertical distances between the ego car and the other car, respectively; and $c_5=7$, $c_6=3$.
    \end{itemize}

\subsection{\emph{Tosser}}
    \begin{itemize}[nosep]
    	\item The maximum distance the object moved forward from the tosser robot.
    	\item The maximum altitude of the object.
    	\item Number of flips (real number) the object does.
    	\item $e^{-c_7d_7}$ where $d_7$ is the final horizontal distance between the object and the center of the closest basket, and $c_7=3$.
    \end{itemize}

\section{Environment Features for Section~\ref{subsec:04_05_corl21_scale_experiments}}\label{app:99_02_corl21_scale_details}

Here, we describe the features of the simulation and user study environments we used. These environments are: \emph{ExtendedDriver}, which we used for the simulations in Section~\ref{subsec:04_05_corl21_scale_experiments}, original \emph{Driver}, which was used in Section~\ref{subsec:04_02_experiments}, and we present the results in Appendix~\ref{app:99_05_corl21_scale_additional_results}, and finally Fetch robot experiment with drink serving (\emph{FetchDrink}), which we used for the user studies in Section~\ref{subsec:04_05_corl21_scale_study}.

\subsection{\emph{ExtendedDriver}}
In Table~\ref{tab:03_06_features_driver_extended} we detail the features of the \emph{ExtendedDriver} scenarios. Notation specific to Table~\ref{tab:03_06_features_driver_extended}: $d_8, d_9, d_{10}$ are the squared distances of the robot car to the center of the left, middle and right lane; $\mathbf{v}$ is the speed profile of the robot trajectory; $\dot{\mathbf{v}}$ the acceleration profile; $\theta$ is the heading of the car, 
${x}(\timestep)$ and ${y}(\timestep)$ are the car's $x$ and $y$ position at a given time $\timestep\in[0,\horizon]$ ($x$ is orthogonal to the road, $y$ is along the road);
$\Delta \mathbf{x}$ and $\Delta \mathbf{y}$ are the ordinal distance between the robot car and the other car; and $c_8, c_9, c_{10}$ are the $x$-coordinates of the lane centers.

\begin{table}
\caption{Features of the \emph{ExtendedDriver} Environment}
\label{tab:03_06_features_driver_extended}
\begin{tabular}{lll}\toprule

           &Description  & Definition\\\midrule
$\trajectoryFeaturesFunction_1$    & Lane keeping: mean distance to closest lane center & $ \frac{\mathrm{mean}[\exp(-30\cdot\min\{d_8, d_9, d_{10}\})] }{0.15343634}$
 \\
 $\trajectoryFeaturesFunction_2$    & Keep speed: mean difference to speed $1$&
 $ \frac{\mathrm{mean}[(1-\mathbf{v})^2] }{ 0.42202643}$
 \\
 $\trajectoryFeaturesFunction_3$    & Driving straight: mean heading $\theta$ &
 $ \frac{\mathrm{mean}[\theta] }{ 0.06112367}$
 \\
  $\trajectoryFeaturesFunction_4$    & Collision avoidance 1: mean distance to other car &
 $ \frac{\mathrm{mean}[\exp(-7\cdot\Delta \mathbf{x}^2)+3\cdot\Delta \mathbf{y}^2] }{0.15258019}$
 \\
   $\trajectoryFeaturesFunction_5$    & Collision avoidance 2: min distance to other car &
 $ \frac{\min[\exp(-7\cdot\Delta \mathbf{x}^2)+3\cdot\Delta \mathbf{y}^2] }{0.10977646}$
 \\
   $\trajectoryFeaturesFunction_6$    & Smoothness: mean jerk &
 $ \frac{\mathrm{mean}[\Delta\dot{\mathbf{v}}] }{0.00317041}$
 \\ 
 
    $\trajectoryFeaturesFunction_7$    & Distance travelled: progress along the road&
 $ \frac{{y}(\horizon)-{y}(0) }{1.01818467}$
 \\ 
  $\trajectoryFeaturesFunction_8$    & Final lane L: robot end in the left lane&
 $\mathrm{int}(\norm{x(\horizon)-c_8} <0.08 )$
 \\ 
  $\trajectoryFeaturesFunction_9$    & Final lane M: robot end in the center lane&
 $\mathrm{int}(\norm{x(\horizon)-c_9} <0.08 )$
 \\ 
  $\trajectoryFeaturesFunction_{10}$    & Final lane R: robot end in the right lane&
 $\mathrm{int}(\norm{x(\horizon)-c_{10}}|| <0.08 )$\\
\bottomrule
\end{tabular}
\end{table}

\subsection{Original \emph{Driver}}
See Appendix~\ref{app:99_04_ijrr_features_driver}.

\subsection{\emph{FetchDrink}}
In the user studies presented in Section~\ref{subsec:04_05_corl21_scale_study} and the simulations presented in Appendix~\ref{app:99_05_corl21_scale_additional_results}, we used the following eight features for the \emph{FetchDrink} robot experiment:
\begin{itemize}[nosep]
    \item Speed of the end-effector $\in \{0, 0.33, 0.67, 1\}$
    \item Maximum height of the end-effector $\in \{0, 0.33, 0.67, 1\}$
    \item Selected drink being the orange juice $\in\{0,1\}$
    \item Selected drink being the water $\in\{0,1\}$
    \item Selected drink being the milk $\in\{0,1\}$
    \item Orientation of the pan $\in\{0,1\}$
    \item Moving the drink behind or over the pan $\in\{0,1\}$
    \item Robot hitting the pan while moving the drink $\in\{0,1\}$
\end{itemize}

\section{Choice of $\scaleNoiseStd$ in the User Studies for Section~\ref{subsec:04_05_corl21_scale_experiments}}\label{app:99_02_corl21_scale_sigma_selection}
In Section~\ref{subsec:04_05_corl21_scale_study}, we stated we took $\scaleNoiseStd=0.35$ in the user studies based on pilot trials with different users. We now describe the procedure that yielded this selection of $\scaleNoiseStd$.

Before all the actual experiments, we recruited 3 participants (3 male, ages 27--40) for a pilot study. In this study, the participants followed the same procedure as in our actual experiments, but responded to only $30$ randomly generated queries. These $30$ queries were formed by three sets: $10$ scale queries, $10$ weak comparison queries and another $10$ scale queries. We randomized the order of these three sets to avoid any bias.

After we collected these data, we repeated the following procedure for $\scaleNoiseStd=0.05, 0.10, \dots, 1.00$. We learned a single posterior for each user by using $10$ scale and $10$ weak comparison query responses under $\scaleNoiseStd$ noise, i.e., the posteriors included both scale and soft choice feedback. We then checked the test set loglikelihood (with the remaining $10$ queries) under the learned posterior and the same $\scaleNoiseStd$.

The $\scaleNoiseStd$ value that yielded the highest test set loglikelihood, $\scaleNoiseStd=0.35$, was then used for all of the actual experiments with real users.

\section{Baselines for Section~\ref{sec:04_06_experiments}} \label{app:99_04_corl21_ranking_baselines}
\subsection{Random}
\label{app:99_04_corl21_ranking_random}
We benchmark against a random agent, wherein at each step the query selected by the agent is a collection of $\abs{\query}$ random items without replacement. We also use the random querying method for comparing the multimodal reward learning with the approaches that assume a unimodal reward (as in Section~\ref{sec:04_02_information_gain}), as it does not introduce any bias in the query selection.
\subsection{Volume Removal}
\label{app:99_04_corl21_ranking_volume}
Volume removal seeks to maximize the difference between the prior distribution over model parameters and the \textit{unnormalized} posterior. Volume removal notably fails to be optimal in domains where there are similar trajectories as we showed in Section~\ref{sec:04_01_volume_removal}. In these settings, querying sets of trajectories with similar features removes a large amount of volume from the unnormalized posterior (since the robot is highly uncertain about their relative quality), yet yields little information about the model parameters (since the human also has high uncertainty). Mutual information based approaches are better able to generate trajectories to query for which the robot has high uncertainty while the human has enough certainty to yield useful information for the robot.

\section{Trajectory Generation in Section~\ref{sec:04_06_experiments}}\label{app:99_04_corl21_ranking_traj}
\subsection{\emph{LunarLander} Trajectories}\label{app:99_04_corl21_ranking_lunar_traj}
\begin{figure}[H]
    \centering
    \begin{minipage}{.45\linewidth}
    \centering
    \includegraphics[width=\linewidth]{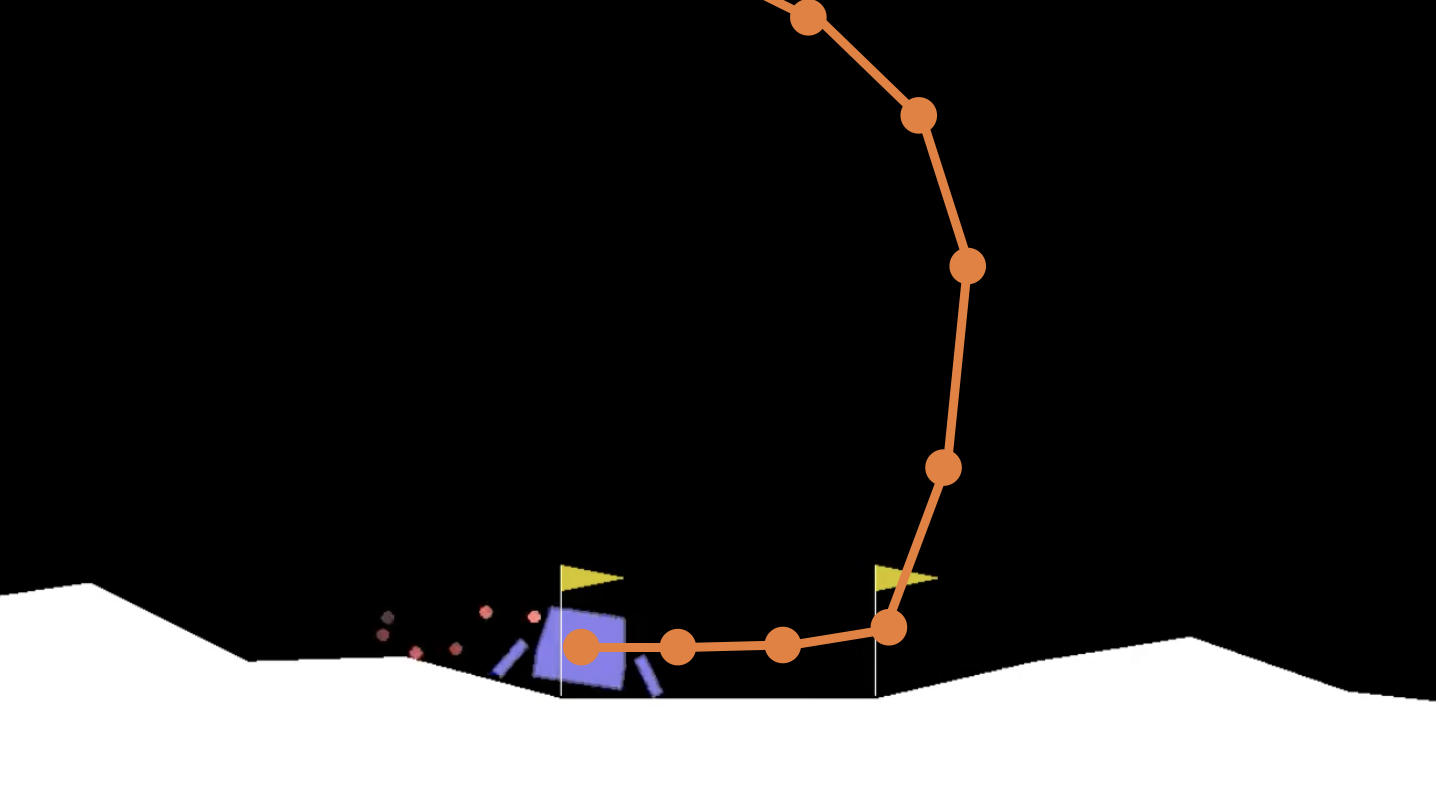}
    \end{minipage}\hskip 2em\begin{minipage}{.45\linewidth}
    \centering
    \begin{tabular}{cr}
    \toprule
         Feature & \multicolumn{1}{c}{Value} \\
     \midrule
        Mean angle & $2.27683634$   \\
        Total angle & $-0.20375356$ \\
        Distance to goal & $5.41860642$ \\
        Total rotation & $0.25948072$ \\
        Path length & $3.71660086$ \\
        Final vertical velocity & $-0.57097337$ \\
        Crash time & $1.11112885$ \\
        Score & $-0.15500268$ \\
     \bottomrule
    \end{tabular}
    \end{minipage}
    \caption{Sample \emph{LunarLander} trajectory (left) with extracted features (right).}
    \label{fig:99_04_corl21_ranking_lunartraj}
\end{figure}

We designed $8$ trajectory features based on: absolute heading angle accumulated over trajectory, final distance to the landing pad, total amount of rotation, path length, task completion (or failure) time, final vertical velocity, whether the lander landed on the landing pad without its body touching the ground, and original environment reward from OpenAI Gym \cite{brockman2016openai}. Using these features, we randomly generated $10$ distinct reward functions based on the linear reward model and trained a DQN policy \cite{mnih2013playing} for each reward. Finally, we generated $100$ trajectories by following each of these $10$ policies in the environment to obtain $1000$ trajectories in total. We used these trajectories as our dataset for the ranking queries. Figure~\ref{fig:99_04_corl21_ranking_lunartraj} presents an example trajectory with extracted, scaled and centered features.

\subsection{\emph{FetchBanana} Trajectories}\label{app:99_04_corl21_ranking_fetch_traj}
To design our $351$ trajectories, we varied the target shelf (3 variations), the movement speed (3), the grasp point on the banana (3) and where in the shelf it is placed (13). We then designed $12$ trajectory features based on these varied parameters and appended another binary feature which indicates whether any object dropped from the shelves on that trajectory.

Specifically, for a trajectory $\trajectory$, let \begin{align*}
y_{\textrm{target},i} = \begin{cases}
1 & i \text{ is the target shelf}\\
0 & \text{otherwise}
\end{cases}\:,
\end{align*}

\begin{figure}[H]
    \centering
    \begin{minipage}{.45\linewidth}
    \centering
    \includegraphics[width=\linewidth]{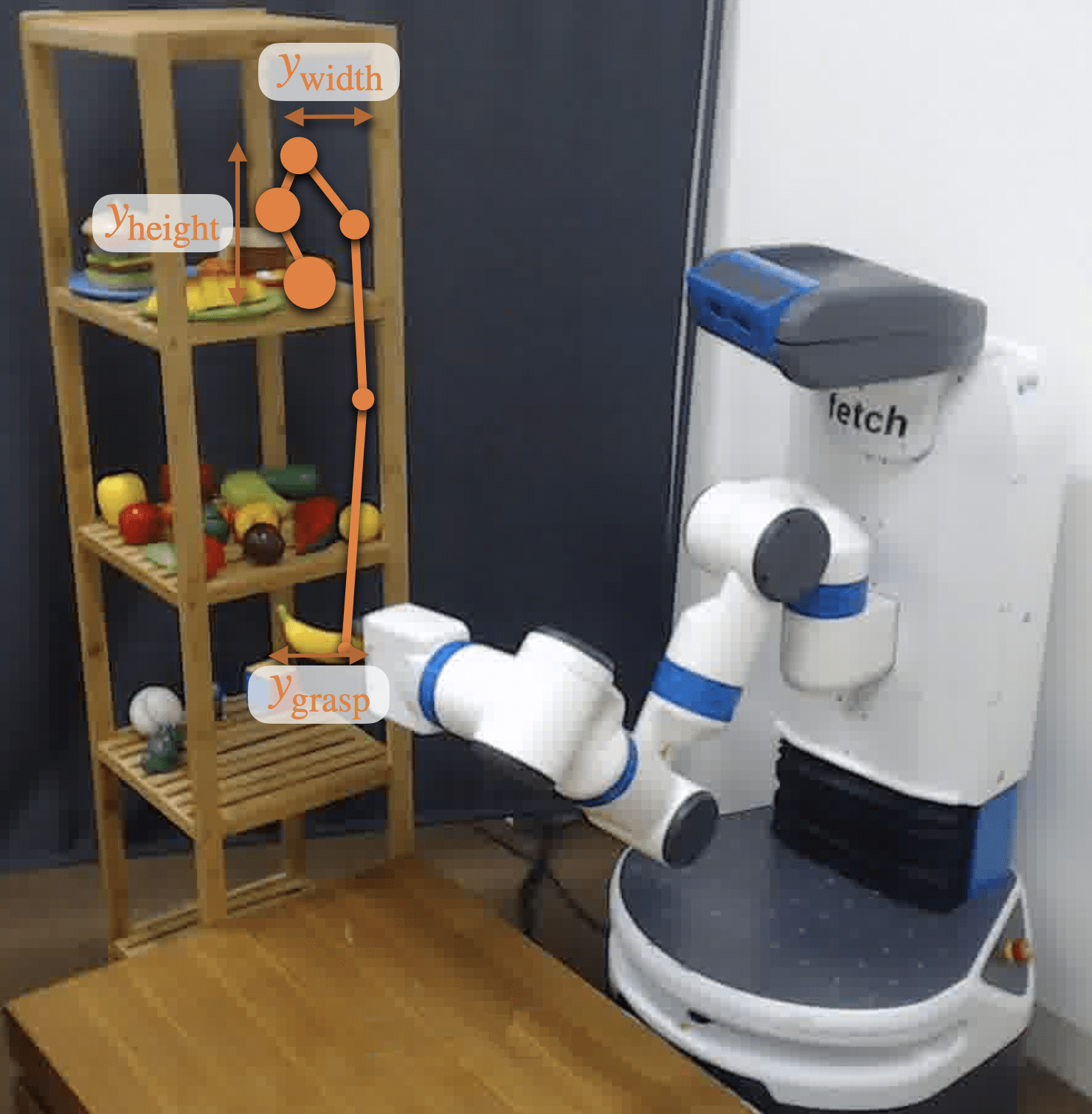}
    \end{minipage}\hskip 2em\begin{minipage}{.45\linewidth}
    \centering
    \begin{tabular}{cr}
    \toprule
         Feature & \multicolumn{1}{c}{Value} \\
     \midrule
        $y_{\textrm{target},1}$ & 1   \\
        $y_{\textrm{target},2}$ & 0 \\
        $y_{\textrm{target},3}$ & 0 \\
        $y_{\textrm{speed}}$ & 0.5\\
        $y_{\textrm{speed}}(1-y_{\textrm{speed}})$ & 0.25\\
        $y_{\rm grasp}$ & 1 \\
        $y_{\rm grasp}(1-y_{\rm grasp})$ & 0 \\
        $y_{\textrm{height}}$ & 0.75 \\
        $y_{\textrm{height}}(1-y_{\textrm{height}})$ & 0.1875 \\
        $y_{\textrm{width}}$ & 0.25 \\
        $y_{\textrm{width}}(1-y_{\textrm{width}})$ & 0.1875 \\
        $1-(y_{\textrm{grasp}} - y_{\textrm{width}})^2$ & 0.4375\\
        $y_{\textrm{success}}$ & 1\\
     \bottomrule
    \end{tabular}
    \end{minipage}
    \caption{Sample \emph{FetchBanana} trajectory (left) with extracted features (right).}
    \label{fig:99_04_corl21_ranking_fetchtraj}
\end{figure}

$y_{\text{grasp}},y_{\text{height}},y_{\text{width}},y_{\text{speed}}$ specify the grasp position and speed, and $y_{\text{success}}$ specifies whether the robot did not drop any objects from the shelves. Our featurization is then
\begin{align*}
\trajectoryFeaturesFunction(\trajectory)&=\big( y_{\textrm{target},1},y_{\textrm{target},2},y_{\textrm{target},3},y_{\text{speed}},y_{\text{speed}}(1-y_{\text{speed}}),y_{\text{grasp}},y_{\text{grasp}}(1-y_{\text{grasp}}),y_{\text{height}},\\&\qquad y_{\text{height}}(1-y_{\text{height}}),y_{\text{width}}, y_{\text{width}}(1-y_{\text{width}}),1-(y_{\text{grasp}}-y_{\text{width}})^2,y_{\text{success}} \big) .
\end{align*}
Figure~\ref{fig:99_04_corl21_ranking_fetchtraj} presents a sample \emph{FetchBanana} trajectory with its featurization.

\section{Metrics in Section~\ref{sec:04_06_experiments}}\label{app:metrics}
\subsection{\texttt{MSE}}
\label{app:99_04_corl21_ranking_mse}
Our metric is \begin{equation}
    \label{eq:99_04_corl21_ranking_mse}
    \texttt{MSE} = \sum_{\modeIndex=1}^\numberOfModes \abs{\weights_\modeIndex^* - \hat\weights_\modeIndex}^2_2
\end{equation}
where the learned reward weights of the experts are matched with the true weights using the Hungarian algorithm. When the learning model assumes a unimodal reward function, as in our simulations for Figure~\ref{fig:04_06_unimodal_vs_bimodal}, we compute the \texttt{MSE} metric as $\sum_{\modeIndex=1}^\numberOfModes \abs{\weights_\modeIndex^*-\hat\weights}^2_2$.

\subsection{\texttt{Log-Likelihood}}
\label{app:99_04_corl21_ranking_ll}
Formally, we define the \texttt{Log-Likelihood} metric as
\begin{equation}
    \label{eq:99_04_corl21_ranking_ll}
    \texttt{Log-Likelihood} = \mathbb{E}_{{\query\sim\mathcal{\query}}}\left[\mathbb{E}_{\queryResponse'\sim \queryResponse\mid \query}\log P(\queryResponse=\queryResponse'\mid \mathcal{\query},\belief^{i-1})\right]
\end{equation} 
for $\mathcal{\query}$ the uniform distribution across all possible queries and $P(\queryResponse\mid \query)$ the distribution over the human's response to query $\query$ (as in Equation~\eqref{eq:03_05_pmf}). We can compute the inner term 
\begin{align*}
    P(\queryResponse\mid \query,\belief^{i-1}) &= \mathbb{E}_{\weights',\mixingCoefficient'\sim\weights,\mixingCoefficient\mid \belief^{i-1}} [P(\queryResponse\mid \query,\weights=\weights',\mixingCoefficient=\mixingCoefficient')]
\end{align*}
using Metropolis-Hastings as in Section~\ref{subsec:04_06_overall_algorithm} to sample from the posterior $P(\weights,\mixingCoefficient\mid \belief^{(i-1)})$ and computing the inner term with Equation~\eqref{eq:03_05_pmf}.

\subsection{\texttt{Learned Policy Reward}}
\label{app:99_04_corl21_ranking_learned_policy_rewards}
Similar to the \texttt{MSE} metric, we match the rewards learned via DQN \cite{mnih2013playing} with the true rewards using the Hungarian algorithm.

\section{Experimental Setup in Section~\ref{sec:04_06_experiments}}
\label{app:expset}

\subsection{Shelf Descriptions for \emph{FetchBanana} Environment}
\label{app:99_04_corl21_ranking_desc}
A picture of each shelf accompanied the following descriptions. 
\begin{itemize}[nosep]
    \item The top shelf has some space, but you usually put cooked meals there.
    \item The middle shelf is for fruits, but it is already full. The robot may accidentally drop other fruits.
    \item The bottom shelf has a lot of space, but you have been using it for toys.
\end{itemize}

\subsection{User Interface}
\label{app:99_04_corl21_ranking_ui}
\begin{figure*}[t]
    \centering
    \includegraphics[width=\textwidth]{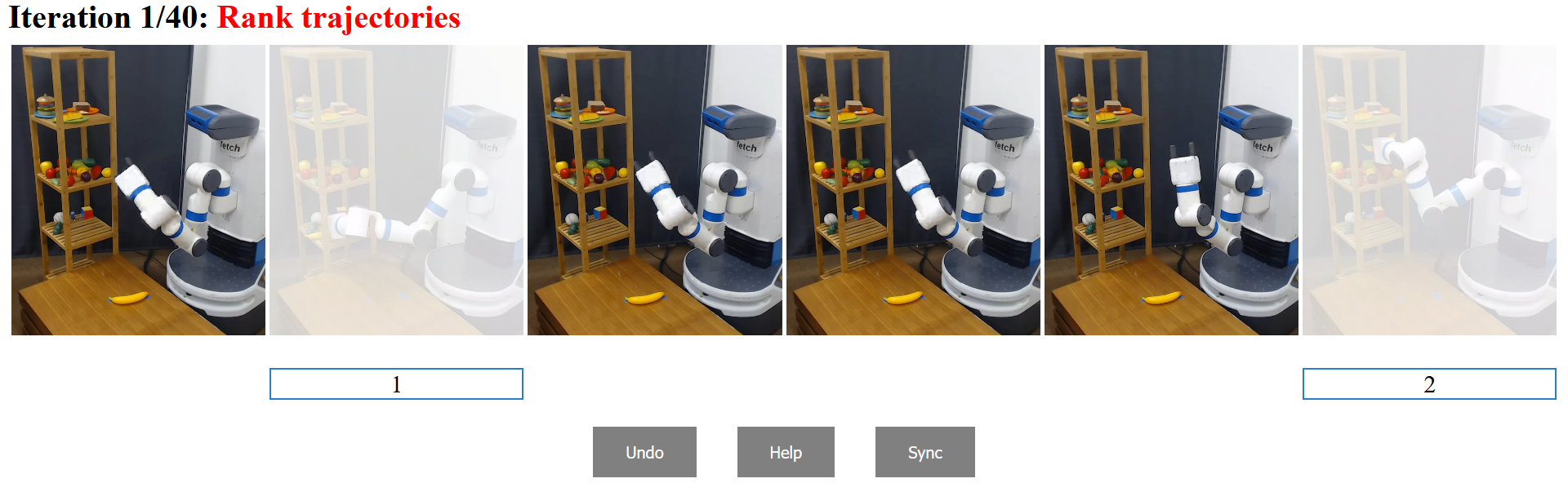}
    \caption{The user interface for the online studies with the real Fetch robot (\emph{FetchBanana} environment). The user selected the $2^{\textrm{nd}}$ trajectory as their top choice and the $6^{\textrm{th}}$ trajectory as the second top.}
    \label{fig:99_04_corl21_ranking_fetch_ui}
\end{figure*}

For both environments, subjects were told they need to rank the six trajectories in each query by clicking on the trajectories starting from the most preferred to the least. The web interface (see Figure~\ref{fig:99_04_corl21_ranking_fetch_ui}) equipped them with ``Undo" and ``Sync" buttons. ``Undo" allowed the subjects to undo a selection they make within a query. ``Sync" enabled them to restart all videos in the query.

%% file: 99_appendix/05_additional_results.tex
\section{Additional Simulation Results for Section~\ref{subsec:04_02_experiments}}\label{app:99_05_ijrr_active_additional_results}
\subsection{Results with User-Specific and Unknown $\minimumPerceivableDifference$}\label{app:99_05_ijrr_active_unknown_delta}
\begin{figure*}[tbh]
	\centering
	\includegraphics[width=\textwidth]{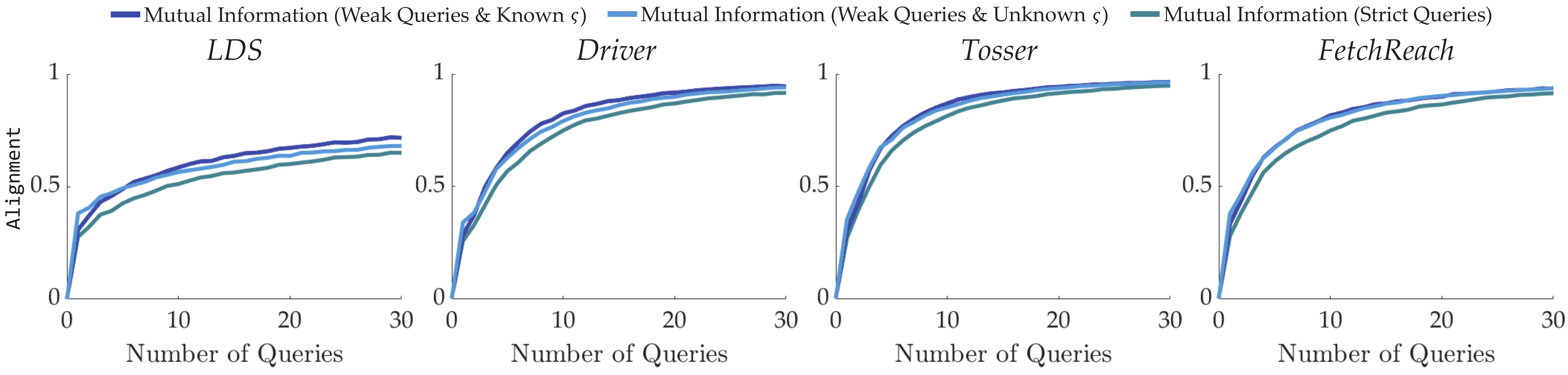}
	\caption{The simulation results with mutual information formulation for unknown $\minimumPerceivableDifference$. Plots are mean$\pm$s.e.}
	\label{fig:99_05_unknown_delta}
\end{figure*}

Using the approximate, but computationally faster optimization we introduced in Appendix~\ref{app:99_02_ijrr_active_unknown_parameter}, we performed additional analysis where we compare the performances of strict pairwise comparison queries, weak pairwise comparison queries with known $\minimumPerceivableDifference$ and weak pairwise comparison queries without assuming any $\minimumPerceivableDifference$ (all with the mutual information formulation). As in the previous simulations, we simulated $100$ users with different random reward functions. Each user is simulated to have a true $\minimumPerceivableDifference$, uniformly randomly taken from $[0,2]$. During the sampling of $\weightsSampleSet^+$, we did not assume any prior knowledge about $\minimumPerceivableDifference$, except the natural condition that $\minimumPerceivableDifference\geq 0$. The comparison results are in Figure~\ref{fig:99_05_unknown_delta}. While knowing $\minimumPerceivableDifference$ increases the performance as expected, weak pairwise comparison queries are still better than strict queries even when $\minimumPerceivableDifference$ is unknown. This supports the advantage of employing weak pairwise comparison queries.

\subsection{Results without Query Space Discretization}\label{app:99_05_ijrr_active_no_discretization}

\begin{figure*}[tbh]
	\centering
	\includegraphics[width=\textwidth]{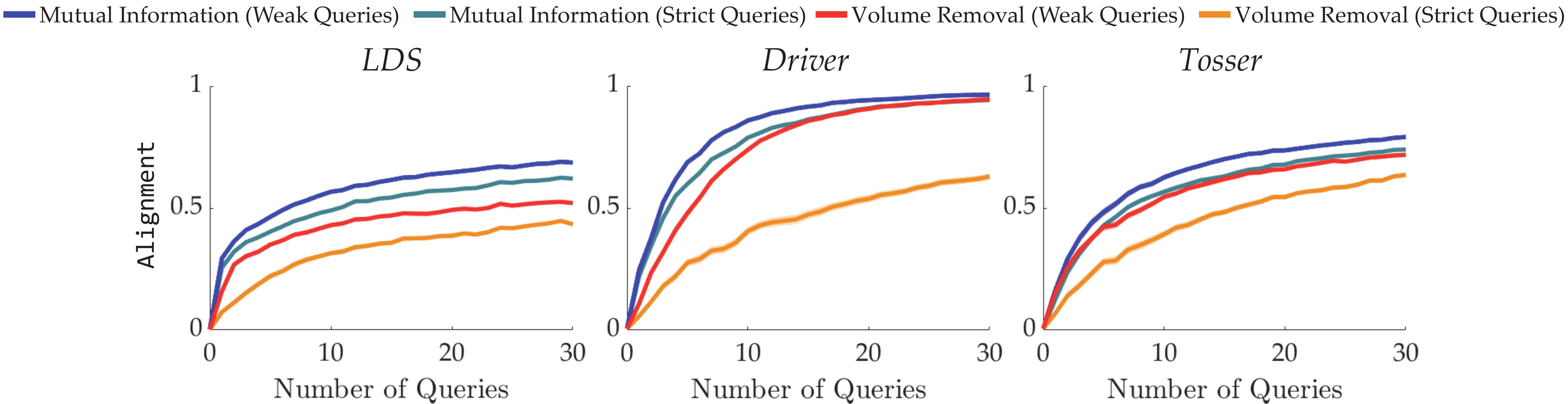}
	\caption{Alignment values are plotted (mean$\pm$s.e.) for the experiments without query space discretization, i.e., with continuous trajectory optimization for active query generation.}
	\label{fig:99_05_continuous_optimization}
\end{figure*}

We repeated the experiment that supports \textbf{H5}, and whose results are shown in Figure~\ref{fig:04_02_simulation_results_m}, without query space discretization. By optimizing over the continuous action space of the environments, we tested mutual information and volume removal formulations with both strict and weak pairwise comparison queries in \emph{LDS}, \emph{Driver} and \emph{Tosser} tasks. We excluded \emph{FetchReach} again in order to avoid prohibitive trajectory optimization due to large action space. Figure~\ref{fig:99_05_continuous_optimization} shows the results. As it is expected, mutual information formulation outperforms the volume removal with both pairwise comparison query types. And, weak pairwise comparison queries lead to faster learning compared to strict pairwise comparison queries.

\subsection{Effect of Information from ``About Equal" Responses}\label{app:99_05_ijrr_active_value_of_idk}
We have seen that weak pairwise comparison queries consistently decrease wrong answers and improve the performance. However, this improvement is not necessarily merely due to the decrease in wrong answers. It can also be credited to the information we acquire thanks to ``About Equal" responses.

\begin{figure*}[tbh]
	\centering
	\includegraphics[width=\textwidth]{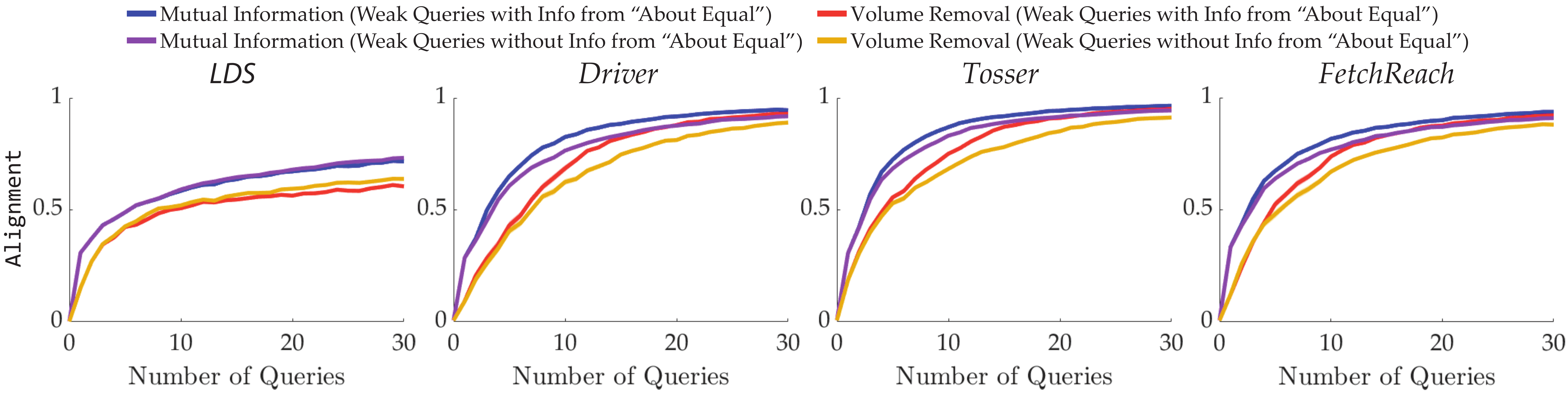}
	\caption{The results (mean$\pm$s.e.) of the simulations with weak pairwise comparison queries where we use the information from ``About Equal" responses (blue and red lines) and where we do not use (purple and orange lines).}
	\label{fig:99_05_value_of_idk}
\end{figure*}

To investigate the effect of this information, we perform two additional experiments with $100$ different simulated human reward functions with weak pairwise comparison queries: First, we use the information by the ``About Equal" responses; and second, we ignore such responses and remove the query from the query set to prevent repetition. Figure~\ref{fig:99_05_value_of_idk} shows the results. It can be seen that for both volume removal and mutual information formulations, the information from ``About Equal" option improves the learning performance in \emph{Driver}, \emph{Tosser} and \emph{FetchReach} tasks, whereas its effect is very small in \emph{LDS}.

\subsection{Optimal Stopping under Query-Independent Costs}\label{app:99_05_ijrr_active_query_independent_stopping}

\begin{figure*}[tbh]
	\centering
	\includegraphics[width=\textwidth]{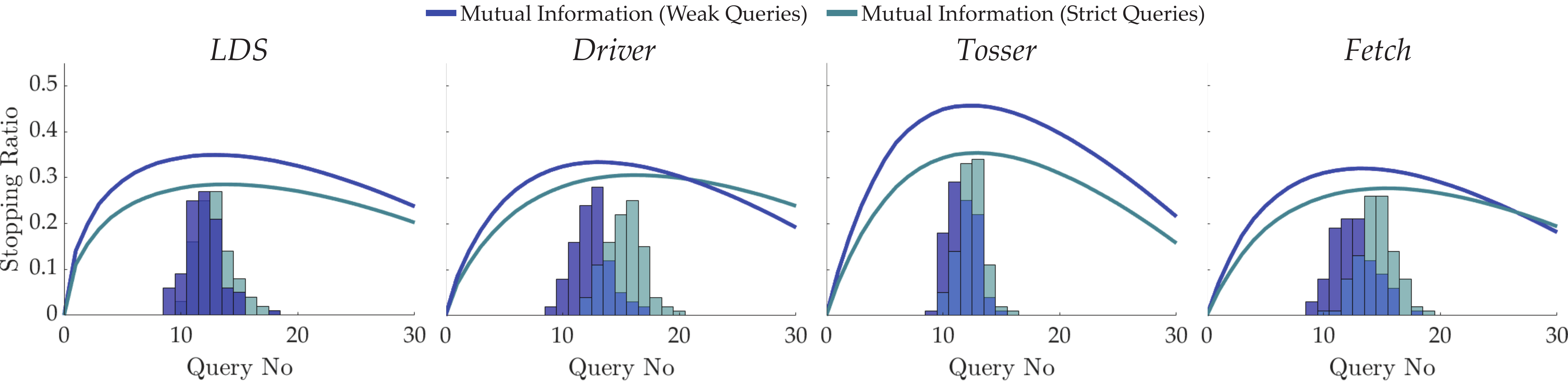}
	\caption{Simulation results for optimal stopping under query-independent costs. Line plots show cumulative active learning rewards (cumulative difference between the information gain values and the query costs), averaged over $100$ test runs and scaled for better appearance. Histograms show when optimal stopping condition is satisfied.}
	\label{fig:99_05_optimal_stopping_query_independent}
\end{figure*}

To investigate optimal stopping performance under query-independent costs, we defined the cost function as $\costFunction(\query) = \queryCostHyperparameter$, which just balances the trade-off between the number of questions and learning performance. Similar to the query-dependent costs case we described in Section~\ref{subsec:04_02_experiments}, we first simulate $100$ random users and tune $\queryCostHyperparameter$ accordingly in the same way. We then use this tuned $\queryCostHyperparameter$ for our tests with $100$ different random users. Figure~\ref{fig:99_05_optimal_stopping_query_independent} shows the results. Optimal stopping rule enables terminating the process with near-optimal cumulative active learning rewards in all environments, which again supports \textbf{H9}.

\section{Additional Simulation Results for Section~\ref{subsec:04_05_corl21_scale_experiments}}\label{app:99_05_corl21_scale_additional_results}
We present additional simulation results to compare the proposed scale feedback with weak pairwise comparisons.
For the \emph{ExtendedDriver} environment from Section~\ref{subsec:04_05_corl21_scale_experiments}, we additionally show data with higher noise, and show results with the log-likelihood measure used in the user study. Further, we show the same analysis for the original \emph{Driver} experiment, and for the simulated version of the \emph{FetchDrink} experiment from the user study.

For all the simulation results in this Appendix, we simulated $40$ different $\weights^*$, each with four different $\sensitivityThreshold^*\in\{.25, .5, .75, 1\}$, making $160$ runs in total. 

\subsection{\emph{ExtendedDriver}}
\begin{figure}[ht]
		\centering
		\begin{subfigure}[t]{0.49\textwidth}
        \includegraphics[width=1\textwidth]{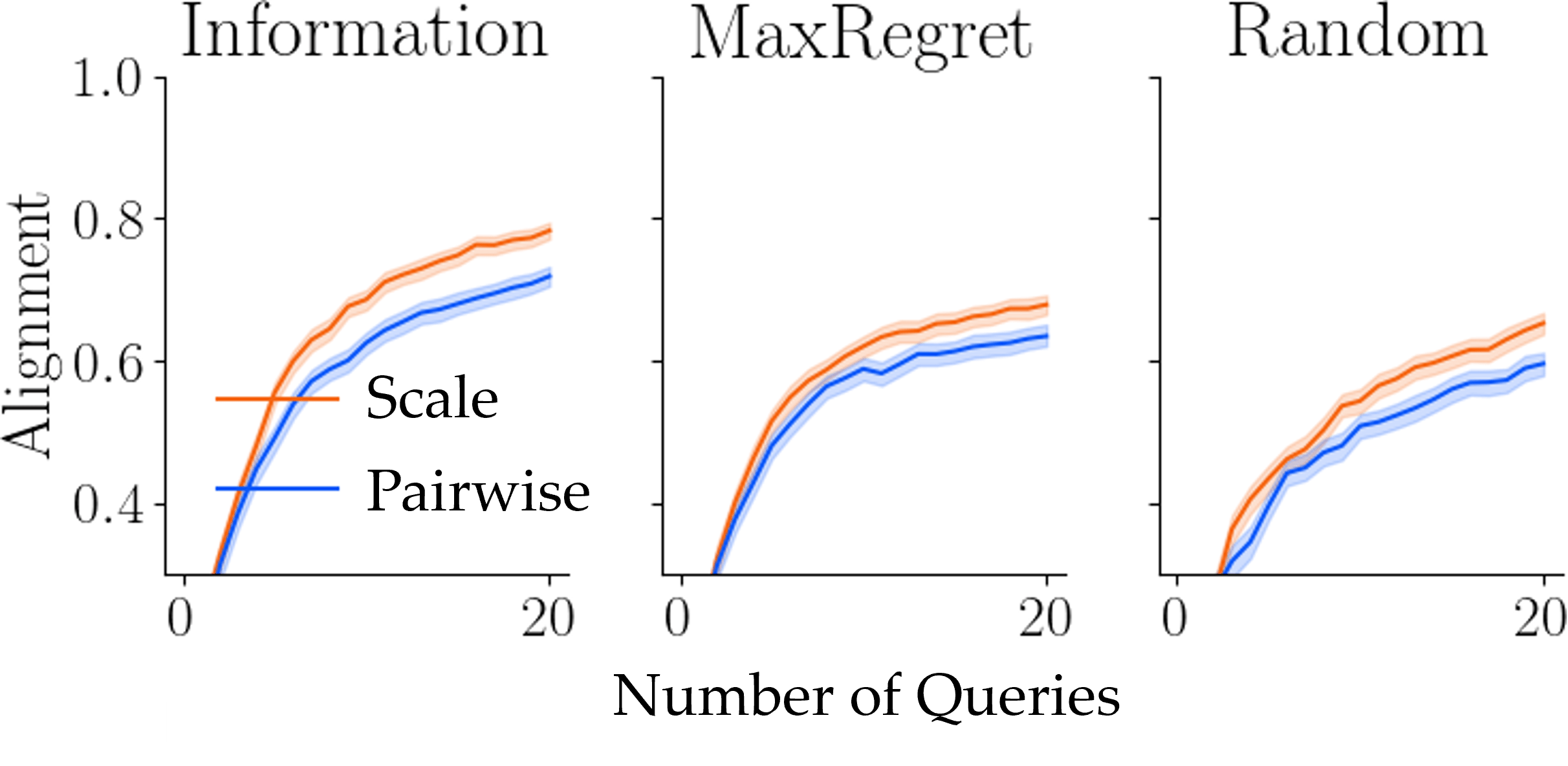}
		\end{subfigure}
		\hfill
		\begin{subfigure}[t]{0.49\textwidth}
        \includegraphics[width=1\textwidth]{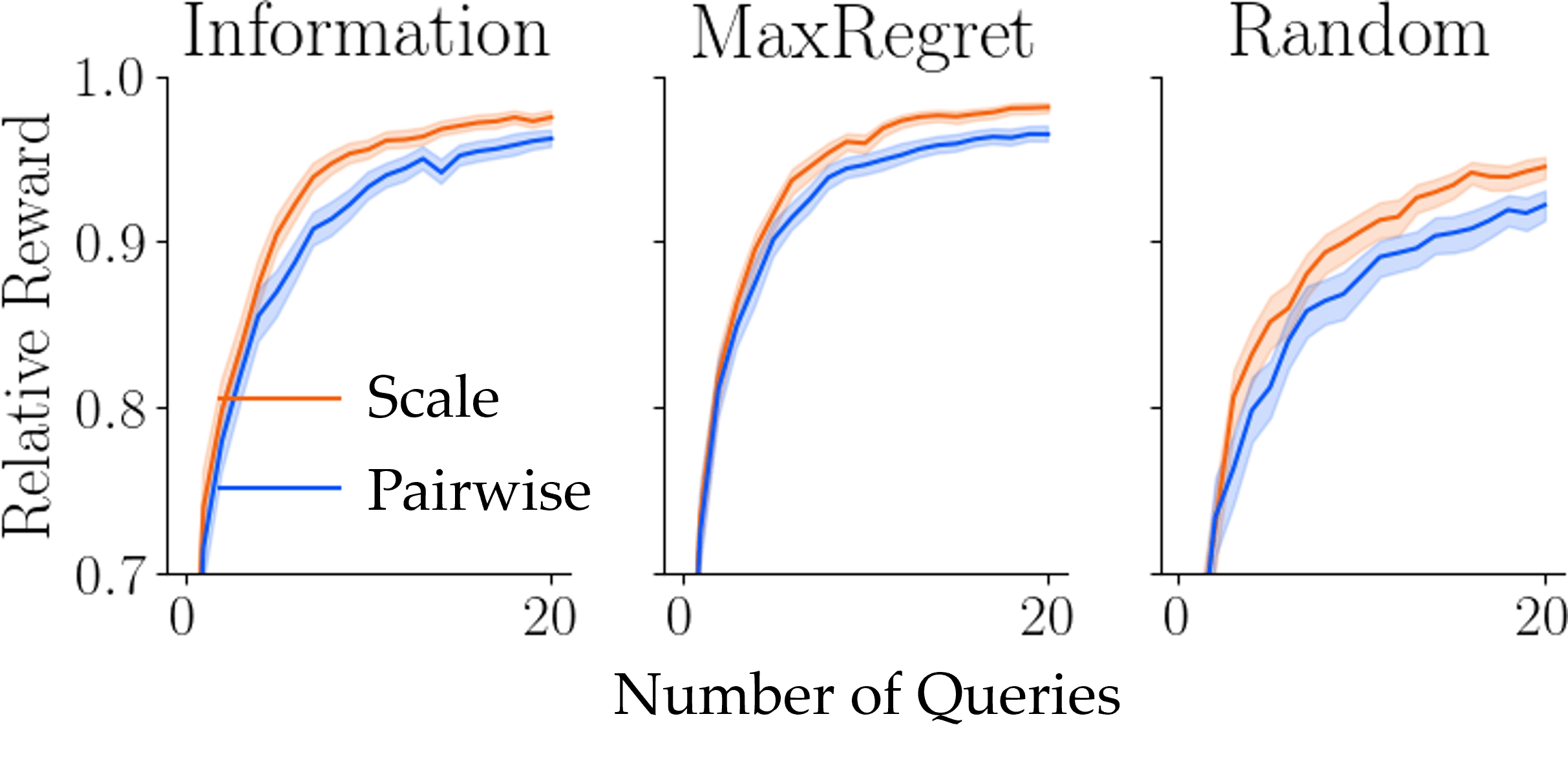}
		\end{subfigure}

		\caption{\texttt{Alignment} (left) and \texttt{Relative\_Reward} (right) for \emph{ExtendedDriver} with $\scaleNoiseStd=0.3$.}
		\label{fig:99_05_corl21_scale_driverExtended_high_noise}
\end{figure}

\noindent\textbf{High Noise.} In Section~\ref{subsec:04_05_corl21_scale_experiments} we showed results for user noise $\scaleNoiseStd=0.1$ in Figure~\ref{fig:04_05_corl21_scale_driver_extended}. In addition, we repeat the same experiment but with $\scaleNoiseStd=0.3$; shown in Figure~\ref{fig:99_05_corl21_scale_driverExtended_high_noise}.
Overall, we observe a poorer performance for all approaches compared to $\scaleNoiseStd=0.1$ -- higher noise in the user feedback makes learning more difficult. Nevertheless, scale feedback still leads to an improvement on both measures, \texttt{Alignment} and \texttt{Relative\_Reward}.

\begin{figure}[ht]
		\centering
		\begin{subfigure}[t]{0.49\textwidth}
        \includegraphics[width=1\textwidth]{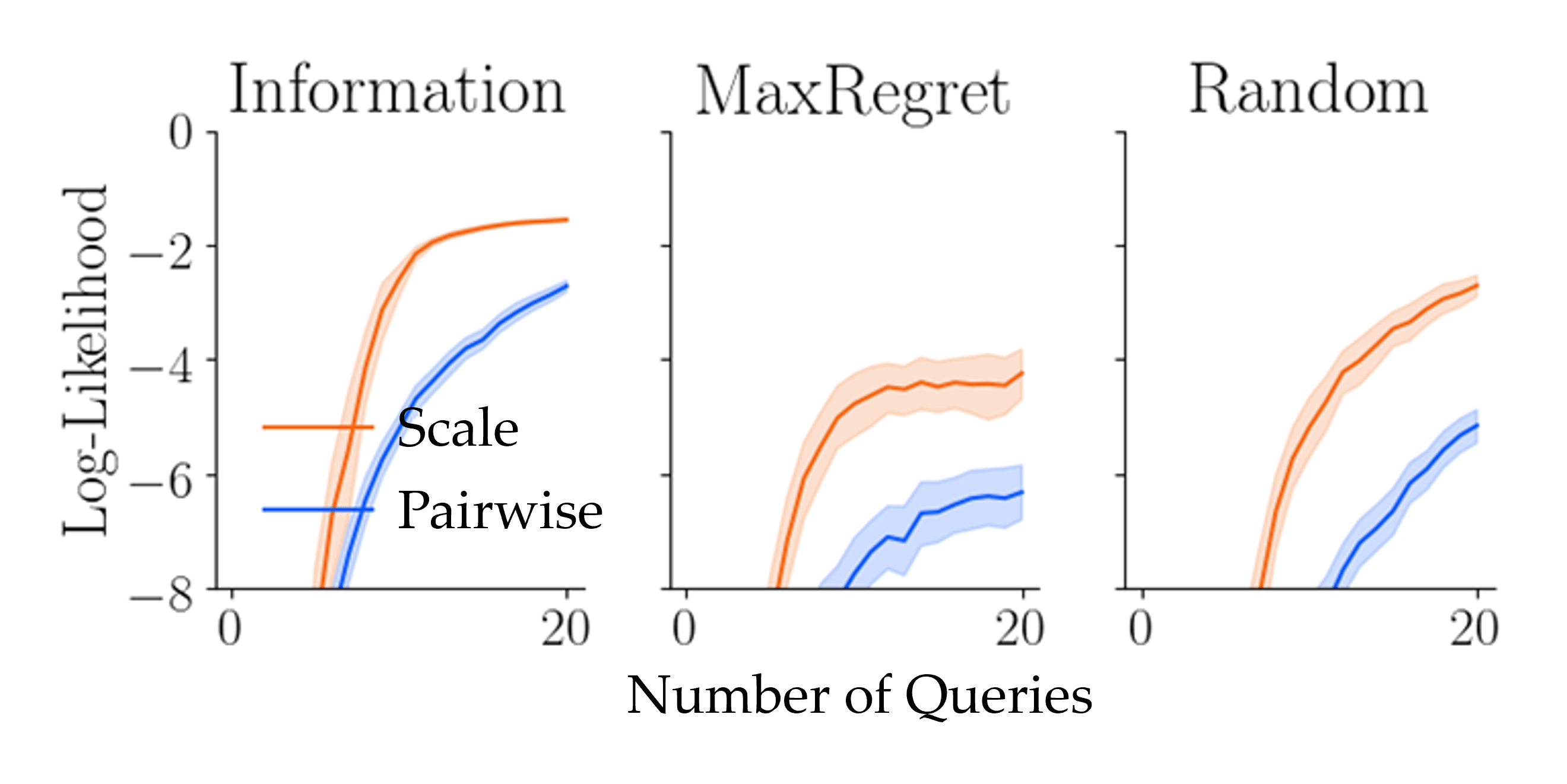}
        \caption{$\scaleNoiseStd=0.1$}
		\end{subfigure}
		\hfill
		\begin{subfigure}[t]{0.49\textwidth}
        \includegraphics[width=1\textwidth]{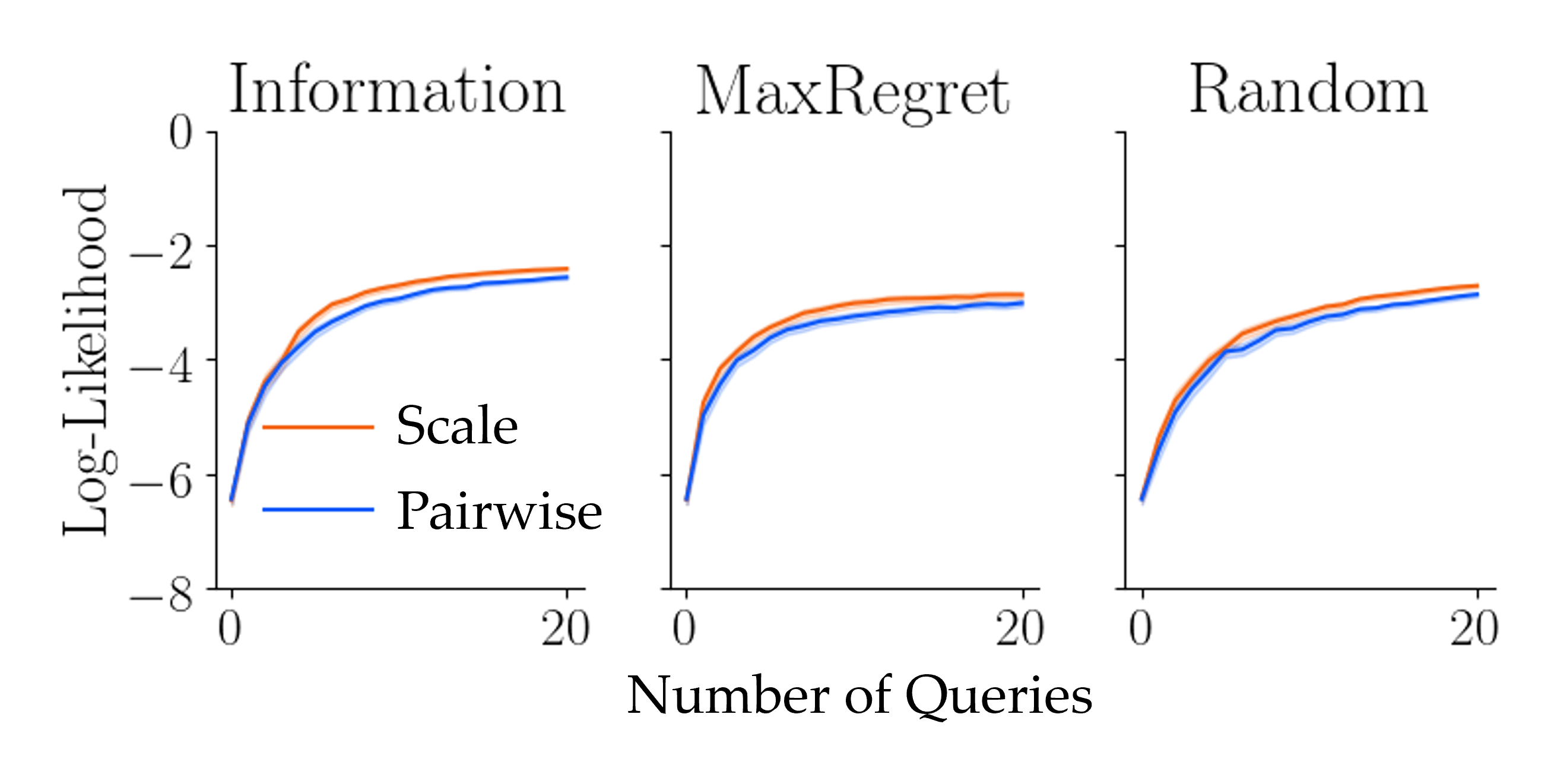}
        \caption{$\scaleNoiseStd=0.3$}
		\end{subfigure}
		\caption{\texttt{Log-Likelihood} for the \emph{ExtendedDriver} simulations.}
		\label{fig:99_05_corl21_scale_driverExtended_log}
\end{figure}

\noindent\textbf{Log-Likelihood.}
Figure~\ref{fig:99_05_corl21_scale_driverExtended_log} shows the \texttt{Log-Likelihood} for the \emph{ExtendedDriver} simulations. When the noise is small, scale feedback significantly outperforms weak pairwise comparisons under all three active querying methods. Further, mutual information based method performs best overall, followed by random. It might be surprising that max regret achieves a lower log-likelihood than random. Max regret greedily tries to find solutions that are close to optimal. Thus, this approach does not gather information about comparably good or bad trajectories (with respect to collected reward). Since the set of test queries is generated randomly, it might contain numerous queries about which the max regret approach is still uncertain since it only focused on finding close to optimal solutions. Mutual information based method, on the other hand, minimizes the uncertainty about weights, regardless of how different the resulting trajectories are. Similarly, random querying is completely unbiased and thus does not focus on a subset of queries as the max regret approach does.

In Figure~\ref{fig:99_05_corl21_scale_driverExtended_log} (b) we show the log-likelihood for high noise. Here all three active querying methods perform nearly identical, and the difference between scale feedback and weak pairwise comparisons is very small. This is because, when the noise is high, i.e., when the Gaussian over the feedback value has high variance, the log-likelihood measure does not heavily penalize bad predictions, which causes all methods to acquire high log-likelihood values. 

\subsection{Original \emph{Driver}}

\begin{figure}[ht]
		\centering
		\begin{subfigure}[t]{0.49\textwidth}
        \includegraphics[width=1\textwidth]{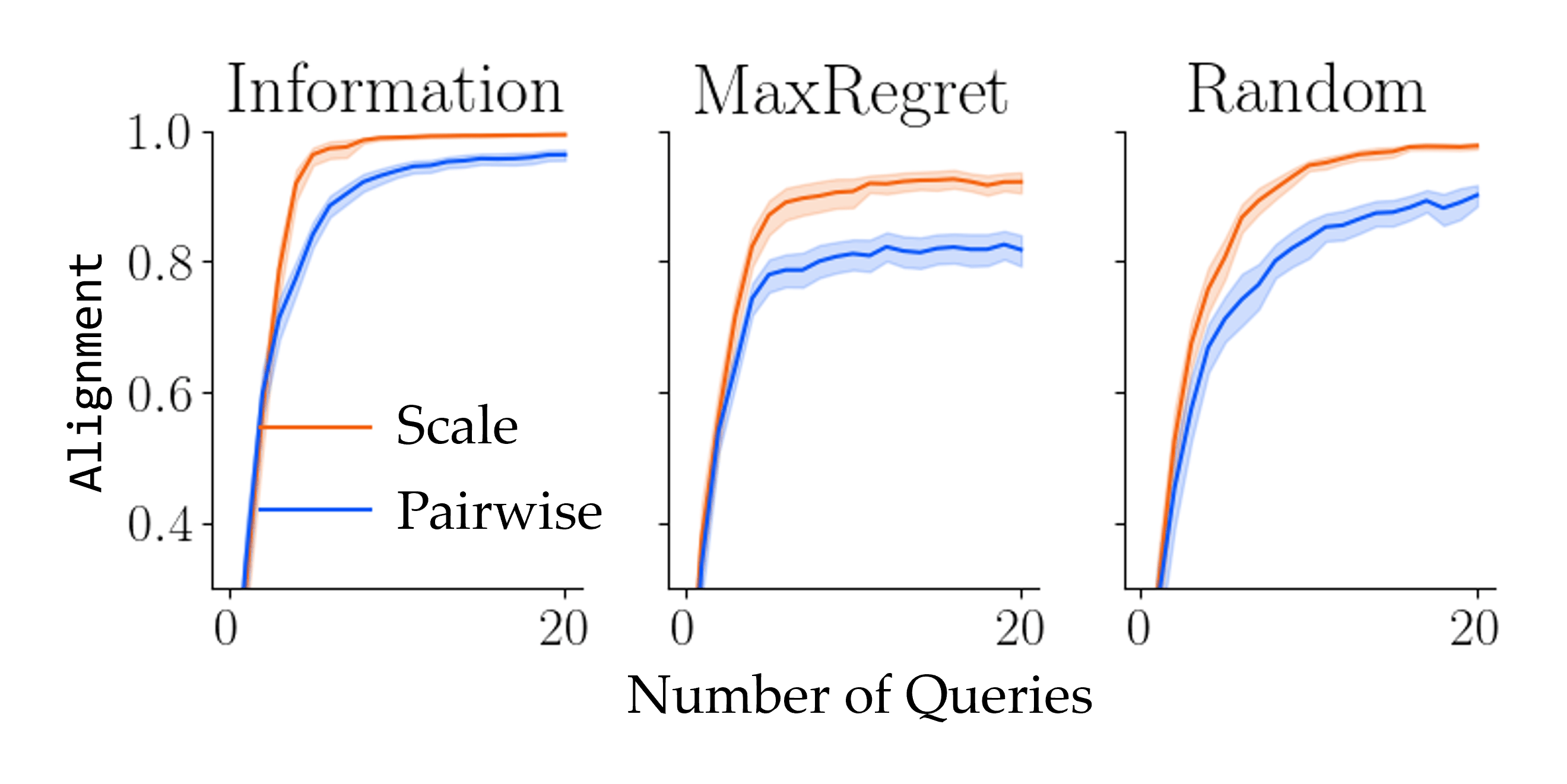}
		\end{subfigure}
		\hfill
		\begin{subfigure}[t]{0.49\textwidth}
        \includegraphics[width=1\textwidth]{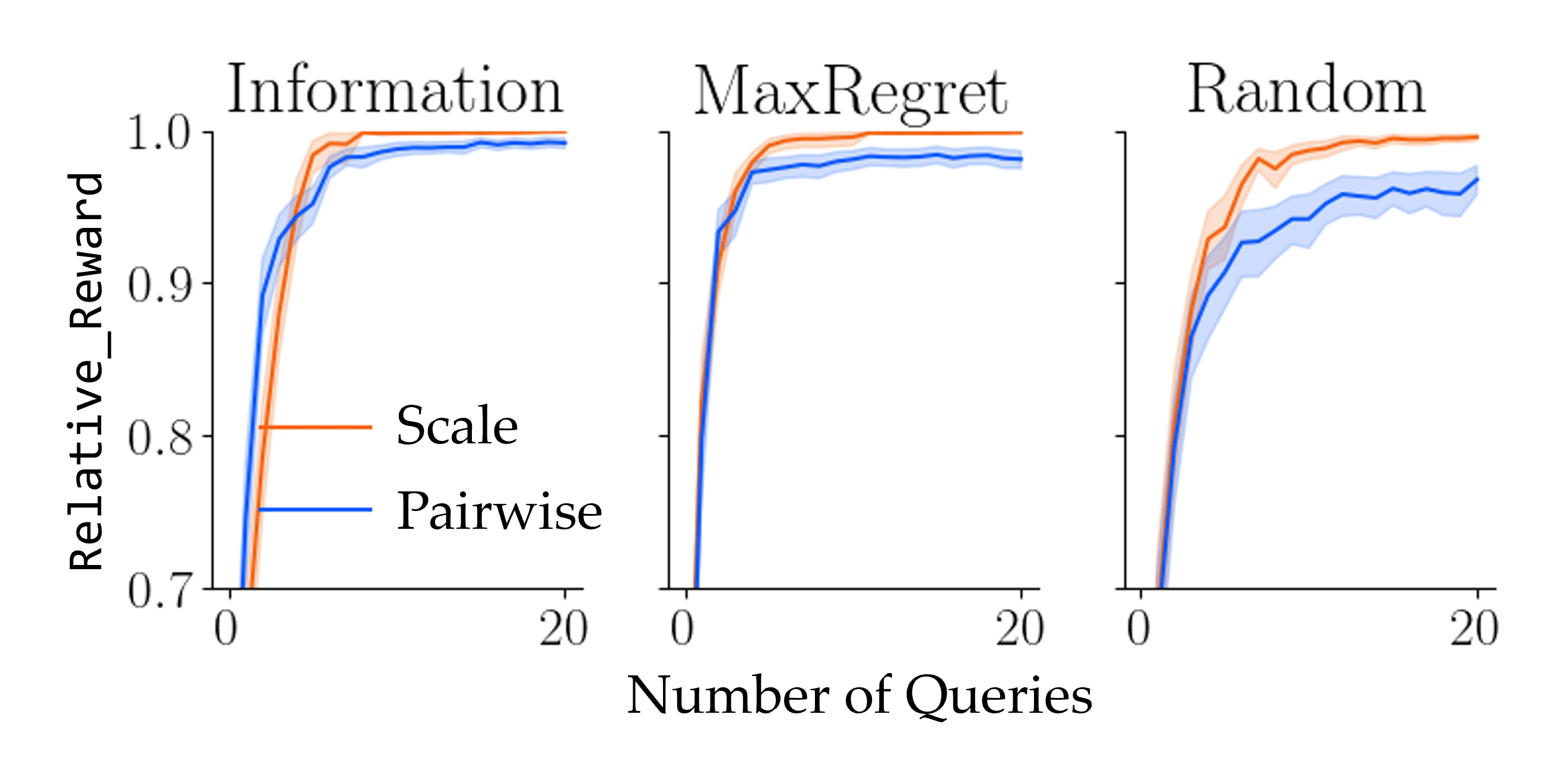}
		\end{subfigure}
		\caption{\texttt{Alignment} and \texttt{Relative\_Reward} for the original \emph{Driver} with $\scaleNoiseStd=0.1$.}
		\label{fig:99_05_corl21_scale_driver_orig_lownoise}

		\centering
		\begin{subfigure}[t]{0.49\textwidth}
        \includegraphics[width=1\textwidth]{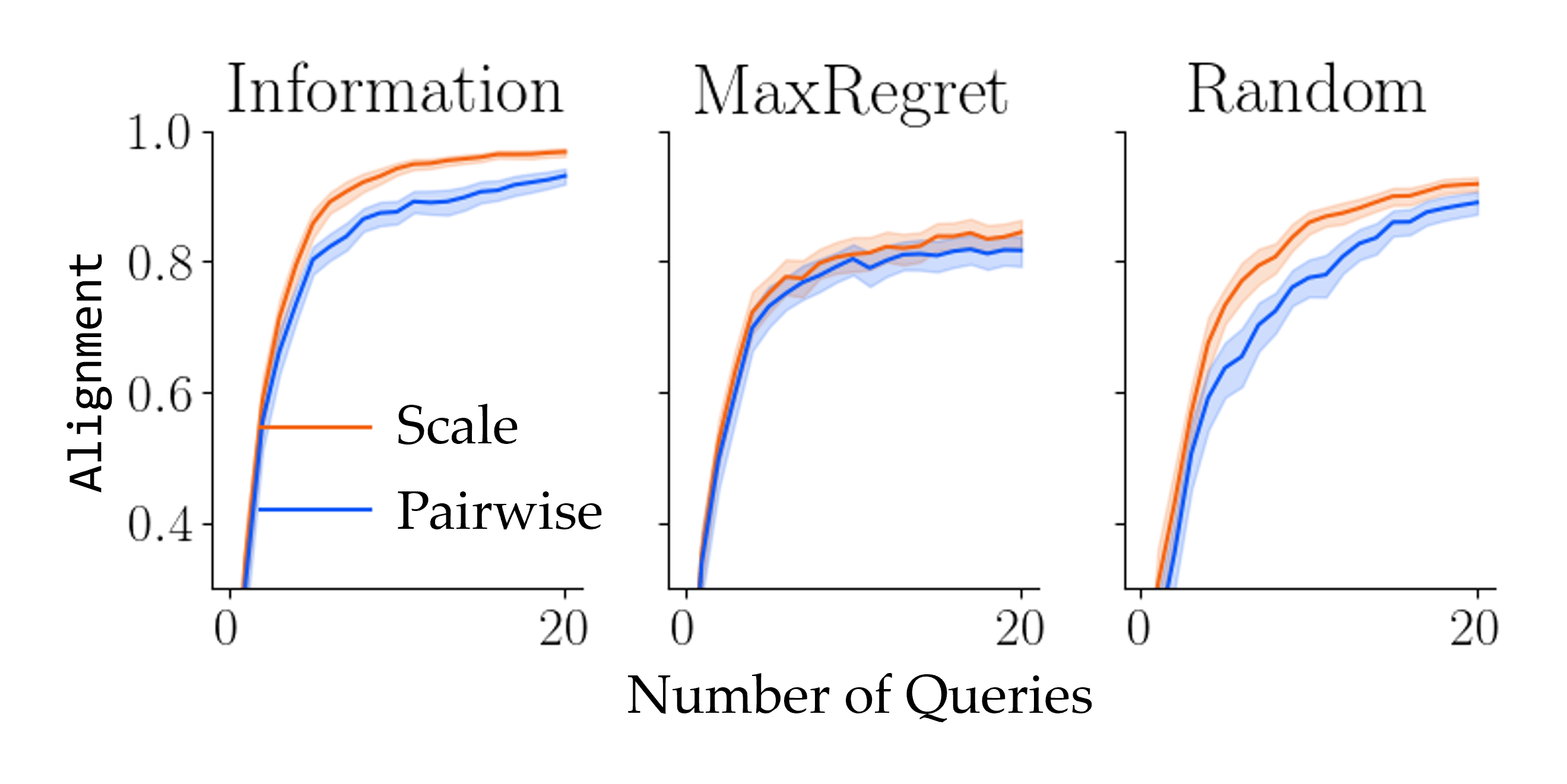}
		\end{subfigure}
		\hfill
		\begin{subfigure}[t]{0.49\textwidth}
        \includegraphics[width=1\textwidth]{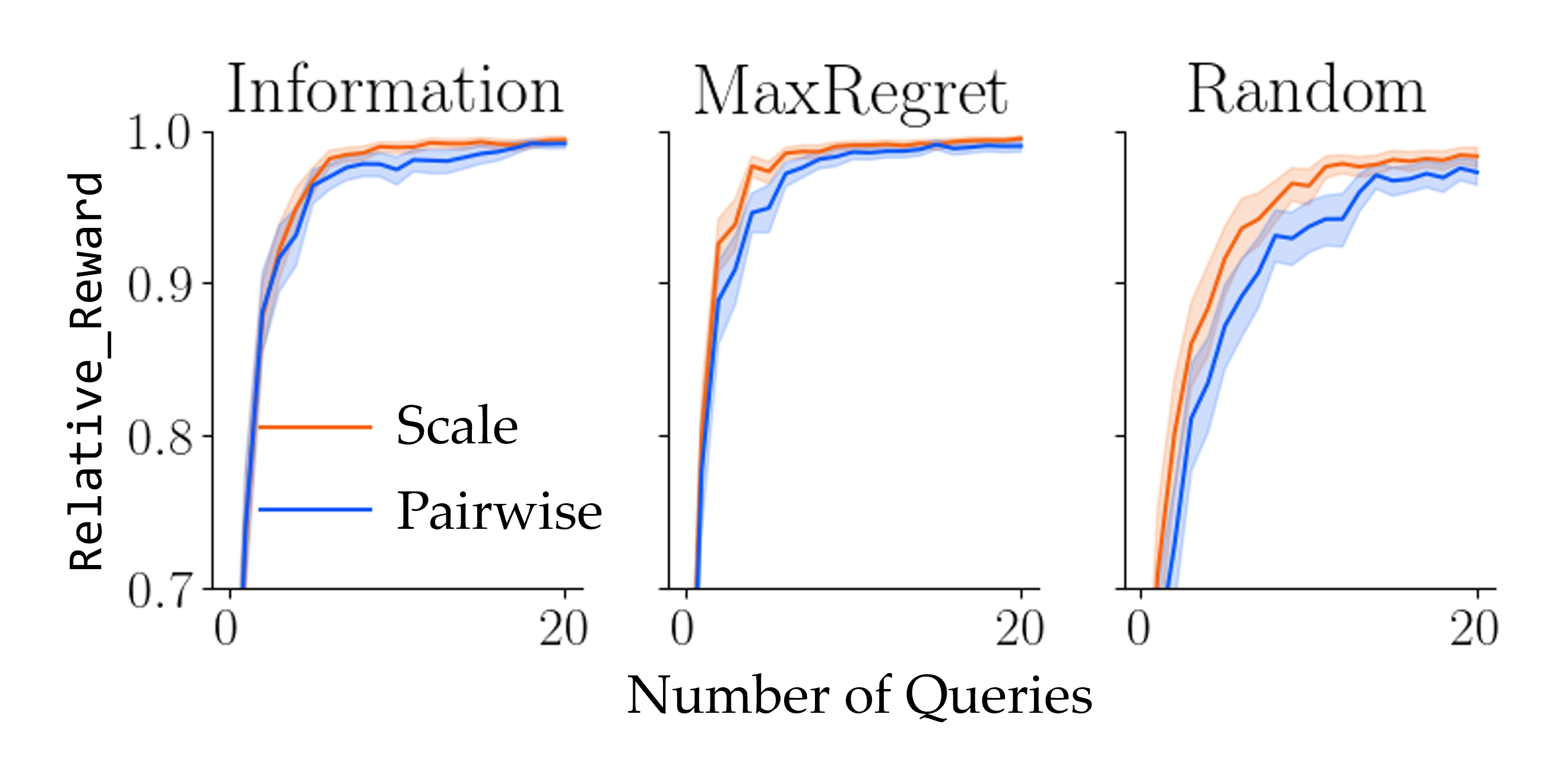}
		\end{subfigure}
		\caption{\texttt{Alignment} and \texttt{Relative\_Reward} for the original \emph{Driver} with $\scaleNoiseStd=0.3$.}
		\label{fig:99_05_corl21_scale_driver_orig_highnoise}

		\centering
		\begin{subfigure}[t]{0.49\textwidth}
        \includegraphics[width=1\textwidth]{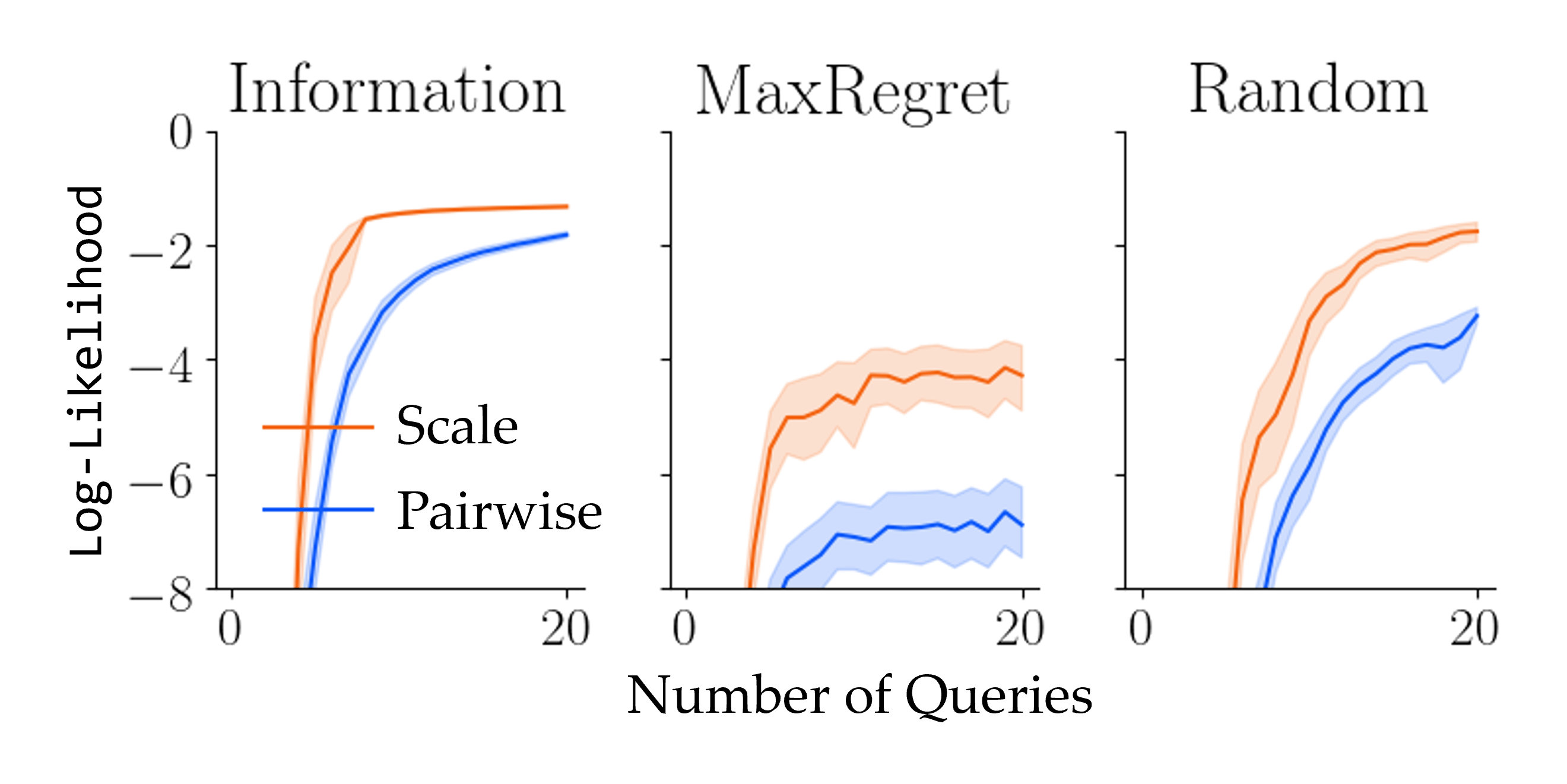}
        \caption{$\scaleNoiseStd=0.1$}
		\end{subfigure}
		\hfill
		\begin{subfigure}[t]{0.49\textwidth}
        \includegraphics[width=1\textwidth]{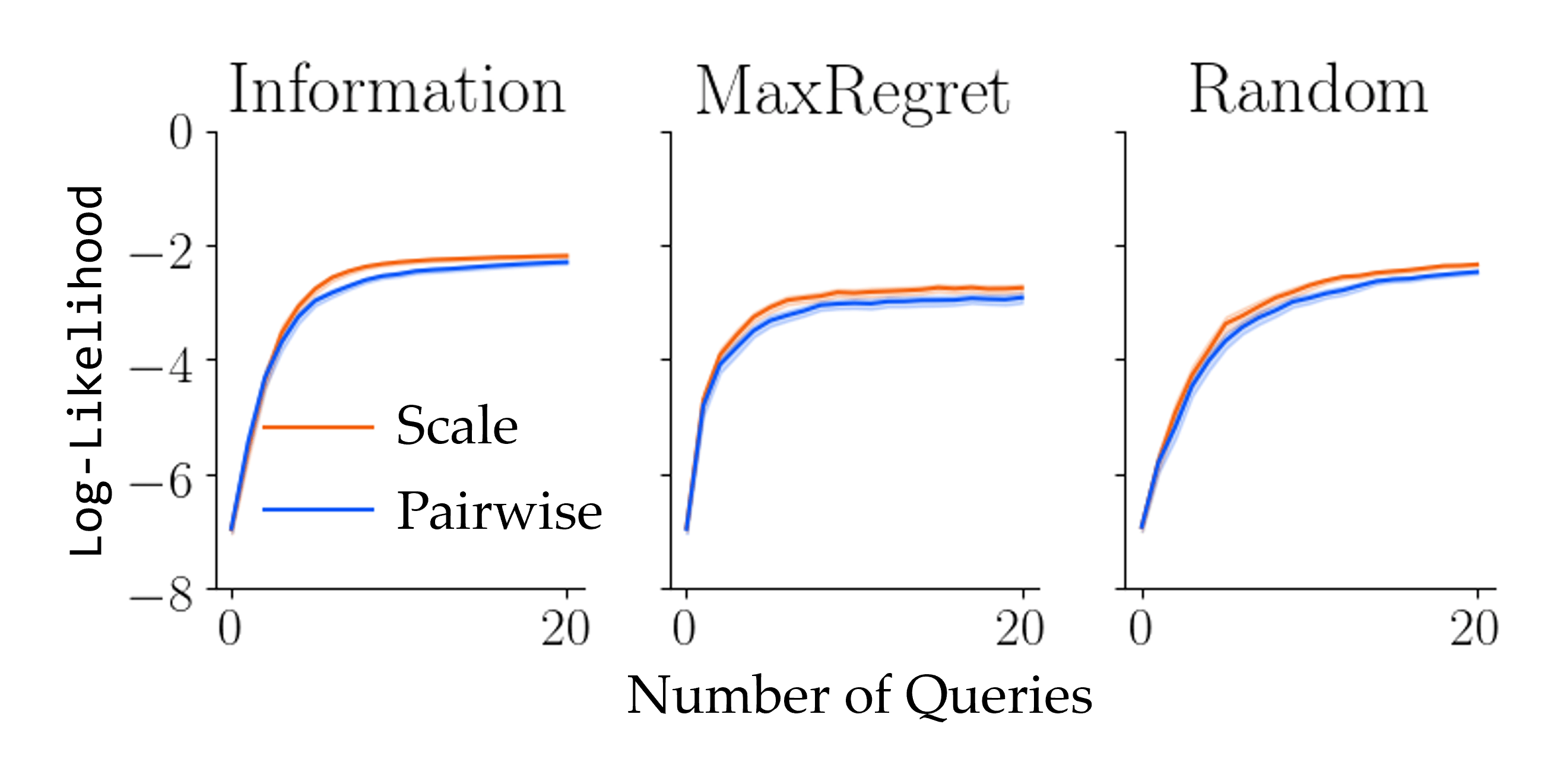}
        \caption{$\scaleNoiseStd=0.3$}
		\end{subfigure}
		\caption{\texttt{Log-Likelihood} for the original \emph{Driver}.}
		\label{fig:99_05_corl21_scale_driverOrig_log}
\end{figure}

\noindent\textbf{Alignment and Relative Reward.} Next, we show results for the original \emph{Driver} experiment. Figure~\ref{fig:99_05_corl21_scale_driver_orig_lownoise} shows the \texttt{Alignment} and \texttt{Relative\_Reward} for low noise ($\scaleNoiseStd=0.1$), Figure\ref{fig:99_05_corl21_scale_driver_orig_highnoise} shows the same measures for high noise ($\scaleNoiseStd=0.3$).
While scale feedback still improves \texttt{Alignment} and \texttt{Relative\_Reward} for all querying methods, the gap to weak pairwise comparison feedback is smaller than for the \emph{ExtendedDriver}. However, we observe that all querying methods achieve a substantially stronger performance than in the \emph{ExtendedDriver} model with $10$ features, indicating that the original \emph{Driver} environment poses a less difficult learning problem with only $4$ features.
We notice that the result for weak pairwise comparisons via mutual information optimization achieves a higher \texttt{Alignment} after $20$ iterations than reported in Section~\ref{sec:04_02_information_gain}. There are two reasons for this: First, we use a Gaussian noise instead of the Boltzmann model (also known as MNL or the softmax model). Second, by emulating weak pairwise comparisons using a slider with step size $1$, we change the model for when users give a neutral (``About Equal") feedback.
Nonetheless, the stronger performance compared to Section~\ref{sec:04_02_information_gain} suggests that these differences do not negatively impact the performance of weak comparison queries with mutual information maximization, and thus that the shown comparisons of scale feedback and weak pairwise comparisons are fair.

\noindent\textbf{Log-Likelihood.} We also report the results in the \texttt{Log-Likelihood} measure in Figure~\ref{fig:99_05_corl21_scale_driverOrig_log}. The results are very similar to the results of the \emph{ExtendedDriver} environment, except the \texttt{Log-Likelihood} values increase faster. This is again because the reward is easier to learn in the original \emph{Driver} environment with the fewer number of features.

\subsection{\emph{FetchDrink}} \label{app:99_05_corl21_scale_simulations_fetch_robot}

\begin{figure}[ht]
		\centering
		\begin{subfigure}[t]{0.49\textwidth}
        \includegraphics[width=1\textwidth]{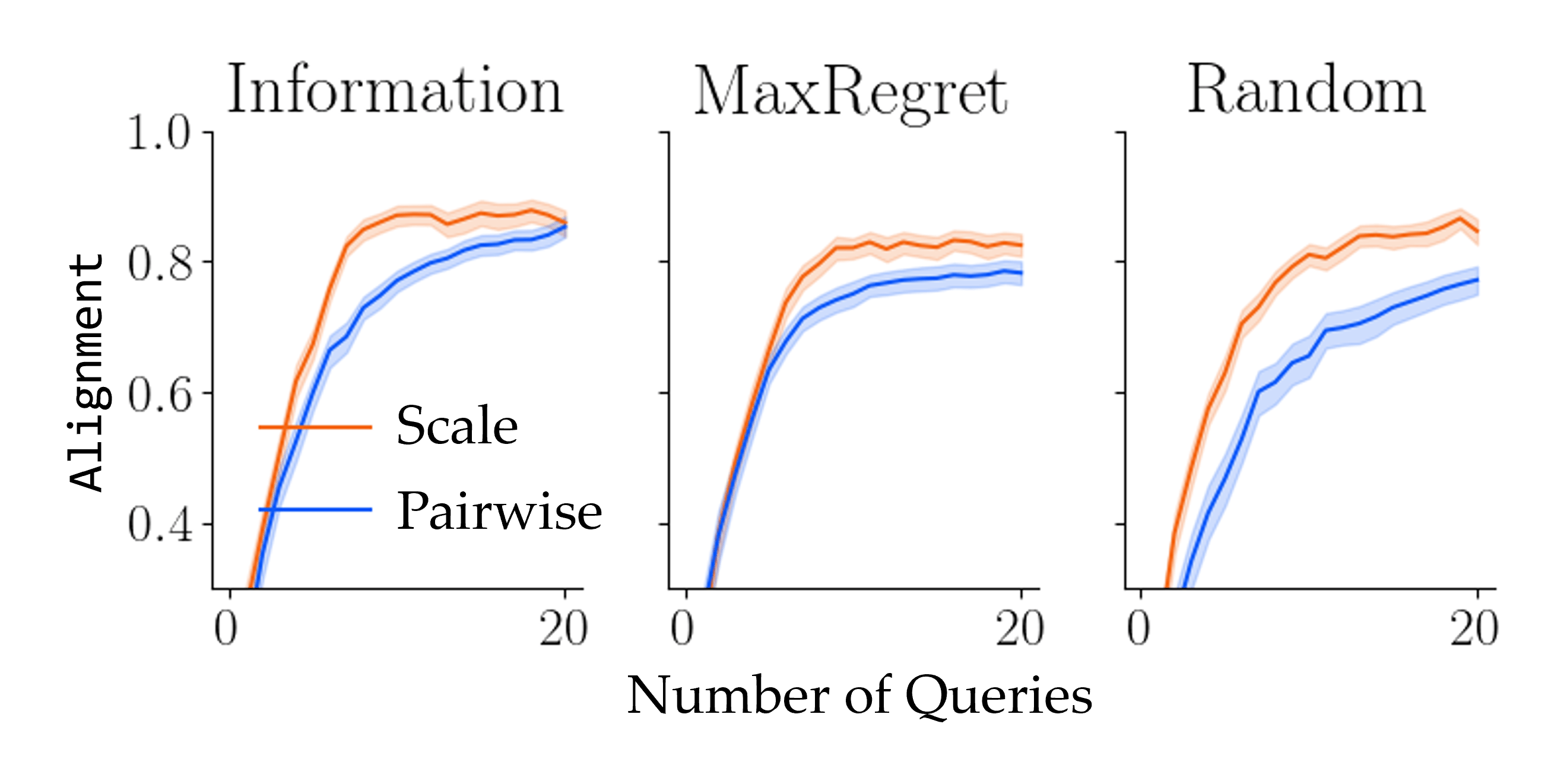}
		\end{subfigure}
		\hfill
		\begin{subfigure}[t]{0.49\textwidth}
        \includegraphics[width=1\textwidth]{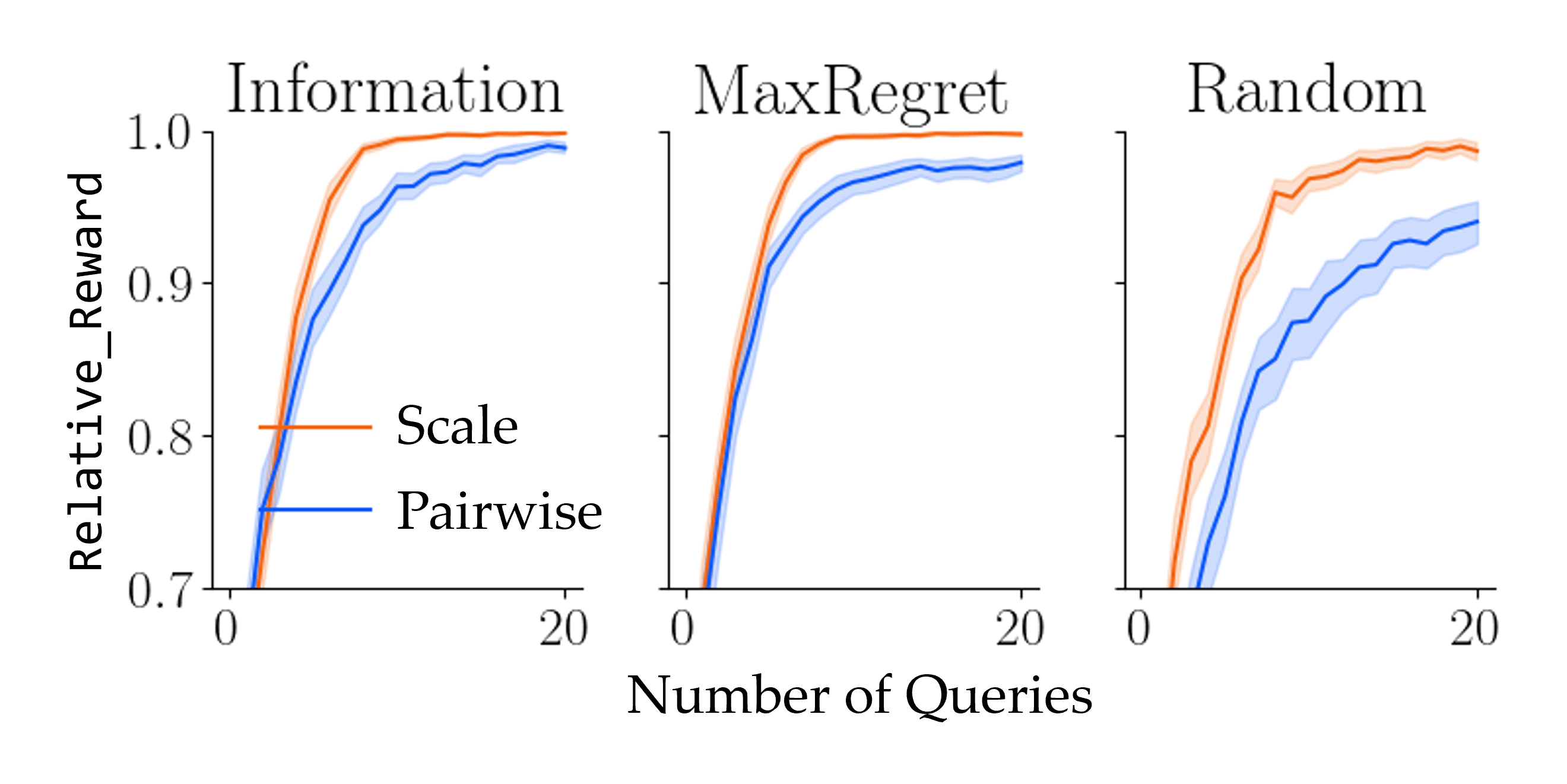}
		\end{subfigure}
		\caption{Fetch robot with drink serving experiment (\emph{FetchDrink}) with $\scaleNoiseStd=0.1$.}
		\label{fig:99_05_corl21_scale_fetch_lownoise}

		\centering
		\begin{subfigure}[t]{0.49\textwidth}
        \includegraphics[width=1\textwidth]{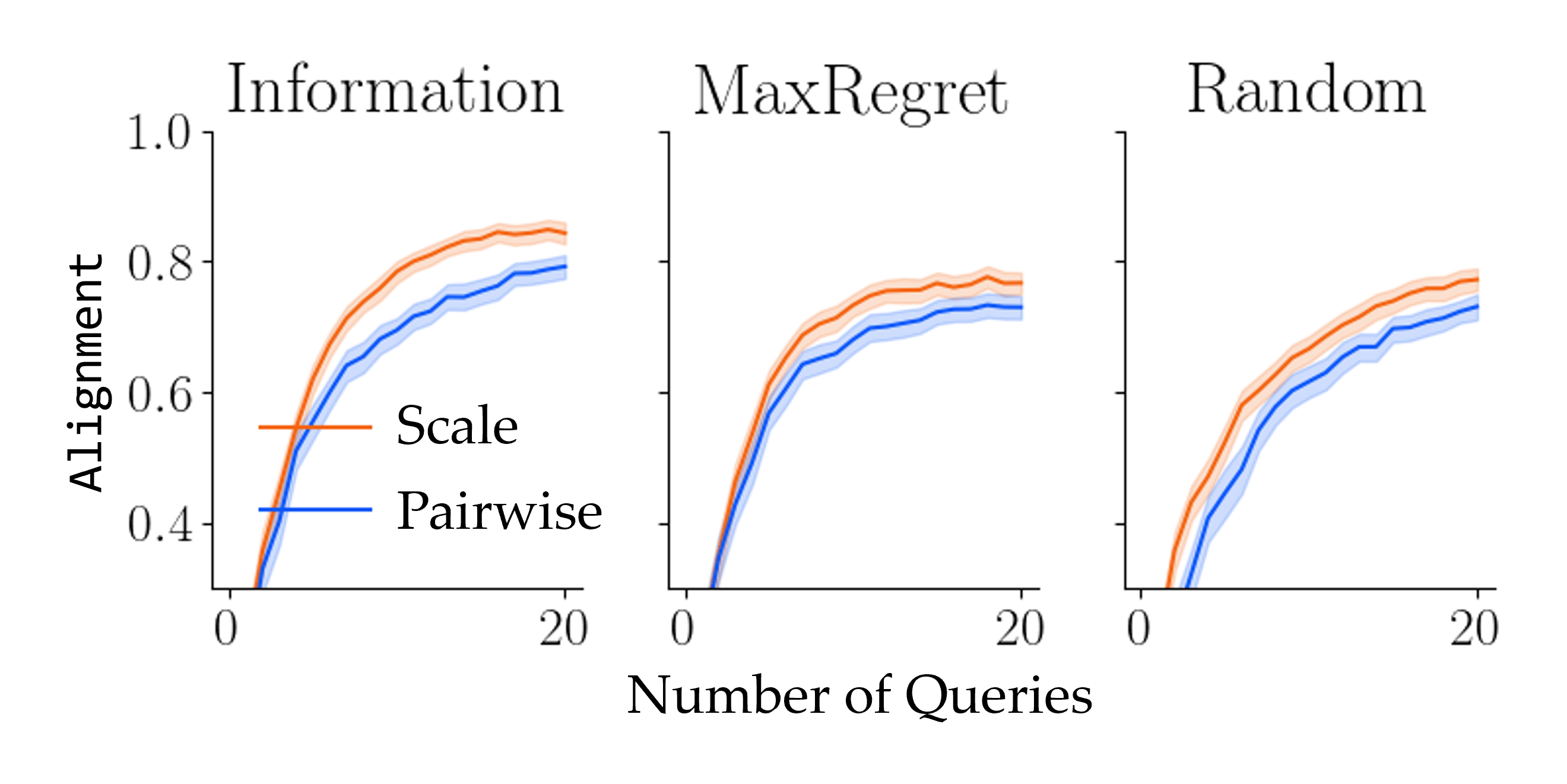}
		\end{subfigure}
		\hfill
		\begin{subfigure}[t]{0.49\textwidth}
        \includegraphics[width=1\textwidth]{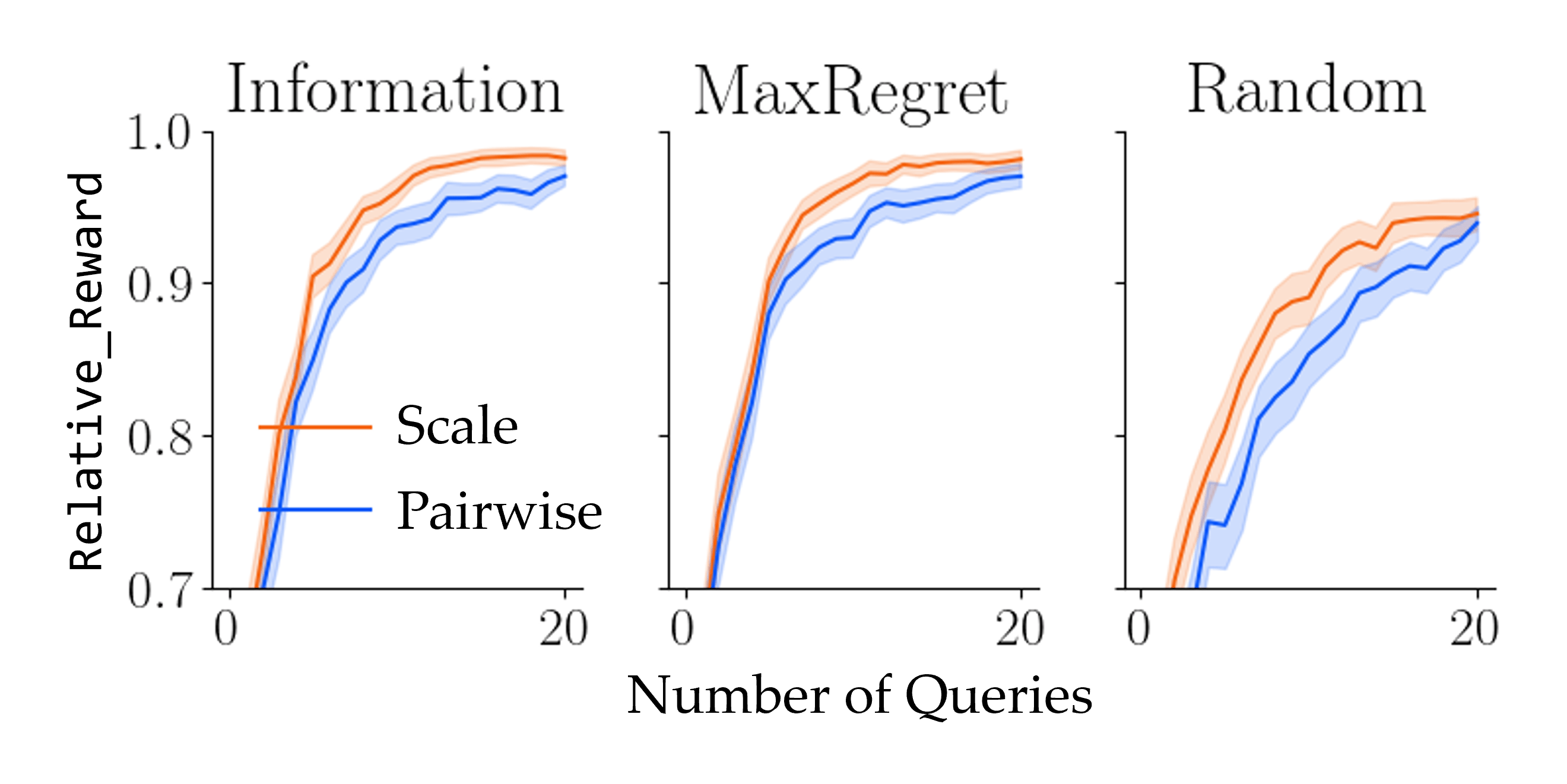}
		\end{subfigure}
		\caption{Fetch robot with drink serving experiment (\emph{FetchDrink}) with $\scaleNoiseStd=0.3$.}
		\label{fig:99_05_corl21_scale_fetch_highnoise}

		\centering
		\begin{subfigure}[t]{0.49\textwidth}
        \includegraphics[width=1\textwidth]{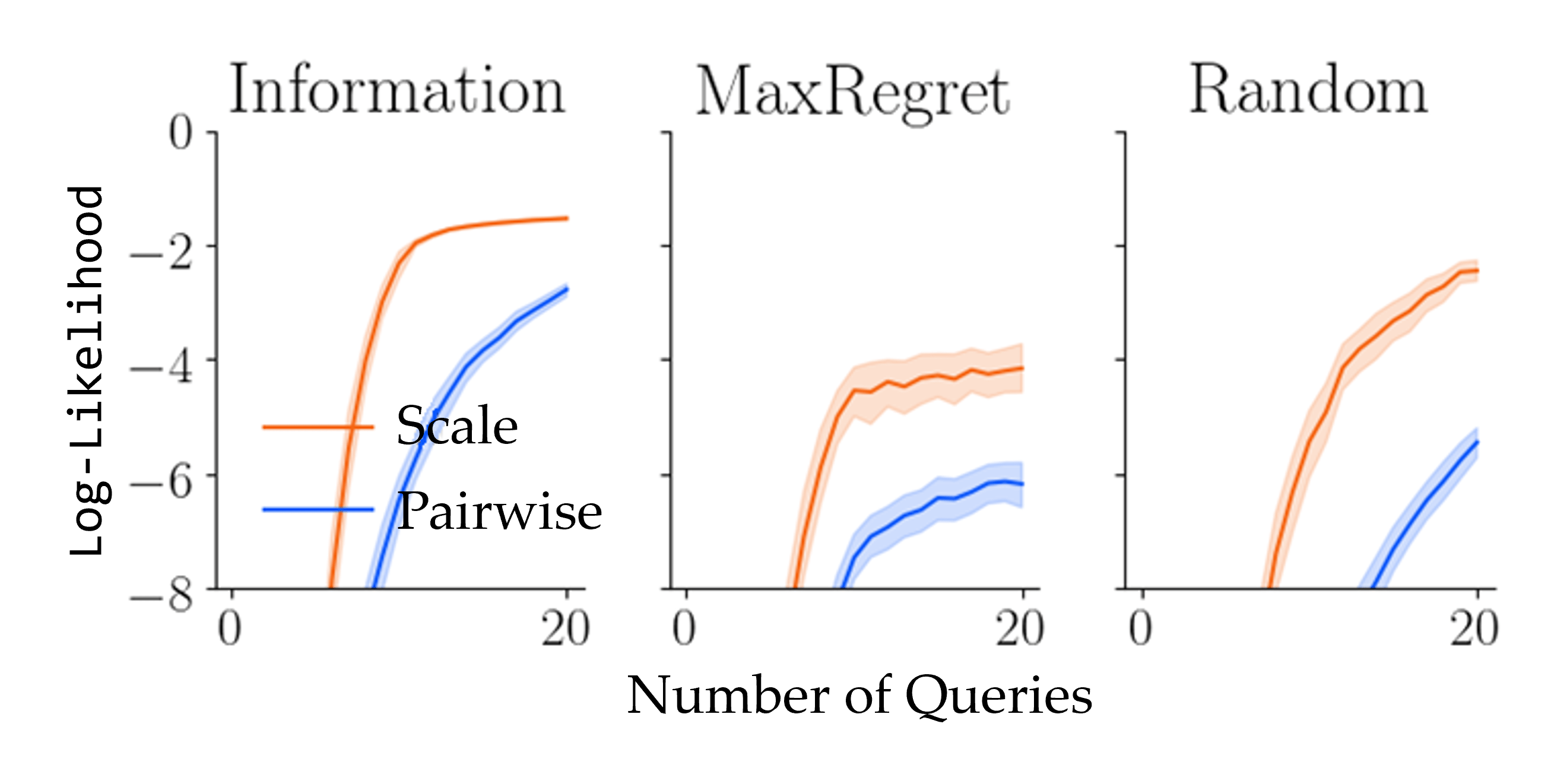}
        \caption{$\scaleNoiseStd=0.1$}
		\end{subfigure}
		\hfill
		\begin{subfigure}[t]{0.49\textwidth}
        \includegraphics[width=1\textwidth]{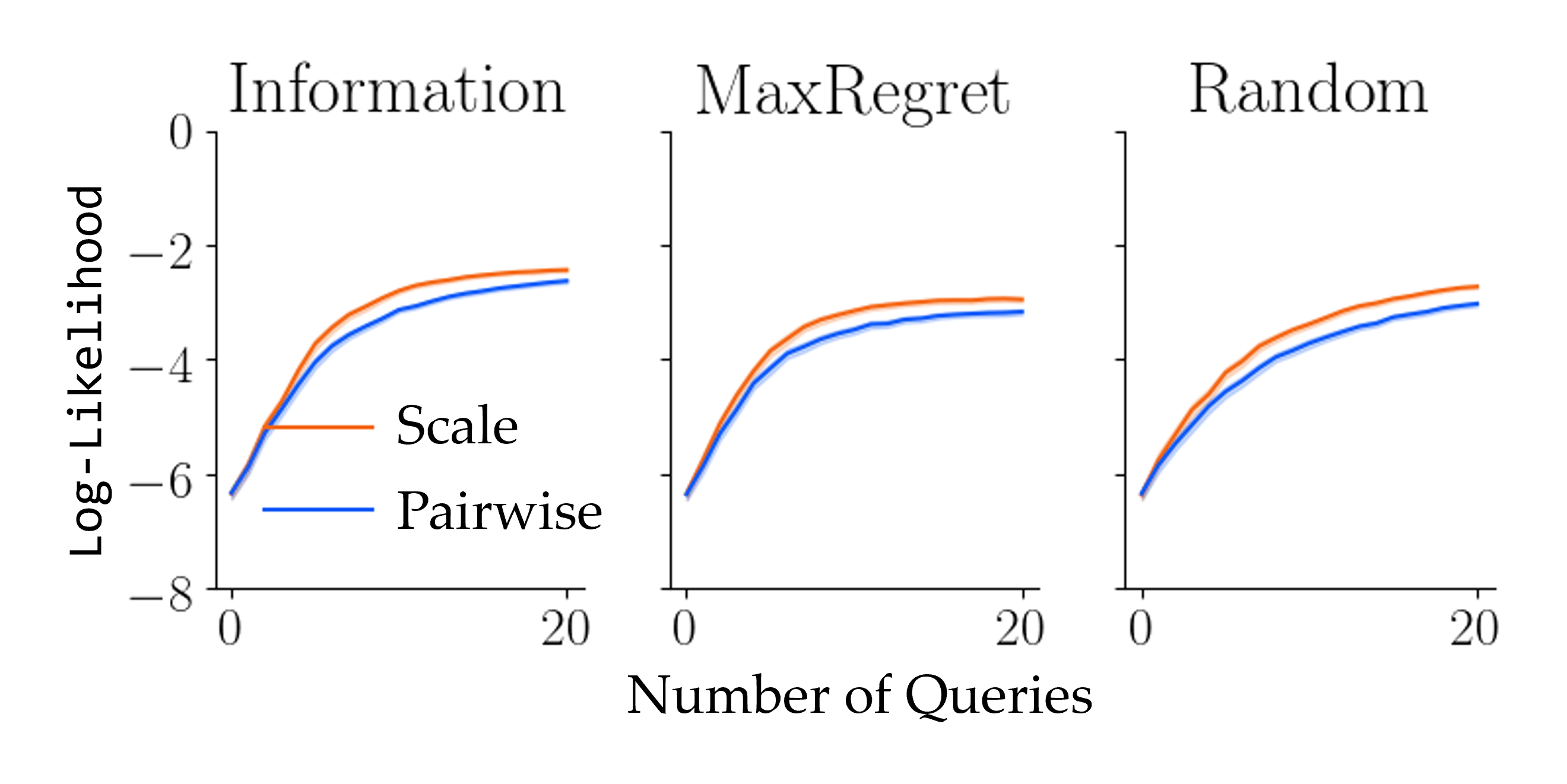}
        \caption{$\scaleNoiseStd=0.3$}
		\end{subfigure}
		\caption{Log-Likelihood for the Fetch robot with drink serving experiment (\emph{FetchDrink}).}
		\label{fig:99_05_corl21_scale_fetch_log}
\end{figure}

We now show simulation results for the experimental setup from the user study, using the Fetch robot: \emph{FetchDrink}. Figure~\ref{fig:99_05_corl21_scale_fetch_lownoise} shows the \texttt{Alignment} and \texttt{Relative\_Reward} for low noise ($\scaleNoiseStd=0.1$), Figure~\ref{fig:99_05_corl21_scale_fetch_highnoise} shows the same measures for high noise ($\scaleNoiseStd=0.3$), and 
Figure~\ref{fig:99_05_corl21_scale_fetch_log} shows the \texttt{Log-Likelihood}.
In terms of the comparisons between different feedback types and different active querying methods, the results have the same trend as the \emph{ExtendedDriver} and the original \emph{Driver} environments.

\section{Results with Test Set with Mixture Data for Section~\ref{subsec:04_05_corl21_scale_experiments}}\label{app:99_05_corl21_scale_test_mixture_data}
In both of our user studies, we used a test set that consists of randomly generated scale questions. Given the fact that the subjective user ratings did not point out a significant difference between learning from scale feedback and pairwise comparison queries, one might argue that the superiority of learning from scale feedback in terms of the \texttt{Log-Likelihood} metric is simply because the test set also consists of scale feedback. Mathematically, this should not happen, because a good posterior should be able to correctly predict any form of user feedback. However, humans have cognitive biases, which makes it possible that the posterior learned with the scale questions captures the bias caused by the scale questions, whereas the posterior learned with the weak pairwise comparisons cannot do this.

To show this is not the case, we present an additional analysis on the same human data as in our first user study. For this analysis, we take the reward posteriors that have been learned with the first $7$ queries (of ``Scale - Mutual Information", ``Scale - Random", and ``Pairwise - Random"). Next, we alter the test set as follows. We take (i) the first 3 scale queries from the original test set, and (ii) the last 3 weak pairwise comparison queries from the original training set of randomly generated weak pairwise comparison queries (and this is why we only take the first $7$ posteriors -- we do not mix the training and test data). Finally, we perform the \texttt{Log-Likelihood} analysis on this modified test set.

\begin{figure}[ht]
    \centering
    \includegraphics[width=0.35\linewidth]{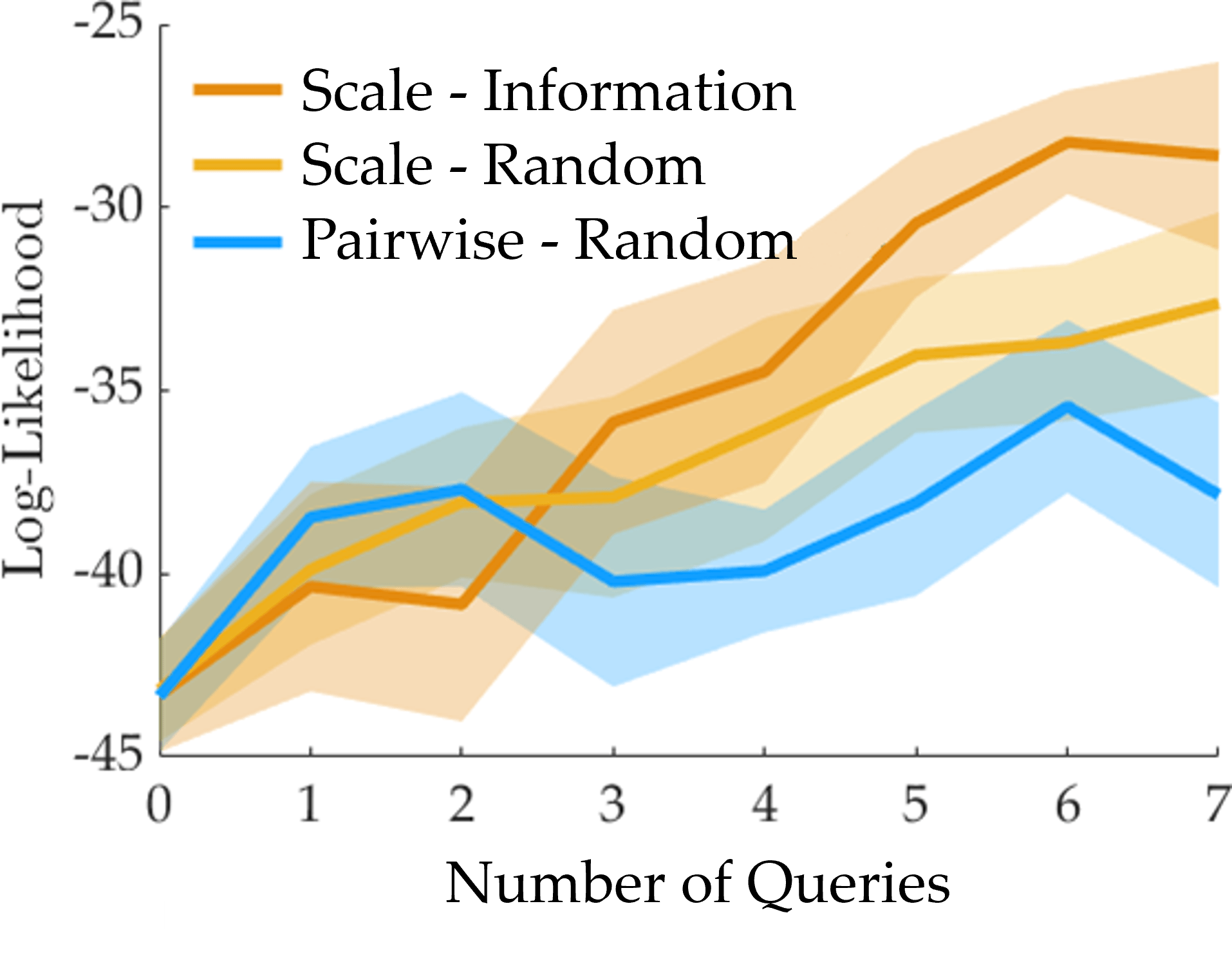}
    \caption{Additional analysis results are shown (mean$\pm$s.e. over $18$ subjects).}
    \label{fig:99_05_corl21_scale_user_study_extra}
\end{figure}

Results are shown in Figure~\ref{fig:99_05_corl21_scale_user_study_extra}. It can be seen that even with a test set that consists of mixture data, the results have the same trend as in the original study results. While having smaller test set ($6$ instead of the $10$ in the original study) causes larger standard errors, ``Scale - Information" and ``Scale - Random" both outperform ``Pairwise - Random" with statistical significance ($p<0.05$ in both comparisons). On the other hand, the comparison between ``Scale - Information" and ``Scale - Random" gives $p=0.098$.

This analysis shows the fact that scale feedback outperforms weak pairwise comparisons in terms of \texttt{Log-Likelihood} is not because of the data in the test set. Even with a test set that consists of both scale and weak pairwise comparison data, we see the benefits of learning from scale queries.

However, this analysis does not answer the question why user ratings did not have a significant difference between the two feedback types. While the answer to this question requires more analysis and possibly more data collection, we speculate the following reason: the mean user ratings are always around $4$, and even higher than $4$ when queries are actively generated with mutual information. This means the users are happy with the optimized trajectories, so we can say that $10$ queries are enough in this task to find the optimal trajectory. However, while user ratings measure how close the optimal trajectory with respect to the robot's posterior is to the optimal trajectory the user has in mind; \texttt{Log-Likelihood} measures the predictive performance of the posterior. Therefore, having a high user rating does not necessarily mean the robot can accurately compare two suboptimal trajectories. On the other hand, a high \texttt{Log-Likelihood} value indicates good predictive performance, which is crucial in many robotics applications, such as behavior modeling. Hence, we claim: (i) learning from scale feedback improves the predictive performance over learning from weak pairwise comparisons, and (ii) a more complex task might be needed to show scale feedback leads to more efficient learning than weak pairwise comparisons, which is also suggested by our simulation studies.

\section{Numerical Results for Section~\ref{subsec:04_05_corl21_scale_experiments}}\label{app:99_05_corl21_scale_numerical_results}
Here, we present Table~\ref{tab:99_05_corl21_scale_numerical_results_sim} where we report the numerical results of the simulations in Section~\ref{subsec:04_05_corl21_scale_experiments} at iterations $0,5,10,20$; and Table~\ref{tab:99_05_corl21_scale_numerical_results} where we report the final numerical results of the user studies. Consistent with the section, the numbers are presented as mean $\pm$ standard deviation (simulations) and standard error (user study).

\begin{table}[ht]
\centering
\caption{Numerical results of the simulations at selected iterations $i$}
\label{tab:99_05_corl21_scale_numerical_results_sim}
\begin{tabular}{lcccc}\toprule

&\multicolumn{4}{r}{\makecell{Mean$\pm$Standard Deviation}} \\
 \cmidrule(lr){2-5}
{Plot}&{$i= 0$}&{$i=5$}&{$i=10$}&{$i=20$}\\\midrule
Fig.~\ref{fig:04_05_corl21_scale_driver_extended} Scale - Information (Alignment) & $-.01\pm.33$&$\mathbf{.62\pm.19}$&$\mathbf{.81\pm.16}$&$\mathbf{.9\pm.08}$\\
Fig.~\ref{fig:04_05_corl21_scale_driver_extended} Pairwise - Information (Alignment) & $-.02\pm.31$&$.52\pm.18$&$.67\pm.16$&$.79\pm.15$\\
Fig.~\ref{fig:04_05_corl21_scale_driver_extended} Scale - MaxRegret (Alignment) & $.01\pm.31$&$.57\pm.19$&$.71\pm.16$&$.75\pm.16$\\
Fig.~\ref{fig:04_05_corl21_scale_driver_extended} Pairwise - MaxRegret (Alignment) & $-.03\pm.3$&$.47\pm.23$&$.59\pm.17$&$.67\pm.18$\\
Fig.~\ref{fig:04_05_corl21_scale_driver_extended} Scale - Random (Alignment) & $.01\pm.33$&$.52\pm.2$&$.67\pm.17$&$.77\pm.17$\\
Fig.~\ref{fig:04_05_corl21_scale_driver_extended} Pairwise - Random (Alignment) & $.02\pm.32$&$.4\pm.21$&$.52\pm.2$&$.63\pm.21$\\
\midrule

Fig.~\ref{fig:04_05_corl21_scale_driver_extended} Scale - Information (Rel. Reward) & $.51\pm.32$&$.92\pm.12$&$.98\pm.04$&$\mathbf{1.0\pm.01}$\\
Fig.~\ref{fig:04_05_corl21_scale_driver_extended} Pairwise - Information (Rel. Reward) & 
$.5\pm.3$&$.89\pm.12$&$.95\pm.07$&$.98\pm.04$\\
Fig.~\ref{fig:04_05_corl21_scale_driver_extended} Scale - MaxRegret (Rel. Reward) & 
$.52\pm.31$&$\mathbf{.96\pm.07}$&$\mathbf{.99\pm.02}$&$\mathbf{1.0\pm.01}$\\
Fig.~\ref{fig:04_05_corl21_scale_driver_extended} Pairwise - MaxRegret (Rel. Reward) & $.51\pm.3$&$.91\pm.12$&$.95\pm.06$&$.96\pm.06$\\
Fig.~\ref{fig:04_05_corl21_scale_driver_extended} Scale - Random (Rel. Reward) & $.52\pm.32$&$.89\pm.14$&$.96\pm.07$&$.99\pm.03$\\
Fig.~\ref{fig:04_05_corl21_scale_driver_extended} Pairwise - Random (Rel. Reward) & $.52\pm.32$&$.85\pm.15$&$.89\pm.12$&$.93\pm.12$\\

\bottomrule
\end{tabular}
\end{table}

\begin{table}[ht]
\centering
\caption{Final numerical results of the user study}
\label{tab:99_05_corl21_scale_numerical_results}
\begin{tabular}{lc}\toprule

Plot  & Mean$\pm$Standard Error\\\midrule
 
Fig.~\ref{fig:04_05_corl21_scale_user_study_results}(a) Scale - Information & $-29.7\pm1.2$\\
Fig.~\ref{fig:04_05_corl21_scale_user_study_results}(a) Scale - Random & $-36.2\pm2.2$\\
Fig.~\ref{fig:04_05_corl21_scale_user_study_results}(a) Pairwise - Random & $-51.2\pm3.5$\\
Fig.~\ref{fig:04_05_corl21_scale_user_study_results}(b) Scale - Information & $4.2\pm0.2$\\
Fig.~\ref{fig:04_05_corl21_scale_user_study_results}(b) Scale - Random & $3.6\pm0.3$\\
Fig.~\ref{fig:04_05_corl21_scale_user_study_results}(b) Pairwise - Random & $3.9\pm0.2$\\
Fig.~\ref{fig:04_05_corl21_scale_user_study_results}(c) Scale (Easiness) & $3.8\pm0.2$\\
Fig.~\ref{fig:04_05_corl21_scale_user_study_results}(c) Pairwise (Easiness) & $4.5\pm0.2$\\
Fig.~\ref{fig:04_05_corl21_scale_user_study_results}(c) Scale (Expressiveness) & $3.8\pm0.3$\\
Fig.~\ref{fig:04_05_corl21_scale_user_study_results}(c) Pairwise (Expressiveness) & $4.1\pm0.2$\\
 
Fig.~\ref{fig:04_05_corl21_scale_user_study2_results}(a) Scale - Information & $-28.8\pm1.3$\\
Fig.~\ref{fig:04_05_corl21_scale_user_study2_results}(a) Pairwise - Information & $-46.0\pm3.1$\\
Fig.~\ref{fig:04_05_corl21_scale_user_study2_results}(b) Scale - Information & $4.5\pm0.2$\\
Fig.~\ref{fig:04_05_corl21_scale_user_study2_results}(b) Pairwise - Information & $4.2\pm0.3$\\
Fig.~\ref{fig:04_05_corl21_scale_user_study2_results}(c) Scale (Easiness) & $3.6\pm0.3$\\
Fig.~\ref{fig:04_05_corl21_scale_user_study2_results}(c) Pairwise (Easiness) & $4.6\pm0.2$\\
Fig.~\ref{fig:04_05_corl21_scale_user_study2_results}(c) Scale (Expressiveness) & $4.3\pm0.2$\\
Fig.~\ref{fig:04_05_corl21_scale_user_study2_results}(c) Pairwise (Expressiveness) & $4.3\pm0.2$\\
 
\bottomrule
\end{tabular}
\end{table}

\section{Synthetic Experiment for Section~\ref{sec:04_06_experiments}}
\label{app:99_04_corl21_ranking_syn}

\subsection{Testing $\numberOfModes>2$}

\begin{figure}[ht]
    \centering
    \includegraphics[width=0.6\linewidth]{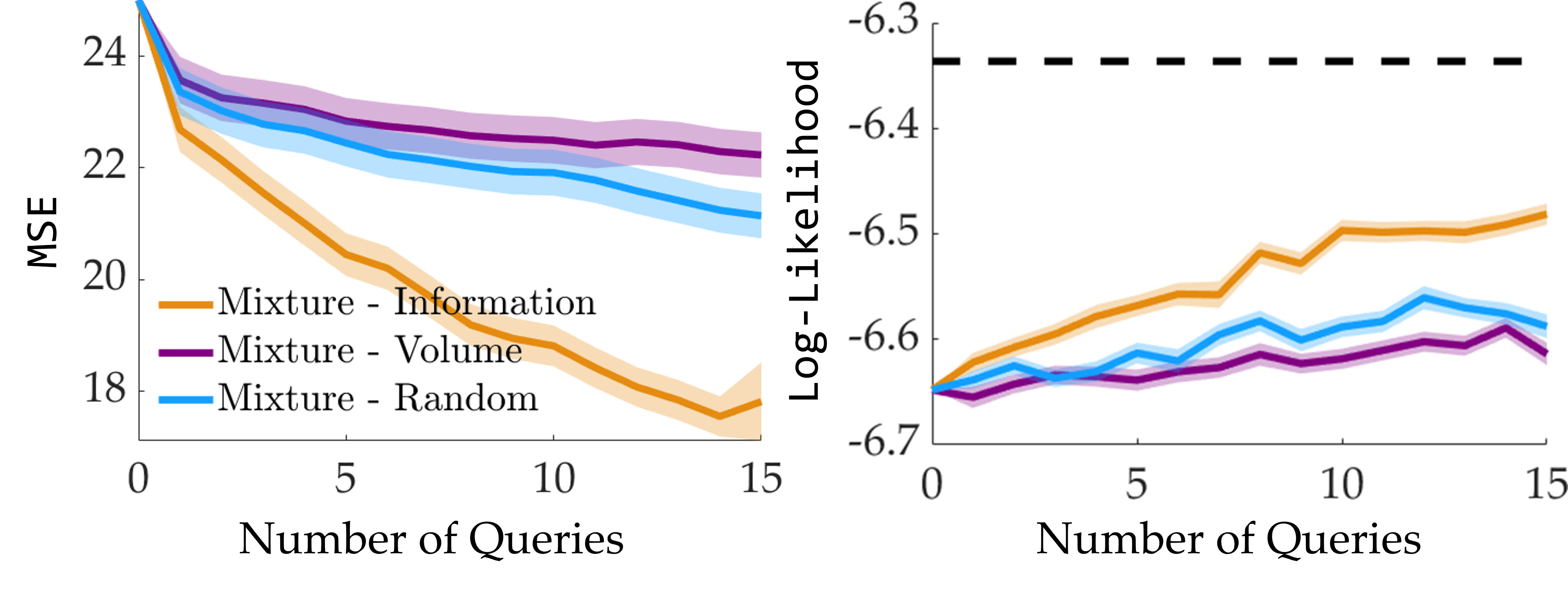}
    \caption{Different querying methods are compared on a synthetic environment (mean$\pm$se over 250 runs).}
    \label{fig:99_04_corl21_ranking_synthetic}
\end{figure}

For our first experiment with synthetic data, we demonstrate effectiveness of our approach for learning mixtures of more than two Plackett-Luce models. In particular, we evaluate our approaches using 250 sets of five randomly simulated reward weights ($\numberOfModes=5$, $\abs{\query}=6$), and trajectory features defined by $\trajectoryFeaturesFunction(\trajectory_{1:10})\sim \mathcal{N}(0,I)$, $\trajectoryFeaturesFunction(\trajectory_{11:110})\sim \mathcal{N}(0,0.1I)$, and $\trajectoryFeaturesFunction(\trajectory_{111:1110})\sim \mathcal{N}(0,0.01I)$ where 
$I$ is the $3\times3$ identity matrix and $\trajectory_{i:i'}$ refers to the $i^\textrm{th}$ through ${i'}^{\textrm{th}}$ trajectory in the trajectory dataset for generating queries. This environment models complex multimodal structure in the trajectory feature space, which is common to many robotic settings.

\begin{figure}[ht]
    \centering
    \includegraphics[width=0.35\linewidth]{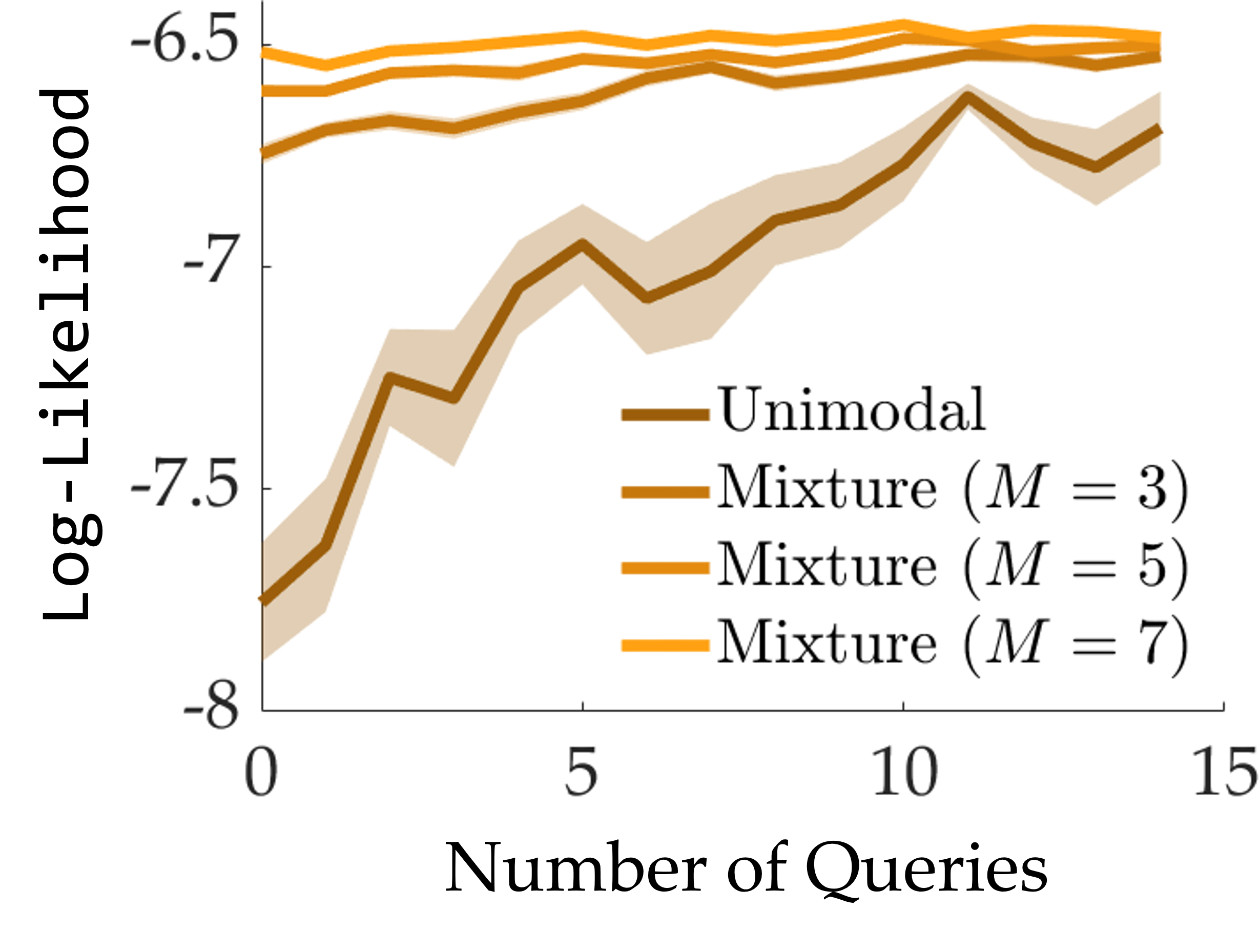}
    \caption{Different values of $\numberOfModes$ for the mutual information maximization approach are compared (mean$\pm$se over 100 runs).}
    \label{fig:99_04_corl21_ranking_vary_M}
\end{figure}

Figure~\ref{fig:99_04_corl21_ranking_synthetic} shows the results of our experiments. We see our approach, Mixture - MI dramatically outperforms the other approaches in both the \texttt{MSE} and \texttt{Log-Likelihood} metrics.

\subsection{Testing Robustness to $M$ Parameter}
\label{app:99_04_corl21_ranking_robust}

We also test the robustness of our Mixture - MI approach to misspecified $\numberOfModes$ values. We repeat the previous experiment, testing the Mixture - MI approach varying the misspecified value of $\numberOfModes$ between $\numberOfModes=1$ (Unimodal) and $\numberOfModes=3,5,7$ (see Figure~\ref{fig:99_04_corl21_ranking_vary_M}). We use the \texttt{Log-Likelihood} metric since \texttt{MSE} is not well-defined for methods with $\numberOfModes\neq 5$ because they learn a mixture of a different number of reward functions from the true synthetic mixture.

We see the best performance occurs for $\numberOfModes=5$ and $\numberOfModes=7$, with only $\numberOfModes=1$ performing significantly worse. We conclude that in this experiment our Mixture - MI approach is relatively robust to the value of $\numberOfModes$, as long as a sufficiently large value $>1$ is selected.

\section{Additional Unimodal Baseline for Section~\ref{sec:04_06_experiments}}
\label{app:99_04_corl21_ranking_new_baseline}
We test an additional baseline on the random queries made during user studies to show the superiority of our learning approach. The additional baseline represents selecting the unimodal reward with fixed norm that maximizes the reward of the top trajectory of each expert-ranked query. We compare this baseline against a learning method that computes the bimodal MLE of the reward function. Formally, for query responses $\rankingDataset=\left\{\query^{(i')},\queryResponse^{(i')}\right\}_{i'=1}^{i}$ with $\trajectory^{(i')}$ the top trajectory in the ranking $\queryResponse^{(i')} = (\trajectory^{(i')}, \ldots)$, we define this baseline to learn the parameters $(\weights,\mixingCoefficient)$ where $\mixingCoefficient=1$ and
\begin{align*}
    \tilde\weights &= \sum_{i'=1}^i \trajectoryFeaturesFunction(\trajectory^{(i')})\\
    \weights &= \frac{\tilde\weights}{\norm{\tilde\weights}_2}.
\end{align*}

Note that we do not vary the querying method in this experiment. Rather, we compare two methods of learning reward weights from the $15$ random human queries that were performed by the Mixture - Random algorithm on the \emph{FetchBanana} and \emph{LunarLander} during our user studies, and then evaluate these methods on the 10 random evaluation queries presented to the humans at the end of the experiment. We compare the two methods in terms of the \texttt{Log-Likelihood} metric. The results are presented in Table~\ref{tab:99_04_corl21_ranking_additional}, with our method denoted as ``Mixture MLE" and the new baseline described above denoted as ``Baseline".

\begin{table}[htbp]
\caption{Additional User Study Reward Learning Baseline}
\centering
\label{tab:99_04_corl21_ranking_additional}
\begin{tabular}{rcc}
& \emph{LunarLander} & \emph{Fetch Robot}\\
\cmidrule[\heavyrulewidth]{2-3}
\begin{tabular}{r}
\\
\arrayrulecolor{white}\midrule
\texttt{Log-Likelihood} \\
\\[-5px]
$p$-value
\end{tabular}
&
\begin{tabular}{cc}
\multicolumn{1}{c}{Baseline} & \multicolumn{1}{c}{Mixture MLE}\\\midrule
$-8.23 \pm 0.31$ & $-5.91 \pm 0.18$ \\
& \\[-5px]
\multicolumn{2}{c}{$6.2\cdot 10^{-7}$}
\end{tabular}
&
\begin{tabular}{cc}
\multicolumn{1}{c}{Baseline} & \multicolumn{1}{c}{Mixture MLE}\\
\midrule
$-5.21 \pm 0.22$ & $-4.70 \pm 0.35$\\
& \\[-5px]
\multicolumn{2}{c}{$0.11$}
\end{tabular}
\\
\cmidrule[\heavyrulewidth]{2-3}
\end{tabular}
\end{table}

We see the Mixture MLE method outperforms the Baseline method on both environments, with statistical significance ($p<0.05$) in \emph{LunarLander} when conducting paired $t$-tests.